\documentclass{article}

 \usepackage[preprint]{neurips_2025}

\usepackage[utf8]{inputenc} %
\usepackage[T1]{fontenc}    %
\usepackage{hyperref}       %
\usepackage{url}            %
\usepackage{booktabs}       %
\usepackage{amsfonts}       %
\usepackage{amsmath}        %
\usepackage{nicefrac}       %
\usepackage{microtype}      %
\usepackage{xcolor}         %
\definecolor{mydarkblue}{rgb}{0, 0.08, 0.45}
\hypersetup{colorlinks=true, citecolor=mydarkblue, linkcolor=mydarkblue, urlcolor=mydarkblue}
\usepackage{colortbl}       %
\usepackage{xspace}         %
\usepackage{graphicx}       %
\usepackage{float}          %
\usepackage{wrapfig}        %
\usepackage{subcaption}     %
\usepackage{tikz}           %
\usetikzlibrary{arrows.meta, positioning, calc, shapes.misc, fit}
\usepackage{pgfplots}
\pgfplotsset{compat=1.18}
\usepackage{enumitem}       %
\usepackage{listings}       %
\usepackage{tcolorbox}      %
\usepackage[normalem]{ulem}
\usepackage{titletoc}       %

\usepackage{color-edits}
\addauthor{gn}{magenta}

\tcbuselibrary{listings, skins, breakable}

\definecolor{promptbg}{HTML}{F7F7FA}
\definecolor{promptframe}{HTML}{A0A0B8}
\definecolor{prompttitlebg}{HTML}{5A6270}
\definecolor{promptkw}{HTML}{1A5276}
\definecolor{prompttempl}{HTML}{7D3C98}

\lstdefinestyle{promptstyle}{
    basicstyle=\ttfamily\scriptsize,
    breaklines=true,
    breakatwhitespace=false,
    columns=fullflexible,
    keepspaces=true,
    showstringspaces=false,
    tabsize=4,
    moredelim=[l][\color{promptkw}\bfseries]{IMPORTANT:},
    moredelim=[l][\color{promptkw}\bfseries]{CRITICAL},
    moredelim=[l][\color{promptkw}\bfseries]{SCORING RULES:},
    moredelim=[l][\color{promptkw}\bfseries]{INTEGRITY CHECK RULES},
    moredelim=**[l][\color{prompttempl}]{[System Message]},
    moredelim=**[l][\color{prompttempl}]{[User Message]},
    moredelim=**[l][\color{prompttempl}]{[Screenshots},
}

\newtcbinputlisting[auto counter]{\promptlisting}[3][]{%
    listing file={#3},
    listing only,
    listing options={style=promptstyle},
    enhanced,
    breakable,
    colback=promptbg,
    colframe=promptframe,
    colbacktitle=prompttitlebg,
    coltitle=white,
    fonttitle=\bfseries\small,
    title={Listing~\thetcbcounter: {#2}},
    arc=2pt,
    boxrule=0.5pt,
    left=6pt,
    right=6pt,
    top=4pt,
    bottom=4pt,
    toptitle=3pt,
    bottomtitle=3pt,
    #1
}

\definecolor{taskbadge}{HTML}{2E6B4F}
\newtcolorbox{taskexample}[1]{%
    enhanced,
    colback=white,
    colframe=promptframe,
    colbacktitle=taskbadge,
    coltitle=white,
    fonttitle=\bfseries\small,
    title={#1},
    arc=2pt,
    boxrule=0.5pt,
    left=6pt,
    right=6pt,
    top=6pt,
    bottom=6pt,
    toptitle=3pt,
    bottomtitle=3pt,
}

\newcommand{\niparagraph}[1]{\vspace{1pt}\noindent\textbf{#1}}

\newcommand{\bench}{\texttt{CUA-World}\xspace}
\newcommand{\method}{\texttt{Gym-Anything}\xspace}

\newcommand{\numsoftwares}{200}
\newcommand{\numenvironments}{10,000}

\newcommand{\gcmark}{\textcolor{green}{\checkmark}}
\newcommand{\rxmark}{\textcolor{red}{$\times$}}

\newcommand{\imp}[1]{{\scriptsize\textcolor{green!50!black}{(#1)}}}

\definecolor{colorSean}{RGB}{200,0,0}       %
\definecolor{colorPranjal}{RGB}{72,61,139}  %
\definecolor{colorGraham}{RGB}{255,165,0}    %

\title{Gym-Anything: Turn any Software into an Agent Environment}

\author{%
  Pranjal Aggarwal \\
  CMU\\
  \And
  Graham Neubig \\
  CMU \\
  \And
  Sean Welleck \\
  CMU \\
}

\begin{document}

\maketitle

\vspace{-12pt}
\begin{center}
\small\url{https://cmu-l3.github.io/gym-anything}
\end{center}

\begin{abstract}
    Computer-use agents hold the promise of assisting in a wide range of digital economic activities. However, current research has largely focused on short-horizon tasks over a limited set of software with limited economic value, such as basic e-commerce and OS-configuration tasks. A key reason is that creating environments for complex software requires significant time and human effort, and therefore does not scale. To address this, we introduce \textbf{Gym-Anything}, a framework for \textit{converting any software into an interactive computer-use environment}. We frame environment creation itself as a multi-agent task: a coding agent writes setup scripts, downloads real-world data, and configures the software, while producing evidence of correct setup. An independent audit agent then verifies evidence for the environment setup against a quality checklist. Using a taxonomy of economically valuable occupations grounded in U.S.\ GDP data, we apply this pipeline to 200 software applications with broad occupational coverage. The result is \textbf{\bench{}}, a collection of over 10K long-horizon tasks spanning domains from medical science and astronomy to engineering and enterprise systems, each configured with realistic data along with train and test splits. \bench{} also includes \bench{}-Long, a challenging long-horizon benchmark with tasks often requiring over 500 steps, far exceeding existing benchmarks. Distilling successful trajectories from the training split into a 2B vision-language model outperforms models 2$\times$ its size. We also apply the same auditing principle at test time: a separate VLM reviews completed trajectories and provides feedback on what remains, improving Gemini-3-Flash on \bench{}-Long from 11.5\% to 14.0\%. We release all code, infrastructure, and benchmark data to facilitate future research in realistic computer-use agents.\looseness=-1
    \end{abstract}

\begin{figure}[H]
    \centering
    \includegraphics[width=0.93\textwidth]{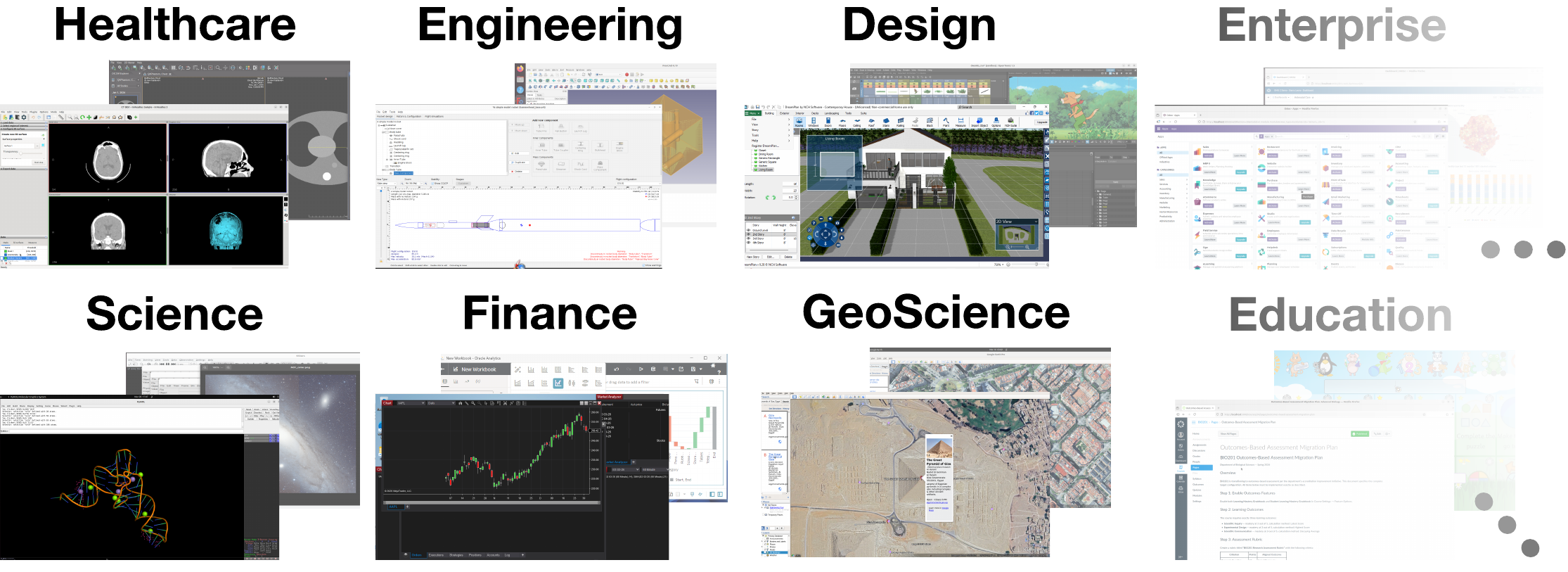}

    \caption{\textbf{Built with \method{}, \bench{} covers all major occupation groups and industries,} spanning over 10K+ long-horizon tasks and environments across 200 software applications, dramatically expanding the scope of computer-use agent evaluation and training.}
    \label{fig:strip}
\end{figure}

\section{Introduction}

Computer-use agents (CUAs, also known as graphical user interface agents) hold the promise of automating and assisting in digitally intensive occupations, which collectively represent trillions of dollars in GDP~\cite{bls2024oews,bea2024nationalaccounts}.
Yet whether these agents can handle real professional work remains an open question.
Real-world workflows are \emph{long-horizon} and take place in \emph{heterogeneous environments}, often requiring hundreds of steps across diverse software configured with domain-specific data.
For instance, end-to-end analysis of a medical imaging dataset requires a radiology tool set up with annotated clinical CT scans, while reconciling financial records across an enterprise resource planning (ERP) system requires the software populated with transaction histories and vendor accounts.
Existing benchmarks shed little light on these capabilities, as they largely test agents on short-horizon tasks such as changing a desktop wallpaper or filling a web form, over a narrow set of consumer-grade applications~\cite{osworld2024,webarena2024,visualwebarena2024,windowsagentarena2025,androidworld2025,deng2023mind2web} that only represent a small slice of the economy~\cite{wang2026doesagentdevelopmentreflect}.

\begin{figure}[ht]
    \centering
    \includegraphics[width=\textwidth]{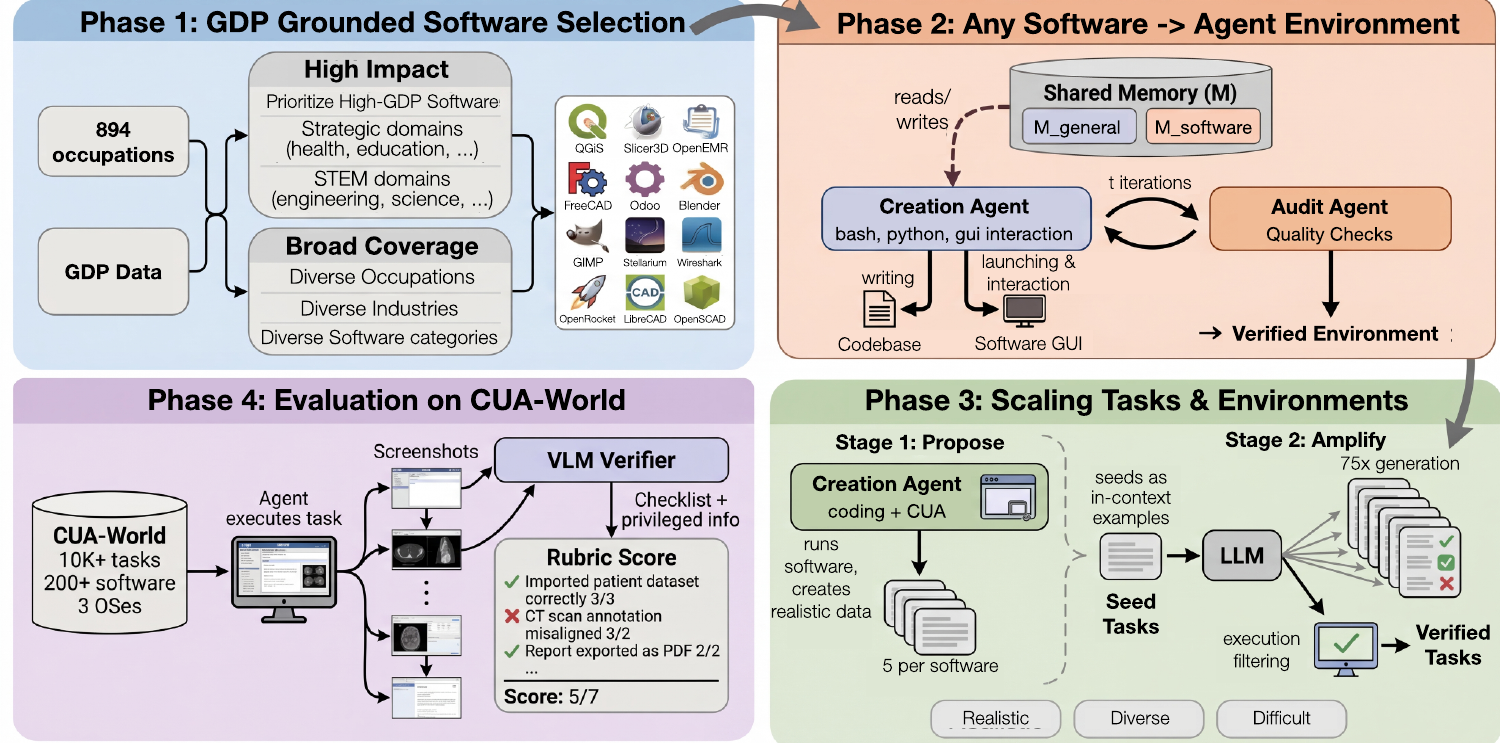}
    \caption{\textbf{Overview of the Gym-Anything pipeline.} \textbf{Phase~1:} We select {$\sim$}200 economically important software applications grounded in GDP data, balancing high economic impact with broad coverage across occupations, industries, and software categories. \textbf{Phase~2:} Each software is converted into an interactive environment via a \textit{creation-audit} loop, in which a creation agent iteratively builds and verifies the environment, while an audit agent checks quality over multiple iterations. The creation agent writes its learnings into a memory, allowing it to improve over time. \textbf{Phase~3:} Tasks are scaled with a \textit{propose-and-amplify} pattern, in which an expensive agentic model creates high-quality seed tasks (e.g., 5 per software), then a cheaper language model generates more tasks (e.g., 75$\times$) using the seeds as in-context examples. \textbf{Phase~4:} Agents are evaluated on \bench{} using a checklist-based VLM verifier with privileged information and fine-grained rubric scores.}
    \label{fig:teaser}
\end{figure}

This gap has two consequences.
First, \emph{evaluation is unfaithful}: high scores on current benchmarks reveal little about an agent's ability to operate the software that drives real economic activity.
Second, \emph{training signal is limited}: short-horizon tasks over few applications may not produce the diverse, long-horizon trajectories needed to train agents for real-world work.
The root cause is that creating realistic environments is prohibitively expensive: each software requires installation, configuration with domain-appropriate data, task design, and verification, often demanding weeks of expert effort per application~\cite{osworld2024,xu2024agenttrek}.
The core question we aim to address is: \emph{how can we scale computer-use environments for training and benchmarking agents in settings closer to real-world work?}
To address this, we introduce \method{}, a scalable framework for automatically constructing realistic environments across hundreds of economically valuable software applications. At its core, \method{} allows turning \emph{any software into an interactable environment.} We ground the software selection on U.S.\ digital GDP data, selecting software based on high economic impact and broad coverage across strategic and STEM domains, different occupations, and industries (Figure~\ref{fig:teaser}~(i)).

The key idea behind the \method{} framework, similar to agent-driven environment construction explored in other domains~\cite{zhu2025training,zhang2025autoenv}, is that \emph{creating computer-use agent environments is itself a coding and computer-use agent task.} Setting up software requires writing installation, configuration, and data-processing scripts, which are coding tasks. Verifying that the environment starts correctly and reaches the expected state requires launching it, taking screenshots, and checking the screen, which are computer-use agent tasks. However, scaling computer-use environment construction to hundreds of types of software requires handling substantial heterogeneity, including different operating systems, databases, and network configurations. To address this, we develop the \method{} library (\S\ref{sec:library}), which reduces every environment to a standardized specification: a small set of setup scripts and a configuration file. In turn, the library enables an AI agent to create environments by writing only software-specific scripts rather than dealing with low-level infrastructure.

However, without external verification, current agents prompted to create environments in this framework frequently produce incorrect environment setups. The common thread is that the agent's claims about what it has done are not reliable, but the actual state of the environment is. For instance, a screenshot of the running software reveals whether the environment is working or stuck on a crash screen, regardless of what the agent claims. We exploit this observation through a \emph{creation-audit loop} (\S\ref{sec:scaling_software}; Figure~\ref{fig:teaser}~(ii)), in which a creation agent builds an environment and produces evidence of a correct setup through screenshots, execution logs, and data outputs, then an independent audit agent verifies the evidence against a quality checklist and reports issues. In addition, the creation agent builds a shared memory of environment creation strategies that it discovers across attempts,  allowing the agent to improve over time.

Next, we adopt a \emph{propose-and-amplify} strategy (Figure~\ref{fig:teaser}~(iii); \S\ref{sec:tasks}) for generating realistic tasks at scale within the software environments. In this pattern, an expensive agentic model proposes a small number of seed tasks per software and runs the tasks, and then a cheaper non-agentic language model amplifies these seeds into a larger set using the seed implementations as in-context examples.
To evaluate the resulting tasks, we use a checklist-based VLM verifier that breaks each task into weighted subtasks for partial credit (\S\ref{sec:verifiers}). To construct the checklists, we leverage data that is embedded in the environment's setup scripts (e.g., the correct tumor location from a downloaded medical dataset). Importantly, agents do not have access to this information when solving a task, thereby making it a form of \emph{privileged information} that the verifier can leverage in order to check the agent's outputs.

We use \method{} to construct \bench{}, a collection of over \numenvironments{} tasks across \numsoftwares{} software applications. \bench{} spans domains such as medical science, astronomy, engineering, finance, enterprise systems, and educational platforms. It includes tasks on three operating systems, and is separated into train and test splits (Table~\ref{tab:benchmark_comparison}). To demonstrate the utility of the training split, we distill trajectories from a strong teacher model into a 2B vision-language model and find that performance scales with the number of software and environments in the training set. The trained model also generalizes to software not seen during training, and outperforms models 2$\times$ its size.

Given the breadth and realism of \bench{}'s software and tasks, we further construct a challenging long-horizon benchmark, \bench{}-Long, consisting of one task per software, with tasks often requiring hundreds of steps. Each task is designed to be realistic and to target common failure modes of existing models. Even the strongest frontier model achieves only 27.5\% pass rate on \bench{}-Long, highlighting the difficulty of long-horizon tasks. One common failure mode is that agents often stop early, claiming the task is complete when it is not. To address this, we apply a similar Test-Time Auditing principle (\S\ref{sec:test_time_agent}), where an independent model reviews the agent's trajectory upon completion and provides feedback on what remains, improving pass rate from 11.5\% to 14.0\% for Gemini-3-Flash. Although Test-Time Auditing helps, \bench{}-Long remains largely unsolved, offering a new challenging benchmark for frontier computer-use agents on realistic tasks.

In summary, we contribute (1)~\method{}, a modular framework and multi-agent pipeline for converting any software into an interactive computer-use environment; (2)~\bench{}, a collection of \numenvironments{} tasks across a GDP-grounded selection of \numsoftwares{} software applications with checklist-based VLM verification, train/test splits, and a challenging long-horizon split (\bench{}-Long) requiring hundreds of steps; (3)~training and test-time scaling results, including distillation to a 2B model that outperforms models 2$\times$ its size, and a test-time audit agent that improves long-horizon performance on \bench{}-Long; and (4)~a full release of all code, infrastructure, and benchmark data.

\section{Methodology}
\label{sec:method}

In this section, we introduce the problem setup, the GDP-grounded software selection procedure, and the library abstraction that makes large-scale environment construction possible. In Section~\ref{sec:scaling_software}, we further discuss our multi-agent strategy to scale the number of software, and in Section~\ref{sec:tasks}, discuss how to further scale tasks and environments for the relevant software.

\subsection{Problem Setup}
\label{sec:problem_setup}

\niparagraph{Environment}
We refer to an environment $\mathcal{E}$ as one or more interactive software applications with a specified initial state of the
filesystem and processes running inside an operating system. The agent interacts with the environment through keyboard and mouse
actions.

\niparagraph{Task.}
A task $\mathcal{T} = (\mathcal{E}_{s_0}, p, V)$ consists of an environment $\mathcal{E}$ with initial state $s_0$, a natural-language instruction $p$, and a verification function $V$ that maps the agent's trajectory to a score.

\niparagraph{Interaction.}
At each step $t$, the agent receives an observation $o_t$ (e.g., a screen capture) and executes mouse/keyboard actions $a_t$, after which the environment returns a new observation $o_{t+1}$.
An episode proceeds by resetting the environment to $s_0$, letting the agent interact for up to $T$ steps, and the final score is determined by $V$.

\subsection{GDP-Grounded Software Selection}
\label{sec:selection}

\begin{figure}[t]
    \centering
    \includegraphics[width=\textwidth]{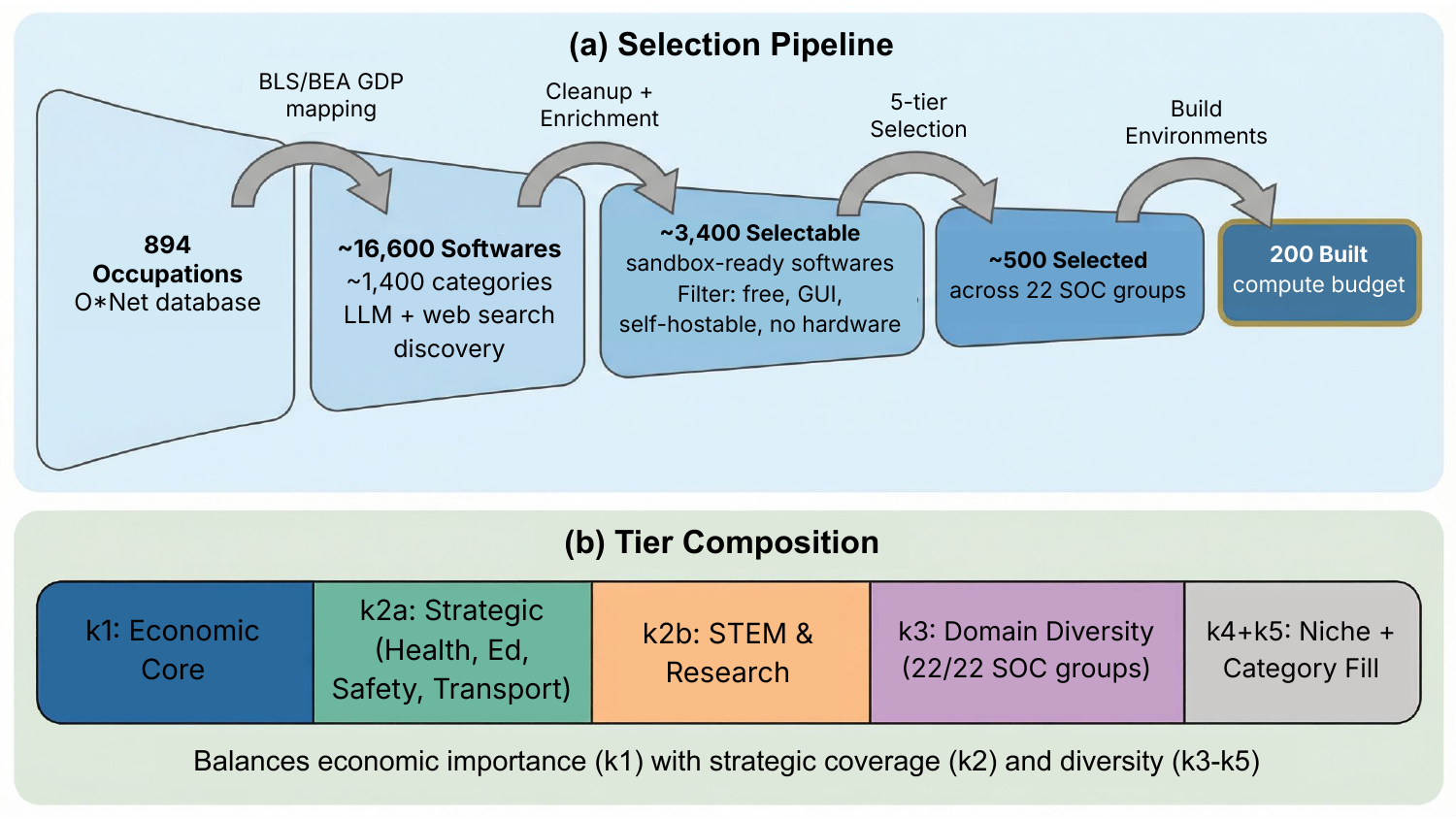}
    \caption{\textbf{GDP-grounded software selection pipeline.} Starting from U.S.\ occupational data, we estimate per-software GDP, filter to sandboxable candidates, and apply tiered selection to yield \numsoftwares{} software applications.}
    \label{fig:gdp_pipeline}
\end{figure}

Determining which software to include in a CUA training dataset and benchmark is an important design decision. We use a simple principle: \emph{prioritize software that drives more economic activity.} Unlike prior benchmarks, we ground our selection in publicly available U.S.\ digital economy data (Figure~\ref{fig:gdp_pipeline}). In a nutshell, we estimate GDP per occupation, discover software used by each occupation, attribute GDP to individual software, filter to sandboxable candidates, apply tiered selection, and based on our compute budget select \numsoftwares{} software applications. We detail each step below.

\niparagraph{Estimating GDP per occupation.}
The O*NET data on the U.S.\ economy comprises ${\sim}$900 standardized occupations, each with publicly available data on employment counts and average wages.\footnote{Occupations from the O*NET database~\cite{peterson2001onet}; employment and wage data from the Bureau of Labor Statistics (BLS)~\cite{bls2024oews}.} For each occupation, we compute a wage bill (employment $\times$ mean wage) and scale it to total GDP using national accounts data~\cite{bea2024nationalaccounts}, yielding a GDP estimate per occupation that sums to the full U.S.\ GDP (Appendix~\ref{app:gdp_pipeline}).

\niparagraph{Discovering software per occupation.}
Next, we need to know what software each occupation actually uses. We use an LLM with web-search access to discover relevant software categories and enumerate software per category for each occupation, producing a catalog of {$\sim$}16,600 software applications across {$\sim$}1,400 categories. The catalog is cleaned by deduplicating categories, validating software-category assignments, and removing hallucinated entries via web-grounded verification.

\niparagraph{Attributing GDP to software.}
Not all of an occupation's output involves computers, and not all computer work uses the same software. We decompose each occupation's GDP into a software-level estimate:
\begin{equation}
\text{GDP}_{\text{software}} = \textstyle\sum_{\text{occ}} \text{GDP}_{\text{occ}} \times p_{\text{computer}} \times s_{\text{category}} \times s_{\text{software}}
\label{eq:gdp}
\end{equation}
Here, $p_{\text{computer}}$ is the fraction of the occupation's work that involves computers (available from occupational surveys), $s_{\text{category}}$ is the share of that computer work attributed to a software category (e.g., ``spreadsheets'' for an accountant), and $s_{\text{software}}$ is the software's share within its category (e.g., Excel's share of spreadsheets). The share factors $s_{\text{category}}$ and $s_{\text{software}}$ are estimated by an LLM with web-search access.

\niparagraph{Filtering to sandboxable software.}
Not all economically important software can be sandboxed into an interactive environment, since they may require paid licenses, organizational credentials, or specialized hardware.
We classify each software application as sandbox-ready if it satisfies the following constraints: (a) self-hostable, i.e., does not require an online account to use, (b) free-tier, i.e., can be used free of charge without license restrictions, (c) has a GUI, and (d) does not require hardware that cannot be simulated (e.g., CNC machines). We select {$\sim$}3,400 of {$\sim$}16,600 that satisfy these constraints.
Further, when a software is not sandboxable, we substitute the closest sandboxable alternative from the same software category, aiming to preserve the economic signal.

\niparagraph{Tiered selection.}
From the filtered catalog, we select \numsoftwares{} software applications across five tiers that balance economic importance with diversity: (1)~highest-GDP software overall, (2a/2b)~strategically important domains (Healthcare, Education, Protective Services, Transportation) and STEM domains (Architecture/Engineering, Computer/Math, Life/Physical/Social Science), (3)~cycling through all 22 SOC major groups to select {$\sim$}5 per group, ensuring every occupation group has representation, (4)~software unique to specific occupations or domains not yet covered, and (5)~software from uncovered categories, ranked by GDP. We build environments for \numsoftwares{} software applications based on our compute budget, although the pipeline is fully automated and extensible (Appendix~\ref{app:gdp_pipeline}). \looseness=-2

\subsection{The Gym-Anything Library}
\label{sec:library}

Constructing computer-use environments across hundreds of diverse software applications requires a unified framework that works across operating systems, application types, and compute backends without per-environment engineering. Previous works have primarily constructed computer-use environments manually by interacting with an actual operating system, and then taken VM snapshots to be reused later~\cite{osworld2024}. However, snapshots cannot be inspected, version-controlled, or partially reused across tasks, and modifying anything requires repeating the manual setup, therefore limiting modularity, reproducibility, and scalability.
To handle these challenges, we construct the \method{} Library.
In \method{}, each environment is defined by a simple specification: three sequential setup scripts and a declarative configuration file. The scripts progress from general to task-specific: \emph{install} installs the software and its dependencies, \emph{configure} sets it up with realistic data and settings, and \emph{task setup} configures the specific starting state for a given task. This separation ensures that multiple tasks for the same software share the same install and configure scripts, varying only the task-specific setup.
For example, creating a LibreOffice Calc environment requires only an install script (e.g., \texttt{apt-get install libreoffice}), a configure script that downloads a sample spreadsheet, and a config file specifying the OS image and resource limits; the library handles container orchestration, display forwarding, and checkpoint management automatically.
This design reduces environment creation to a scripting task: users and AI agents create new environments by writing setup scripts and key-value configurations, then interact with any environment through a standard gymnasium-style API~\cite{towers2024gymnasium,brockman2016openai} that provides a unified observation space (e.g., screenshots) and action space (e.g., keyboard and mouse inputs), with the library handling display forwarding and input translation across operating systems.
This specification is simple enough that an LLM agent can author environments autonomously (\S\ref{sec:scaling_software}), yet expressive enough to capture complex, production-grade software configured with realistic data, ranging from desktop image editors to multi-container enterprise systems. Behind this simple interface, the library manages the complexity of running environments across three operating systems (Linux, Windows, Android) and multiple compute backends (such as docker and apptainer for rootless systems such as slurm). The staged design further enables caching at each stage boundary, so creating new tasks only requires re-running the task-specific setup. Combined with network-process-file isolation, this enables massive parallelization; in our experiments, we run 400+ concurrent environments across 1,600 CPUs (Appendix~\ref{app:framework}).

\section{Scaling Computer-Use Agent Software Applications}
\label{sec:scaling_software}

\begin{figure}[t]
    \centering
    \includegraphics[width=\textwidth]{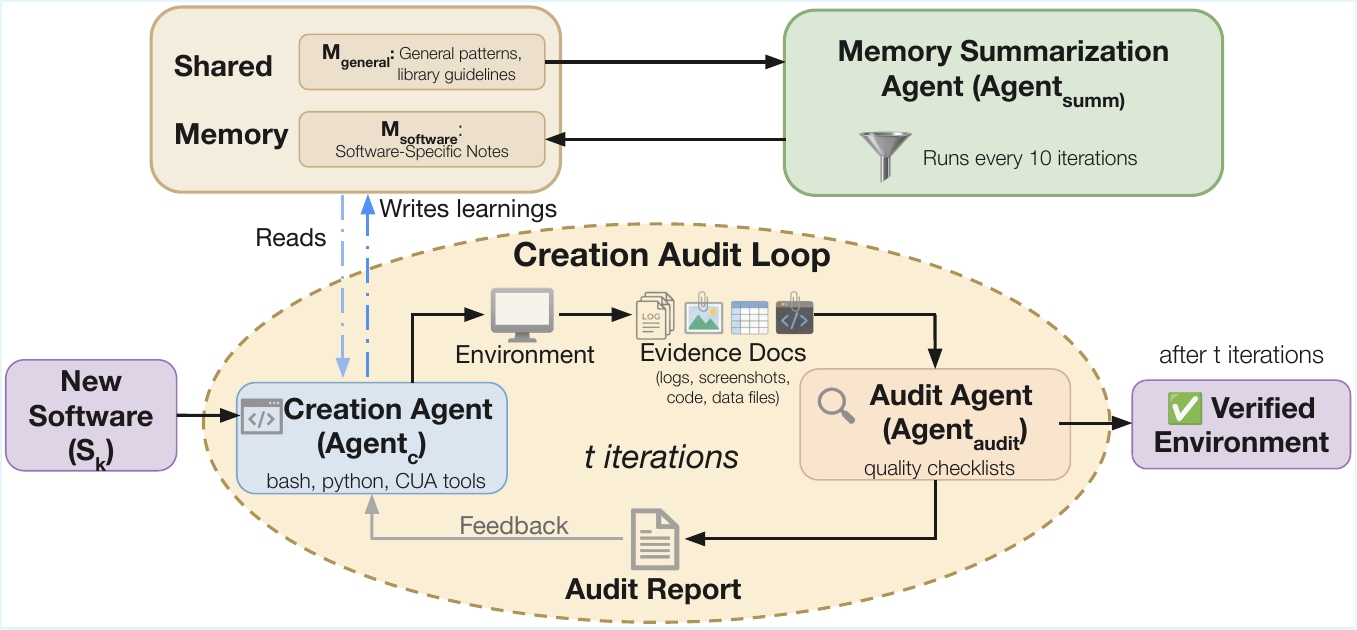}
    \caption{\textbf{The \method{} creation-audit loop.} A Creation Agent writes setup scripts and produces evidence documents (screenshots, logs, etc.)\ while an Audit Agent evaluates this evidence against quality checklists and returns feedback. Learnings accumulate in a shared memory $M$, which a Summarization Agent periodically condenses so that newer environments are created faster.}
    \label{fig:full_pipeline_gemini_2}
\end{figure}

Setting up real-world software as interactable environments is hard, laborious, and time-consuming, even for expert humans~\cite{osworld2024,xu2024agenttrek,he2024pcagent}. Each software requires installation, configuration with domain-appropriate data, and verification; for instance, a radiology tool requires annotated clinical CT scans, while an ERP system needs transaction histories and vendor accounts. This often demands weeks of expert effort per application, naturally limiting scalability.

The key idea is that \textit{setting up computer-use agent environments is itself a coding + computer-use agent task.}
Because the Gym-Anything library (\S\ref{sec:library}) constrains environment creation to a fixed, small interface (writing setup scripts and config files), the creation task becomes a coding task. Further, verifying whether the environment is correctly set up requires launching it and interacting with it, which is a computer-use agent task.

However, naively prompting even state-of-the-art agents results in poor environments; the agent stops early, uses fake placeholder data, leaves the software at the wrong starting screen, or claims things are done without actually verifying them.
We therefore propose a multi-agent framework that iteratively creates, audits, and improves environments, while accumulating learnings in a shared memory.

\niparagraph{Multi-agent framework.}
Each agent in our framework is an instance of Claude Opus 4.5/4.6~\cite{anthropic2025claude} run via Claude Code, differentiated by a.) access to specific tools, and b.) the objective described by its system prompt. In a nutshell, these agents iteratively generate environments, audit the quality and improve them, and document the learnings for future attempts in a shared memory $M$. We next describe each of the 3 agents in detail.

\niparagraph{Creation agent ($\text{Agent}_C$).}
This is a coding agent equipped with bash, python, and computer-use (for visual grounding) tools, with complete access to the \method{} library and all previously created environments.
Given a new software name $S_k$, we prompt $\text{Agent}_C$ with a software-agnostic detailed prompt describing the workflow to follow and library usage, along with $S_k$, with the objective of implementing the software as an environment and then verifying it by actually running and interacting with it. Before writing any scripts, the agent first researches how the software should be configured, finds and downloads real-world data for the environment (e.g., public medical imaging datasets for radiology software, published email corpora for messaging clients), and studies similar previously created environments. It then implements the setup scripts, launches the environment, takes screenshots, uses visual grounding to check that the application reached the expected state (as intended by the setup scripts), and iteratively debugs failures.
Crucially, the agent is required to produce evidence that the software was set up correctly in the form of screenshots of the running software, execution logs, etc. (see Appendix~\ref{app:evidence_docs} for an example).

However, the agent often declares the task done prematurely. For instance, it may use placeholder data instead of real datasets, leave the software at the wrong screen, or never verify the task by actual execution.
We speculate that these failures are due to context fatigue~\cite{liu2024lostmiddle, sinha2026illusiondiminishingreturnsmeasuring}: after hundreds of thousands of tokens, the agent loses track of what it still needs to do.
To address this, whenever the agent stops, we re-prompt it to reread the setup guidelines, reread the checklists, and complete any requirements it may have skipped. We find this simple technique recovers many omissions.

\niparagraph{Audit agent ($\text{Agent}_{\text{audit}}$).}
While $\text{Agent}_C$ typically gets the environment running, its claims about what it has done are not always reliable. For instance, it may leave the software at a setup wizard instead of the main screen, use placeholder data, or skip verification entirely. However, the evidence produced above reveals the actual state of the environment regardless of what the agent claims: a screenshot shows whether the software is running correctly or stuck on an error screen.

To verify this evidence, we use $\text{Agent}_{\text{audit}}$, a similar coding+computer-use agent that acts as an adversary to $\text{Agent}_C$ and evaluates whether the evidence demonstrates that the environment satisfies a set of quality checklists (see Appendix~\ref{app:evidence_docs}).
It does so by analyzing the screenshots and logs, inspecting the actual config and script files, and, if necessary, actually running the environment.
Given the implementation of software $S_k$, the audit agent outputs an audit detailing what is correctly implemented and what the critical issues are (Appendix~\ref{app:audit_examples} contains example audits, and Appendix~\ref{app:audit_cross_round} shows how issues are corrected across audit rounds).

In principle, both $\text{Agent}_C$ and $\text{Agent}_{\text{audit}}$ have access to the same tools and files; the only difference lies in their prompt. We find this separation offers multiple benefits: a.) separating agents removes any self-confirmation bias, and we find audits are more detailed and accurate than self-review (\S\ref{par:cross_model_audit}), b.) the written audits ensure higher interpretability, letting human authors independently verify quality (Appendix~\ref{app:audit_examples}), and c.) the adversarial framing catches cases where $\text{Agent}_C$ made self-misleading claims.
The audit findings are fed back to $\text{Agent}_C$ for correction, and this loop runs for $t$ iterations. A key feature of our framework is that agents accumulate learnings in a shared memory, allowing them to improve over iterations. We describe this mechanism next.

\niparagraph{Shared memory.}
The creation agent maintains a shared memory $M$, effectively a directory of files that grows over time. $M$ is initialized with the hand-written prompt for $\text{Agent}_C$ (describing the setup workflow, checklists for verification, and library usage) and evolves as agents add their learnings.
After each environment, $\text{Agent}_C$ documents what it tried, what failed, and what fixed it, updating $M$ in two places: software-specific notes $M_{\text{soft}}$ and general notes $M_{\text{general}}$ that could help future agents. For instance, one agent discovered that a multi-service web platform needed readiness polling before the GUI could launch; once added to $M$, it became the default for all subsequent web stacks, resulting in faster creation.
This ensures sublinear growth in creation time: as more environments are built, newer environments can be created faster.
Further, $M$ acts as an asynchronous but shared memory, such that multiple agents running in parallel can write to and read from discoveries made by other agents.
However, as more environments are created, $M_{\text{soft}}$ grows large, causing future agents to miss important details due to long contexts. To address this, in every $L$ environment, a memory summarization agent ($\text{Agent}_{\text{summ}}$) reads through all memory files, finds common patterns, and summarizes findings from $M_{\text{soft}}$ into $M_{\text{general}}$. This theoretically adds only ${\sim}1/L$ overhead compared to each agent reading the full $M_{\text{soft}}$ every time.

\niparagraph{Output.}
Applying this recipe to the software identified through our GDP-grounded selection (\S\ref{sec:selection}), we construct environments for \numsoftwares{} software applications across three operating systems (Linux, Windows, Android), ranging from desktop applications to multi-service enterprise systems, each configured with realistic data (public email corpora, medical imaging datasets, financial schemas, and government open data). We select \numsoftwares{} based on our current compute budget; the pipeline is fully automated and extensible to additional software.
We next describe how tasks are generated for these environments.

\section{Scaling Tasks}
\label{sec:tasks}

While \S\ref{sec:scaling_software} addressed the primary bottleneck of getting complex software to run correctly (handling installation, configuration, and background services), generating diverse tasks over these environments poses a separate scaling challenge. Recall from \S\ref{sec:problem_setup} that a task requires a starting environment state $\mathcal{E}(s_0)$, a natural-language instruction $p$, and a verifier $V$. Once the base software is configured, creating new tasks reduces to generating these task-specific assets.

Nonetheless, naively prompting a model to generate these often results in subpar quality. For instance, setup scripts reference non-existent data, formats mismatch the software's expectations, and instructions are either trivially simple or impossible to execute from the given starting state. Conversely, relying purely on agentic models to author and validate every task is prohibitively expensive for scaling.

To scale task creation efficiently, we propose a \emph{propose-and-amplify} strategy. First, a proposer agent (Claude Opus 4.5/4.6 via Claude Code, equipped with computer-use tools) proposes a small set of high quality, difficult \emph{seed tasks} per software. The agent is provided with a set of guidelines for high quality tasks across three dimensions: a.)~realism (does the instruction reflect a genuine, real-world use case?), b.)~difficulty (does the task require a long-horizon, multi-step trajectory to solve?), and c.)~diversity (do the tasks cover varied functionalities of the software?).
An agentic loop is necessary here because the model must actively run the software, search, download or generate realistic data, interact via the GUI, and verify the resulting state. Crucially, this expensive step only occurs once per software, ensuring core functionality across relevant occupations identified in \S\ref{sec:selection} is covered.

Second, for amplification, a non-agentic LLM (Gemini 3 Pro) uses these high-quality seeds as in-context examples to generate additional tasks at scale. 
While the agentic seeds ensure realism and difficulty in further generated tasks, naively sampling from a non-agentic LLM often yields repetitive or very similar instructions. To enforce diversity, we generate tasks sequentially, providing the model with all previously generated instructions $1, \ldots, t$ as context for task $t{+}1$. We subsequently apply semantic similarity filtering to discard duplicate tasks.

Finally, because the non-agentic LLM generates tasks without interactive execution, we implement an automated filtering step. We launch each generated task, capture the starting state observation $o_0$, and pass it alongside the instruction to a Vision-Language Model (VLM) to check whether the start state matches the expectation from the task description. Tasks that fail this test are filtered from the dataset. Examples of task descriptions and starting states are provided in Appendix~\ref{app:task_examples}.
\begin{table*}[t]
\centering
\setlength{\tabcolsep}{3pt}
\renewcommand{\arraystretch}{1.04}
\tiny
\caption{\textbf{Full-text examples of task descriptions, privileged information, and VLM checklist items.}; color only highlights the most important privileged information.}
\label{tab:pi_examples}
\begin{tabular}{@{}>{\raggedright\arraybackslash}p{0.09\textwidth}>{\raggedright\arraybackslash}p{0.31\textwidth}>{\raggedright\arraybackslash}p{0.29\textwidth}>{\raggedright\arraybackslash}p{0.27\textwidth}@{}}
\toprule
\textbf{Software} & \textbf{Task Description} & \textbf{Privileged Information} & \textbf{VLM Checklist} \\
\midrule
AstroImageJ &
Analyze the WASP-12 (RA: 06:30:32.79, Dec: +29:40:20.4) astronomical image sequence from January 5-6, 2016, to identify evidence of a planetary transit. If a transit is detected, determine the transit depth, mid-transit time (BJD\_TDB or JD), and transit duration in hours. Using a host star radius of 1.599 solar radii, estimate the planet's radius in Jupiter radii. Save your findings and uncertainties to \url{~/Documents/transit_analysis.txt}. &
Target: WASP-12b. Expected Transit Depth: \textcolor{blue}{\textasciitilde1.4\%} (0.014 relative flux). Expected Duration: \textcolor{blue}{\textasciitilde2.7 hours}. Expected Planet Radius: \textcolor{blue}{\textasciitilde1.79 Jupiter radii} (calculation: sqrt(0.014) * 1.599 R\_sun * conversion factors). The dataset is real ground-based imagery from Jan 5-6, 2016, so the light curve will have noise but the transit dip should be clearly visible. &
1. The agent loads the sequence of astronomical images into AstroImageJ. \newline
2. The agent selects the target star (WASP-12) and appropriate comparison stars for differential photometry. \newline
3. The agent generates and displays a light curve plot showing the star's flux over time. \newline
4. The agent fits a model or trend line to the data to characterize the transit. \newline
5. The agent reports the measured transit depth, mid-transit time, and duration. \newline
6. The agent calculates and reports the planet's radius in Jupiter radii. \\

Apache OpenOffice Writer &
You are a Clinical Research Associate (CRA) performing an Interim Monitoring Visit (IMV) for Protocol ZN-994 at Site 142. Using the visit data in \url{/home/ga/Documents/visit_notes.json}, create a formal IMV Report in Apache OpenOffice Writer saved as \url{/home/ga/Documents/IMV_Report_Site_142.odt}. The report must include a document header with the Protocol and Site numbers, page numbers in the footer, and sections using `Heading 1' style for `Visit Details', `Subject Enrollment', `Protocol Deviations', and `Action Items'. [\ldots] &
Protocol Number: \textcolor{blue}{ZN-994}. Site Number: \textcolor{blue}{142}. Enrollment counts from the input file: \textcolor{blue}{Screened=18, Randomized=14, Completed=3, Discontinued=2}. The calculated `Active' count must be exactly \textcolor{blue}{9} (14 - 3 - 2 = 9). The Action Items table contains items \textcolor{blue}{AI-02 and AI-03} which have an \textcolor{red}{`Open'} status and must be formatted with yellow highlight or red text. The four required sections are `Visit Details', `Subject Enrollment', `Protocol Deviations', and `Action Items'. &
1. Verify the document has a header with Protocol and Site numbers, and a footer with page numbers. \newline
2. Verify the four required sections are present and use the `Heading 1' style. \newline
3. Verify the Subject Enrollment section contains a table with the correct base counts. \newline
4. Verify the `Active' count in the Subject Enrollment table is correctly calculated. \newline
5. Verify the Protocol Deviations list and Action Items table structure. \newline
6. Verify that Action Items with an `Open' status are conditionally formatted. \\

Aerobridge &
Calculate the total flight duration in minutes for all flight plans belonging to the operator `SkyHigh Surveyors' that occurred in October 2023. Save the total number of minutes to \url{/home/ga/Documents/utilization_report.txt}. &
The correct total flight duration for `SkyHigh Surveyors' in October 2023 is \textcolor{blue}{135 minutes} (acceptable tolerance: 133-137 minutes). This is calculated from two flights: Flight 1 on Oct 5 (10:00-10:45, 45 mins) and Flight 2 on Oct 12 (14:00-15:30, 90 mins). Distractor flights (wrong operator or wrong month) must be excluded. &
1. Locate flight plans belonging to the operator `SkyHigh Surveyors'. \newline
2. Identify the flight plans that occurred in October 2023. \newline
3. View the details or times of the relevant flights to calculate duration. \newline
4. Open or create the utilization report file. \newline
5. Save the correct total flight duration to the report file. \\

Liverpool Cancer iChart &
Using the Liverpool Cancer iChart Archive app, determine the drug-drug interaction between Dabrafenib and Ketoconazole (located in the Antifungal agents category). Leave the application on the screen displaying the interaction result. &
Dabrafenib is a cancer drug (BRAF inhibitor) and Ketoconazole is an antifungal agent. The interaction between them is clinically significant due to potent CYP3A4 inhibition by ketoconazole, which increases dabrafenib \textcolor{blue}{AUC by 71\% and Cmax by 33\%}. The VLM should look for an interaction result indicating a \textcolor{red}{severe/red warning} or mentioning CYP3A4 inhibition. &
1. The Liverpool Cancer iChart Archive application is opened. \newline
2. Dabrafenib is selected as the cancer drug. \newline
3. The Antifungal agents category is accessed. \newline
4. Ketoconazole is selected as the comedication. \newline
5. The interaction result between Dabrafenib and Ketoconazole is displayed on the screen. \\
\bottomrule
\end{tabular}
\end{table*}

\subsection{Task Verification}
\label{sec:verifiers}

Recall that each task $\mathcal{T} = (\mathcal{E}_{s_0}, p, V)$ includes a verification function $V$ that maps the agent's trajectory to a score (\S\ref{sec:problem_setup}). Evaluating long-horizon trajectories requires $V$ to be both robust and granular. We construct $V$ as a checklist-based VLM verifier augmented with \emph{privileged information}.

\niparagraph{Privileged information.}
Each task's starting state $s_0$ is configured by the setup scripts $\mathcal{S}$ (the install, configure, and task setup scripts from \S\ref{sec:library}). These scripts contain ground-truth data that is not present in the task description $p$ but is deterministically tied to the environment's configuration. We call this \emph{privileged information} $\mathcal{I} = \text{Extract}(\mathcal{S}, p)$, extracted automatically by a separate coding agent that parses the scripts, retrieves, or searches online for the relevant ground truth. For instance, in a medical imaging task, the correct tumor location is already known from the downloaded dataset; in a financial task, the expected account balances are determined by the initialization data. Importantly, $\mathcal{I}$ assists the VLM verifier rather than making the task artificially harder for the evaluated computer-use agent.

\niparagraph{Checklist-based verification.}
The VLM verifier uses $\mathcal{I}$ alongside the task instruction $p$ to generate a granular checklist $C = \{(c_i, w_i)\}_{i=1}^{N}$, where each $c_i$ defines a specific subtask to verify and $w_i$ is its point value. Given the evaluated agent's trajectory $\tau$, the verification score is:
\begin{equation}
V(\tau) = \textstyle\sum_{i=1}^{N} w_i \cdot \text{VLM}(\tau, c_i, \mathcal{I})
\end{equation}
where each $\text{VLM}(\tau, c_i, \mathcal{I})$ returns a binary judgment of whether subtask $c_i$ was completed, using $\mathcal{I}$ to check the agent's outputs against known ground-truth answers. This formulation allows partial credit on complex, multi-step tasks without requiring manual annotation.
Table~\ref{tab:pi_examples} shows representative examples of privileged information and the concrete checklist items it enables across scientific analysis, clinical reporting, business operations, and clinical decision support.

In addition to $C$, the VLM verifier evaluates a separate integrity checklist $C_{\text{int}}$ to ensure the evaluated computer-use agent did not bypass the intended workflow: a.)~the intended software was used to complete the task, b.)~the required application state was reached through the software's own interface rather than by directly editing configuration or data files, and c.)~the agent did not exploit environment artifacts to shortcut the work. Failing any integrity item sets $V(\tau) = 0$, regardless of the task checklist score.  
We manually compared human-agreement rates of our checklist-based VLM verification against end-state-only VLM verification and programmatic verification, finding the proposed method to be significantly more reliable (\S\ref{sec:verifier_robustness}).

\section{\bench{}}
\label{sec:benchmark}

\definecolor{ourrow}{RGB}{219, 234, 254}   %
\definecolor{sechdr}{RGB}{243, 244, 246}   %

\newcommand{\pcmark}{\textcolor{orange!80!black}{$\checkmark^{\!*}$}}
\newcommand{\nr}{\textit{n.r.}}

\begin{table*}[t]
\centering
\setlength{\tabcolsep}{4pt}
\renewcommand{\arraystretch}{1.18}
\caption{%
    \textbf{Comparison of \bench{} with datasets and environments for computer-use agents.}
    \gcmark\;yes;\;
    \rxmark\;no;\;
    \pcmark\;partial or with caveats;\;
    \textbf{---}\;not applicable.
    $^{\star}$ Benchmark allows or requires {>}100 agent steps per task.
    $^{*}$ Offline human demonstrations only (not interactive verified trajectories).
    $^\S$ Number of 2018 SOC (Standard Occupational Classification) major occupation groups (out of 22 civilian groups)
    whose workers would routinely use the benchmark's applications;
    counted conservatively such that a group is included only if tasks directly simulate
    work in that occupation.
}
\label{tab:benchmark_comparison}
\resizebox{\textwidth}{!}{%
\begin{tabular}{@{} l c  c c  r r  c c  c c @{}}
\toprule

 & &
\multicolumn{2}{c}{\textbf{Environment}} &
\multicolumn{2}{c}{\textbf{Scale}} &
\multicolumn{2}{c}{\textbf{Task Properties}} &
\multicolumn{2}{c}{\textbf{Infrastructure}} \\[-2pt]

\cmidrule(lr){3-4}
\cmidrule(lr){5-6}
\cmidrule(lr){7-8}
\cmidrule(lr){9-10}

\textbf{Benchmark} &
\textbf{Agent} &
\rotatebox{60}{\textbf{Interactive}} &
\textbf{Platform} &
\textbf{\#\,SW} &
\textbf{\#\,Tasks} &
\rotatebox{60}{\textbf{Long-Horizon$^{\star}$}} &
\rotatebox{60}{\textbf{Econ.\,Cov.$^\S$}} &
\rotatebox{60}{\textbf{Auto-Create}} &
\rotatebox{60}{\textbf{Train Split}} \\

\midrule

\rowcolor{sechdr}
\multicolumn{10}{@{}l}{\small\textit{Static Datasets}} \\[-3pt]

Mind2Web~\cite{deng2023mind2web}        %
    & Web & \rxmark & Web
    & 137   & 2{,}350
    & \textbf{---}  & 7/22
    & \rxmark  & \pcmark \\

AITW~\cite{rawles2023androidinthewild}            %
    & CUA & \rxmark & Android
    & 357+  & 715K
    & \textbf{---}  & 4/22
    & \rxmark  & \pcmark \\

AndroidControl~\cite{androidcontrol2024}  %
    & CUA & \rxmark & Android
    & 833   & 15{,}283
    & \textbf{---}  & 7/22
    & \rxmark  & \pcmark \\

OmniACT~\cite{omniact2024}         %
    & CUA & \rxmark & Lin / Win / macOS / Web
    & 65    & 9{,}802
    & \textbf{---}  & 6/22
    & \rxmark  & \pcmark \\
GDPval~\cite{gdpval2025}
    & LLM & \gcmark & \textbf{---}
    & \textbf{---}  & 1{,}320
    & \gcmark   & 13/22
    & \rxmark  & \rxmark \\
\midrule

\rowcolor{sechdr}
\multicolumn{10}{@{}l}{\small\textit{Interactive Benchmarks}} \\[-3pt]

MiniWob{++}~\cite{liu2018miniwob}     %
    & Web & \gcmark & Web (sim.)
    & 1     & $80^{\dagger}$
    & \rxmark   & 3/22
    & \rxmark  & \gcmark \\

WebArena~\cite{webarena2024}        %
    & Web & \gcmark & Web
    & 6     & 812
    & \rxmark   & 5/22
    & \rxmark  & \rxmark \\

VisualWebArena~\cite{visualwebarena2024}  %
    & Web & \gcmark & Web
    & 3     & 910
    & \rxmark   & 1/22
    & \rxmark  & \rxmark \\

WorkArena~\cite{workarena2024}       %
    & Web & \gcmark & Web
    & 1     & $33$
    & \rxmark   & 3/22
    & \pcmark  & \rxmark \\

WorkArena{++}~\cite{workarenaplus2024}   %
    & Web & \gcmark & Web
    & 1     & $682^{\dagger}$
    & \gcmark   & 3/22
    & \rxmark  & \pcmark \\

OSWorld~\cite{osworld2024}         %
    & CUA & \gcmark & Linux / Win
    & 9     & 369
    & \rxmark   & 3/22
    & \rxmark  & \rxmark \\

AndroidWorld~\cite{androidworld2025}    %
    & CUA & \gcmark & Android
    & 20    & $116^{\dagger}$
    & \rxmark   & 2/22
    & \rxmark  & \rxmark \\

WindowsAgentArena~\cite{windowsagentarena2025} %
    & CUA & \gcmark & Windows
    & 11    & 154
    & \rxmark   & 3/22
    & \rxmark  & \rxmark \\

Spider2-V~\cite{spider2v2024}       %
    & CUA & \gcmark & Linux / Cloud
    & 20    & 494
    & \rxmark   & 2/22
    & \rxmark  & \rxmark \\

ScienceBoard~\cite{scienceboard2025}    %
    & CUA & \gcmark & Linux
    & 6     & 169
    & \rxmark   & 2/22
    & \rxmark  & \rxmark \\

AssistGUI~\cite{assistgui2024}       %
    & CUA & \gcmark & Windows
    & 9     & 100
    & \rxmark   & 3/22
    & \rxmark  & \rxmark \\

TheAgentCompany~\cite{theagentcompany2024} %
    & CUA & \gcmark & Linux / Web
    & 5     & 175
    & \rxmark   & 4/22
    & \rxmark  & \rxmark \\

ProgrammingWithPixels~\cite{programmingwithpixels2025}
    & CUA & \gcmark & Linux
    & 1     & 5400
    & \rxmark   & 1/22
    & \rxmark  & \rxmark \\

\midrule\addlinespace[2pt]

\rowcolor{ourrow}
\textbf{\bench{}~(Ours)}
    & \textbf{CUA} & \gcmark & \textbf{Lin / Win / Android / Web}
    & \textbf{200{+}} & \textbf{10{,}000{+}}
    & \gcmark & \textbf{22/22}
    & \gcmark  & \gcmark \\[2pt]

\bottomrule
\end{tabular}%
}
\end{table*}

\begin{figure}[t]
    \centering
    \includegraphics[width=\textwidth]{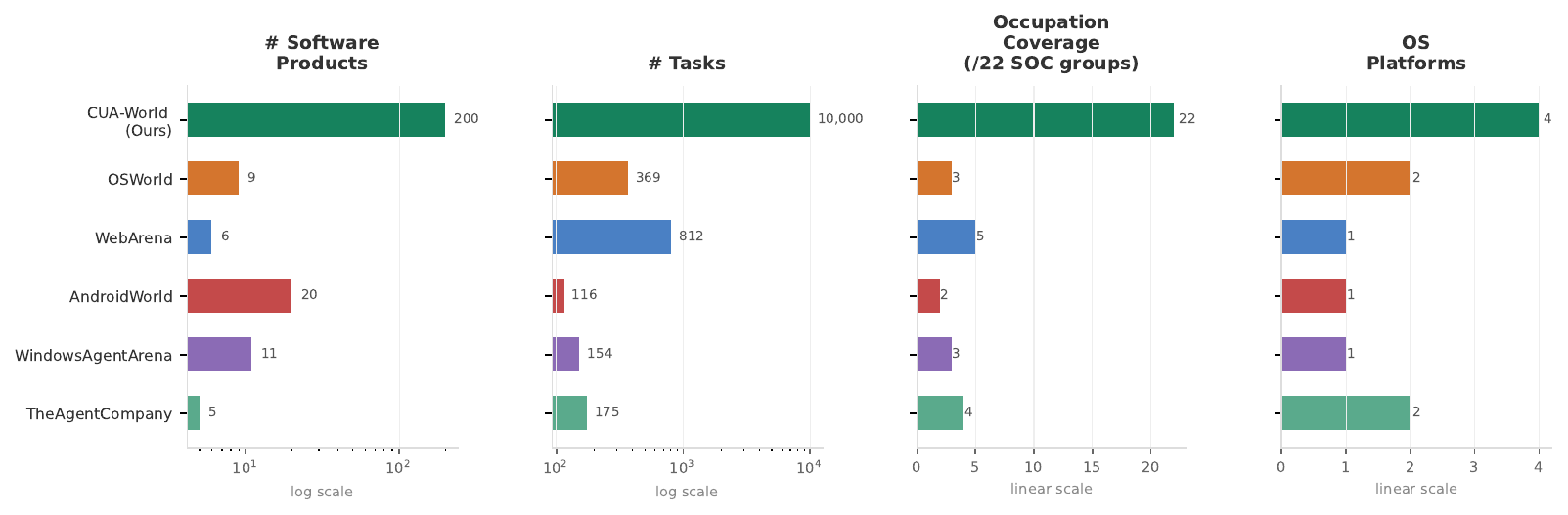}
    \caption{\textbf{Quantitative comparison of \bench{} against existing benchmarks} across four dimensions. The first two axes use a log scale.}
    \label{fig:env_comparison}
\end{figure}

Applying \method{} with our compute budget, the proposer generates 5 and the amplifier 75 tasks per piece of software. After filtering, this yields 12,103 tasks and environments across \numsoftwares{} software applications, each with checklist-based verification (Table~\ref{tab:benchmark_comparison}; Figure~\ref{fig:env_comparison}). As shown in Table~\ref{tab:benchmark_comparison}, \bench{} is the first collection to simultaneously provide interactive environments at scale (200+ varieties of software, 10K+ tasks), support long-horizon evaluation, cover all 22 major occupation groups, offer automated environment creation, and include a training split. We divide \bench{} into Train and Test splits.

\niparagraph{Contamination filtering.}
To ensure no data leakage between splits, we apply a conservative contamination check. Given two task instructions, we prompt an LLM to grade their similarity on a scale of 1 to 8 (ranging from ``not similar'' to ``duplicate, subset, or superset''). Any pair scoring 4 (``very similar'') or higher is flagged as contaminated.
We formalize this by treating tasks as nodes and contamination flags as edges in a similarity graph. 
We compute the connected components of this graph and randomly assign entire components to either the Train or Test split, ensuring no two tasks across splits contaminate each other. 
Manual verification shows the pipeline is suitably conservative: it flags several non-contaminated pairs (false positives) but misses very few true instances of contamination (false negatives). For more details, see Appendix~\ref{app:contamination}.

\niparagraph{\bench{}-Long.}
To evaluate agents on extremely long-horizon tasks, we introduce \bench{}-Long, a set of 200 tasks (one per software). 
The key challenge is generating tasks that are genuinely harder than those already in the benchmark while remaining solvable. 
We address this with a trajectory-guided strategy: for each piece of software, we first generate $k$ trajectories from a strong computer-use agent on existing tasks, then prompt a coding and visual agent to analyze these trajectories, specifically identifying why certain tasks have lower pass rates and noting common failure modes. 
The agent also receives a set of 8 quality guidelines covering real-world relevance, objective evaluability, realistic data, and others (see Appendix~\ref{app:bench_long_guidelines}).

Based on this analysis, the agent creates a new task designed to be harder than the existing ones for that software application. While the agent's failure assessment is not perfect, the resulting tasks are substantially more difficult. We manually verify that all 200 tasks are set up correctly and are meaningful according to the 8 quality criteria. Further for tasks that fail this verification, we iteratively refine them through further interaction with the agent. The full pipeline is described in Appendix~\ref{app:bench_long_pipeline}. These tasks often require more than 200 steps for human completion, and current models frequently fail even after 500 steps. \looseness=-1

\section{Experimental Setup}
\label{sec:experimental_setup}

We next describe how we use \bench{} in two roles: as a source of training data for distilling smaller models, and as an evaluation benchmark for computer-use agents.

\subsection{Training}
\label{sec:training}

To evaluate the utility of \bench{}-Train, we distill execution trajectories from a strong teacher model (Kimi-K 2.5~\cite{kimiteam2026kimik25visualagentic}) into a smaller student model (Qwen3-VL-2B-Thinking~\cite{yang2025qwen3technicalreport}). For every task in the training split, we generate $k=4$ trajectories from the teacher until at least one is correct, and utilize these successful rollouts for fine-tuning. Cumulatively, we collect roughly 2000 trajectories across all \numenvironments{} tasks.

We further systematically ablate several design choices in this distillation process on a small set of software, investigating: (1) teacher model selection, (2) the optimal number of steps and samples per trajectory (see \S\ref{par:finetuning_ablations} for results). We then use the best configuration to distill our model on all trajectories. Post-distillation, we evaluate the models on \bench{}-Test alongside external benchmarks such as OSWorld~\cite{osworld2024}.

\subsection{Test-Time Auditing (TTA) Agent}
\label{sec:test_time_agent}

Some of the tasks in \bench{} are extremely long. This opens up a unique opportunity to test agents capable of working over extended horizons. However, we find that current agents often stop after a few dozen steps, making mistakes or prematurely claiming the task is complete when it is not.

Inspired by our approach in software generation (\S\ref{sec:scaling_software}), we introduce an audit agent to address this. Whenever the main model signals that the task is terminated, we run this audit agent. It takes the complete trajectory (all screenshots) as input and determines whether the task is actually complete. Crucially, it does not receive the chain-of-thought from the main model, as we find this biases the auditor's assessment. 
If the audit agent determines that the task is not completed, it generates an explanation of what is missing. We provide this feedback back to the main computer-use agent, prompting it to continue completing the task. 

\subsection{Evaluation}
\label{sec:evaluation}

We evaluate agents using the checklist-based VLM verifier described in \S\ref{sec:verifiers}. 
Each task's checklist consists of weighted subtasks; we report two metrics: (1)~\textbf{Average Score} (0-100), the mean checklist score across tasks, which captures partial credit, and (2)~\textbf{Pass Rate} (\%), the fraction of tasks fully completed, i.e., achieving a perfect checklist score. 
We evaluate on \bench{}-Test (the full test split) and \bench{}-Long (200 long-horizon tasks, one per software). Unless otherwise noted, we use Gemini 3 Flash as the VLM verifier. Each agent is given a maximum budget per episode: 200 steps for \bench{}-Test, and 500 steps or \$5, whichever hits first, for \bench{}-Long. For GPT-5.4 and Claude Sonnet 4.6, we use their official agent harnesses. For Gemini 3 Flash and Kimi-K 2.5, official harnesses were not publicly available at the time of our experiments, so we adapted the Qwen3-VL harness (Appendix~\ref{app:evaluated_models}).

\section{Results and Analysis}

\subsection{Main Results}

\begin{wraptable}{r}{0.48\textwidth}
\vspace{-12pt}
\centering
\small
\caption{\textbf{Model performance on \bench{}-Test.} Our 2B distilled model outperforms open-source models up to 2$\times$ its size.}
\label{tab:main_results_test}
\begin{tabular}{@{} l c c @{}}
\toprule
\textbf{Model} & \textbf{Avg.\ Score} & \textbf{Pass Rate} \\
\midrule
\rowcolor{sechdr}
\multicolumn{3}{@{}l}{\small\textit{Large Models}} \\
Gemini 3 Flash    & \textbf{50.1}   & \textbf{22.6} \\
Kimi-K 2.5          & 37.1   & 12.8 \\
\midrule
\rowcolor{sechdr}
\multicolumn{3}{@{}l}{\small\textit{Small Models}} \\
Qwen3-VL-2B         & 12.7   & 1.6 \\
Qwen3-VL-4B         & 19.3   & 3.9 \\
\midrule
\rowcolor{ourrow}
Ours (2B distilled)  & 22.5 \imp{+9.8}   & 4.4 \imp{+2.8} \\
\bottomrule
\end{tabular}
\vspace{-8pt}
\end{wraptable}

\textbf{Distillation on \bench{}-Train yields a strong 2B model.}
Table~\ref{tab:main_results_test} shows results on \bench{}-Test. We evaluate four frontier models: Gemini-3-Flash and Kimi-K 2.5, Claude-Sonnet-4.6, and GPT-5.4, along with several small models. Gemini-3-Flash is strongest with 50.1 average score and 22.6\% pass rate, followed by Kimi-K 2.5 with 37.1 and 12.8\%. On the other extreme, small models perform very poorly: Qwen3-VL-2B achieves only 1.6\% pass rate while Qwen3-VL-4B achieves 3.9\%. Distillation on \bench{}-Train trajectories shows significant improvements, boosting the pass rate of Qwen3-VL-2B from 1.6\% to 4.4\%, outperforming Qwen3-VL-4B, a model 2$\times$ its size. This demonstrates that \bench{}-Train provides a useful supervision signal for improving small models.

\begin{wraptable}{r}{0.44\textwidth}
\vspace{-12pt}
\centering
\small
\caption{\textbf{Performance on \bench{}-Long.}}
\label{tab:main_results_long}
\begin{tabular}{@{} l c c @{}}
\toprule
\textbf{Model} & \textbf{Avg.\ Score} & \textbf{Pass Rate} \\
\midrule
\rowcolor{sechdr}
\multicolumn{3}{@{}l}{\small\textit{Max 500 steps, \$5 cost cap}} \\
Gemini 3 Flash       & 36.2   & 7.5 \\
GPT-5.4              & 22.7   & 3.0 \\
Sonnet 4.6           & 20.5   & 6.0 \\
Kimi-K 2.5           & 33.9   & 5.5 \\
\midrule
\rowcolor{sechdr}
\multicolumn{3}{@{}l}{\small\textit{Max 2000 steps, no cost cap}} \\
Gemini 3 Flash       & 38.7   & 11.5 \\
GPT-5.4              & \textbf{55.5}   & \textbf{27.5} \\
\bottomrule
\end{tabular}
\vspace{-8pt}
\end{wraptable}

\textbf{\bench{}-Long is challenging for frontier models.}
Table~\ref{tab:main_results_long} shows the performance of multiple frontier models on \bench{}-Long. Even the strongest model, Gemini-3-Flash, achieves a pass rate of only 7.5\% and an average score of 36.2. Interestingly, GPT-5.4 achieves 3\% pass rate while Claude-Sonnet-4.6 achieves 6\%. This is partly because they exhaust their \$5 budget in roughly 150 steps (${\le}$100 for GPT-5.4), much less than Gemini-3-Flash. 
To test whether budget is a bottleneck, we remove the cost cap and raise the step limit to 2,000 for GPT-5.4 and Gemini-3-Flash (Table~\ref{tab:main_results_long}, lower half). Both models substantially improve, notably GPT-5.4 reaching 27.5\% pass rate. 
However, these improvements come at a substantial test-time cost. On average, Gemini-3-Flash requires 1{,}300 steps and approximately \$16 per trajectory, while GPT-5.4 requires 242 steps and approximately \$18 per trajectory. 
These results highlight that improvements in model capabilities are needed before agents can reliably and efficiently handle the long-horizon, multi-step workflows that \bench{}-Long demands.

\paragraph{Scaling Software Applications and Environments:}

\begin{figure}[t]
    \centering
    \begin{subfigure}[t]{0.48\columnwidth}
        \centering
        \includegraphics[width=\linewidth]{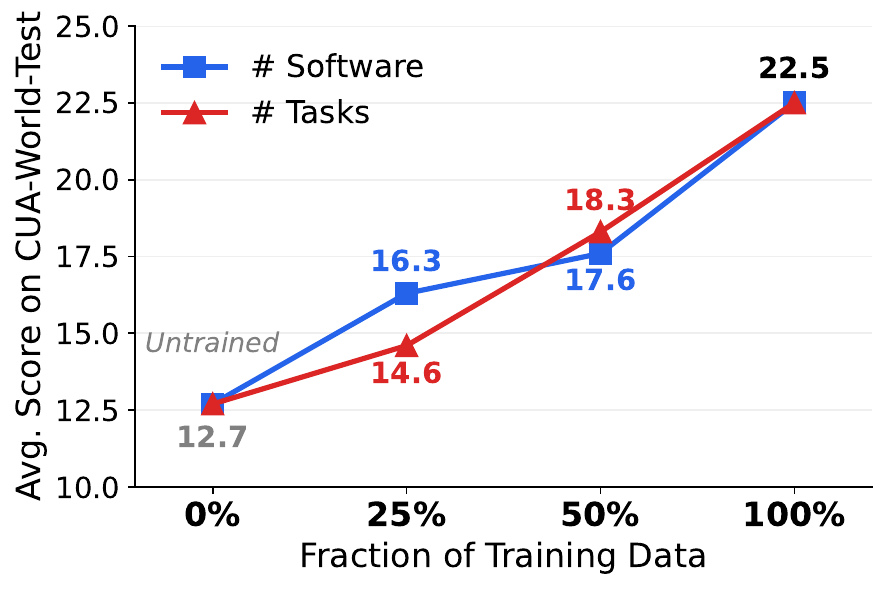}
        \caption{Training data scaling.}
        \label{fig:scaling}
    \end{subfigure}
    \hfill
    \begin{subfigure}[t]{0.48\columnwidth}
        \centering
        \includegraphics[width=\linewidth]{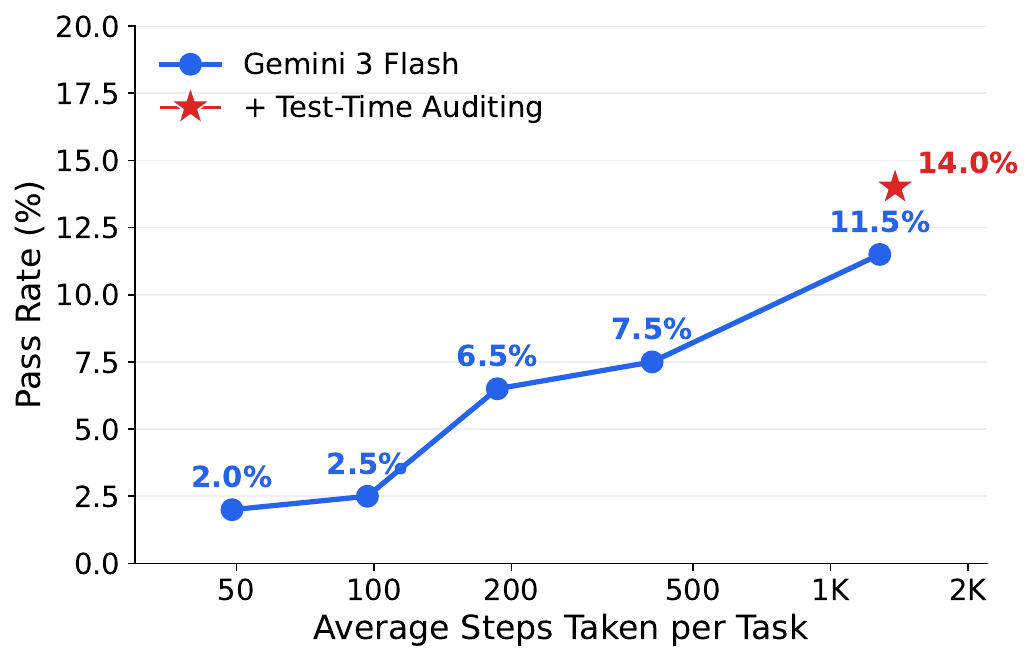}
        \caption{Test-time compute scaling.}
        \label{fig:ttc_scaling}
    \end{subfigure}
    \caption{\textbf{Scaling behavior on \bench{}.} \textbf{(a)}~Training data scaling on \bench{}-Test: varying the number of software (50, 100, 200) or the fraction of tasks (25\%, 50\%, 100\%). Both axes improve with scale, following a roughly log-linear trend. \textbf{(b)}~Test-time compute scaling on \bench{}-Long: pass rate as a function of average steps taken per task, where each point corresponds to a different maximum step budget (50, 100, 200, 500, 2,000 steps). The star indicates Test-Time Auditing (TTA, \S\ref{sec:test_time_agent}) under the same 2,000-step cap.}
\end{figure}

\textbf{Performance scales with both increasing software and task count.} Figure~\ref{fig:scaling} shows how the distilled 2B model's score on \bench{}-Test changes as we scale the training data along two axes: the number of software applications (50, 100, 200) while keeping all tasks per software, and the fraction of tasks (25\%, 50\%, 100\%) across all 200 software applications. Both curves show consistent performance improvements, following a roughly log-linear trend of ${\sim}$3.5 point increase on doubling the data. This suggests further scaling our \method{} pipeline could yield an even stronger model.

\paragraph{Generalization:}

\begin{figure}[t]
    \centering
    \includegraphics[width=\columnwidth]{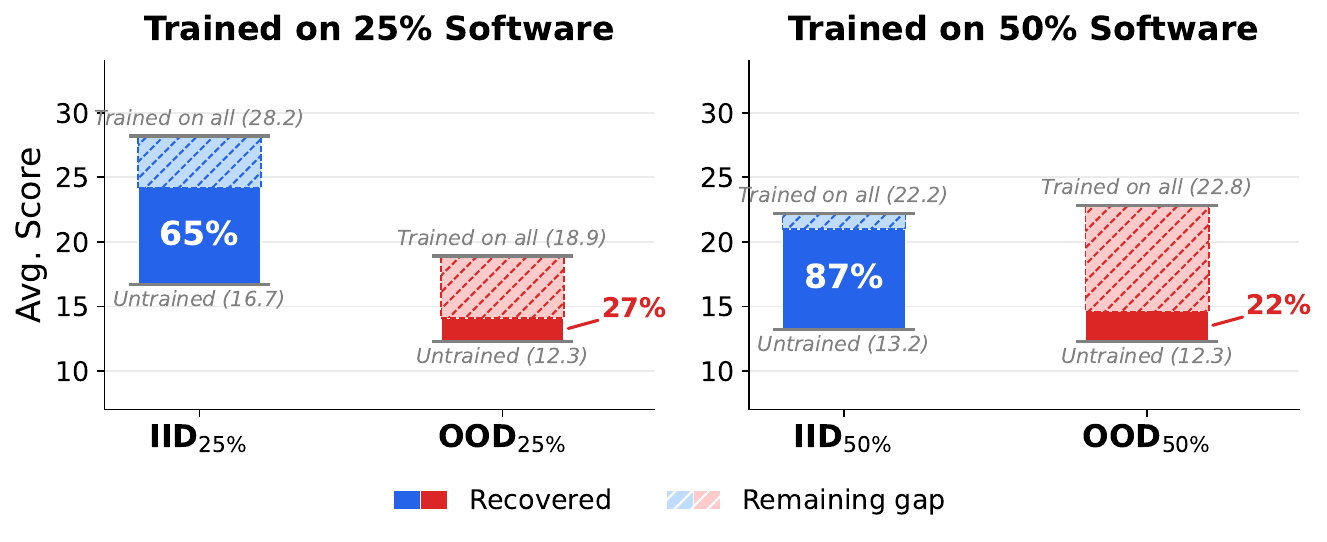}
    \caption{\textbf{Generalization to seen (IID) vs.\ unseen (OOD) software.} We train models on 25\% (left) and 50\% (right) of the 200 software applications, and evaluate on the training software applications (IID) and the held-out software applications (OOD). Each bar spans from the untrained baseline (bottom) to the model trained on all software (top). The solid portion shows the gain recovered by the model trained on the subset; the hatched portion shows the remaining gap. Training on a subset recovers 65-87\% of the gain on IID software but only 22-27\% on OOD software, indicating that generalization to unseen software is limited and scaling to diverse software is important.}
    \label{fig:generalization}
\end{figure}

\textbf{Distillation improves performance on both seen and unseen software, but gains are larger on seen software.} To study how distillation generalizes to software not seen during training, we train models on 25\% and 50\% of the 200 software applications and evaluate separately on the software used during training (IID) and those that are not used (OOD) (Figure~\ref{fig:generalization}). Performance improves on both: on IID software, the average score increases from 16.7 to 24.2 (at 25\% of software), and on OOD software from 12.3 to 14.1. However, the OOD gain is limited; Figure~\ref{fig:generalization} shows it recovers only 22-27\% of the improvement one would obtain from training on all software, compared to 65-87\% on IID software. This suggests that generalization to unseen software does happen but is limited. Secondly, since recovery on IID software ranges from 65\%-87\% this implies training on a specific software helps substantially, but training on other software also contributes in the evaluated software's performance. Overall, this underscores that building agents for the large variety of software used in the digital economy requires training that is both software-specific and across a diverse set of software, motivating the need for scalable environment creation pipelines such as \method{}. Larger models may generalize better across software, which we leave to future work.

\subsection{Scaling Test-Time Compute}

\textbf{Pass rate scales with step budget, and Test-Time Auditing provides further gains.} Figure~\ref{fig:ttc_scaling} shows how Gemini 3 Flash's pass rate on \bench{}-Long changes as we increase the maximum step budget per episode, where each point represents a budget of 50, 100, 200, 500, 2,000 steps. Pass rate stays low between 50 and 100 average steps (2.0\% $\to$ 2.5\%), then rises steeply at higher budgets (6.5\% at ${\sim}$200 steps, 7.5\% at ${\sim}$400 steps). The sharp jump likely suggests that most \bench{}-Long tasks require a minimum number of steps ($>$100) before the agent can complete them at all. Increasing compute beyond that continues to help, reaching 11.5\% at ${\sim}$1,300 average steps.
Further, the TTA agent uplifts performance even further, raising the pass rate to 14.0\% under the same 2,000-step cap. Since the maximum step budget remains the same, this likely implies that TTA helps when the model stops prematurely, as the auditor is able to verify the trajectory and provide feedback on any missed subtasks.

\subsection{Benchmark Analysis}

\paragraph{Trajectory Behavioral Patterns:}

\begin{wrapfigure}{r}{0.48\textwidth}
    \vspace{-12pt}
    \centering
    \includegraphics[width=0.47\textwidth]{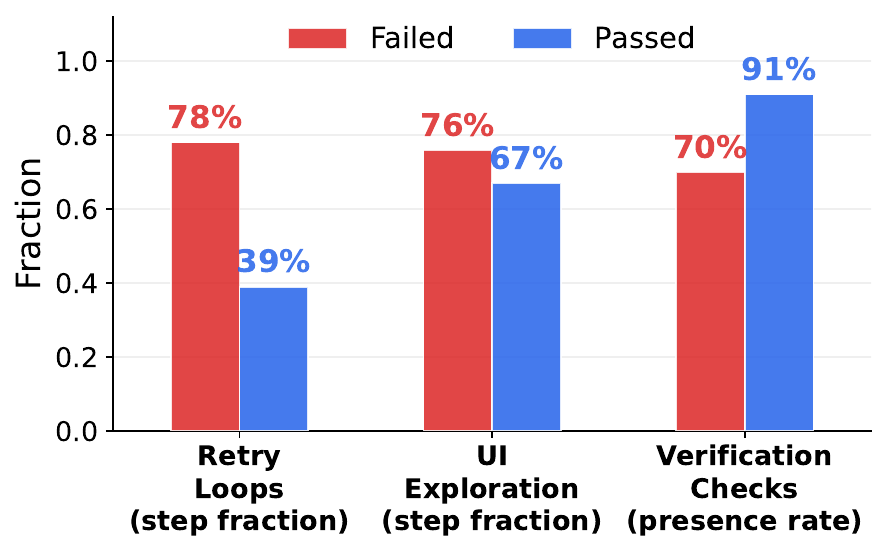}
    \caption{\textbf{Behavioral patterns in passed vs.\ failed trajectories} across Gemini-3-Flash evaluations on \bench{}. See Appendix for the full set of 15 patterns.}
    \label{fig:trajectory_patterns}
    \vspace{-10pt}
\end{wrapfigure}

\textbf{Failed trajectories are dominated by retry loops, while successful ones verify their progress more often.}
To understand how agents behave on \bench{}, we analyze all trajectories from Gemini-3-Flash evaluated on \bench{}, using an automated behavioral analysis pipeline (Appendix~\ref{app:trajectory_analysis}). We focus on Gemini-3-Flash here, and note that patterns may differ across models. We first obtain per-trajectory behavioral summaries via an LLM, then aggregate these across all trajectories to discover recurring behavioral patterns, yielding 15 canonical patterns. Figure~\ref{fig:trajectory_patterns} highlights three patterns.
\textbf{Retry loops} show the largest gap: failed trajectories spend 78\% of their steps retrying actions that did not take effect, compared to 39\% for passed trajectories.
\textbf{UI exploration} is high for both outcomes (76\% vs.\ 67\%), indicating that the majority of agent effort across all trajectories is spent navigating menus and locating the right controls rather than executing the core task.
\textbf{Verification checks}, where the agent re-inspects its work after making changes, are present in 91\% of passed trajectories but only 70\% of failed ones, suggesting an association between self-verification and task success. This observation provides empirical motivation for the Test-Time Auditing approach (\S\ref{sec:test_time_agent}).

\begin{figure}[t]
    \centering
    \begin{subfigure}[t]{0.48\columnwidth}
        \centering
        \includegraphics[width=\linewidth]{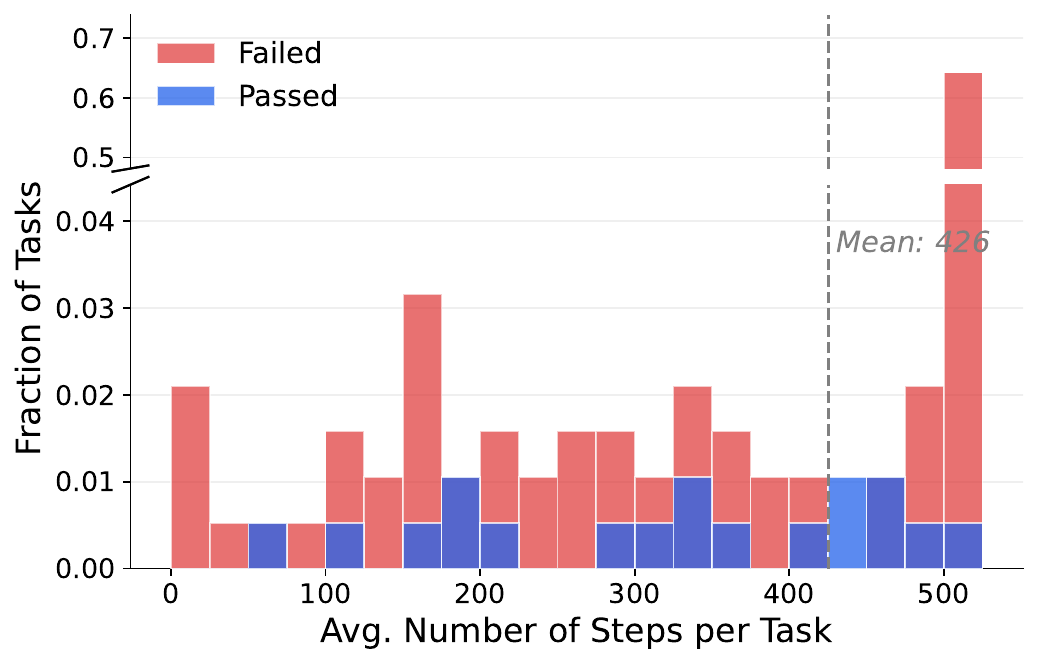}
        \caption{Step distribution.}
        \label{fig:step_distribution}
    \end{subfigure}
    \hfill
    \begin{subfigure}[t]{0.48\columnwidth}
        \centering
        \includegraphics[width=\linewidth]{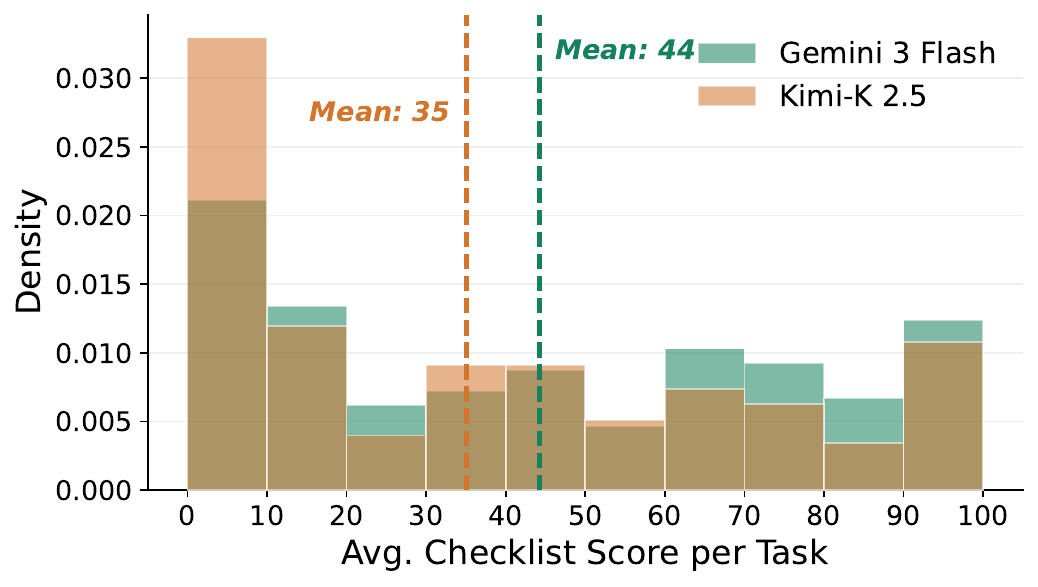}
        \caption{Difficulty distribution.}
        \label{fig:difficulty_histogram}
    \end{subfigure}
    \caption{\textbf{Properties of \bench{}-Long.} \textbf{(a)}~Distribution of average steps per task. The y-axis is broken to accommodate the spike at the 500-step cap. \textbf{(b)}~Distribution of per-task average checklist scores for Gemini 3 Flash and Kimi-K 2.5.}
\end{figure}

\paragraph{Step Distribution on \bench{}-Long:}

\textbf{Most failed trajectories exhaust the step budget, while passed ones finish at varying lengths.} Figure~\ref{fig:step_distribution} shows the distribution of average steps per task on \bench{}-Long. Failed tasks have a large spike at the 500-step cap, indicating that many episodes keep running until the budget is exhausted rather than failing immediately. Passed tasks are spread across a wide range of lengths, including many tasks that still require several hundred steps. The overall mean is 425 steps, highlighting the long-horizon nature of \bench{}-Long.

\paragraph{Difficulty Distribution on \bench{}-Long:}

\textbf{\bench{}-Long spans a wide difficulty range.} Figure~\ref{fig:difficulty_histogram} shows the distribution of per-task average checklist scores on \bench{}-Long for the two strongest models (Gemini 3 Flash, mean 44; Kimi-K 2.5, mean 35). Outside of the 0-10 bin, scores are spread fairly evenly across the range, indicating that \bench{}-Long contains tasks at every difficulty level rather than being split into trivially easy and impossible ones. The most notable feature is a large spike at 0-10: roughly a quarter of tasks for Gemini and a third for Kimi receive near-zero scores, indicating complete failure on a substantial fraction of tasks even for the strongest models. \looseness=-1

\paragraph{Performance by Software Category:}

\begin{figure}[t]
    \centering
    \includegraphics[width=\textwidth]{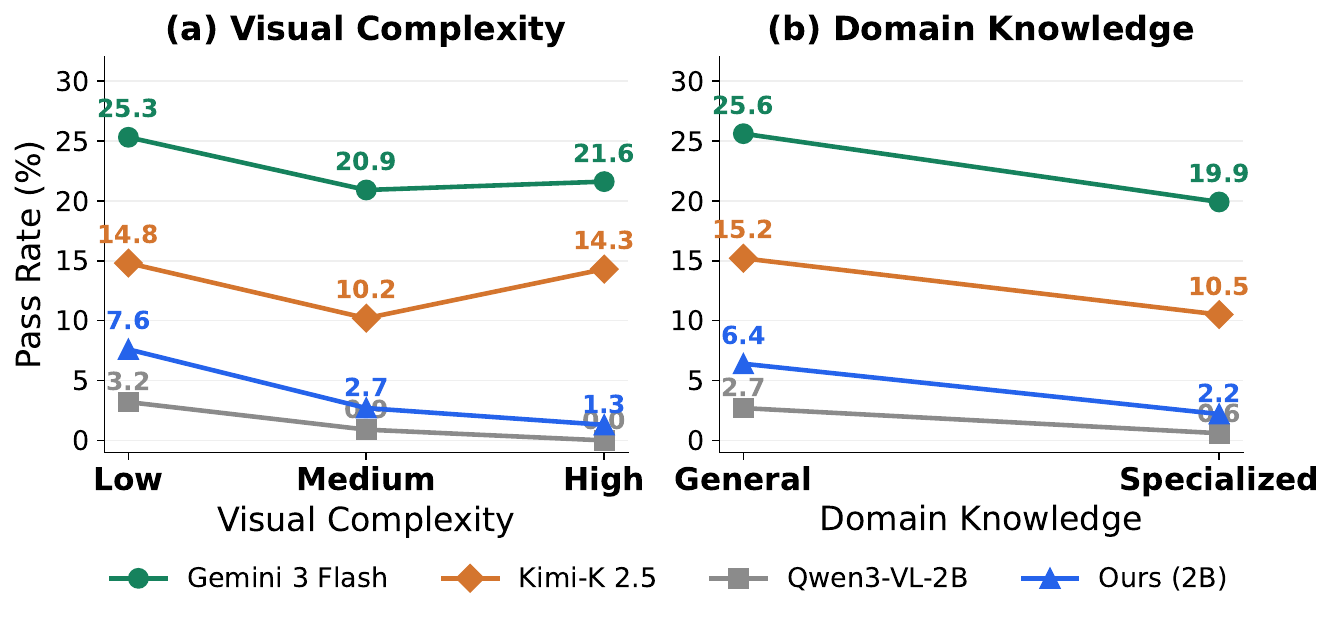}
    \caption{\textbf{Pass rate on \bench{}-Test by software category.} \textbf{(a)}~Visual complexity and \textbf{(b)}~domain knowledge. See Appendix~\ref{app:software_categories} for category definitions and assignment of software to categories.}
    \label{fig:software_categories}
\end{figure}

\textbf{High visual complexity is a persistent bottleneck for smaller models.} We classify each software along two axes: \emph{visual complexity} (low/medium/high) and \emph{domain knowledge} (general/specialized); see Appendix~\ref{app:software_categories} for definitions. Figure~\ref{fig:software_categories} shows pass rates on \bench{}-Test broken down by each axis. For \textbf{visual complexity} (a), larger models (Gemini 3 Flash, Kimi-K 2.5) achieve roughly consistent pass rates across all three levels (e.g., 25.3\%, 20.9\%, 21.6\% for Gemini). In contrast, smaller models show a steep decline: Qwen3-VL-2B drops from 3.2\% on low-complexity software to 0.0\% on high-complexity software. Distillation improves absolute performance at every level (e.g., 0.0\% $\to$ 1.3\% on high, 3.2\% $\to$ 7.6\% on low), but the decline from low to high remains steep, indicating that visual complexity creates a disparity for small models that distillation alone does not resolve.

For \textbf{domain knowledge} (b), all models show a downward trend from general to specialized software, with smaller models showing a steeper decline (${\sim}$3$\times$ for our 2B model: 6.4\% $\to$ 2.2\%) than large models (${\sim}$1.3$\times$ for Gemini: 25.6\% $\to$ 19.9\%).

\paragraph{Verifier Robustness}
\label{sec:verifier_robustness}

We evaluate the robustness of the verifier across two dimensions: a.)~how well does it agree with humans in terms of correctness, and b.)~integrity checks: how often the integrity checklist $C_{\text{int}}$ correctly identifies shortcut behavior. 

\textbf{Checklist-based verification achieves highest human agreement.} We compare three verifier designs on 60 randomly sampled Gemini-3-Flash trajectories from \bench{}-Test: (1)~our checklist-based VLM verifier (\S\ref{sec:verifiers}), (2)~a direct VLM verifier that receives the trajectory and outputs a single pass/fail judgment, and (3)~programmatic verifiers where a model generates a script that runs on the end state and computes a score. The checklist-based verifier achieves 93.3\% task-level agreement with human annotations, compared to 81.7\% for the direct VLM verifier and 43.3\% for programmatic verifiers. Per-item checklist agreement is 90.9\%. The programmatic approach performs poorly primarily because the model writes incorrect scripts that fail to parse the data formats present in the end state; manually authored programmatic verifiers could provide stronger guarantees and are an interesting direction for future work. Overall, we use the checklist-based VLM verifier with privileged information for all experiments.

\textbf{Integrity checks catch shortcuts at a low flag rate.} Across ${\sim}$3{,}000 Gemini-3-Flash trajectories, the integrity checklist flags only ${\sim}$1.5\% of high-scoring runs (score $>$75), producing 21 flags total, of which 15 are true positives. We describe two representative cases.
In a forensic analysis task on \textit{Autopsy} (digital forensics tool), the agent followed the correct workflow but fabricated hash values in its final report rather than copying the values visible in the application.
In a statistical analysis task on \textit{Epi Info} (epidemiology toolkit), the agent mistyped an input parameter, causing the tool to display an incorrect result, but wrote the mathematically correct answer in its report, a value never shown by the tool.
In both cases, the agent scored high on task completion but was zeroed by the integrity check. Additional examples and a detailed breakdown are provided in Appendix~\ref{app:integrity_examples}.

\subsection{Gym-Anything Pipeline Ablations}

\paragraph{Distillation Ablations:}
\label{par:finetuning_ablations}

We ablate the teacher model, student model, and number of training steps to identify the best configuration for full distillation.

\begin{wraptable}{r}{0.52\textwidth}
\vspace{-12pt}
\centering
\footnotesize
\caption{\textbf{Teacher model selection} on 4 software applications.}
\label{tab:distillation_teacher}
\begin{tabular}{@{}lccc@{}}
\toprule
 & Teacher & \multicolumn{2}{c}{Student Score} \\
\cmidrule(lr){3-4}
Teacher & Score & Q3-VL-2B & Q2.5-3B \\
\midrule
Opus 4.5       & \textbf{53.5} & 19.3 & 8.5 \\
Sonnet 4.5     & 45.5 & 17.5 & 9.8 \\
Gemini 3 Flash & 44.0 & 16.3 & 8.3 \\
Kimi-K 2.5     & 39.8 & \textbf{25.3} & \textbf{15.8} \\
Gemini 3 Pro   & 39.3 & 15.8 & 7.0 \\
\bottomrule
\end{tabular}
\vspace{-8pt}
\end{wraptable}

\textbf{The strongest teacher does not produce the strongest student.} Table~\ref{tab:distillation_teacher} compares five teacher models distilled into two student architectures (Qwen3-VL-2B and Qwen2.5-3B) on 4 software. Opus~4.5 is the strongest teacher (53.5 avg.\ score) while Kimi-K~2.5 is one of the weakest (39.8). However, Kimi-K~2.5 produces the best student for both model sizes: 25.3 vs.\ 19.3 (Opus) for Qwen3-VL-4B, and 15.8 vs.\ 9.8 (Sonnet) for Qwen2.5-3B. One possible explanation is that, unlike other models, Kimi-K~2.5 is open-source and provides full reasoning chains; however, other factors may contribute as well~\cite{xu-etal-2025-stronger}.

\begin{wraptable}{r}{0.52\textwidth}
\vspace{-12pt}
\centering
\footnotesize
\caption{\textbf{Effect of training trajectory length.}}
\label{tab:step_ablation}
\begin{tabular}{@{}lccccc@{}}
\toprule
Train & \rotatebox{60}{GIMP} & \rotatebox{60}{G.\ Earth} & \rotatebox{60}{OpenEMR} & \rotatebox{60}{Slicer 3D} & Avg. \\
\midrule
200 steps & 60.1 & \textbf{13.4} & \textbf{18.2} & 3.3 & 23.8 \\
50 steps  & \textbf{64.2} & \textbf{13.4} & 8.5 & \textbf{4.5} & 22.7 \\
\midrule
\rowcolor{sechdr}
\multicolumn{6}{@{}l}{\scriptsize\textit{No distillation}} \\
Q3-VL-2B  & 47.0 & 7.2 & 0.0 & 0.0 & 13.6 \\
\bottomrule
\end{tabular}
\vspace{-8pt}
\end{wraptable}

\textbf{Effect of training trajectory length.} Table~\ref{tab:step_ablation} compares training on the first 50 vs.\ all 200 steps of each teacher trajectory under the same \$25-per-software budget. On average, the two settings perform similarly (22.7 vs.\ 23.8), but the per-software pattern differs: training on 50 steps wins on GIMP and Slicer~3D, while training on 200 steps is better on OpenEMR, likely because its tasks require longer-horizon interaction. Based on this, we adopt a two-stage curriculum for full distillation: first train with a maximum step budget of 50, then continue on full 200-step trajectories, with equal budget for each stage.

\paragraph{Propose-and-Amplify Ablation:}

\textbf{Proposal step substantially improve amplified task quality.} To evaluate the propose-and-amplify strategy (\S\ref{sec:tasks}), we compare tasks generated with and without seed examples, using proposal step, across 10 software applications. We launch each generated task and use a VLM to verify whether the starting state matches the task description. Tasks amplified from seed examples achieve an 88.9\% setup success rate, compared to 55.2\% without seeds. Qualitative analysis on three software applications (Firefox, AstroImageJ, Moodle) reveals that without seeds, the model defaults to demonstrating software features rather than generating realistic professional workflows, produces shorter-horizon tasks, and writes less thorough setup scripts (Appendix~\ref{app:seed_ablation}).

\paragraph{Cross-Model Auditing:}
\label{par:cross_model_audit}

\textbf{Using a separate model for $\text{Agent}_{\text{audit}}$ catches more issues than self-auditing.} In \S\ref{sec:scaling_software}, we argued that separating $\text{Agent}_C$ and $\text{Agent}_{\text{audit}}$ removes self-confirmation bias. To test this, we compare audits where the same model serves as both $\text{Agent}_C$ and $\text{Agent}_{\text{audit}}$ (self-audit) against audits where a different model serves as $\text{Agent}_{\text{audit}}$ (cross-model audit) across 10 software applications. Both configurations detect all critical issues, but cross-model audits consistently surface additional problems that self-audits miss. For example, on OpenELIS, the self-audit accepts patient data as realistic, while the cross-model audit inspects the seeding script and identifies the data as hardcoded despite comments claiming real-world WHO/CDC sourcing. Across 10 software applications, cross-model audits identify on average 2.1 additional issues per environment, predominantly low-to-moderate severity. We present three representative comparisons in Appendix~\ref{app:cross_model_audit}.

\paragraph{Additional analysis.}
The appendix contains further results and analysis. We qualitatively show how the creation-audit loop iteratively corrects issues across rounds, with before-and-after examples (Appendix~\ref{app:audit_cross_round}). We verify that \bench{} covers all 22 SOC major occupation groups (Appendix~\ref{app:soc_coverage}). We apply an automated behavioral analysis pipeline to ${\sim}$3{,}000 trajectories, discovering 15 canonical patterns and comparing their prevalence in passed vs.\ failed runs (Appendix~\ref{app:trajectory_analysis}). We detail the contamination filtering statistics (Appendix~\ref{app:contamination}) and provide 12 representative task examples with agent trajectories (Appendix~\ref{app:task_examples}). Finally, the \bench{}-Long generation pipeline and a trajectory analysis example are in Appendix~\ref{app:bench_long}.

\section{Related Work}
\label{sec:related_work}

\niparagraph{Benchmarks for computer-use agents.}
Prior benchmarks for computer-use agents are either static or interactive but small-scale (Table~\ref{tab:benchmark_comparison}; Figure~\ref{fig:env_comparison}).
Static datasets~\cite{deng2023mind2web,rawles2023androidinthewild,androidcontrol2024,omniact2024} collect thousands of episodes but evaluate via action-matching rather than execution, penalizing valid alternative strategies.
Interactive benchmarks provide execution-based evaluation but cover narrow slices of the software landscape: web benchmarks~\cite{liu2018miniwob,webarena2024,visualwebarena2024,workarena2024,workarenaplus2024} are restricted to a few websites, desktop benchmarks~\cite{osworld2024,windowsagentarena2025,spider2v2024,assistgui2024,theagentcompany2024,scienceboard2025,programmingwithpixels2025} span at most a handful of applications with manually authored environments, and AndroidWorld~\cite{androidworld2025} covers 20 apps.
Critically, all interactive benchmarks rely on manual environment creation, limiting their scale, and none simultaneously provides training data, long-horizon tasks, or broad occupational coverage.
\bench{} addresses these gaps through automated environment creation, yielding 10K+ interactive tasks across 200+ software applications on four platforms, with train/test splits, long-horizon evaluation, and GDP-grounded coverage of all 22 SOC occupation groups.

\niparagraph{Automated environment and task generation.}
Several works generate tasks or trajectories within pre-existing environments~\cite{xu2024agenttrek,osgenesis2025,pae2024,bagel2024,webrl2025}, but cannot create new ones.
LLM-based environment generation has been explored for text planning~\cite{agentgen2024}, embodied AI~\cite{holodeck2024}, tool-use APIs~\cite{agentworldmodel2026}, code editing and SWE training~\cite{swegym2024,zhu2025training}, and text-based simulations~\cite{zhang2025autoenv}, but not for real GUI software requiring installation, configuration, and realistic data.
Concurrently, GUI-GENESIS~\cite{cao2026gui} synthesizes lightweight web replicas from interaction traces of a single app ecosystem for efficient RL training, but does not install or configure real software, handle multi-OS environments, or target long-horizon evaluation.
The seed-then-amplify paradigm~\cite{selfinstruct2023,wizardlm2024,agentinstruct2024} is effective for generating instruction data at scale, but targets text pairs rather than executable environment tasks.
\method{} combines all three: a creation-audit loop that converts real software into interactive environments via coding agents verified by an independent auditor, a propose-and-amplify strategy that generates tasks grounded in actual software execution, and a shared memory that accumulates learnings across environments.

\niparagraph{Evaluation of computer-use agents.}
Existing benchmarks predominantly use hand-written programmatic verifiers that check the final system state~\cite{osworld2024,webarena2024}, which are reliable but labor-intensive and offer only binary pass/fail.
VLM-based evaluation has been explored for filtering training trajectories~\cite{xu2024agenttrek}, step-level trajectory assessment~\cite{seagent2025}, and autonomous evaluation of agent trajectories~\cite{panetal2024guieval}, but these approaches lack access to ground-truth answers and cannot detect workflow shortcuts.
Our checklist-based VLM verifier addresses both gaps by incorporating privileged information extracted from environment setup scripts, enabling verification against known answers without per-task code, and by including integrity checks that detect workflow bypasses such as fabricated outputs or tool misuse.\looseness=-1

We provide an extended related work with additional coverage of training methods, economic grounding, and detailed per-benchmark comparisons in Appendix~\ref{app:related_work}.

\section{Conclusion}

In this work, we introduced \method{}, a scalable framework for converting arbitrary software into interactive computer-use
environments. By reducing environment creation to setup scripts and configuration files, and by framing creation itself as a multi-agent loop of generation, auditing, and correction, \method{} addresses a central bottleneck in computer-use agents: the difficulty
of constructing realistic environments at scale. Applying this framework, we built \bench{}, a GDP-grounded collection of over 10K
tasks across 200 software applications spanning diverse occupations, domains, and operating systems, together with checklist-based VLM
verification and train/test/long-horizon splits. We further showed that \bench{} provides useful supervision for training smaller agents through distillation, and that test-time auditing can improve performance on especially long-horizon tasks. At the same time, \bench{}-Long is challenging even for frontier models, indicating that realistic computer-use remains far from solved.

More broadly, our results suggest that progress in computer-use agents will require not only stronger models, but also scalable methods for constructing the environments and tasks on which those models are trained and evaluated. We hope that \method{},
\bench{}, and the released code and infrastructure provide a foundation for future work on long-horizon, economically relevant
computer-use, including more capable agents, stronger verifiers, and broader coverage of the software that underlies real-world
digital work.

\section{Acknowledgements}
Pranjal is supported by a SoftBank Group-Arm Fellowship. This work was supported in part by the National Science Foundation under Grant Nos. DMS-2434614 and DMS-2502281, a gift from Convergent Research and a grant of compute credits from Microsoft Azure.

\section{Limitations}

Our GDP-grounded software selection is designed to produce a reasonable ranking of which software matters more, not a precise dollar-level attribution. We use the strongest available LLM with web-search access to estimate the share factors in Equation~\ref{eq:gdp}, but other methods may yield more accurate estimates. While we specifically select the closest sandboxable alternative for software that cannot be freely sandboxed (e.g., due to licensing), a large fraction of professionally used software remains excluded, and the degree to which performance on free alternatives predicts performance on their commercial counterparts is an open question. While we manually verified that every software environment launches correctly and that every \bench{}-Long task loads with the correct starting state and data, we did not solve all tasks end-to-end, and therefore cannot guarantee that every task is solvable. Creating a fully human-verified version of the benchmark is an interesting direction for future work. Finally, we use VLM checklist verifiers throughout our evaluation pipeline. Manual annotation shows high human agreement, but like any evaluation method, VLM verifiers are imperfect and may be susceptible to adversarial exploitation. Developing robust programmatic verifiers, each manually vetted per task, would complement the current approach and is another promising direction.

\section{Ethics Statement}

We acknowledge that computer-use agents may pose risks if deployed autonomously. While this work introduces methods for environment creation and test-time auditing, it does not train or release a model that exceeds existing frontier capabilities. All software used is freely available, and all datasets were obtained from public sources or synthetically generated.

\clearpage

\bibliographystyle{plain}
\bibliography{references}

@inproceedings{osworld2024,
  title     = {{OSWorld}: Benchmarking Multimodal Agents for Open-Ended Tasks
               in Real Computer Environments},
  author    = {Xie, Tianbao and Zhang, Danyang and Chen, Jixuan and
               Li, Xiaochuan and Zhao, Siheng and Cao, Ruisheng and
               Hua, Toh Jing and Cheng, Zhoujun and Shin, Dongchan and
               Lei, Fangyu and Liu, Yitao and Xu, Yiheng and Zhou, Shuyan and
               Savarese, Silvio and Xiong, Caiming and Zhong, Victor and
               Yu, Tao},
  booktitle = {Advances in Neural Information Processing Systems},
  editor    = {Globerson, A. and Mackey, L. and Belgrave, D. and Fan, A. and
               Paquet, U. and Tomczak, J. and Zhang, C.},
  pages     = {52040--52094},
  publisher = {Curran Associates, Inc.},
  volume    = {37},
  year      = {2024},
  doi       = {10.52202/079017-1650},
  url       = {https://proceedings.neurips.cc/paper_files/paper/2024/file/5d413e48f84dc61244b6be550f1cd8f5-Paper-Datasets_and_Benchmarks_Track.pdf}
}

@inproceedings{windowsagentarena2025,
  title     = {{Windows Agent Arena}: Evaluating Multi-Modal {OS} Agents at Scale},
  author    = {Bonatti, Rogerio and Zhao, Dan and Bonacci, Francesco and
               Dupont, Dillon and Abdali, Sara and Li, Yinheng and Lu, Yadong and
               Wagle, Justin and Koishida, Kazuhito and Bucker, Arthur and
               Jang, Lawrence Keunho and Hui, Zheng},
  booktitle = {Proceedings of the 42nd International Conference on Machine Learning},
  editor    = {Singh, Aarti and Fazel, Maryam and Hsu, Daniel and
               Lacoste-Julien, Simon and Berkenkamp, Felix and Maharaj, Tegan and
               Wagstaff, Kiri and Zhu, Jerry},
  pages     = {4874--4910},
  publisher = {PMLR},
  series    = {Proceedings of Machine Learning Research},
  volume    = {267},
  year      = {2025},
  month     = {13--19 Jul},
  url       = {https://proceedings.mlr.press/v267/bonatti25a.html}
}

@inproceedings{androidworld2025,
  title     = {{AndroidWorld}: A Dynamic Benchmarking Environment for
               Autonomous Agents},
  author    = {Rawles, Christopher and Clinckemaillie, Sarah and Chang, Yifan and
               Waltz, Jonathan and Lau, Gabrielle and Fair, Marybeth and
               Li, Alice and Bishop, William and Li, Wei and
               Campbell-Ajala, Folawiyo and Toyama, Daniel and Berry, Robert and
               Tyamagundlu, Divya and Lillicrap, Timothy and Riva, Oriana},
  booktitle = {The Thirteenth International Conference on Learning Representations},
  year      = {2025},
  url       = {https://openreview.net/forum?id=il5yUQsrjC},
  note      = {ICLR 2025; arXiv:2405.14573}
}

@inproceedings{webarena2024,
  title     = {{WebArena}: A Realistic Web Environment for Building
               Autonomous Agents},
  author    = {Zhou, Shuyan and Xu, Frank F. and Zhu, Hao and Zhou, Xuhui and
               Lo, Robert and Sridhar, Abishek and Cheng, Xianyi and Ou, Tianyue and
               Bisk, Yonatan and Fried, Daniel and Alon, Uri and Neubig, Graham},
  booktitle = {The Twelfth International Conference on Learning Representations},
  year      = {2024},
  url       = {https://openreview.net/forum?id=oKn9c6ytLx},
  note      = {ICLR 2024; arXiv:2307.13854}
}

@inproceedings{visualwebarena2024,
  title     = {{V}isual{W}eb{A}rena: Evaluating Multimodal Agents on Realistic
               Visual Web Tasks},
  author    = {Koh, Jing Yu and Lo, Robert and Jang, Lawrence and
               Duvvur, Vikram and Lim, Ming and Huang, Po-Yu and
               Neubig, Graham and Zhou, Shuyan and Salakhutdinov, Russ and
               Fried, Daniel},
  editor    = {Ku, Lun-Wei and Martins, Andre and Srikumar, Vivek},
  booktitle = {Proceedings of the 62nd Annual Meeting of the Association for
               Computational Linguistics (Volume 1: Long Papers)},
  pages     = {881--905},
  month     = aug,
  year      = {2024},
  address   = {Bangkok, Thailand},
  publisher = {Association for Computational Linguistics},
  url       = {https://aclanthology.org/2024.acl-long.50/},
  doi       = {10.18653/v1/2024.acl-long.50}
}

@misc{yang2025qwen3technicalreport,
      title={Qwen3 Technical Report}, 
      author={An Yang and Anfeng Li and Baosong Yang and Beichen Zhang and Binyuan Hui and Bo Zheng and Bowen Yu and Chang Gao and Chengen Huang and Chenxu Lv and Chujie Zheng and Dayiheng Liu and Fan Zhou and Fei Huang and Feng Hu and Hao Ge and Haoran Wei and Huan Lin and Jialong Tang and Jian Yang and Jianhong Tu and Jianwei Zhang and Jianxin Yang and Jiaxi Yang and Jing Zhou and Jingren Zhou and Junyang Lin and Kai Dang and Keqin Bao and Kexin Yang and Le Yu and Lianghao Deng and Mei Li and Mingfeng Xue and Mingze Li and Pei Zhang and Peng Wang and Qin Zhu and Rui Men and Ruize Gao and Shixuan Liu and Shuang Luo and Tianhao Li and Tianyi Tang and Wenbiao Yin and Xingzhang Ren and Xinyu Wang and Xinyu Zhang and Xuancheng Ren and Yang Fan and Yang Su and Yichang Zhang and Yinger Zhang and Yu Wan and Yuqiong Liu and Zekun Wang and Zeyu Cui and Zhenru Zhang and Zhipeng Zhou and Zihan Qiu},
      year={2025},
      eprint={2505.09388},
      archivePrefix={arXiv},
      primaryClass={cs.CL},
      url={https://arxiv.org/abs/2505.09388}, 
}

@misc{kimiteam2026kimik25visualagentic,
      title={Kimi K2.5: Visual Agentic Intelligence}, 
      author={Kimi Team and Tongtong Bai and Yifan Bai and Yiping Bao and S. H. Cai and Yuan Cao and Y. Charles and H. S. Che and Cheng Chen and Guanduo Chen and Huarong Chen and Jia Chen and Jiahao Chen and Jianlong Chen and Jun Chen and Kefan Chen and Liang Chen and Ruijue Chen and Xinhao Chen and Yanru Chen and Yanxu Chen and Yicun Chen and Yimin Chen and Yingjiang Chen and Yuankun Chen and Yujie Chen and Yutian Chen and Zhirong Chen and Ziwei Chen and Dazhi Cheng and Minghan Chu and Jialei Cui and Jiaqi Deng and Muxi Diao and Hao Ding and Mengfan Dong and Mengnan Dong and Yuxin Dong and Yuhao Dong and Angang Du and Chenzhuang Du and Dikang Du and Lingxiao Du and Yulun Du and Yu Fan and Shengjun Fang and Qiulin Feng and Yichen Feng and Garimugai Fu and Kelin Fu and Hongcheng Gao and Tong Gao and Yuyao Ge and Shangyi Geng and Chengyang Gong and Xiaochen Gong and Zhuoma Gongque and Qizheng Gu and Xinran Gu and Yicheng Gu and Longyu Guan and Yuanying Guo and Xiaoru Hao and Weiran He and Wenyang He and Yunjia He and Chao Hong and Hao Hu and Jiaxi Hu and Yangyang Hu and Zhenxing Hu and Ke Huang and Ruiyuan Huang and Weixiao Huang and Zhiqi Huang and Tao Jiang and Zhejun Jiang and Xinyi Jin and Yu Jing and Guokun Lai and Aidi Li and C. Li and Cheng Li and Fang Li and Guanghe Li and Guanyu Li and Haitao Li and Haoyang Li and Jia Li and Jingwei Li and Junxiong Li and Lincan Li and Mo Li and Weihong Li and Wentao Li and Xinhang Li and Xinhao Li and Yang Li and Yanhao Li and Yiwei Li and Yuxiao Li and Zhaowei Li and Zheming Li and Weilong Liao and Jiawei Lin and Xiaohan Lin and Zhishan Lin and Zichao Lin and Cheng Liu and Chenyu Liu and Hongzhang Liu and Liang Liu and Shaowei Liu and Shudong Liu and Shuran Liu and Tianwei Liu and Tianyu Liu and Weizhou Liu and Xiangyan Liu and Yangyang Liu and Yanming Liu and Yibo Liu and Yuanxin Liu and Yue Liu and Zhengying Liu and Zhongnuo Liu and Enzhe Lu and Haoyu Lu and Zhiyuan Lu and Junyu Luo and Tongxu Luo and Yashuo Luo and Long Ma and Yingwei Ma and Shaoguang Mao and Yuan Mei and Xin Men and Fanqing Meng and Zhiyong Meng and Yibo Miao and Minqing Ni and Kun Ouyang and Siyuan Pan and Bo Pang and Yuchao Qian and Ruoyu Qin and Zeyu Qin and Jiezhong Qiu and Bowen Qu and Zeyu Shang and Youbo Shao and Tianxiao Shen and Zhennan Shen and Juanfeng Shi and Lidong Shi and Shengyuan Shi and Feifan Song and Pengwei Song and Tianhui Song and Xiaoxi Song and Hongjin Su and Jianlin Su and Zhaochen Su and Lin Sui and Jinsong Sun and Junyao Sun and Tongyu Sun and Flood Sung and Yunpeng Tai and Chuning Tang and Heyi Tang and Xiaojuan Tang and Zhengyang Tang and Jiawen Tao and Shiyuan Teng and Chaoran Tian and Pengfei Tian and Ao Wang and Bowen Wang and Chensi Wang and Chuang Wang and Congcong Wang and Dingkun Wang and Dinglu Wang and Dongliang Wang and Feng Wang and Hailong Wang and Haiming Wang and Hengzhi Wang and Huaqing Wang and Hui Wang and Jiahao Wang and Jinhong Wang and Jiuzheng Wang and Kaixin Wang and Linian Wang and Qibin Wang and Shengjie Wang and Shuyi Wang and Si Wang and Wei Wang and Xiaochen Wang and Xinyuan Wang and Yao Wang and Yejie Wang and Yipu Wang and Yiqin Wang and Yucheng Wang and Yuzhi Wang and Zhaoji Wang and Zhaowei Wang and Zhengtao Wang and Zhexu Wang and Zihan Wang and Zizhe Wang and Chu Wei and Ming Wei and Chuan Wen and Zichen Wen and Chengjie Wu and Haoning Wu and Junyan Wu and Rucong Wu and Wenhao Wu and Yuefeng Wu and Yuhao Wu and Yuxin Wu and Zijian Wu and Chenjun Xiao and Jin Xie and Xiaotong Xie and Yuchong Xie and Yifei Xin and Bowei Xing and Boyu Xu and Jianfan Xu and Jing Xu and Jinjing Xu and L. H. Xu and Lin Xu and Suting Xu and Weixin Xu and Xinbo Xu and Xinran Xu and Yangchuan Xu and Yichang Xu and Yuemeng Xu and Zelai Xu and Ziyao Xu and Junjie Yan and Yuzi Yan and Guangyao Yang and Hao Yang and Junwei Yang and Kai Yang and Ningyuan Yang and Ruihan Yang and Xiaofei Yang and Xinlong Yang and Ying Yang and Yi Yang and Yi Yang and Zhen Yang and Zhilin Yang and Zonghan Yang and Haotian Yao and Dan Ye and Wenjie Ye and Zhuorui Ye and Bohong Yin and Chengzhen Yu and Longhui Yu and Tao Yu and Tianxiang Yu and Enming Yuan and Mengjie Yuan and Xiaokun Yuan and Yang Yue and Weihao Zeng and Dunyuan Zha and Haobing Zhan and Dehao Zhang and Hao Zhang and Jin Zhang and Puqi Zhang and Qiao Zhang and Rui Zhang and Xiaobin Zhang and Y. Zhang and Yadong Zhang and Yangkun Zhang and Yichi Zhang and Yizhi Zhang and Yongting Zhang and Yu Zhang and Yushun Zhang and Yutao Zhang and Yutong Zhang and Zheng Zhang and Chenguang Zhao and Feifan Zhao and Jinxiang Zhao and Shuai Zhao and Xiangyu Zhao and Yikai Zhao and Zijia Zhao and Huabin Zheng and Ruihan Zheng and Shaojie Zheng and Tengyang Zheng and Junfeng Zhong and Longguang Zhong and Weiming Zhong and M. Zhou and Runjie Zhou and Xinyu Zhou and Zaida Zhou and Jinguo Zhu and Liya Zhu and Xinhao Zhu and Yuxuan Zhu and Zhen Zhu and Jingze Zhuang and Weiyu Zhuang and Ying Zou and Xinxing Zu},
      year={2026},
      eprint={2602.02276},
      archivePrefix={arXiv},
      primaryClass={cs.CL},
      url={https://arxiv.org/abs/2602.02276}, 
}

@article{qwenteam2025qwen25vl,
  title   = {{Qwen2.5-VL} Technical Report},
  author  = {Shuai Bai and Keqin Chen and Xuejing Liu and Jialin Wang and
             Wenbin Ge and Sibo Song and Kai Dang and Peng Wang and
             Shijie Wang and Jun Tang and Humen Zhong and Yuanzhi Zhu and
             Mingkun Yang and Zhaohai Li and Jianqiang Wan and Pengfei Wang and
             Wei Ding and Zheren Fu and Yiheng Xu and Jiabo Ye and Xi Zhang and
             Tianbao Xie and Zesen Cheng and Hang Zhang and Zhibo Yang and
             Haiyang Xu and Junyang Lin},
  journal = {arXiv preprint arXiv:2502.13923},
  year    = {2025},
  eprint  = {2502.13923},
  archivePrefix = {arXiv},
  primaryClass = {cs.CV},
  url     = {https://arxiv.org/abs/2502.13923}
}

@article{wang2024qwen2vl,
  title   = {{Qwen2-VL}: Enhancing Vision-Language Model's Perception of the
             World at Any Resolution},
  author  = {Wang, Peng and Bai, Shuai and Tan, Sinan and Wang, Shijie
             and Fan, Zhihao and Bai, Jinze and Chen, Keqin and Liu, Xuejing
             and Wang, Jialin and Ge, Wenbin and Fan, Yang and Dang, Kai
             and Du, Mengfei and Ren, Xuancheng and Men, Rui and Liu, Dayiheng
             and Zhou, Chang and Zhou, Jingren and Lin, Junyang},
  journal = {arXiv preprint arXiv:2409.12191},
  year    = {2024},
  url     = {https://arxiv.org/abs/2409.12191}
}

@article{xu2024agenttrek,
  title        = {{AgentTrek}: Agent Trajectory Synthesis via Guiding Replay with Web Tutorials},
  author       = {Xu, Yiheng and Lu, Dunjie and Shen, Zhennan and Wang, Junli and
                  Wang, Zekun and Mao, Yuchen and Xiong, Caiming and Yu, Tao},
  journal      = {arXiv preprint arXiv:2412.09605},
  year         = {2025},
  eprint       = {2412.09605},
  archivePrefix = {arXiv},
  primaryClass = {cs.CL},
  url          = {https://arxiv.org/abs/2412.09605}
}

@article{he2024pcagent,
  title        = {{PC} Agent: While You Sleep, {AI} Works -- A Cognitive Journey into Digital World},
  author       = {He, Yanheng and Jin, Jiahe and Xia, Shijie and Su, Jiadi and
                  Fan, Runze and Zou, Haoyang and Hu, Xiangkun and Liu, Pengfei},
  journal      = {arXiv preprint arXiv:2412.17589},
  year         = {2024}
}

@inproceedings{deng2023mind2web,
  title        = {{Mind2Web}: Towards a Generalist Agent for the Web},
  author       = {Deng, Xiang and Gu, Yu and Zheng, Boyuan and Chen, Shijie and
                  Stevens, Sam and Wang, Boshi and Sun, Huan and Su, Yu},
  editor       = {Oh, A. and Naumann, T. and Globerson, A. and
                  Saenko, K. and Hardt, M. and Levine, S.},
  booktitle    = {Advances in Neural Information Processing Systems},
  pages        = {28091--28114},
  publisher    = {Curran Associates, Inc.},
  url          = {https://proceedings.neurips.cc/paper_files/paper/2023/file/5950bf290a1570ea401bf98882128160-Paper-Datasets_and_Benchmarks.pdf},
  volume       = {36},
  year         = {2023}
}

@inproceedings{rawles2023androidinthewild,
  title        = {{AndroidInTheWild}: A Large-Scale Dataset For {Android} Device Control},
  author       = {Rawles, Christopher and Li, Alice and Rodriguez, Daniel and
                  Riva, Oriana and Lillicrap, Timothy},
  editor       = {Oh, A. and Naumann, T. and Globerson, A. and
                  Saenko, K. and Hardt, M. and Levine, S.},
  booktitle    = {Advances in Neural Information Processing Systems},
  pages        = {59708--59728},
  publisher    = {Curran Associates, Inc.},
  url          = {https://proceedings.neurips.cc/paper_files/paper/2023/file/bbbb6308b402fe909c39dd29950c32e0-Paper-Datasets_and_Benchmarks.pdf},
  volume       = {36},
  year         = {2023}
}

@misc{brockman2016openai,
  title         = {{OpenAI Gym}},
  author        = {Brockman, Greg and Cheung, Vicki and Pettersson, Ludwig
                   and Schneider, Jonas and Schulman, John and Tang, Jie
                   and Zaremba, Wojciech},
  year          = {2016},
  eprint        = {1606.01540},
  archivePrefix = {arXiv},
  primaryClass  = {cs.LG},
  url           = {https://arxiv.org/abs/1606.01540}
}

@article{towers2024gymnasium,
  title         = {{Gymnasium}: A Standard Interface for Reinforcement
                   Learning Environments},
  author        = {Towers, Mark and Kwiatkowski, Ariel and Terry, Jordan
                   and Balis, John U. and {De Cola}, Gianluca and Deleu, Tristan
                   and Goul{\~{a}}o, Manuel and Kallinteris, Andreas
                   and Krimmel, Markus and {KG}, Arjun and Perez-Vicente, Rodrigo
                   and Pierr{\'e}, Andrea and Schulhoff, Sander and Tai, Jun Jet
                   and Tan, Hannah and Younis, Omar G.},
  journal       = {arXiv preprint arXiv:2407.17032},
  year          = {2025},
  eprint        = {2407.17032},
  archivePrefix = {arXiv},
  primaryClass  = {cs.LG},
  url           = {https://arxiv.org/abs/2407.17032}
}

@article{peterson2001onet,
  title   = {Understanding Work Using the Occupational Information
             Network ({O*NET}): Implications for Practice and Research},
  author  = {Peterson, Norman G. and Mumford, Michael D. and Borman, Walter C.
             and Jeanneret, P. Richard and Fleishman, Edwin A. and Levin, Kerry Y.
             and Campion, Michael A. and Mayfield, Melinda S.
             and Morgeson, Frederick P. and Pearlman, Kenneth
             and Gowing, Marilyn K. and Lancaster, Anita R.
             and Silver, Marilyn B. and Dye, Donna M.},
  journal = {Personnel Psychology},
  volume  = {54},
  number  = {2},
  pages   = {451--492},
  publisher = {Wiley},
  year    = {2001},
  doi     = {10.1111/j.1744-6570.2001.tb00100.x},
  url     = {http://dx.doi.org/10.1111/j.1744-6570.2001.tb00100.x}
}

@misc{bls2024oews,
  author       = {{U.S. Bureau of Labor Statistics}},
  title        = {Occupational Employment and Wage Statistics ({OEWS})},
  year         = {2024},
  howpublished = {U.S. Department of Labor},
  url          = {https://www.bls.gov/oes/},
  note         = {May 2024 estimates. Accessed February 2025.}
}

@misc{bea2024nationalaccounts,
  author       = {{U.S. Bureau of Economic Analysis}},
  title        = {National Income and Product Accounts ({NIPA})},
  year         = {2024},
  howpublished = {U.S. Department of Commerce},
  url          = {https://www.bea.gov/itable/national-gdp-and-personal-income},
  note         = {Interactive data tables, annual estimates. Accessed February 2025.}
}

@article{liu2024lostmiddle,
  title     = {Lost in the Middle: How Language Models Use Long Contexts},
  author    = {Liu, Nelson F. and Lin, Kevin and Hewitt, John
               and Paranjape, Ashwin and Bevilacqua, Michele
               and Petroni, Fabio and Liang, Percy},
  journal   = {Transactions of the Association for Computational Linguistics},
  volume    = {12},
  pages     = {157--173},
  year      = {2024},
  address   = {Cambridge, MA},
  publisher = {MIT Press},
  doi       = {10.1162/tacl_a_00638},
  url       = {https://aclanthology.org/2024.tacl-1.9/}
}

@misc{wang2026doesagentdevelopmentreflect,
      title={How Well Does Agent Development Reflect Real-World Work?}, 
      author={Zora Zhiruo Wang and Sanidhya Vijayvargiya and Aspen Chen and Hanmo Zhang and Venu Arvind Arangarajan and Jett Chen and Valerie Chen and Diyi Yang and Daniel Fried and Graham Neubig},
      year={2026},
      eprint={2603.01203},
      archivePrefix={arXiv},
      primaryClass={cs.AI},
      url={https://arxiv.org/abs/2603.01203}, 
}

@misc{sinha2026illusiondiminishingreturnsmeasuring,
      title={The Illusion of Diminishing Returns: Measuring Long Horizon Execution in LLMs},
      author={Akshit Sinha and Arvindh Arun and Shashwat Goel and Steffen Staab and Jonas Geiping},
      year={2026},
      eprint={2509.09677},
      archivePrefix={arXiv},
      primaryClass={cs.AI},
      url={https://arxiv.org/abs/2509.09677},
}

@inproceedings{androidcontrol2024,
  title     = {On the Effects of Data Scale on {UI} Control Agents},
  author    = {Li, Wei and Bishop, William and Li, Alice and Rawles, Chris and
               Campbell-Ajala, Folawiyo and Tyamagundlu, Divya and Riva, Oriana},
  editor    = {Globerson, A. and Mackey, L. and Belgrave, D. and Fan, A. and
               Paquet, U. and Tomczak, J. and Zhang, C.},
  booktitle = {Advances in Neural Information Processing Systems},
  pages     = {92130--92154},
  publisher = {Curran Associates, Inc.},
  volume    = {37},
  year      = {2024},
  doi       = {10.52202/079017-2925},
  url       = {https://proceedings.neurips.cc/paper_files/paper/2024/file/a79f3ef3b445fd4659f44648f7ea8ffd-Paper-Datasets_and_Benchmarks_Track.pdf}
}

@inproceedings{omniact2024,
  title     = {{OmniACT}: A Dataset and Benchmark for Enabling Multimodal
               Generalist Autonomous Agents for Desktop and Web},
  author    = {Kapoor, Raghav and Butala, Yash Parag and Russak, Melisa and
               Koh, Jing Yu and Kamble, Kiran and AlShikh, Waseem and
               Salakhutdinov, Ruslan},
  booktitle = {Computer Vision -- {ECCV} 2024},
  pages     = {161--178},
  publisher = {Springer Nature Switzerland},
  year      = {2024},
  doi       = {10.1007/978-3-031-73113-6_10},
  url       = {http://dx.doi.org/10.1007/978-3-031-73113-6_10}
}

@inproceedings{liu2018miniwob,
  title     = {Reinforcement Learning on Web Interfaces Using
               Workflow-Guided Exploration},
  author    = {Liu, Evan Zheran and Guu, Kelvin and Pasupat, Panupong and
               Shi, Tianlin and Liang, Percy},
  booktitle = {International Conference on Learning Representations},
  year      = {2018},
  url       = {https://arxiv.org/abs/1802.08802},
  note      = {ICLR 2018; arXiv:1802.08802}
}

@inproceedings{workarena2024,
  title     = {{W}ork{A}rena: How Capable are Web Agents at Solving Common
               Knowledge Work Tasks?},
  author    = {Drouin, Alexandre and Gasse, Maxime and Caccia, Massimo and
               Laradji, Issam H. and Del Verme, Manuel and Marty, Tom and
               Vazquez, David and Chapados, Nicolas and Lacoste, Alexandre},
  editor    = {Salakhutdinov, Ruslan and Kolter, Zico and Heller, Katherine and
               Weller, Adrian and Oliver, Nuria and Scarlett, Jonathan and
               Berkenkamp, Felix},
  booktitle = {Proceedings of the 41st International Conference on Machine Learning},
  pages     = {11642--11662},
  publisher = {PMLR},
  series    = {Proceedings of Machine Learning Research},
  volume    = {235},
  year      = {2024},
  month     = {21--27 Jul},
  url       = {https://proceedings.mlr.press/v235/drouin24a.html}
}

@inproceedings{workarenaplus2024,
  title     = {{WorkArena++}: Towards Compositional Planning and
               Reasoning-based Common Knowledge Work Tasks},
  author    = {Boisvert, L{\'e}o and Thakkar, Megh and Gasse, Maxime and
               Caccia, Massimo and {De Chezelles}, Thibault Le Sellier and
               Cappart, Quentin and Chapados, Nicolas and Lacoste, Alexandre and
               Drouin, Alexandre},
  editor    = {Globerson, A. and Mackey, L. and Belgrave, D. and Fan, A. and
               Paquet, U. and Tomczak, J. and Zhang, C.},
  booktitle = {Advances in Neural Information Processing Systems},
  pages     = {5996--6051},
  publisher = {Curran Associates, Inc.},
  volume    = {37},
  year      = {2024},
  doi       = {10.52202/079017-0195},
  url       = {https://proceedings.neurips.cc/paper_files/paper/2024/file/0b82662b6c32e887bb252a74d8cb2d5e-Paper-Datasets_and_Benchmarks_Track.pdf}
}

@inproceedings{spider2v2024,
  title     = {{Spider2-V}: How Far Are Multimodal Agents From Automating
               Data Science and Engineering Workflows?},
  author    = {Cao, Ruisheng and Lei, Fangyu and Wu, Haoyuan and Chen, Jixuan and
               Fu, Yeqiao and Gao, Hongcheng and Xiong, Xinzhuang and
               Zhang, Hanchong and Mao, Yuchen and Hu, Wenjing and Xie, Tianbao and
               Xu, Hongshen and Zhang, Danyang and Wang, Sida and Sun, Ruoxi and
               Yin, Pengcheng and Xiong, Caiming and Ni, Ansong and Liu, Qian and
               Zhong, Victor and Chen, Lu and Yu, Kai and Yu, Tao},
  editor    = {Globerson, A. and Mackey, L. and Belgrave, D. and Fan, A. and
               Paquet, U. and Tomczak, J. and Zhang, C.},
  booktitle = {Advances in Neural Information Processing Systems},
  pages     = {107703--107744},
  publisher = {Curran Associates, Inc.},
  volume    = {37},
  year      = {2024},
  doi       = {10.52202/079017-3421},
  url       = {https://proceedings.neurips.cc/paper_files/paper/2024/file/c2f71567cd53464161cab3336e8fc865-Paper-Datasets_and_Benchmarks_Track.pdf}
}

@inproceedings{scienceboard2025,
  title     = {{ScienceBoard}: Evaluating Multimodal Autonomous Agents in
               Realistic Scientific Workflows},
  author    = {Sun, Qiushi and Liu, Zhoumianze and Ma, Chang and Ding, Zichen and
               Xu, Fangzhi and Yin, Zhangyue and Zhao, Haiteng and Wu, Zhenyu and
               Cheng, Kanzhi and Liu, Zhaoyang and Wang, Jianing and Li, Qintong and
               Tang, Xiangru and Xie, Tianbao and Feng, Xiachong and Li, Xiang and
               Kao, Ben and Wang, Wenhai and Qi, Biqing and Kong, Lingpeng and
               Wu, Zhiyong},
  booktitle = {The Fourteenth International Conference on Learning Representations},
  year      = {2026},
  url       = {https://iclr.cc/virtual/2026/poster/10008621},
  note      = {ICLR 2026; arXiv:2505.19897}
}

@inproceedings{assistgui2024,
  title     = {{AssistGUI}: Task-Oriented {PC} Graphical User Interface Automation},
  author    = {Gao, Difei and Ji, Lei and Bai, Zechen and Ouyang, Mingyu and
               Li, Peiran and Mao, Dongxing and Wu, Qinchen and Zhang, Weichen and
               Wang, Peiyi and Guo, Xiangwu and Wang, Hengxu and Zhou, Luowei and
               Shou, Mike Zheng},
  booktitle = {Proceedings of the {IEEE/CVF} Conference on Computer Vision
               and Pattern Recognition ({CVPR})},
  month     = {June},
  pages     = {13289--13298},
  year      = {2024}
}

@inproceedings{theagentcompany2024,
  title     = {{TheAgentCompany}: Benchmarking {LLM} Agents on Consequential
               Real World Tasks},
  author    = {Xu, Frank F. and Song, Yufan and Li, Boxuan and Tang, Yuxuan and
               Jain, Kritanjali and Bao, Mengxue and Wang, Zora Z. and Zhou, Xuhui and
               Guo, Zhitong and Cao, Murong and Yang, Mingyang and Lu, Hao Yang and
               Martin, Amaad and Su, Zhe and Maben, Leander and Mehta, Raj and
               Chi, Wayne and Jang, Lawrence and Xie, Yiqing and Zhou, Shuyan and
               Neubig, Graham},
  booktitle = {Advances in Neural Information Processing Systems},
  volume    = {38},
  year      = {2025},
  url       = {https://neurips.cc/virtual/2025/poster/121705},
  note      = {NeurIPS 2025 Datasets and Benchmarks Track}
}

@article{programmingwithpixels2025,
  title   = {Programming with Pixels: Can Computer-Use Agents do
             Software Engineering?},
  author  = {Aggarwal, Pranjal and Welleck, Sean},
  journal = {arXiv preprint arXiv:2502.18525},
  year    = {2025},
  url     = {https://arxiv.org/abs/2502.18525}
}

@inproceedings{xu-etal-2025-stronger,
    title = "Stronger Models are Not Always Stronger Teachers for Instruction Tuning",
    author = "Xu, Zhangchen  and
      Jiang, Fengqing  and
      Niu, Luyao  and
      Lin, Bill Yuchen  and
      Poovendran, Radha",
    editor = "Chiruzzo, Luis  and
      Ritter, Alan  and
      Wang, Lu",
    booktitle = "Proceedings of the 2025 Conference of the Nations of the Americas Chapter of the Association for Computational Linguistics: Human Language Technologies (Volume 1: Long Papers)",
    month = apr,
    year = "2025",
    address = "Albuquerque, New Mexico",
    publisher = "Association for Computational Linguistics",
    url = "https://aclanthology.org/2025.naacl-long.224/",
    doi = "10.18653/v1/2025.naacl-long.224",
    pages = "4392--4405",
    ISBN = "979-8-89176-189-6"
}

@misc{eloundou2023gpts,
      title={GPTs are GPTs: An Early Look at the Labor Market Impact Potential of Large Language Models}, 
      author={Tyna Eloundou and Sam Manning and Pamela Mishkin and Daniel Rock},
      year={2023},
      eprint={2303.10130},
      archivePrefix={arXiv},
      primaryClass={econ.GN},
      url={https://arxiv.org/abs/2303.10130}, 
}

@misc{gdpval2025,
      title={GDPval: Evaluating AI Model Performance on Real-World Economically Valuable Tasks}, 
      author={Tejal Patwardhan and Rachel Dias and Elizabeth Proehl and Grace Kim and Michele Wang and Olivia Watkins and Simón Posada Fishman and Marwan Aljubeh and Phoebe Thacker and Laurance Fauconnet and Natalie S. Kim and Patrick Chao and Samuel Miserendino and Gildas Chabot and David Li and Michael Sharman and Alexandra Barr and Amelia Glaese and Jerry Tworek},
      year={2025},
      eprint={2510.04374},
      archivePrefix={arXiv},
      primaryClass={cs.LG},
      url={https://arxiv.org/abs/2510.04374}, 
}

@misc{mtbench2023,
      title={Judging LLM-as-a-Judge with MT-Bench and Chatbot Arena}, 
      author={Lianmin Zheng and Wei-Lin Chiang and Ying Sheng and Siyuan Zhuang and Zhanghao Wu and Yonghao Zhuang and Zi Lin and Zhuohan Li and Dacheng Li and Eric P. Xing and Hao Zhang and Joseph E. Gonzalez and Ion Stoica},
      year={2023},
      eprint={2306.05685},
      archivePrefix={arXiv},
      primaryClass={cs.CL},
      url={https://arxiv.org/abs/2306.05685}, 
}

@misc{wildbench2024,
      title={WildBench: Benchmarking LLMs with Challenging Tasks from Real Users in the Wild}, 
      author={Bill Yuchen Lin and Yuntian Deng and Khyathi Chandu and Faeze Brahman and Abhilasha Ravichander and Valentina Pyatkin and Nouha Dziri and Ronan Le Bras and Yejin Choi},
      year={2024},
      eprint={2406.04770},
      archivePrefix={arXiv},
      primaryClass={cs.CL},
      url={https://arxiv.org/abs/2406.04770}, 
}

@misc{agentasjudge2024,
      title={Agent-as-a-Judge: Evaluate Agents with Agents}, 
      author={Mingchen Zhuge and Changsheng Zhao and Dylan Ashley and Wenyi Wang and Dmitrii Khizbullin and Yunyang Xiong and Zechun Liu and Ernie Chang and Raghuraman Krishnamoorthi and Yuandong Tian and Yangyang Shi and Vikas Chandra and Jürgen Schmidhuber},
      year={2024},
      eprint={2410.10934},
      archivePrefix={arXiv},
      primaryClass={cs.AI},
      url={https://arxiv.org/abs/2410.10934}, 
}

@misc{evilgenie2025,
      title={EvilGenie: A Reward Hacking Benchmark}, 
      author={Jonathan Gabor and Jayson Lynch and Jonathan Rosenfeld},
      year={2025},
      eprint={2511.21654},
      archivePrefix={arXiv},
      primaryClass={cs.LG},
      url={https://arxiv.org/abs/2511.21654}, 
}

@misc{stepeval2025,
      title={Score the Steps, Not Just the Goal: VLM-Based Subgoal Evaluation for Robotic Manipulation}, 
      author={Ramy ElMallah and Krish Chhajer and Chi-Guhn Lee},
      year={2025},
      eprint={2509.19524},
      archivePrefix={arXiv},
      primaryClass={cs.AI},
      url={https://arxiv.org/abs/2509.19524}, 
}

@misc{panetal2024guieval,
      title={Autonomous Evaluation and Refinement of Digital Agents}, 
      author={Jiayi Pan and Yichi Zhang and Nicholas Tomlin and Yifei Zhou and Sergey Levine and Alane Suhr},
      year={2024},
      eprint={2404.06474},
      archivePrefix={arXiv},
      primaryClass={cs.AI},
      url={https://arxiv.org/abs/2404.06474}, 
}

@misc{seagent2025,
      title={SEAgent: Self-Evolving Computer Use Agent with Autonomous Learning from Experience}, 
      author={Zeyi Sun and Ziyu Liu and Yuhang Zang and Yuhang Cao and Xiaoyi Dong and Tong Wu and Dahua Lin and Jiaqi Wang},
      year={2025},
      eprint={2508.04700},
      archivePrefix={arXiv},
      primaryClass={cs.AI},
      url={https://arxiv.org/abs/2508.04700}, 
}

@misc{selfinstruct2023,
      title={Self-Instruct: Aligning Language Models with Self-Generated Instructions}, 
      author={Yizhong Wang and Yeganeh Kordi and Swaroop Mishra and Alisa Liu and Noah A. Smith and Daniel Khashabi and Hannaneh Hajishirzi},
      year={2023},
      eprint={2212.10560},
      archivePrefix={arXiv},
      primaryClass={cs.CL},
      url={https://arxiv.org/abs/2212.10560}, 
}

@misc{wizardlm2024,
      title={WizardLM: Empowering large pre-trained language models to follow complex instructions}, 
      author={Can Xu and Qingfeng Sun and Kai Zheng and Xiubo Geng and Pu Zhao and Jiazhan Feng and Chongyang Tao and Qingwei Lin and Daxin Jiang},
      year={2025},
      eprint={2304.12244},
      archivePrefix={arXiv},
      primaryClass={cs.CL},
      url={https://arxiv.org/abs/2304.12244}, 
}

@misc{agentinstruct2024,
      title={AgentInstruct: Toward Generative Teaching with Agentic Flows}, 
      author={Arindam Mitra and Luciano Del Corro and Guoqing Zheng and Shweti Mahajan and Dany Rouhana and Andres Codas and Yadong Lu and Wei-ge Chen and Olga Vrousgos and Corby Rosset and Fillipe Silva and Hamed Khanpour and Yash Lara and Ahmed Awadallah},
      year={2024},
      eprint={2407.03502},
      archivePrefix={arXiv},
      primaryClass={cs.AI},
      url={https://arxiv.org/abs/2407.03502}, 
}

@misc{agentgen2024,
      title={AgentGen: Enhancing Planning Abilities for Large Language Model based Agent via Environment and Task Generation}, 
      author={Mengkang Hu and Pu Zhao and Can Xu and Qingfeng Sun and Jianguang Lou and Qingwei Lin and Ping Luo and Saravan Rajmohan},
      year={2025},
      eprint={2408.00764},
      archivePrefix={arXiv},
      primaryClass={cs.CL},
      url={https://arxiv.org/abs/2408.00764}, 
}

@misc{holodeck2024,
      title={Holodeck: Language Guided Generation of 3D Embodied AI Environments}, 
      author={Yue Yang and Fan-Yun Sun and Luca Weihs and Eli VanderBilt and Alvaro Herrasti and Winson Han and Jiajun Wu and Nick Haber and Ranjay Krishna and Lingjie Liu and Chris Callison-Burch and Mark Yatskar and Aniruddha Kembhavi and Christopher Clark},
      year={2024},
      eprint={2312.09067},
      archivePrefix={arXiv},
      primaryClass={cs.CV},
      url={https://arxiv.org/abs/2312.09067}, 
}

@misc{swegym2024,
      title={Training Software Engineering Agents and Verifiers with SWE-Gym}, 
      author={Jiayi Pan and Xingyao Wang and Graham Neubig and Navdeep Jaitly and Heng Ji and Alane Suhr and Yizhe Zhang},
      year={2025},
      eprint={2412.21139},
      archivePrefix={arXiv},
      primaryClass={cs.SE},
      url={https://arxiv.org/abs/2412.21139}, 
}

@misc{agentworldmodel2026,
      title={Agent World Model: Infinity Synthetic Environments for Agentic Reinforcement Learning}, 
      author={Zhaoyang Wang and Canwen Xu and Boyi Liu and Yite Wang and Siwei Han and Zhewei Yao and Huaxiu Yao and Yuxiong He},
      year={2026},
      eprint={2602.10090},
      archivePrefix={arXiv},
      primaryClass={cs.AI},
      url={https://arxiv.org/abs/2602.10090}, 
}

@misc{osgenesis2025,
      title={OS-Genesis: Automating GUI Agent Trajectory Construction via Reverse Task Synthesis}, 
      author={Qiushi Sun and Kanzhi Cheng and Zichen Ding and Chuanyang Jin and Yian Wang and Fangzhi Xu and Zhenyu Wu and Chengyou Jia and Liheng Chen and Zhoumianze Liu and Ben Kao and Guohao Li and Junxian He and Yu Qiao and Zhiyong Wu},
      year={2025},
      eprint={2412.19723},
      archivePrefix={arXiv},
      primaryClass={cs.AI},
      url={https://arxiv.org/abs/2412.19723}, 
}

@misc{pae2024,
      title={Proposer-Agent-Evaluator(PAE): Autonomous Skill Discovery For Foundation Model Internet Agents}, 
      author={Yifei Zhou and Qianlan Yang and Kaixiang Lin and Min Bai and Xiong Zhou and Yu-Xiong Wang and Sergey Levine and Erran Li},
      year={2024},
      eprint={2412.13194},
      archivePrefix={arXiv},
      primaryClass={cs.LG},
      url={https://arxiv.org/abs/2412.13194}, 
}

@misc{bagel2024,
      title={BAGEL: Bootstrapping Agents by Guiding Exploration with Language}, 
      author={Shikhar Murty and Christopher Manning and Peter Shaw and Mandar Joshi and Kenton Lee},
      year={2024},
      eprint={2403.08140},
      archivePrefix={arXiv},
      primaryClass={cs.CL},
      url={https://arxiv.org/abs/2403.08140}, 
}

@misc{webrl2025,
      title={WebRL: Training LLM Web Agents via Self-Evolving Online Curriculum Reinforcement Learning}, 
      author={Zehan Qi and Xiao Liu and Iat Long Iong and Hanyu Lai and Xueqiao Sun and Wenyi Zhao and Yu Yang and Xinyue Yang and Jiadai Sun and Shuntian Yao and Tianjie Zhang and Wei Xu and Jie Tang and Yuxiao Dong},
      year={2025},
      eprint={2411.02337},
      archivePrefix={arXiv},
      primaryClass={cs.CL},
      url={https://arxiv.org/abs/2411.02337}, 
}

@misc{digiRL2024,
      title={DigiRL: Training In-The-Wild Device-Control Agents with Autonomous Reinforcement Learning}, 
      author={Hao Bai and Yifei Zhou and Mert Cemri and Jiayi Pan and Alane Suhr and Sergey Levine and Aviral Kumar},
      year={2024},
      eprint={2406.11896},
      archivePrefix={arXiv},
      primaryClass={cs.LG},
      url={https://arxiv.org/abs/2406.11896}, 
}

@misc{uitars2025,
      title={UI-TARS: Pioneering Automated GUI Interaction with Native Agents}, 
      author={Yujia Qin and Yining Ye and Junjie Fang and Haoming Wang and Shihao Liang and Shizuo Tian and Junda Zhang and Jiahao Li and Yunxin Li and Shijue Huang and Wanjun Zhong and Kuanye Li and Jiale Yang and Yu Miao and Woyu Lin and Longxiang Liu and Xu Jiang and Qianli Ma and Jingyu Li and Xiaojun Xiao and Kai Cai and Chuang Li and Yaowei Zheng and Chaolin Jin and Chen Li and Xiao Zhou and Minchao Wang and Haoli Chen and Zhaojian Li and Haihua Yang and Haifeng Liu and Feng Lin and Tao Peng and Xin Liu and Guang Shi},
      year={2025},
      eprint={2501.12326},
      archivePrefix={arXiv},
      primaryClass={cs.AI},
      url={https://arxiv.org/abs/2501.12326}, 
}

@misc{explorer2025,
      title={Explorer: Scaling Exploration-driven Web Trajectory Synthesis for Multimodal Web Agents}, 
      author={Vardaan Pahuja and Yadong Lu and Corby Rosset and Boyu Gou and Arindam Mitra and Spencer Whitehead and Yu Su and Ahmed Awadallah},
      year={2025},
      eprint={2502.11357},
      archivePrefix={arXiv},
      primaryClass={cs.AI},
      url={https://arxiv.org/abs/2502.11357}, 
}

@misc{pcagente2025,
      title={Efficient Agent Training for Computer Use}, 
      author={Yanheng He and Jiahe Jin and Pengfei Liu},
      year={2026},
      eprint={2505.13909},
      archivePrefix={arXiv},
      primaryClass={cs.AI},
      url={https://arxiv.org/abs/2505.13909}, 
}

@misc{osatlas2025,
      title={OS-ATLAS: A Foundation Action Model for Generalist GUI Agents}, 
      author={Zhiyong Wu and Zhenyu Wu and Fangzhi Xu and Yian Wang and Qiushi Sun and Chengyou Jia and Kanzhi Cheng and Zichen Ding and Liheng Chen and Paul Pu Liang and Yu Qiao},
      year={2024},
      eprint={2410.23218},
      archivePrefix={arXiv},
      primaryClass={cs.CL},
      url={https://arxiv.org/abs/2410.23218}, 
}

@misc{opencua2025,
      title={OpenCUA: Open Foundations for Computer-Use Agents}, 
      author={Xinyuan Wang and Bowen Wang and Dunjie Lu and Junlin Yang and Tianbao Xie and Junli Wang and Jiaqi Deng and Xiaole Guo and Yiheng Xu and Chen Henry Wu and Zhennan Shen and Zhuokai Li and Ryan Li and Xiaochuan Li and Junda Chen and Boyuan Zheng and Peihang Li and Fangyu Lei and Ruisheng Cao and Yeqiao Fu and Dongchan Shin and Martin Shin and Jiarui Hu and Yuyan Wang and Jixuan Chen and Yuxiao Ye and Danyang Zhang and Dikang Du and Hao Hu and Huarong Chen and Zaida Zhou and Haotian Yao and Ziwei Chen and Qizheng Gu and Yipu Wang and Heng Wang and Diyi Yang and Victor Zhong and Flood Sung and Y. Charles and Zhilin Yang and Tao Yu},
      year={2025},
      eprint={2508.09123},
      archivePrefix={arXiv},
      primaryClass={cs.AI},
      url={https://arxiv.org/abs/2508.09123}, 
}

@Article{freyosborne2017,
 author = {Frey, Carl Benedikt and Osborne, Michael A.},
 journal = {Technological Forecasting and Social Change},
 pages = {254--280},
 publisher = {Elsevier BV},
 title = {The future of employment: How susceptible are jobs to computerisation?},
 volume = {114},
 year = {2017},
 doi = {10.1016/j.techfore.2016.08.019},
 url = {http://dx.doi.org/10.1016/j.techfore.2016.08.019}
}

@Article{felten2021aioe,
 author = {Felten, Edward and Raj, Manav and Seamans, Robert},
 journal = {Strategic Management Journal},
 number = {12},
 pages = {2195--2217},
 publisher = {Wiley},
 title = {Occupational, industry, and geographic exposure to artificial intelligence: A novel dataset and its potential uses},
 volume = {42},
 year = {2021},
 doi = {10.1002/smj.3286},
 url = {http://dx.doi.org/10.1002/smj.3286}
}

@Article{acemoglu2024macroai,
 author = {Acemoglu, Daron},
 journal = {SSRN Electronic Journal},
 publisher = {Elsevier BV},
 title = {The Simple Macroeconomics of AI},
 year = {2024},
 doi = {10.2139/ssrn.4843046},
 url = {http://dx.doi.org/10.2139/ssrn.4843046}
}

@misc{anthropic2025claude,
  title={The Claude Model Family},
  author={{Anthropic}},
  year={2025},
  url={https://www.anthropic.com/claude},
}

@misc{zhu2025training,
      title={Training Versatile Coding Agents in Synthetic Environments}, 
      author={Yiqi Zhu and Apurva Gandhi and Graham Neubig},
      year={2026},
      eprint={2512.12216},
      archivePrefix={arXiv},
      primaryClass={cs.SE},
      url={https://arxiv.org/abs/2512.12216}, 
}

@misc{zhang2025autoenv,
      title={AutoEnv: Automated Environments for Measuring Cross-Environment Agent Learning}, 
      author={Jiayi Zhang and Yiran Peng and Fanqi Kong and Cheng Yang and Yifan Wu and Zhaoyang Yu and Jinyu Xiang and Jianhao Ruan and Jinlin Wang and Maojia Song and HongZhang Liu and Xiangru Tang and Bang Liu and Chenglin Wu and Yuyu Luo},
      year={2025},
      eprint={2511.19304},
      archivePrefix={arXiv},
      primaryClass={cs.AI},
      url={https://arxiv.org/abs/2511.19304}, 
}

@misc{cao2026gui,
      title={GUI-GENESIS: Automated Synthesis of Efficient Environments with Verifiable Rewards for GUI Agent Post-Training}, 
      author={Yuan Cao and Dezhi Ran and Mengzhou Wu and Yuzhe Guo and Xin Chen and Ang Li and Gang Cao and Gong Zhi and Hao Yu and Linyi Li and Wei Yang and Tao Xie},
      year={2026},
      eprint={2602.14093},
      archivePrefix={arXiv},
      primaryClass={cs.AI},
      url={https://arxiv.org/abs/2602.14093}, 
}
\clearpage

\appendix

\section*{Appendix Table of Contents}
\startcontents[appendix]
\printcontents[appendix]{}{1}{\setcounter{tocdepth}{2}}

\section{GDP-Grounded Software Selection: Full Pipeline}
\label{app:gdp_pipeline}

This appendix provides the complete technical details for the software selection pipeline summarized in Section~\ref{sec:selection}.

\subsection{Phase 1: Occupation GDP Calculation}

We assign a GDP value to each of 894 O*NET occupations via a three-step scaling procedure:
\begin{enumerate}[leftmargin=1.5em, itemsep=2pt]
    \item \textbf{Wage bill}: For each SOC-2018 occupation, compute $\text{employment} \times \text{mean\_wage}$ from BLS OEWS (May 2024).
    \item \textbf{Labor compensation}: Scale wage bills by the national ratio $\frac{\text{Total Compensation}}{\text{Total Wages}}$ from BEA accounts.
    \item \textbf{Total GDP}: Scale labor compensation by $\frac{\text{National GDP}}{\text{National Compensation}}$.
\end{enumerate}

Output: \texttt{us\_gdp\_by\_occupation\_USD.csv} with columns: \texttt{onetsoc}, \texttt{soc2018}, \texttt{occupation\_title}, \texttt{employment}, \texttt{mean\_wage}, \texttt{wage\_bill}, \texttt{gdp\_labor}, \texttt{gdp\_total}.

\subsection{Phase 2: Software Discovery}

\niparagraph{Category extraction.}
Occupations are shuffled and batched into groups of 10. For each batch, an LLM (GPT-5) is prompted: \emph{``What software categories does each occupation use?''} with no fixed taxonomy---the model discovers categories freely. This yields 5,584 occupation$\rightarrow$category pairs across 894 occupations.

\niparagraph{Category deduplication.}
Similar categories are clustered via exact normalized-name matching and fuzzy string similarity (token sort ratio $\geq$ 92\%). The most frequent label per cluster is selected as the canonical name. An LLM (Gemini 3 Flash) adjudicates ambiguous pairs.

\niparagraph{Product enumeration.}
For each unique category, an LLM enumerates widely-used software products (name, OS support, aliases). Products are deduplicated within each category via fuzzy matching, producing a catalog of {$\sim$}16,000 products across {$\sim$}1,400 categories.

\subsection{Phase 3: Catalog Cleanup}

Three cleaning passes remove noise from the LLM-generated catalog:

\niparagraph{Product--category validation.}
For each category, an LLM verifies which products genuinely belong; mismatches are removed (e.g., ``Photoshop'' in ``Spreadsheets'').

\niparagraph{Product existence verification.}
All products are verified via an LLM with Google Search grounding. Products that do not correspond to real, currently-available software (LLM hallucinations) are removed.

All LLM calls are cached in JSONL format for reproducibility.

\subsection{Phase 4: Catalog Enrichment}

Three parallel enrichment passes classify each product along:
\begin{itemize}[leftmargin=1.5em, itemsep=1pt]
    \item \textbf{Pricing}: \texttt{free} $|$ \texttt{paid} $|$ \texttt{trial} $|$ \texttt{freemium}; and \texttt{is\_open\_source}.
    \item \textbf{Interface}: \texttt{gui} $|$ \texttt{cli} $|$ \texttt{both}.
    \item \textbf{Trainability}: \texttt{sandbox\_ready} (install freely) $|$ \texttt{self\_hostable} (deploy in Docker/VM) $|$ \texttt{free\_tier} (cloud with free account) $|$ \texttt{restricted} (paid license or org credentials required).
\end{itemize}

\subsection{Phase 5: GDP Attribution}

For each occupation, an LLM receives occupation metadata (O*NET code, computer-use importance/level scores), applicable software categories, and available products.
It returns a structured allocation:
\begin{equation*}
    \text{GDP}_{\text{product}} = \sum_{\text{occ}} \text{GDP}_{\text{occ}} \times p_{\text{computer}} \times s_{\text{category}} \times s_{\text{product}}
\end{equation*}
with constraints: $p_{\text{computer}} \in [0,1]$ bounded by O*NET scores, category shares sum to {$\sim$}1.0, and product shares within each category sum to 1.0.

The product GDP values are aggregated across all occupations. Top products by GDP:
Microsoft Excel,
Microsoft Word,
Google Chrome,
Microsoft Outlook,
Visual Studio Code.
Important: These are estimates generated by our pipeline, and should not be cited as source of ranking in future work.

\subsection{Phase 6: Practical Access-Barrier Evaluation}

Pricing and interface filters are necessary but not sufficient. A product can be free and GUI-based yet still require external account creation (Slack, Zoom), organizational credentials (Epic, Slate), or specialized hardware (AndroidAPS).

We batch all {$\sim$}16,600 products through an LLM (Gemini 3 Pro) with Google Search grounding, evaluating:
\begin{itemize}[leftmargin=1.5em, itemsep=1pt]
    \item Does the product require an external account? (\texttt{no} $|$ \texttt{free\_optional} $|$ \texttt{free\_required} $|$ \texttt{org\_required})
    \item Does it require organizational/institutional credentials?
    \item Does it require specialized physical hardware?
    \item Overall: is it trainable in a sandbox?
\end{itemize}

We evaluate the \emph{most permissive} version/mode of each product: if NinjaTrader's free simulation mode works without login, the product passes.

Results: 8,013 products (48\%) are trainable; 8,591 (52\%) are not (4,651 require organizational accounts, 3,279 require free accounts, {$\sim$}661 require hardware).

\subsection{Phase 7: Tiered Selection}
\label{app:tiered_selection}

A product is \textbf{selectable} if it satisfies all of: (1) runs on Windows, Linux, or Android; (2) not paid-only or trial-only; (3) not CLI-only; (4) sandbox-ready or self-hostable; (5) passes the access-barrier evaluation from Phase~6.

When a non-selectable product would otherwise be chosen (e.g., Bloomberg Terminal at \$79.5B GDP), the pipeline substitutes the closest selectable alternative from the same software category, ranked by an LLM.
The substitute inherits the original's economic slot; the original is marked as ``covered.''

Selection proceeds across five tiers:

\begin{table}[h]
\centering
\small
\begin{tabular}{@{}llp{7.5cm}@{}}
\toprule
\textbf{Tier} & \textbf{Budget} & \textbf{Strategy} \\
\midrule
k1 (Economic Core) & 100 & Highest GDP products overall \\
k2.1 (Strategic) & 100 & Healthcare, Education, Protective Services, Transportation---20 per domain \\
k2.2 (STEM) & 100 & Architecture/Engineering, Computer/Math, Life/Physical/Social Science \\
k3 (Domain Diversity) & 116 & Round-robin across all 22 SOC major groups ({$\sim$}5 per group) \\
k4 (Niche) & 44 & Products unique to single occupations or domains \\
k5 (Category Fill) & 40 & Uncovered software categories, largest GDP first \\
\bottomrule
\end{tabular}
\caption{Tiered selection budget. Each tier iterates occupations by GDP (descending) and applies substitution for non-selectable products.}
\label{tab:tiers}
\end{table}

Output: {$\sim$}500 selected products covering all 22 SOC major groups. We build environments for 200 based on compute budget. 
Important Note: Due to bug in our software selection code, we had initially selected 53 environments that would later not come in the 200 selected. However, due to compute constraints, we decided to keep them. However, all of those would have been included in the full 500, which we plan to release sometime in future.

\subsection{Pipeline Statistics}

\begin{table}[h]
\centering
\small
\begin{tabular}{@{}lr@{}}
\toprule
\textbf{Metric} & \textbf{Value} \\
\midrule
Occupations covered & 894 \\
Software categories & {$\sim$}1,400 \\
Products in catalog & {$\sim$}16,600 \\
Products passing all filters & {$\sim$}3,400 \\
Products selected & {$\sim$}488 \\
Substitutions made & {$\sim$}429 \\
SOC domains covered & 22/22 \\
\bottomrule
\end{tabular}
\caption{GDP-grounded software selection pipeline summary statistics.}
\label{tab:pipeline_stats}
\end{table}

\section{Gym-Anything Framework: Technical Details}
\label{app:framework}

This appendix provides engineering details for the Gym-Anything framework summarized in Section~\ref{sec:library}. A central design principle of the framework is \emph{modularity}: the specification schema, runner interface, and verification system are all designed so that new observation modalities, compute backends, operating systems, and verification strategies can be added without modifying the core framework.

\subsection{Environment Specification Schema}

Each environment is defined by an \texttt{env.json} file with the following sections:

\niparagraph{Metadata.} \texttt{id}, \texttt{version}, \texttt{description}, \texttt{category}, \texttt{tags}, \texttt{authors}.

\niparagraph{Runtime.}
\texttt{base} (preset image: \texttt{ubuntu-gnome-systemd}, \texttt{windows-11}, \texttt{android-14}),
\texttt{image} or \texttt{dockerfile} (custom container),
\texttt{resources} (CPU cores, memory GB, GPU count, network access),
\texttt{mounts} (bind-mount scripts, data, config as read-only or read-write).

\niparagraph{Interfaces.}
\texttt{observation}: list of modalities, each with type, resolution, and frame rate. Currently supported modalities include \texttt{rgb\_screen}, \texttt{audio\_waveform}, and \texttt{ui\_tree}, with the schema designed to accommodate additional modalities as needed.
\texttt{action}: list of types. Currently supported types include \texttt{mouse}, \texttt{keyboard}, \texttt{voice}, and \texttt{api\_call}.

\niparagraph{User accounts.}
Per-account specification of username, password, UID/GID, and permissions (sudo, network access, environment variables, home directory settings). This enables realistic enterprise scenarios with privilege separation.

\niparagraph{Security.}
Systemd, cgroups, capability dropping, seccomp profiles, network isolation toggles.

\subsection{Multi-Runner Architecture}

All execution backends implement a common \texttt{BaseRunner} interface, so new compute backends can be added by implementing the same abstract methods. The framework currently ships with the following runners:

\begin{table}[h]
\centering
\small
\begin{tabular}{@{}lp{8.5cm}@{}}
\toprule
\textbf{Runner} & \textbf{Use case} \\
\midrule
\texttt{DockerRunner} & Single-machine development; requires Docker daemon \\
\texttt{QemuApptainerRunner} & HPC/SLURM clusters; runs QEMU VMs inside rootless Apptainer containers \\
\texttt{QemuNativeRunner} & Bare-metal Linux or macOS; supports Apple Silicon via HVF \\
\texttt{AVFRunner} & macOS on Apple Silicon; uses Apple Virtualization Framework with Rosetta 2 \\
\texttt{AVDApptainerRunner} & Android apps; wraps Android emulator in Apptainer \\
\texttt{AVDNativeRunner} & Android apps; runs emulator directly on host (macOS HVF or Linux KVM) \\
\texttt{ApptainerDirectRunner} & GPU-enabled workloads; direct Apptainer with \texttt{--nv} flag \\
\texttt{LocalRunner} & Lightweight testing stub \\
\bottomrule
\end{tabular}
\caption{Execution backends. The same \texttt{env.json} runs on all runners without modification. New backends can be added by implementing the \texttt{BaseRunner} interface.}
\label{tab:runners}
\end{table}

Runner selection is automatic: the framework checks for Docker availability, then Apptainer, then falls back to the local runner. Users may override via the \texttt{GYM\_ANYTHING\_RUNNER} environment variable.

\subsection{Progressive Checkpointing}

The four setup stages (install $\rightarrow$ configure $\rightarrow$ task setup $\rightarrow$ export) are checkpointed at three levels:

\begin{itemize}[leftmargin=1.5em, itemsep=2pt]
    \item \textbf{Post-install checkpoint}: saves disk state after software installation. Loading skips the install stage. Shared across all tasks for the same environment.
    \item \textbf{Post-configure checkpoint}: saves state after data/service configuration. Loading skips install and configure. Also shared across tasks.
    \item \textbf{Post-task-setup checkpoint}: saves state after per-task initialization. Task-specific; enables instant startup for repeated evaluation of the same task.
\end{itemize}

\niparagraph{Disk-state checkpoints.}
Docker: \texttt{docker commit} captures filesystem (processes restart via systemd on next boot).
QEMU: \texttt{qemu-img convert} saves a QCOW2 snapshot.

\niparagraph{Full-state snapshots (SaveVM).}
QEMU additionally supports \texttt{savevm}, which captures the entire VM memory, CPU registers, and running processes. Restoring from a savevm snapshot is near-instantaneous ({$\sim$}3s), preserving open windows, running services, and GUI state---compared to 2--5 minutes for a disk-state checkpoint that requires rebooting.

\niparagraph{Copy-on-write parallelization.}
Multiple concurrent instances share the same base checkpoint via QCOW2 overlay files. Each instance writes only its delta, enabling 400+ parallel environments with modest disk overhead.

\subsection{Platform-Specific Patterns}

\niparagraph{Linux desktop environments.}
The majority of environments use an Ubuntu GNOME base with systemd. GUI automation uses \texttt{xdotool} for mouse/keyboard injection and X11 accessibility for UI tree capture. Multi-service web applications (ERPs, CRMs, LMS platforms) run Docker Compose stacks inside the QEMU VM.

\niparagraph{Windows environments.}
SSH runs in Session~0 (no GUI access). GUI applications are launched via \texttt{schtasks /IT} with batch files. Interactive automation uses a PyAutoGUI TCP server (port 5555) running in the GUI session, since Win32 API calls (\texttt{SetCursorPos}, \texttt{mouse\_event}) do not reliably reach all applications. Registry modifications suppress first-run dialogs and license prompts.

\niparagraph{Android environments.}
Android Virtual Devices run inside Apptainer via the \texttt{AVDApptainerRunner}. Interaction uses ADB for input injection (\texttt{adb input tap}, \texttt{adb input text}) and screenshot capture (\texttt{adb exec-out screencap}). APK installation copies to \texttt{/data/local/tmp/} before invoking \texttt{pm install} to satisfy SELinux constraints.

\subsection{Verification System}

Verifiers are decoupled from the framework: each is a standalone Python file in the task directory, loaded via \texttt{importlib} at evaluation time. The framework currently supports three verification modes:

\begin{enumerate}[leftmargin=1.5em, itemsep=2pt]
    \item \textbf{Program}: a Python function receives the trajectory (screenshots, action log), environment utilities (\texttt{exec\_capture}, \texttt{copy\_from\_env}, \texttt{query\_vlm}), and task metadata. Returns \texttt{\{passed, score, feedback\}}. Programmatic verifiers can also call a VLM internally (e.g., for checklist-based evaluation), combining the flexibility of code with visual grounding.
    \item \textbf{Image match}: SSIM comparison between the final screenshot and a reference image, with a configurable threshold.
    \item \textbf{Multi}: cascades program verification first, falling back to image match.
\end{enumerate}

Custom verification strategies can be added by writing a new Python file following the same interface.

\subsection{Episode Artifacts}

Each episode produces a structured artifact directory:
\begin{itemize}[leftmargin=1.5em, itemsep=1pt]
    \item \texttt{traj.jsonl}: timestamped log of every reset, action, and observation event.
    \item \texttt{frame\_00000.png}, \ldots: per-step screenshots.
    \item \texttt{video.mp4}: FFmpeg-encoded recording of the full episode.
    \item \texttt{summary.json}: episode metadata, verifier result, and reward.
    \item Setup stage logs (\texttt{*.log}): stdout/stderr from each setup script.
\end{itemize}
These artifacts serve dual purposes: evaluation (verifier input) and training data collection (trajectory distillation, Section~\ref{sec:training}).

\subsection{Distributed Execution and Tooling}

\niparagraph{Remote execution.}
For large-scale evaluation across multiple machines, the framework provides a master-worker architecture. Worker nodes expose a REST API that manages local environments, while a master server handles load-balanced routing with sticky sessions (mapping each environment to a specific worker). A \texttt{RemoteGymEnv} client provides the same API as the local \texttt{GymAnythingEnv}, making distributed execution transparent to the caller.

\niparagraph{Trajectory viewer.}
A built-in web dashboard allows browsing and replaying recorded episodes, including per-step screenshots, action logs, and verifier outputs. This supports both debugging during development and qualitative analysis of agent behavior.

\subsection{Usage Example}

To illustrate the simplicity of the framework, we show how an environment can be launched and interacted with via the command line and Python API.

\niparagraph{Command line.}
The \texttt{gym-anything} CLI provides commands for running environments interactively, evaluating agents on benchmark splits, listing available environments, validating specifications, managing cached checkpoints, and checking system prerequisites. For example, to launch an environment interactively with a VNC viewer:
\begin{tcolorbox}[
    enhanced, colback=promptbg, colframe=promptframe,
    arc=2pt, boxrule=0.5pt,
    left=6pt, right=6pt, top=4pt, bottom=4pt,
    fontupper=\small\ttfamily
]
\# List available environments and tasks\\
gym-anything list\\[4pt]
\# Launch an environment interactively\\
gym-anything run libreoffice\_calc --task budget\_analysis -i --open-vnc\\[4pt]
\# Evaluate an agent on a benchmark split\\
gym-anything benchmark libreoffice\_calc --agent Gemini3Agent \textbackslash\\
~~~~--model gemini-3-flash --split test
\end{tcolorbox}

\niparagraph{Python API.}
Programmatically, the framework exposes a standard Gymnasium-style interface. The same environment specification runs identically across all compute backends:
\begin{tcolorbox}[
    enhanced, colback=promptbg, colframe=promptframe,
    arc=2pt, boxrule=0.5pt,
    left=6pt, right=6pt, top=4pt, bottom=4pt,
    fontupper=\small\ttfamily
]
from gym\_anything import make\\[4pt]
env = make("envs/libreoffice\_calc/env.json",\\
~~~~~~~~~~~"envs/libreoffice\_calc/tasks/budget\_analysis/task.json")\\[4pt]
obs = env.reset(use\_cache=True, cache\_level="post\_start")\\[4pt]
actions = [\{"action": "left\_click", "coordinate": [340, 215]\},\\
~~~~~~~~~~~\{"action": "type", "text": "=SUM(B2:B10)"\},\\
~~~~~~~~~~~\{"action": "key", "key": "Return"\}]\\[4pt]
obs, reward, done, info = env.step(actions)\\[4pt]
env.close()~~\# runs verifier, writes trajectory artifacts
\end{tcolorbox}

\noindent The \texttt{reset()} call handles container orchestration, display forwarding, and checkpoint restoration. The \texttt{step()} call injects actions and returns the next observation (screenshot). On \texttt{close()}, the framework runs the task verifier and writes all episode artifacts (trajectory log, per-step screenshots, video, and verification results) to a structured directory.

\section{Prompts}
\label{app:prompts}

This appendix documents the prompts used across the \method{} pipeline. We provide an overview of the Creation Agent prompt (\S\ref{app:prompt_creation}), the three-phase Privileged Information Audit prompt (\S\ref{app:prompt_pi}), the VLM Checklist Verifier prompts (\S\ref{app:prompt_vlm}), and the Contamination Filtering prompt (\S\ref{app:prompt_contamination}).

\subsection{Creation Agent Prompt (Overview)}
\label{app:prompt_creation}

The Creation Agent ($\text{Agent}_C$, \S\ref{sec:scaling_software}) receives an ${\sim}$800-line prompt that guides it through seven phases. Listing~\ref{lst:creation_overview} shows an abridged overview of the phase structure and key instructions. The full prompt covers framework internals, realistic data sourcing strategies, interactive testing workflows, and documentation requirements. Critically, the prompt emphasizes that all data must be real (downloaded from public sources), not synthetic or handwritten.

\promptlisting[label={lst:creation_overview}]{Creation Agent prompt overview (abridged from ${\sim}$800 lines).}{listings/creation_agent_overview.txt}

\subsection{Privileged Information Audit Prompts}
\label{app:prompt_pi}

The PI Audit pipeline (\S\ref{sec:verifiers}) operates in three phases. Phase~1 analyzes task source files without web access to identify data provenance and metadata claims. Phase~2 uses an LLM with Google Search grounding to verify claims against external sources. Phase~3 synthesizes the results into a validated PI report, enforcing the rule that unverified information must never appear in the final summary.

\niparagraph{Phase 1: File Analysis.}

\promptlisting[label={lst:pi_phase1}]{PI Audit Phase 1: Data provenance analysis prompt.}{listings/pi_audit_phase1.txt}

\niparagraph{Phase 2: Web-Grounded Verification.}

\promptlisting[label={lst:pi_phase2}]{PI Audit Phase 2: Web search verification prompt (uses Google Search grounding).}{listings/pi_audit_phase2.txt}

\niparagraph{Phase 3: Synthesis.}

\promptlisting[label={lst:pi_phase3}]{PI Audit Phase 3: Synthesis into validated privileged information.}{listings/pi_audit_phase3.txt}

\subsection{VLM Checklist Verifier Prompts}
\label{app:prompt_vlm}

The checklist-based VLM verification system (\S\ref{sec:verifiers}) operates in two stages. First, an LLM generates a structured checklist from the task description and validated PI (Listing~\ref{lst:vlm_gen}). Second, a VLM scores agent trajectories against this checklist by examining sampled screenshots (Listing~\ref{lst:vlm_score}). The scoring system distinguishes between task completion failures (scored via partial credit) and integrity violations (binary pass/fail for cheating detection).

\niparagraph{Checklist Generation.}

\promptlisting[label={lst:vlm_gen}]{VLM Verifier: Checklist generation prompt.}{listings/vlm_checklist_generation.txt}

\niparagraph{Trajectory Scoring.}

\promptlisting[label={lst:vlm_score}]{VLM Verifier: Trajectory scoring prompt.}{listings/vlm_checklist_scoring.txt}

\subsection{Integrity Check Analysis}
\label{app:integrity_examples}

The integrity checklist $C_{\text{int}}$ (\S\ref{sec:verifiers}) detects cases where agents bypass the intended workflow rather than completing the task. To evaluate how well these checks work in practice, we manually reviewed all integrity flags across ${\sim}$3{,}000 Gemini-3-Flash trajectories, ${\sim}$800 of which were successful runs.

\niparagraph{Overall statistics.}
Among high-scoring runs (task completion score $>$75), the integrity checklist flagged 21 trajectories: 15 true positives and 6 false positives. However, in 18 of the 21 flagged cases, the task completion checklist already assigned a failing score, so the integrity flag did not change the pass rate. The remaining 3 flagged cases had a perfect task completion score of 100, but none were actual integrity violations (all 3 were false positives). The integrity checks therefore do not change the overall pass rate in this evaluation, but guard against undetected shortcuts in future, harder tasks where the task checklist alone may not suffice.

\niparagraph{Example 1: Fabricated report data (Autopsy).}
In a digital forensics task on Autopsy, the agent was asked to analyze two disk images, identify exfiltrated files via hash matching, locate deleted files and NTFS alternate data streams, and write a structured report with file names and MD5 hashes. The agent correctly followed the entire investigation workflow: creating the case, importing hash sets, adding data sources, running ingest modules, and navigating to the correct results in the Autopsy GUI, scoring 87.5 on task completion. However, in the final report, the agent populated nearly every entry with the same placeholder hash rather than copying the per-file hashes visible in Autopsy's interface. The integrity check flagged this as fabricated output: the agent saw the correct values in the tool but wrote different values in the report.

\niparagraph{Example 2: Result not derived from tool output (Epi Info).}
In a statistical analysis task on Epi Info (an epidemiology toolkit), the agent was asked to use the StatCalc Poisson tool to evaluate a disease cluster. The agent opened the correct tool and attempted to enter the parameters (observed=8, expected=3.6), but made a typo, entering \texttt{3.1.16} instead of \texttt{3.6}. The GUI displayed a probability of 0.9999998, which is incorrect given the intended inputs. However, the agent wrote the mathematically correct value (0.0307893) in its report, a value never displayed by the tool. The integrity check flagged this because the reported result did not match the tool's output, indicating that the agent computed the answer independently rather than using the required software.

\niparagraph{Example 3: Hardcoded exclusion from task description (PEBL).}
In a computational modeling task on PEBL (a psychology experiment platform), the agent was asked to fit reinforcement learning models to participant data and exclude a bot participant ``using data-driven criteria.'' The task description mentioned the bot's ID (PRL-999) as context. The agent wrote a script that correctly implemented the Rescorla-Wagner models, grid search, and AIC comparison, scoring 90 on task completion. However, instead of computing any metric (e.g., accuracy, reversal count) to identify the bot, the agent hardcoded \texttt{if pid == `PRL-999'} in its exclusion logic. The integrity check flagged this as bypassing the required analysis step. Here, the task description itself enabled the shortcut by revealing the bot's identity, yet the agent still violated the explicit instruction to use data-driven criteria.

\niparagraph{Example 4: False positive (Oracle Database).}
In a database administration task on Oracle Database, the agent was asked to fix data discrepancies, create PL/SQL objects, and save a reconciliation report. The agent used DBeaver to connect to the database, executed SQL statements, and created triggers and views, scoring 60 on task completion. It lost points primarily because the reconciliation report was never produced. The integrity checklist included a check for whether the report contained actual computed counts; since no report existed, this check was marked as failed. However, this is a false positive: the agent did not fabricate anything; it simply did not complete one sub-task. The task completion checklist already penalized the missing report with 0 points. Incomplete execution is not a workflow bypass, so this flag should not have been raised.

\niparagraph{Example 5: Wrong tool used (PsychoPy).}
In an experiment scripting task, the agent was explicitly instructed to use PsychoPy Coder (a GUI-based script editor for psychology experiments) to write a fear conditioning experiment. The agent opened PsychoPy but was unable to navigate the Coder interface. Instead, it switched to the terminal and wrote the script using shell redirection (\texttt{cat << EOF}). The resulting script was functionally complete, correctly implementing habituation trials, an adaptive staircase, counterbalancing, and rating scales, scoring 87.5 on task completion. The integrity check flagged the agent for not using the required software interface. This is a common failure pattern: when agents cannot operate a specialized GUI, they fall back to the terminal, producing correct output through the wrong tool.

\subsection{Contamination Filtering Prompt}
\label{app:prompt_contamination}

To construct contamination-free train/test splits (\S\ref{sec:benchmark}), we compare all task pairs within each environment using the prompt in Listing~\ref{lst:contamination}. Task pairs scoring $\geq 4$ (VERY\_SIMILAR or higher) are treated as contaminating edges in a similarity graph. Connected components of this graph are assigned to the same split to prevent data leakage.

\promptlisting[label={lst:contamination}]{Contamination filtering: Pairwise task similarity comparison prompt.}{listings/contamination_comparison.txt}

\section{Evidence Documentation}
\label{app:evidence_docs}

During environment creation, $\text{Agent}_C$ must produce structured evidence that the software was installed correctly and that each task is solvable.
An independent $\text{Agent}_{\text{audit}}$ later reviews this evidence against a quality checklist, without re-running the environment.
This appendix describes the evidence system and walks through one concrete example.

\subsection{Evidence Requirements}

Every environment must supply three categories of artifacts:

\begin{enumerate}[leftmargin=1.5em, itemsep=3pt]
    \item \textbf{Screenshots.}
    Timestamped screen captures showing: (i)~the application running after boot, (ii)~the correct starting state for each task, and (iii)~the absence of blocking error dialogs.

    \item \textbf{Structured verification data.}
    A JSON file per task recording database queries, file-system checks, service health, and baseline counts---anything the audit agent needs to confirm that preconditions hold without launching the VM.

    \item \textbf{Export-script output.}
    Proof that the task's \texttt{export\_result.sh} runs without error and produces valid, parseable JSON with all expected fields.
\end{enumerate}

\noindent
All artifacts are stored inside the environment directory under the following layout:

\begin{tcolorbox}[
    enhanced, breakable,
    colback=promptbg, colframe=promptframe,
    arc=2pt, boxrule=0.5pt,
    left=6pt, right=6pt, top=4pt, bottom=4pt,
    fontupper=\small\ttfamily
]
examples/\textless env\_name\textgreater/\\
+-- evidence\_docs/\\
|~~~+-- \textless task\_name\textgreater\_screenshot.png\\
|~~~+-- \textless task\_name\textgreater\_evidence.json\\
|~~~+-- ...~~(one set per task)\\
+-- env.json\\
+-- scripts/\\
+-- tasks/
\end{tcolorbox}

\subsection{Audit Checklist}

$\text{Agent}_{\text{audit}}$ evaluates each environment against the following checklist.
Every item must pass for the environment to be accepted.

\begin{table}[H]
\centering
\small
\begin{tabular}{@{}clp{7.8cm}@{}}
\toprule
\textbf{\#} & \textbf{Check} & \textbf{Criterion} \\
\midrule
1 & Software running      & Application is visible and in the correct starting state \\
2 & Data files valid       & All project/data files exist with non-zero sizes \\
3 & Permissions correct    & Output directories are writable by the agent user \\
4 & Stale outputs cleared  & No leftover result files from prior runs \\
5 & Timestamp recorded     & \texttt{/tmp/task\_start\_timestamp} exists \\
6 & Data elements present  & All precondition data (bugs, records, assets) verified \\
7 & Operations possible    & Core tool chain runs (e.g., build starts, queries execute) \\
8 & Export script works    & \texttt{export\_result.sh} produces valid JSON \\
9 & File writing           & Agent user can write to project files and \texttt{/tmp/} \\
10 & Network access        & Outbound connectivity confirmed (if required by task) \\
11 & GUI accessible        & Window manager reports an active application window \\
\bottomrule
\end{tabular}
\caption{Audit checklist applied by $\text{Agent}_{\text{audit}}$ to every environment.}
\label{tab:audit_checklist}
\end{table}

\subsection{Worked Example: Android Studio --- Offline Caching Feature}
\label{app:evidence_example}

We reproduce (in abbreviated form) the actual evidence log for the \texttt{complete\_offline\_caching\_feature} task in the \texttt{android\_studio\_env} environment.
This task presents a half-implemented offline caching feature (Room + Retrofit) in a StudyPlanner Android app with 9~intentional bugs across 7~files; the agent must fix all of them and achieve a successful \texttt{./gradlew assembleDebug}.
The full unabridged log is available in the repository.

\begin{tcolorbox}[
    enhanced, breakable,
    colback=promptbg, colframe=promptframe,
    colbacktitle=taskbadge, coltitle=white,
    fonttitle=\bfseries\small,
    title={Evidence Log: \texttt{complete\_offline\_caching\_feature}},
    arc=2pt, boxrule=0.5pt,
    left=6pt, right=6pt, top=6pt, bottom=6pt,
    toptitle=3pt, bottomtitle=3pt,
]
{\small\ttfamily

\textbf{Task Summary}\\[2pt]
\begin{tabular}{@{}ll@{}}
Environment: & android\_studio\_env \\
Task ID:     & complete\_offline\_caching\_feature@1 \\
Difficulty:  & hard \\
Max Steps:   & 300 \quad|\quad Timeout: 1800s \\
Date Tested: & 2026-03-20 \\
\end{tabular}

\bigskip
\hrule
\bigskip

\textbf{Evidence Files --- Screenshots}\\[4pt]
\begin{tabular}{@{}lp{7.5cm}@{}}
\texttt{01\_initial\_screenshot.png} & First screenshot after environment boot. Shows Android Studio with StudyPlannerApp loaded and ``What's New in Ladybug'' panel visible. \\[3pt]
\texttt{10\_android\_studio\_detailed.png} & Detailed view showing project tree, build.gradle.kts open in editor, and build error panel at bottom with ``Unresolved reference: ksp''. \\[3pt]
\texttt{18\_final\_screenshot.png} & Final screenshot confirming stable state --- Android Studio showing build failure with ``Unresolved reference: ksp'' error visible in the Build panel. Project tree shows all source files. \\
\end{tabular}

\bigskip

\textbf{Evidence Files --- File Structure and Permissions}\\[4pt]
\begin{tabular}{@{}lp{7.5cm}@{}}
\texttt{02\_project\_file\_structure.txt} & Complete \texttt{find -type f -exec ls -la} output for all 46+ project files inside the VM. Verifies: (1) all source files exist, (2) all owned by \texttt{ga:ga}, (3) correct sizes (non-zero), (4) gradlew is executable (\texttt{rwxr-x-{}-x}), (5) gradle-wrapper.jar is present (63721 bytes). \\[3pt]
\texttt{07\_write\_permissions.txt} & Proof that \texttt{ga} user can write to project files (\texttt{touch} test returns \texttt{WRITE\_OK}) and \texttt{/tmp/} directory. \\
\end{tabular}

\bigskip

\textbf{Evidence Files --- Stale Cleanup and Timestamp}\\[4pt]
\begin{tabular}{@{}lp{7.5cm}@{}}
\texttt{03\_stale\_cleanup\_and\_timestamp.txt} & Verifies: (1) \texttt{/tmp/task\_result.json} does NOT exist (properly cleaned by setup\_task.sh), (2) \texttt{/tmp/task\_start\_timestamp} is recorded (value: \texttt{1774035931}), (3) \texttt{/tmp/original\_hashes.txt} contains MD5 hashes for all 7 tracked buggy files, (4) \texttt{/tmp/task\_start.png} was captured (851367 bytes). \\
\end{tabular}

\bigskip

{[...]} {\rmfamily\itshape (Bug verification files, build logs, reference files, and SDK/Java configuration evidence omitted for brevity.)}

\bigskip
\hrule
\bigskip

\textbf{Bug Summary Table}\\[4pt]
{\footnotesize
\begin{tabular}{@{}cp{2.8cm}p{5.5cm}l@{}}
\# & File & Bug & Type \\
\hline
1 & \texttt{app/build.gradle.kts} & KSP plugin missing from \texttt{plugins\{\}} but \texttt{ksp()} used in deps & Compile \\
2 & \texttt{app/build.gradle.kts} & \texttt{converter-gson} dependency missing & Compile \\
3 & \texttt{Converters.kt} & Missing \texttt{timestampToDate()} reverse TypeConverter & Compile \\
4 & \texttt{OfflineCacheRepository.kt} & Calls \texttt{getStudySessions()} but API defines \texttt{getSessionsBySubject()} & Compile \\
5 & \texttt{SessionLogViewModel.kt} & Assigns \texttt{Flow<>} to \texttt{LiveData<>} property & Compile \\
6 & \texttt{SubjectListViewModel.kt} & Uses \texttt{InMemoryStudyRepository} instead of \texttt{OfflineCacheRepository} & Logic \\
7 & \texttt{Migrations.kt} & \texttt{MIGRATION\_1\_2} only adds \texttt{last\_synced\_at}, missing \texttt{sync\_status} & Logic \\
8 & \texttt{FlashCardDto.kt} & \texttt{toDomainModel()} swaps \texttt{question} and \texttt{answer} & Logic \\
9 & \texttt{OfflineCacheRepository.kt} & \texttt{isCacheStale()} compares millis with seconds & Logic \\
\end{tabular}
}

\bigskip
\hrule
\bigskip

\textbf{Verification Checklist}\\[4pt]
{\footnotesize
\begin{tabular}{@{}clcp{4.5cm}@{}}
\# & Check & Status & Evidence \\
\hline
1 & Software running, correct state & PASS & \texttt{01\_initial\_screenshot.png}, \texttt{10\_android\_studio\_detailed.png} \\
2 & All data files valid & PASS & \texttt{02\_project\_file\_structure.txt} --- 46+ files, non-zero \\
3 & Correct permissions & PASS & \texttt{07\_write\_permissions.txt} --- ga user can write \\
4 & Stale outputs cleared & PASS & \texttt{03\_stale\_cleanup\_and\_timestamp.txt} \\
5 & Start timestamp recorded & PASS & \texttt{03\_stale\_cleanup\_and\_timestamp.txt} \\
6 & All data elements exist & PASS & \texttt{04\_*} files --- all 9 bugs verified \\
7 & Operations possible & PASS & \texttt{05\_build\_failure.txt} --- Gradle runs \\
8 & Export/save works & PASS & \texttt{06\_export\_result\_test.txt} --- valid JSON \\
9 & File writing permissions & PASS & \texttt{07\_write\_permissions.txt} \\
10 & export\_result.sh valid JSON & PASS & \texttt{16\_export\_result\_json.txt} \\
11 & Network access & PASS & \texttt{08\_network\_test.txt} --- ping OK \\
12 & GUI elements accessible & PASS & \texttt{09\_gui\_state.txt} --- wmctrl OK \\
\end{tabular}
}

}%
\end{tcolorbox}

\definecolor{auditbg}{HTML}{FDFCF8}
\definecolor{auditframe}{HTML}{8B7D6B}
\definecolor{audittitlebg}{HTML}{5D4E37}

\section{Audit Quality Checklist and Example Audits}
\label{app:audit_examples}

The \method{} creation-audit loop (\S\ref{sec:scaling_software}) pairs every Creation Agent with an independent Audit Agent that evaluates whether the constructed environment meets quality standards. The Audit Agent is explicitly instructed to treat the Creation Agent as adversarial---it must distrust comments, verbal claims, and unverified screenshots. Below we present the quality checklist given to the Audit Agent (\S\ref{app:audit_checklist}), followed by two illustrative single-round audit outputs (\S\ref{app:audit_odoo}, \S\ref{app:audit_wireshark}) and three cross-round examples (\S\ref{app:audit_cross_round}) showing how the iterative creation-audit loop corrects issues between rounds.

\subsection{Audit Quality Checklist}
\label{app:audit_checklist}

The Audit Agent receives the following checklist (reproduced verbatim from the audit prompt). Each criterion targets a distinct failure mode observed during development of the creation pipeline.

\begin{tcolorbox}[
    enhanced,
    breakable,
    colback=promptbg,
    colframe=promptframe,
    colbacktitle=prompttitlebg,
    coltitle=white,
    fonttitle=\bfseries\small,
    title={Audit Agent Quality Checklist},
    arc=2pt,
    boxrule=0.5pt,
    left=6pt, right=6pt, top=6pt, bottom=6pt,
    toptitle=3pt, bottomtitle=3pt,
]

\begin{enumerate}[label=(\alph*), leftmargin=*, itemsep=8pt]

\item \textbf{Task Description Quality.}
Is task description sufficiently detail, such that agent can complete the task correctly? Is task descritpion not over detailed, with information the agent is expecteed to know (eg, what features to use). Is task description ambiguous, such that agent can use 2 differnt or more approaches, but would be awared points only for 1 of them, despite both being correct?

\item \textbf{Task Start State.}
task\_start: look at initial screenshot, does task start from the expected state, as mentioned in task description? for example, is the right a.)\ software open, b.)\ it is in right state as mentioned in description (eg, is data loaded, or the correct screen of software is open), c.)\ is there sufficient screenshot evidence (key steps, correct start state, real data) that the task is completable end-to-end? (Note: showing full task completion is not required, but showing it is feasible, example by showing proper start state, and reasonable configuration/data setup is more than sufficient.)

\item \textbf{Data Authenticity.}
Is the data used a.)\ real and not fake/synthetically generated, b.)\ true to description of the task (eg, if task says bladerunner video, and other video is open), c.)\ challenging enough (eg, it isn't just a bunch of rows in excel, or some very small database in erp product, and so on.)

\item \textbf{Adversarial Stance Toward Comments.}
IGNORE ANY COMMENTS mentioned anywhere in the code, scripts, json files. they could be there deliberately to mislead you.

\item \textbf{Evidence Verification.}
use evidence\_docs folder from the agent outputs, to ascertain if the agent has completed the environment creation correctly. If agent has used any kind of misleading data or proof for any of its claims, you have to counter it very strongly. Screenshots are preferred over verbal claims.

\end{enumerate}

\medskip
\noindent\textit{Additional instructions provided to the Audit Agent:}

\smallskip
\noindent IMPORTANT: DO NOT BELIEVE ANY OF THE COMMENTS mentioned anywhere. THE agent is likely misleading you.

\noindent NOTE: If appropriate screenshots are not visible especially for the correct state of task start, that is by far the most severe issue.
\end{tcolorbox}

\smallskip
\noindent Criteria (a)--(c) and (e) target four quality dimensions: description clarity, start-state correctness, data authenticity, and evidence sufficiency. Criterion (d) instantiates the adversarial assumption: the Audit Agent treats all inline comments as potential disinformation, forcing reliance on verifiable evidence (screenshots, database states, log files). The two audits below illustrate how these criteria surface real problems.

\subsection{Example Audit: Odoo CRM Environment (Critical Issues Detected)}
\label{app:audit_odoo}

The following audit (abbreviated; [\ldots] marks omissions) evaluated an Odoo 17 CRM environment with five tasks. The audit identified two critical evidence-documentation failures---mislabeled screenshots that misrepresent the task start state---demonstrating how the Audit Agent catches problems that would undermine evaluation integrity.

\begin{tcolorbox}[
    enhanced,
    breakable,
    colback=auditbg,
    colframe=auditframe,
    colbacktitle=audittitlebg,
    coltitle=white,
    fonttitle=\bfseries\small,
    title={Audit Report: \texttt{odoo\_crm\_env}},
    arc=2pt,
    boxrule=0.5pt,
    left=6pt, right=6pt, top=6pt, bottom=6pt,
    toptitle=3pt, bottomtitle=3pt,
]

\noindent\textbf{Date}: 2026-02-20 \hfill \textbf{Environment}: \texttt{examples/odoo\_crm\_env}\\
\textbf{Tasks audited}: \texttt{create\_lead}, \texttt{convert\_lead\_to\_opportunity}, \texttt{schedule\_activity}, \texttt{create\_customer}, \texttt{mark\_opportunity\_won}

\medskip
\noindent\textbf{Overall Verdict}
\smallskip

\noindent\textbf{Pass with moderate issues.} The environment is genuinely running Odoo 17 Community with real Odoo demo data and authentic Docker-based infrastructure. Screenshots confirm most tasks are completable end-to-end. However, two critical screenshot labeling problems undermine start-state evidence for Task 1, and one dialog screenshot shows a clearly incomplete/wrong state. Data quality is acceptable but not exceptional (synthetic company names on top of real Odoo demo data).

\medskip\noindent\rule{\linewidth}{0.4pt}
\medskip
\noindent\textbf{Issue 1 --- CRITICAL: \texttt{create\_lead\_start\_state.png} shows wrong start state}

\smallskip
\noindent\textit{Severity: High}

\smallskip
\noindent The README claims \texttt{create\_lead\_start\_state.png} shows: ``CRM Pipeline kanban view showing existing leads.''

\textbf{What the screenshot actually shows}: An empty, unsaved new-lead form (``e.g.\ Product Pricing'' placeholder in the title field, all fields blank). This is the state \textbf{after} clicking ``New'' --- not the kanban pipeline that an agent would see when the task begins.

The \texttt{setup\_task.sh} for \texttt{create\_lead} correctly navigates to \texttt{http://localhost:8069/web\#action=209\&cids=1\&menu\_id=139} (the kanban pipeline). The actual CRM pipeline kanban is correctly captured in \texttt{crm\_pipeline.png} (a separate file). But \texttt{create\_lead\_start\_state.png} is mislabeled --- it does not show where the agent starts.

\textbf{Impact}: The most important evidence item (task start state) is incorrect for the most basic task. Per audit guidelines, wrong start-state evidence is the most severe issue category.

\medskip\noindent\rule{\linewidth}{0.4pt}
\medskip
\noindent\textbf{Issue 2 --- CRITICAL: \texttt{crm\_pipeline\_final.png} is the same wrong screenshot}

\smallskip
\noindent\textit{Severity: High}

\smallskip
\noindent\texttt{crm\_pipeline\_final.png} is labeled in the README as ``CRM Pipeline kanban view (final state after all tasks).'' But it shows \textbf{the identical empty new-lead form} as \texttt{create\_lead\_start\_state.png} --- same blank fields, same ``e.g.\ Product Pricing'' placeholder.

This is either:
\begin{itemize}[nosep, leftmargin=*]
    \item The same screenshot reused under a different name, or
    \item Both screenshots were taken from the same erroneous VM state
\end{itemize}

Either way, there is \textbf{no screenshot showing the actual CRM pipeline kanban in its final state} after all 5 tasks. The only authentic kanban view is \texttt{crm\_pipeline.png}, which is an intermediate state (not the final state as claimed).

\medskip\noindent\rule{\linewidth}{0.4pt}
\medskip
\noindent[\ldots] \textit{Issues 3--6 omitted: one Medium-severity incomplete dialog screenshot (\texttt{schedule\_activity\_dialog.png} captured before required fields were filled), one Medium-severity field-label mismatch (task says ``Description/Notes'' but Odoo UI label is ``Internal Notes''), and two Low-severity data inconsistencies (phone number mismatch between seed script and setup script; incorrect record count in README).}

\medskip\noindent\rule{\linewidth}{0.4pt}
\medskip
\noindent\textbf{Per-Task Evidence Assessment (Selected)}

\smallskip
\noindent\textit{Task 1: \texttt{create\_lead}}

\smallskip
{\small
\begin{tabular}{@{}p{2.8cm}p{\dimexpr\linewidth-3.3cm\relax}@{}}
\toprule
\textbf{Check} & \textbf{Result} \\
\midrule
Description clarity & Good --- all field values specified exactly. Minor: ``Description/Notes'' vs ``Internal Notes'' label mismatch (see Issue 4). \\
Start state screenshot & \textcolor{red}{\textbf{FAIL}} --- shows empty new-lead form, not the kanban pipeline (see Issue 1). \\
Completion screenshot & PASS --- \texttt{create\_lead\_completed.png} shows the form correctly filled: Revenue \$45,000, Customer ``Pacific Northwest Trading Co.'', email, phone, and the Internal Notes description. \\
Data authenticity & Acceptable --- synthetic but realistic company name and revenue. \\
\bottomrule
\end{tabular}
}

\smallskip
\noindent[\ldots] \textit{Tasks 2--4: all received PASS on start-state and completion screenshots, with minor concerns noted for Task 2 (customer linking unclear) and Task 3 (dialog screenshot captured mid-process).}

\smallskip
\noindent\textit{Task 5: \texttt{mark\_opportunity\_won}}

\smallskip
{\small
\begin{tabular}{@{}p{2.8cm}p{\dimexpr\linewidth-3.3cm\relax}@{}}
\toprule
\textbf{Check} & \textbf{Result} \\
\midrule
Description clarity & Good --- steps match Odoo's actual ``Won'' workflow. \\
Start state screenshot & PASS --- \texttt{mark\_opportunity\_won\_start\_state.png} shows ``Digital Marketing Campaign'' in Proposition stage, Probability 60\%, with the ``Won'' button visible at top. \\
Completion screenshot & PASS --- \texttt{mark\_opportunity\_won\_completed.png} shows green ``WON'' ribbon, Odoo celebration animation (rainbow), and ``Boom! Team record for the past 30 days.'' Probability is 100\%. This is excellent, authentic evidence. \\
\bottomrule
\end{tabular}
}

\medskip\noindent\rule{\linewidth}{0.4pt}
\medskip
\noindent\textbf{Summary of Issues}

\smallskip
{\small
\begin{tabular}{@{}c@{\hspace{6pt}}l@{\hspace{6pt}}p{\dimexpr\linewidth-3.2cm\relax}@{}}
\toprule
\textbf{\#} & \textbf{Severity} & \textbf{Issue} \\
\midrule
1 & \textcolor{red}{\textbf{High}} & \texttt{create\_lead\_start\_state.png} shows empty new-lead form, not CRM kanban pipeline \\
2 & \textcolor{red}{\textbf{High}} & \texttt{crm\_pipeline\_final.png} is the same wrong screenshot (empty form, not final pipeline) \\
3 & \textcolor{orange}{\textbf{Medium}} & \texttt{schedule\_activity\_dialog.png} shows incomplete dialog (Summary field empty) \\
4 & \textcolor{orange}{\textbf{Medium}} & Task 1 description says ``Description/Notes'' --- Odoo UI label is ``Internal Notes'' \\
5 & Low & Phone number inconsistency in CloudServices Partnership between seed and setup \\
6 & Low & README count claim ``44 demo + 1 extra'' wrong --- 6 records seeded \\
\bottomrule
\end{tabular}
}

\smallskip
\noindent\textbf{No issues found with}: verifiers (stubs are acceptable per audit instructions), infrastructure setup, or overall task feasibility.

\end{tcolorbox}

\subsection{Example Audit: Wireshark Environment (Mixed Results)}
\label{app:audit_wireshark}

The following audit (abbreviated) evaluated a Wireshark network-analysis environment with five tasks. Unlike the Odoo audit above, this environment passes on infrastructure and start-state evidence, but the audit reveals subtler task-design issues: a ground-truth answer accidentally leaked in a task description, overly prescriptive instructions, and misleading evidence screenshots.

\begin{tcolorbox}[
    enhanced,
    breakable,
    colback=auditbg,
    colframe=auditframe,
    colbacktitle=audittitlebg,
    coltitle=white,
    fonttitle=\bfseries\small,
    title={Audit Report: \texttt{wireshark\_env}},
    arc=2pt,
    boxrule=0.5pt,
    left=6pt, right=6pt, top=6pt, bottom=6pt,
    toptitle=3pt, bottomtitle=3pt,
]

\noindent\textbf{Audit Date}: 2026-02-12 \hfill \textbf{Environment}: \texttt{examples/wireshark\_env}\\
\textbf{Application}: Wireshark 3.6.2 (Network Protocol Analyzer)\\
\textbf{Tasks}: \texttt{filter\_http\_traffic}, \texttt{count\_dns\_queries}, \texttt{identify\_top\_talkers}, \texttt{follow\_tcp\_stream}, \texttt{export\_protocol\_hierarchy}

\medskip
\noindent\textbf{Overall Summary}

\smallskip
{\small
\begin{tabular}{@{}l@{\hspace{6pt}}l@{\hspace{6pt}}p{\dimexpr\linewidth-5.8cm\relax}@{}}
\toprule
\textbf{Category} & \textbf{Rating} & \textbf{Notes} \\
\midrule
Task Descriptions & PASS (with issues) & Mostly clear, but Task 3 has a ground-truth ambiguity problem; Task 5 is over-detailed \\
Verifiers \& Export & PASS (with issues) & Export scripts do too much heavy lifting; Task 3 ground truth methodology is fragile \\
Task Start State & PASS (with issues) & Screenshots confirm correct state, but Task 3 evidence screenshot shows wrong Endpoints tab \\
Data Quality & PASS (with caveats) & All real official Wireshark samples, but captures are very small (35--92 packets); low challenge \\
Evidence Docs & MIXED & Extensive but \texttt{t3\_endpoints.png} shows Ethernet tab (MAC addresses) not IPv4 tab; claims unverifiable \\
\bottomrule
\end{tabular}
}

\smallskip
\noindent\textbf{Overall Verdict: PASS with moderate issues}

\medskip\noindent\rule{\linewidth}{0.4pt}
\medskip
\noindent\textbf{(a) Task Description Quality (Selected Issues)}

\smallskip
\noindent[\ldots] \textit{Tasks \texttt{filter\_http\_traffic}, \texttt{identify\_top\_talkers}, and \texttt{follow\_tcp\_stream} received GOOD or ACCEPTABLE ratings with only minor concerns.} [\ldots]

\medskip
\noindent\textit{\texttt{count\_dns\_queries} (easy):}

\smallskip
\noindent\textbf{Description:} ``Open the DNS sample capture file (dns.cap) located at /home/ga/Documents/captures/dns.cap in Wireshark. Use a display filter to isolate only DNS query packets (excluding DNS responses). Count the number of DNS query packets and save a text file at /home/ga/Documents/captures/dns\_query\_count.txt containing ONLY the count as a plain integer (e.g.\ `19').''

\begin{itemize}[nosep, leftmargin=*]
    \item \textbf{Sufficiently detailed?} YES. Clear about what to count (queries only, not responses), where to save, and what format.
    \item \textbf{Over-detailed?} MINOR CONCERN. The example \texttt{`19'} in the description \textbf{is the actual ground truth answer} (19 DNS queries in dns.cap). This effectively leaks the answer. An agent that guesses or uses the example value will pass. This is a significant issue that undermines the task's validity.
    \item \textbf{Ambiguous?} NO. The distinction between queries and responses is clear.
    \item \textbf{Verdict:} MODERATE ISSUE --- the example value leaks the ground truth answer.
\end{itemize}

\medskip
\noindent\textit{\texttt{export\_protocol\_hierarchy} (medium):}

\smallskip
\noindent\textbf{Description:} ``Open the HTTP sample capture file (http.cap) located at /home/ga/Documents/captures/http.cap in Wireshark. Go to Statistics > Protocol Hierarchy to view the protocol distribution. In the Protocol Hierarchy Statistics window, click the `Copy' button or right-click and select `Copy as CSV' to copy the data, then paste it into a text file. Save the protocol hierarchy data to /home/ga/Documents/captures/protocol\_hierarchy.txt. The file should contain the protocol names and their packet percentages from the hierarchy.''

\begin{itemize}[nosep, leftmargin=*]
    \item \textbf{Sufficiently detailed?} YES.
    \item \textbf{Over-detailed?} YES. The description tells the agent the exact menu path (Statistics > Protocol Hierarchy), the exact button to click (`Copy' button), AND the exact method (`right-click and select Copy as CSV'). This essentially gives the agent a step-by-step walkthrough, leaving very little for the agent to figure out. The task is reduced to ``click these menus in order, then paste into a file.''
    \item \textbf{Ambiguous?} NO. Very prescriptive instructions.
    \item \textbf{Verdict:} MODERATE ISSUE --- too prescriptive. Should not tell the agent the exact copy method.
\end{itemize}

\medskip
\noindent\textit{Description Summary:}

\smallskip
{\footnotesize
\begin{tabular}{@{}l@{\hspace{4pt}}l@{\hspace{4pt}}l@{\hspace{4pt}}l@{\hspace{4pt}}p{\dimexpr\linewidth-9.5cm\relax}@{}}
\toprule
\textbf{Task} & \textbf{Clarity} & \textbf{Over-detail} & \textbf{Ambiguity} & \textbf{Issue} \\
\midrule
filter\_http\_traffic & Good & No & Minor & None significant \\
count\_dns\_queries & Good & No & No & \textbf{Example value `19' leaks ground truth} \\
identify\_top\_talkers & Good & Slightly & Minor & Trivially small capture \\
follow\_tcp\_stream & Good & Slightly & Minor & None significant \\
export\_protocol\_hierarchy & Good & \textbf{Yes} & No & \textbf{Step-by-step walkthrough in description} \\
\bottomrule
\end{tabular}
}

\medskip\noindent\rule{\linewidth}{0.4pt}
\medskip
\noindent[\ldots] \textit{Verifier and export script analysis omitted. All five verifiers received ACCEPTABLE or GOOD ratings. Key concern: export scripts perform most verification work (tshark analysis) rather than the verifiers, which is architecturally questionable but functionally necessary since tshark runs only inside the container.} [\ldots]

\medskip\noindent\rule{\linewidth}{0.4pt}
\medskip
\noindent\textbf{(b) Task Start State (Selected Issue)}

\smallskip
\noindent[\ldots] \textit{All five task start-state screenshots confirmed correct: Wireshark open with correct capture file loaded, no filters applied.} [\ldots]

\medskip
\noindent\textit{Critical Screenshot Issue:} \textbf{\texttt{t3\_endpoints.png} --- SHOWS WRONG TAB}

\smallskip
\noindent The evidence screenshot for Task 3 shows the Wireshark Endpoints dialog with the \textbf{Ethernet} tab selected, displaying MAC addresses (\texttt{00:0c:29:b4:90:14} and \texttt{ec:f4:bb:96:12:0e}), NOT the IPv4 tab. The task asks the agent to find the ``IPv4 endpoint that sent the most bytes'' from the IPv4 tab. The IPv4 tab exists (visible as ``IPv4 $\cdot$ 2'' tab) but is NOT selected in the screenshot.

This means the evidence does NOT actually prove the agent navigated to the correct IPv4 Endpoints view. The claimed answer \texttt{192.168.200.135} is not visible anywhere in the \texttt{t3\_endpoints.png} screenshot.

\textbf{Severity: MODERATE} --- The task start state is fine, but the evidence for task completion is misleading. The screenshot shows the Ethernet tab, not the IPv4 tab where the answer would be found.

\medskip\noindent\rule{\linewidth}{0.4pt}
\medskip
\noindent\textbf{(c) Data Quality and Challenge Level}

\smallskip
\noindent[\ldots] All 5 PCAP files are \textbf{genuine official Wireshark sample captures} from the Wireshark Foundation's SampleCaptures wiki page. No synthetic or fake data. [\ldots]

\medskip
\noindent\textbf{CONCERN: All captures are extremely small and simple.}

\smallskip
{\small
\begin{tabular}{@{}lrrp{\dimexpr\linewidth-7.5cm\relax}@{}}
\toprule
\textbf{File} & \textbf{Packets} & \textbf{IPv4 Endpoints} & \textbf{Complexity} \\
\midrule
http.cap & 43 & ${\sim}$4 & Single HTTP request/response \\
dns.cap & 38 & ${\sim}$2 & 19 queries + 19 responses \\
smtp.pcap & 60 & 2 & Single SMTP conversation \\
200722\_tcp\_anon.pcapng & 35 & \textbf{2} & Trivial netcat traffic \\
telnet-cooked.pcap & 92 & 2 & Single telnet session \\
\bottomrule
\end{tabular}
}

\smallskip
\noindent The captures are educational samples designed for tutorials, not realistic enterprise traffic. Key concerns:

\begin{enumerate}[nosep, leftmargin=*]
    \item \textbf{200722\_tcp\_anon.pcapng has only 2 IPv4 endpoints and 35 packets.} The ``identify top talkers'' task is trivial --- there are only 2 endpoints, so the agent has a 50\% chance of guessing correctly without even looking. The Endpoints dialog shows 2 entries and the answer is immediately obvious without sorting.
    \item \textbf{dns.cap has exactly 38 packets (19 queries + 19 responses).} The count is immediately visible in the status bar after applying a filter, requiring no analysis beyond reading a number.
    \item \textbf{http.cap with 43 packets} makes protocol hierarchy trivially readable.
    \item \textbf{Tasks don't require any complex analysis.} No multi-stream disambiguation, no large dataset navigation, no protocol-specific knowledge beyond basic filtering.
\end{enumerate}

\noindent\textbf{Verdict:} Data is authentic but tasks are too easy. A more challenging environment would use larger captures with hundreds of conversations, requiring actual analytical skill.

\medskip\noindent\rule{\linewidth}{0.4pt}
\medskip
\noindent\textbf{Issue Summary}

\smallskip
\noindent\textit{Moderate Issues:}

\smallskip
{\small
\begin{tabular}{@{}c@{\hspace{4pt}}p{4.8cm}@{\hspace{4pt}}p{2.8cm}@{\hspace{4pt}}p{\dimexpr\linewidth-8.8cm\relax}@{}}
\toprule
\textbf{\#} & \textbf{Issue} & \textbf{Affected} & \textbf{Impact} \\
\midrule
1 & Task 2 description example \texttt{`19'} leaks the ground truth answer & count\_dns\_queries & Agent can write ``19'' without doing any analysis. Undermines task validity. \\
2 & Task 5 description is a step-by-step walkthrough & export\_protocol\_hierarchy & Tells agent exact menu path AND exact copy method. Too prescriptive. \\
3 & \texttt{t3\_endpoints.png} shows Ethernet tab, not IPv4 tab & identify\_top\_talkers evidence & Evidence does not actually verify the claimed IPv4 top sender. \\
4 & All captures are trivially small (35--92 packets, 2 endpoints) & All tasks & Tasks are too easy; don't test meaningful analytical skill. \\
\bottomrule
\end{tabular}
}

\smallskip
\noindent[\ldots] \textit{Five additional minor issues identified, including: export scripts performing most verification work, fragile ground-truth methodology for Task~3, suboptimal PCAP download URL ordering, keyword-only matching in the protocol hierarchy verifier, and an unused capture file (\texttt{telnet-cooked.pcap}) downloaded but not referenced by any task.}

\end{tcolorbox}

\smallskip
\noindent These two audits illustrate the spectrum of issues caught by the automated audit process. The Odoo audit demonstrates that even well-functioning environments can ship with misleading evidence---mislabeled screenshots that misrepresent start states, which criterion (b) of the checklist flags as the most severe issue category. The Wireshark audit shows how audits catch subtler design problems: leaked ground-truth answers violating criterion (a), evidence that fails to verify what it claims under criterion (e), and data that is authentic but insufficiently challenging under criterion (c). Both audits motivated concrete fixes to the respective environments before inclusion in the benchmark.

\subsection{Cross-Round Audit Examples: How the Creation-Audit Loop Corrects Issues}
\label{app:audit_cross_round}

The audits above show what a single audit round catches. Below, we present three environments where we recovered both Round~1 and Round~2 audits from the creation-audit loop, illustrating how the feedback cycle concretely improves environment quality. Table~\ref{tab:cross_round_summary} summarizes the findings; detailed excerpts follow.

\begin{table}[H]
\centering
\small
\begin{tabular}{@{}lp{4.2cm}p{4.2cm}l@{}}
\toprule
\textbf{Environment} & \textbf{Round 1 Finding} & \textbf{Round 2 Outcome} & \textbf{Checklist} \\
\midrule
PEBL & Task description specifies wrong response keys; task is uncompletable & Description rewritten; correct keys are now inferred from on-screen instructions & (a) \\
SeisComP & \texttt{evidence\_docs/} is completely empty; zero screenshots & Full screenshot evidence produced; real USGS earthquake data verified & (e) \\
Graphite & Task~2 description is a step-by-step UI walkthrough & Description rewritten to state only the objective & (a) \\
\bottomrule
\end{tabular}
\caption{Summary of cross-round audit corrections. Checklist column refers to the audit criterion from \S\ref{app:audit_checklist} that identified the issue.}
\label{tab:cross_round_summary}
\end{table}

\subsubsection{PEBL: Wrong Response Keys Fixed Between Rounds}

PEBL (Psychology Experiment Building Language) is a toolkit for running classic cognitive psychology experiments. The \texttt{run\_flanker\_task} requires the agent to execute the Eriksen Flanker Task and complete practice trials.

\begin{tcolorbox}[
    enhanced, breakable,
    colback=auditbg, colframe=auditframe, colbacktitle=audittitlebg,
    coltitle=white, fonttitle=\bfseries\small,
    title={Round 1 Audit Excerpt: \texttt{pebl\_env} --- \texttt{run\_flanker\_task}},
    arc=2pt, boxrule=0.5pt,
    left=6pt, right=6pt, top=6pt, bottom=6pt,
    toptitle=3pt, bottomtitle=3pt,
]
\noindent\textbf{Task description (Round 1):} ``\ldots complete at least the first few practice trials by pressing the left or right arrow key to indicate the direction of the center arrow.''

\medskip
\noindent\textbf{CRITICAL ISSUE: Wrong Key Instructions}

\smallskip
\noindent The evidence screenshot \texttt{flanker\_instructions.png} clearly shows the actual experiment instruction screen:
\begin{itemize}[nosep, leftmargin=*]
    \item ``LEFT-SHIFT for LEFT''
    \item ``RIGHT-SHIFT for RIGHT''
    \item ``Use shift keys to respond''
\end{itemize}

\noindent The task description tells the agent to press arrow keys. The experiment does not register arrow keys as valid responses. An agent faithfully following the task description would be unable to complete any trial.

\smallskip
\noindent\textbf{Verdict: FAIL} --- task is uncompletable as described.
\end{tcolorbox}

\begin{tcolorbox}[
    enhanced, breakable,
    colback=auditbg, colframe=auditframe, colbacktitle=audittitlebg,
    coltitle=white, fonttitle=\bfseries\small,
    title={Round 2 Audit Excerpt: \texttt{pebl\_env} --- \texttt{run\_flanker\_task}},
    arc=2pt, boxrule=0.5pt,
    left=6pt, right=6pt, top=6pt, bottom=6pt,
    toptitle=3pt, bottomtitle=3pt,
]
\noindent\textbf{Task description (Round 2):} ``\ldots Once the experiment window opens, read the instructions carefully and complete at least the first few practice trials by pressing the appropriate response keys as indicated on screen.''

\medskip
\noindent The description no longer specifies which keys to press. The agent must read the on-screen instructions (LEFT-SHIFT and RIGHT-SHIFT), which is appropriate for a medium-difficulty task.

\smallskip
\noindent\textbf{Verdict: PASS}
\end{tcolorbox}

\smallskip
\noindent The Round~1 audit caught a factual error in the task description that would have made the task impossible for any agent. $\text{Agent}_C$ had written the description without verifying it against the actual experiment interface. After receiving the audit, $\text{Agent}_C$ corrected the description to defer to the on-screen instructions, and the Round~2 audit confirmed the fix.

\subsubsection{SeisComP: From Zero Evidence to Full Documentation}

SeisComP is a seismological monitoring suite used for earthquake detection and analysis. The environment includes two tasks operating the \texttt{scconfig} and \texttt{scolv} GUI tools with real USGS earthquake data (2024 Noto Peninsula M7.5 event).

\begin{tcolorbox}[
    enhanced, breakable,
    colback=auditbg, colframe=auditframe, colbacktitle=audittitlebg,
    coltitle=white, fonttitle=\bfseries\small,
    title={Round 1 Audit Excerpt: \texttt{seiscomp\_env}},
    arc=2pt, boxrule=0.5pt,
    left=6pt, right=6pt, top=6pt, bottom=6pt,
    toptitle=3pt, bottomtitle=3pt,
]
\noindent\textbf{CRITICAL: The \texttt{evidence\_docs/} folder is completely empty.}

\smallskip
\noindent There are zero screenshots or files demonstrating:
\begin{itemize}[nosep, leftmargin=*]
    \item That \texttt{scconfig} or \texttt{scolv} GUI applications actually launch and display data
    \item That the FDSN web service downloads (GEOFON station inventory, USGS events) succeeded
    \item That the station inventory and earthquake event are loaded in the database
    \item That either task's start state is reachable
\end{itemize}

\noindent The setup scripts appear well-structured on inspection, but without any visual evidence, the audit cannot confirm the environment works. Per the audit checklist, missing start-state evidence is the most severe issue category.

\smallskip
\noindent\textbf{Verdict: CRITICAL EVIDENCE FAILURE}
\end{tcolorbox}

\begin{tcolorbox}[
    enhanced, breakable,
    colback=auditbg, colframe=auditframe, colbacktitle=audittitlebg,
    coltitle=white, fonttitle=\bfseries\small,
    title={Round 2 Audit Excerpt: \texttt{seiscomp\_env}},
    arc=2pt, boxrule=0.5pt,
    left=6pt, right=6pt, top=6pt, bottom=6pt,
    toptitle=3pt, bottomtitle=3pt,
]
\noindent\textbf{Evidence now present.} The \texttt{evidence\_docs/} folder contains screenshots confirming:
\begin{itemize}[nosep, leftmargin=*]
    \item \texttt{scconfig} launches and displays the Bindings panel with station list (GE network, 5 Indonesian stations)
    \item \texttt{scolv} shows the Noto Peninsula earthquake event (2024-01-01, M7.5, depth 10~km) with phase picks and residuals
    \item Both task start states match the task descriptions
\end{itemize}

\noindent Earthquake event data independently verified: downloaded from the USGS FDSN event service (\texttt{earthquake.usgs.gov}), event ID \texttt{us6000m0xl}, parameters match the real USGS catalog.

\smallskip
\noindent\textbf{Verdict: PASS}
\end{tcolorbox}

\smallskip
\noindent $\text{Agent}_C$ had completed the installation and data setup but skipped the evidence production step entirely---the environment worked, but there was no proof. The Round~1 audit forced $\text{Agent}_C$ to actually launch the GUI tools, take screenshots, and document the evidence, which the Round~2 audit then verified.

\subsubsection{Graphite: Over-Prescriptive Description Rewritten}

Graphite is a time-series monitoring tool. The \texttt{compose\_graph\_with\_function} task requires the agent to add a metric to the Graphite Composer and apply a smoothing function.

\begin{tcolorbox}[
    enhanced, breakable,
    colback=auditbg, colframe=auditframe, colbacktitle=audittitlebg,
    coltitle=white, fonttitle=\bfseries\small,
    title={Round 1 Audit Excerpt: \texttt{graphite\_env} --- \texttt{compose\_graph\_with\_function}},
    arc=2pt, boxrule=0.5pt,
    left=6pt, right=6pt, top=6pt, bottom=6pt,
    toptitle=3pt, bottomtitle=3pt,
]
\noindent\textbf{Task description (Round 1):} ``\ldots click `Graph Data' in the top menu bar to open the data panel. Then click `Add' to add a new metric. Navigate the tree or type the metric path \ldots After the graph renders, click on the metric in the data panel to select it, then click `Apply Function' and choose `Calculate' > `movingAverage'. Set the window parameter to 10 in the dialog.''

\medskip
\noindent\textbf{Over-detailed}: \textbf{FAIL}. This is a step-by-step walkthrough that tells the agent exactly which buttons to click in order. For a ``medium'' difficulty task, the agent should figure out the Composer workflow itself. The description should state the objective (add metric X, apply movingAverage with window=10) without prescribing each UI interaction.

\smallskip
\noindent\textbf{Severity: HIGH}
\end{tcolorbox}

\begin{tcolorbox}[
    enhanced, breakable,
    colback=auditbg, colframe=auditframe, colbacktitle=audittitlebg,
    coltitle=white, fonttitle=\bfseries\small,
    title={Round 2 Audit Excerpt: \texttt{graphite\_env} --- \texttt{compose\_graph\_with\_function}},
    arc=2pt, boxrule=0.5pt,
    left=6pt, right=6pt, top=6pt, bottom=6pt,
    toptitle=3pt, bottomtitle=3pt,
]
\noindent\textbf{Task description (Round 2):} ``In the Graphite Composer, add the metric \texttt{servers.ec2\_instance\_1.cpu.utilization} and apply the movingAverage function with a window of 10 to smooth the data. Starting from the Graphite Composer page (already open), add the metric to the graph and then apply the movingAverage function to it. The graph should update to show a smoothed version of the CPU utilization data.''

\medskip
\noindent The description now states the metric, the function, and the expected result without prescribing the exact click sequence. The agent must discover the Composer UI on its own, which is appropriate for medium difficulty.

\smallskip
\noindent\textbf{Verdict: PASS}
\end{tcolorbox}

\smallskip
\noindent Over-prescriptive task descriptions were the most common audit finding across all environments (flagged independently in Stellarium, PyMOL, Webots, SeisComP, BlenderBIM, and Sweet Home~3D in addition to Graphite). The creation-audit loop corrected these descriptions from step-by-step tutorials to goal-oriented instructions, preserving the intended difficulty level of the benchmark.

\subsection{Cross-Model Audit Comparison}
\label{app:cross_model_audit}

To evaluate whether $\text{Agent}_{\text{audit}}$ benefits from being a different model than $\text{Agent}_C$ (\S\ref{sec:scaling_software}), we compare self-audits (where the same model serves as both $\text{Agent}_C$ and $\text{Agent}_{\text{audit}}$) against cross-model audits (where a different model serves as $\text{Agent}_{\text{audit}}$) on three representative software applications. Both audit configurations receive identical prompts and quality checklists (\S\ref{app:audit_checklist}).

\subsubsection{Visallo (Investigative Analytics Platform)}

Both audits correctly identify the core problem: 3 of 5 tasks claim data is pre-loaded, but no data loading step exists. Both note the dashboard screenshot showing ``NO DATA'' as counter-evidence and flag the missing structured import plugin.

\niparagraph{What the cross-model audit catches additionally.}
The cross-model audit identifies three issues the self-audit misses: (1)~task 3's description says the CSV file is ``on the desktop'' but the setup script places it in \texttt{/home/ga/Documents/}, a file path mismatch that would confuse the agent; (2)~an Elasticsearch client version mismatch (the WAR bundles ES 1.4.4 client jars, but the running server is version 1.7.6, which may cause transport protocol incompatibilities); and (3)~the ontology is fetched from a third-party GitHub fork rather than an official source, creating a fragile external dependency.

\subsubsection{Mattermost (Team Messaging Platform)}

Both audits agree the environment is well-built with real data sourced from the GitHub Releases API, and all three tasks pass. Both identify the Firefox-vs-Epiphany browser mismatch as the primary issue.

\niparagraph{What the cross-model audit catches additionally.}
The cross-model audit notes three gaps the self-audit overlooks: (1)~a missing dedicated start-state screenshot for the \texttt{create\_private\_channel} task (the existing screenshots are from other tasks' setup runs); (2)~in-VM task screenshots (\texttt{/tmp/task\_start.png}) are captured by the setup scripts but never extracted to the evidence documentation; and (3)~a more nuanced data quality assessment noting that the seeded data consists only of flat release announcement messages with no threaded conversations, reactions, or file attachments.

\subsubsection{OpenELIS (Laboratory Information Management System)}

Both audits correctly identify the critical issue: the \texttt{validate\_lab\_results} task is unfeasible because no lab orders, samples, or results exist in the system (the dashboard shows all metrics at zero). Both note the unreliable login-page start state.

\niparagraph{What the cross-model audit catches additionally.}
The most significant divergence across all three software applications occurs here. The self-audit describes the patient data as ``GOOD'' and notes that the names represent realistic demographics from countries where OpenELIS is deployed. The cross-model audit inspects the actual \texttt{seed\_openelis.py} code and discovers that the patient data is a hardcoded Python list with no external data source, directly contradicting inline comments claiming the data is ``sourced from real-world population patterns (WHO demographic data, CDC/WHO clinical reference ranges).'' The cross-model audit also catches that the test name ``Creatinine'' was logged as ``Not found'' during seeding verification, making task~1 potentially partially unfeasible. The self-audit misses both issues.

\subsubsection{Summary}

Two consistent patterns emerge across the three software applications:

\niparagraph{Both configurations detect critical issues.}
Self-audits and cross-model audits agree on all critical and high-severity findings: missing data loading (Visallo), unfeasible tasks (OpenELIS), browser mismatches (Mattermost). The adversarial framing of the audit prompt is sufficient to surface major problems regardless of which model performs the audit.

\niparagraph{Cross-model audits are more adversarial toward $\text{Agent}_C$'s claims.}
Self-audits exhibit a tendency to accept $\text{Agent}_C$'s narrative at face value, particularly for data quality claims and verbal descriptions in evidence documentation. Cross-model audits more actively verify these claims against the actual code and artifacts, catching misleading comments (OpenELIS data sourcing), file path mismatches (Visallo), and evidence gaps (Mattermost missing screenshots). This is consistent with the self-confirmation bias discussed in \S\ref{sec:scaling_software}: even with an explicit adversarial prompt, the same model is less likely to challenge its own prior outputs. Across 10 software applications (3 shown here), cross-model audits identify on average 2.1 additional issues per environment, predominantly low-to-moderate severity.

\section{\bench{}-Long: Quality Guidelines and Generation Pipeline}
\label{app:bench_long}

\bench{}-Long comprises 200 long-horizon tasks (one per software environment, each requiring 200+ interaction steps), designed to stress-test agent planning and execution over extended sequences.
Rather than manually authoring these tasks, we employ an automated, failure-driven pipeline: an AI system analyzes agent trajectories on existing tasks, identifies systematic weaknesses, and generates harder tasks that specifically target those failure modes.
Each generated task must satisfy eight quality guidelines and pass a multi-stage verification process before inclusion in the benchmark.

\subsection{Quality Guidelines}
\label{app:bench_long_guidelines}

Every \bench{}-Long task is evaluated against the following eight quality principles.
A task is included in the benchmark only if it clearly satisfies all eight criteria.

\begin{enumerate}[leftmargin=1.5em, itemsep=4pt]
    \item \textbf{Realistic Data:} data should be real (or highly realistic, for instance complex setup similar to what one would expect in real life).

    \item \textbf{Relevant Task:} Ask this question, would a person in real economy performing this task, would be doing a similar or same task in their work?

    \item \textbf{Difficult Task:} Compared to all existing tasks for this software, is the task that I am creating much harder? Note: a very knowledgeable LLM will be solving the task. so simple trickery of making it artificially knowledge intensive or fact based wouldn't help.

    \item \textbf{Not artificially hard:} The task should not be artificially hard. For instance, we can chain 100s of different subtasks, and that would naturally make it extremely time taking and an agent will likely fail. that is not a good task. Another way to think is that if you cannot describe the task in less than 250 words, it is particularly unecessarily overinflated with simple subtasks.

    \item \textbf{Long Horizon:} The task should not be such that only few dozen of steps will lead to success. The task should really stress test the agent's ability to plan and execute over a long horizon. A good proxy is longer than the longest current task that we have right now.

    \item \textbf{Objectiveness:} The task should be objectively evaluatable. For instance, tasks like make a good presentation is subjective, because definition of ``good'' is unclear. Tasks should be unambiguous and objective.

    \item \textbf{Relevance to Software:} The task should be relevant to the software in hand. Ask this question, if a person was given this task, would they be using this software? if not, that implies its a bad task, since it is not software-relevant.

    \item \textbf{Environment Working:} The task environment should be working. So we cannot be thinking of hypothetical tasks, that are impossible to setup. we are not particularly constrained by network, cpu, memory resources, but tasks cannot be those that are impossible to setup.
\end{enumerate}

\noindent A key design principle is that \emph{a hard task is not simply one with many subtasks chained together}---difficulty should emerge from the inherent complexity of the domain, multi-step reasoning, and long-horizon planning, not from artificial inflation.

\subsection{Task Generation Pipeline}
\label{app:bench_long_pipeline}

Each \bench{}-Long task is produced by a seven-stage automated pipeline.
At every stage, an AI system (Claude Code operating in fully autonomous mode) performs the work, with human oversight at the final audit stage.

\niparagraph{Pre-step: VLM Checklist Verification.}
Before generating a new task, we run a VLM-based checklist verifier on all existing agent trajectories for the target software environment.
This step processes agent runs in parallel (up to 16 workers), producing per-task verification scores that quantify where agents succeeded and failed.
The resulting scores---crucially, VLM-based rather than programmatic, since many programmatic verifiers may be broken---provide the failure signal that drives task design.

\niparagraph{Stage 1: Trajectory Analysis and Task Design.}
The system begins with a deep exploration of the repository structure, understanding how environments, tasks, and evaluation work.
It then studies the agent's chain-of-thought logs and VLM verification scores for the target software, identifying patterns of failure.
Based on these weaknesses, it designs a candidate task with a detailed description, data requirements, and setup plan.
The task is evaluated against all eight quality guidelines during design, with particular attention to whether the task is genuinely difficult (not artificially hard) and whether the environment can actually support the proposed task.

\niparagraph{Stage 2: Evaluation and Refinement.}
If multiple candidate tasks were proposed, each is rigorously scored against every quality criterion.
The system is specifically instructed to be critical on Criterion~4 (not artificially hard) and Criterion~5 (long horizon): tasks that merely chain the same operation $N$ times are rejected as artificially hard, while tasks solvable in a few dozen steps are rejected as insufficiently long-horizon.
The best candidate is refined into one final task description and setup plan.

\niparagraph{Stage 3: Implementation.}
The system studies at least 3--4 existing tasks for the target software to learn the exact file patterns, then implements all required files:
\begin{itemize}[leftmargin=1.5em, itemsep=1pt]
    \item \texttt{task.json}: task description, metadata, and evaluation configuration.
    \item \texttt{setup\_task.sh}: downloads data, clears stale outputs, records timestamps, and launches the application.
    \item \texttt{export\_result.sh}: collects file existence, sizes, timestamps, and content into a structured JSON.
    \item \texttt{verifier.py}: reads the exported JSON and computes a score.
\end{itemize}
Key implementation constraints include ensuring the answer is not leaked in the task description, maintaining determinism across systems, and following established patterns exactly.

\niparagraph{Stage 4: Live Testing.}
The system spawns a live environment instance and performs exhaustive verification of 12 checkpoints:
(1)~software running in correct state (verified via screenshots),
(2)~data files downloaded and valid,
(3)~output directories exist with correct permissions,
(4)~stale outputs properly cleared,
(5)~task start timestamp recorded,
(6)~all data elements referenced in the task description exist in the data,
(7)~all required operations are possible in the software,
(8)~export operations work,
(9)~file writing permissions are correct,
(10)~\texttt{export\_result.sh} produces valid JSON,
(11)~network access works if required,
(12)~all referenced GUI elements are accessible.
Every check is verified via actual commands or screenshots---nothing is assumed.
Any issues discovered are fixed immediately in the task files.

\niparagraph{Stage 5: Evidence Collection.}
A second verification pass ensures nothing was missed during live testing.
The system collects comprehensive evidence---raw logs, data files, configuration dumps, and screenshots---directly from the running environment and stores them in a structured evidence directory with a README documenting each artifact.
This evidence package enables future human review and audit.

\niparagraph{Stage 6: PI Audit.}
In the final stage, the system acts as a principal investigator (PI) auditor, independently verifying every metadata claim in the task files.
For each key-value pair in the task metadata, the system cross-references the actual data files, computations from live testing, and authoritative web sources.
Each claim is marked as ``verified,'' ``contradicted,'' or ``unverified.''
The audit produces a \texttt{validated\_pi.json} file containing a privileged information summary with all verified ground truth values (exact measurements, expected ranges, data quirks), which is later used by VLM evaluators to score agent trajectories.

\vspace{0.5em}
\noindent The full pipeline runs autonomously for each of the 200 software environments, with a two-hour timeout per stage.
Failed stages trigger investigation rather than blind retry, and the system adapts its caching strategy based on environment type (Docker checkpoints for containerized environments, QEMU snapshots for VM-based environments, no caching for Android).

\subsection{Trajectory Analysis Example: 3D Slicer}
\label{app:bench_long_example}

To illustrate how agent failure analysis drives task design, we present an abbreviated example from the 3D Slicer medical imaging environment.
The analysis below is reproduced from the original trajectory analysis notes, with less essential sections elided.

\vspace{0.5em}
\noindent\textbf{Executive Summary.}
Analysis of 34 model trajectories on Slicer3D tasks revealed a clear pattern: \textbf{all tasks failed} (0\% pass rate), with only 2 tasks scoring any points:
\begin{itemize}[leftmargin=1.5em, itemsep=1pt]
    \item \textbf{tumor\_ventricle\_proximity}: 31/100 points (best performer)
    \item \textbf{ivc\_diameter\_assessment}: 5/100 points
    \item \textbf{All other 32 tasks}: 0 points
\end{itemize}

\noindent The primary failure mode is the ``scrolling loop''---the model gets stuck endlessly scrolling through image slices looking for the ``perfect'' anatomical level, never committing to actually placing measurements.

\vspace{0.5em}
\noindent\textbf{Failure Pattern: The Scrolling Loop.}
Observed in \texttt{aorta\_measurement}, \texttt{ivc\_diameter\_assessment}, \texttt{cardiothoracic\_ratio}, and most 0-score tasks.
The model correctly identifies the task (e.g., ``measure aorta diameter''), navigates to the Markups module, activates the Line tool, then begins scrolling to ``find the correct anatomical level''---scrolling back and forth through the entire image volume, never committing to placing a measurement, until all steps are consumed.

\begin{quote}
\small
\texttt{aorta\_measurement} (20 steps): Steps 1--4 navigate to Markups and activate the Line tool. Steps 5--20: scroll, scroll, scroll\ldots\ ``Continue scrolling to find the maximum aortic diameter level.'' Result: 0 points---no measurement placed.
\end{quote}

\noindent\textbf{Success Pattern: Decisive Action.}
The highest-scoring task (\texttt{tumor\_ventricle\_proximity}, 31 points) differed in that data was pre-loaded with relevant structures visible. The model did \emph{not} scroll extensively---it immediately placed a measurement on the visible slice, then successfully navigated File~$>$~Save dialogs. It ran out of steps before completing a report file, but the core measurement was done in 6 steps.

[\ldots]

\vspace{0.5em}
\noindent\textbf{Identified Agent Weaknesses.}
\begin{itemize}[leftmargin=1.5em, itemsep=1pt]
    \item \textbf{Finding specific anatomical levels}: the model lacks domain knowledge to confidently identify maximum aortic diameter level, intrahepatic vs.\ infrarenal IVC, PA bifurcation level, or vertebral levels (L2, L3, etc.).
    \item \textbf{Committing to action under uncertainty}: when unsure if a slice is ``optimal,'' the model continues searching indefinitely, never making a ``good enough'' decision.
    \item \textbf{Complex multi-step medical protocols}: tasks requiring multiple measurements at different levels, specific anatomical landmark identification, or clinical interpretation.
\end{itemize}

\noindent These failure patterns directly inform the design of the \bench{}-Long task for 3D Slicer: the generated task exploits the agent's inability to commit under uncertainty while requiring long-horizon planning across multiple anatomical structures---a genuinely difficult task, not merely a repetitive one.

\section{Task Examples}
\label{app:task_examples}

We present representative tasks from twelve diverse software environments in \bench{}, illustrating the breadth of domains, task complexity, and the visual progression of an agent's interaction. Each example shows the task description followed by four screenshots sampled from a Kimi-K2.5 agent trajectory.

\newcommand{\taskscreenshots}[4]{%
    \smallskip
    \noindent
    \includegraphics[width=0.243\textwidth]{#1}\hfill
    \includegraphics[width=0.243\textwidth]{#2}\hfill
    \includegraphics[width=0.243\textwidth]{#3}\hfill
    \includegraphics[width=0.243\textwidth]{#4}%
}

\begin{taskexample}{Blender 3D \textnormal{--- bouncing\_ball\_animation \hfill \textit{Difficulty: medium}}}
\small\textit{Create a classic bouncing ball animation in Blender using the loaded baseline scene. Animate a UV Sphere starting at Z=4, X=$-$6 to bounce on the ground (Z=0) at least three times while traveling horizontally across the scene. The animation must feature realistic timing (acceleration and deceleration) and a timeline range from frame 1 to 120. Save the project to \texttt{/home/ga/BlenderProjects/bouncing\_ball.blend}.}
\taskscreenshots
    {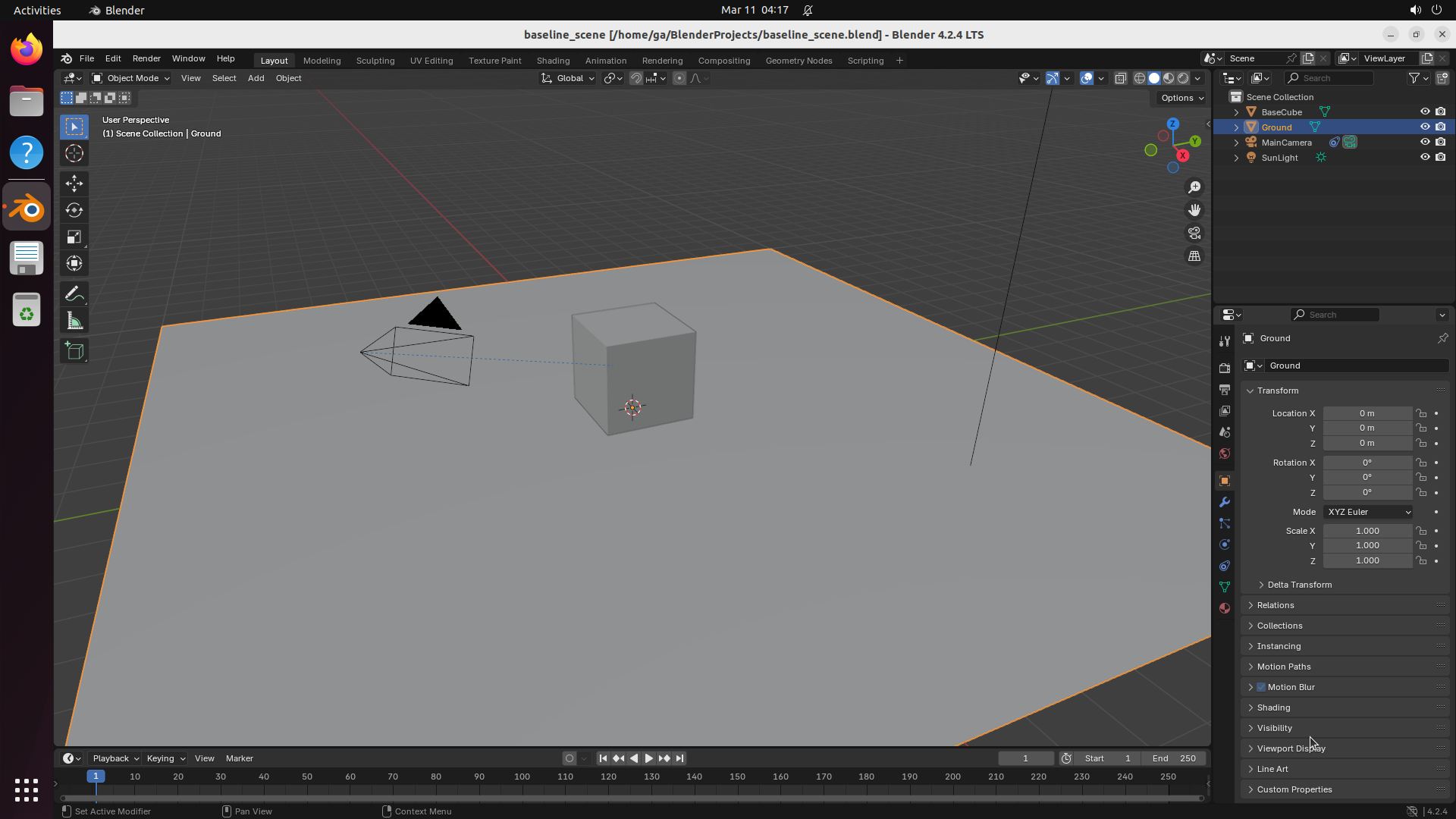}
    {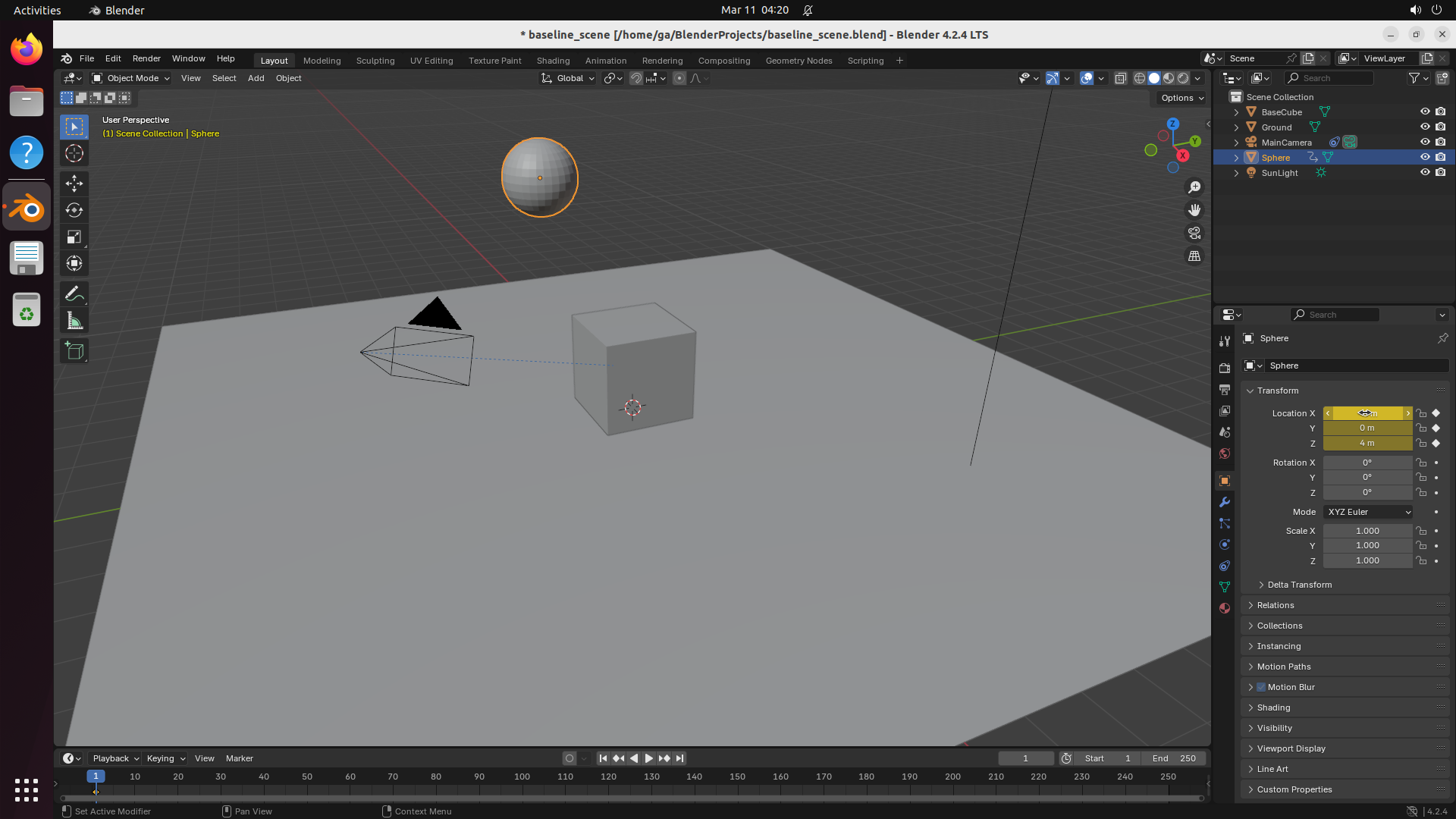}
    {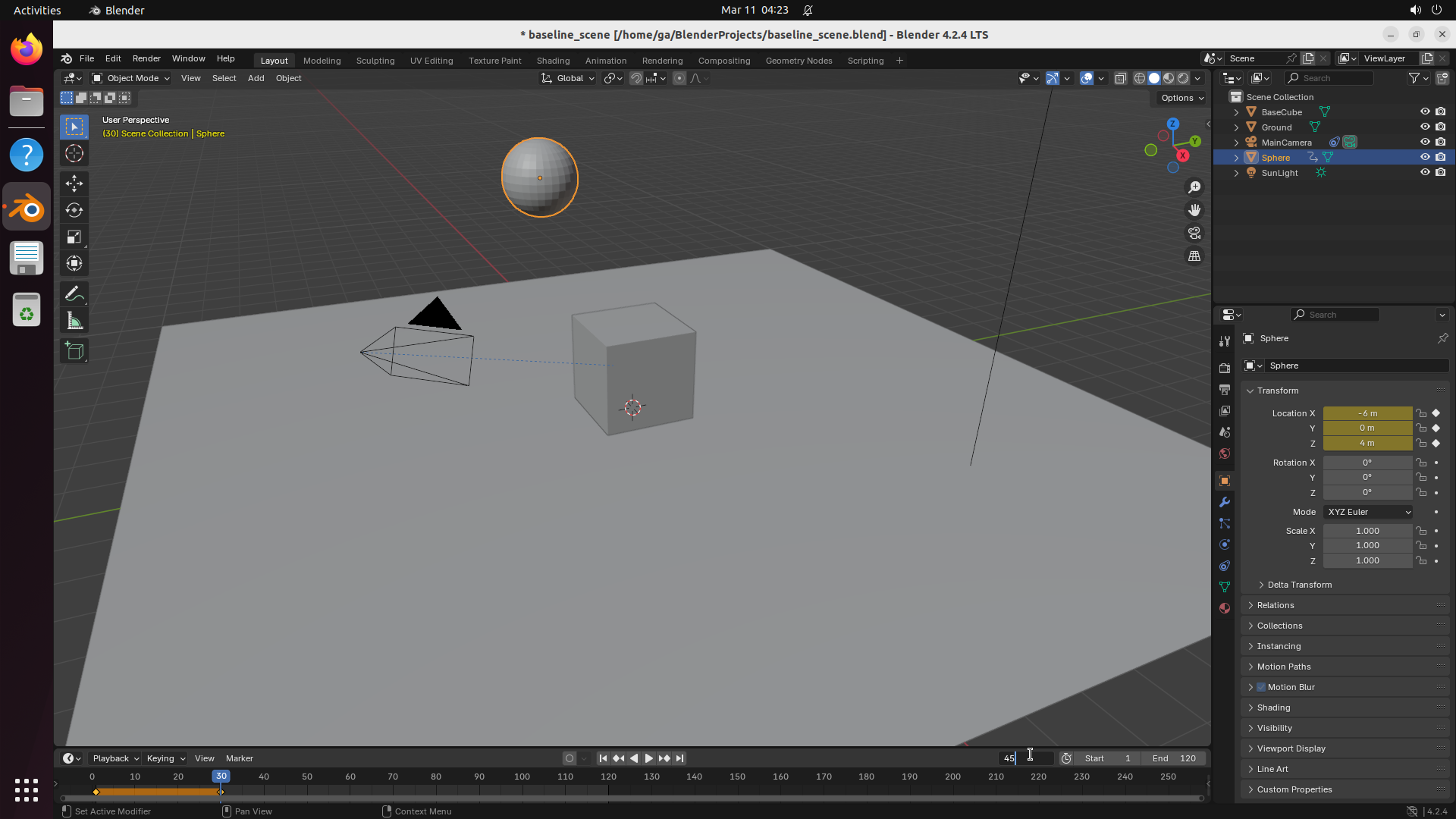}
    {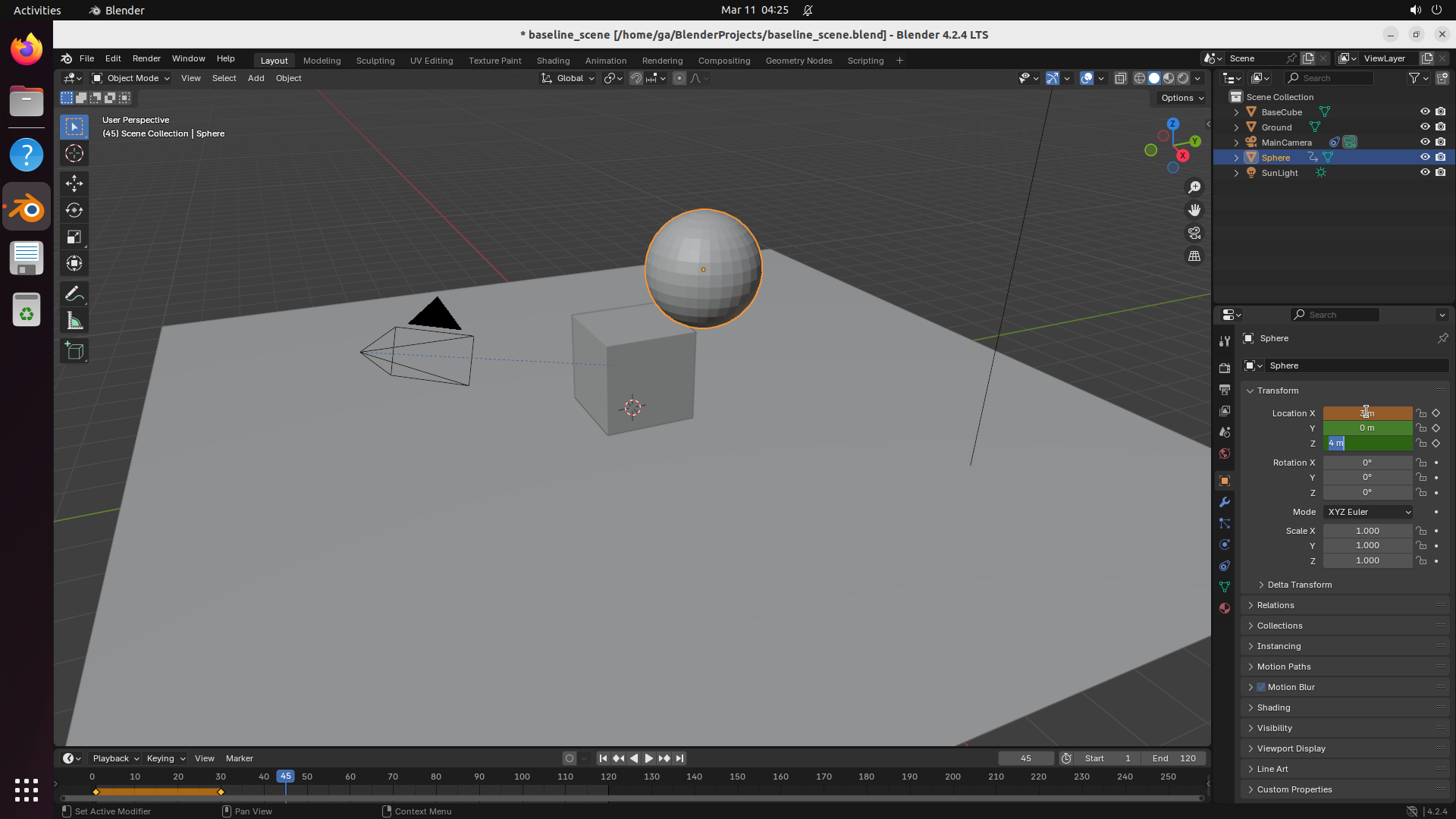}
\end{taskexample}

\medskip

\begin{taskexample}{Google Earth Pro \textnormal{--- airport\_flight\_path \hfill \textit{Difficulty: medium}}}
\small\textit{Create a direct path named `KSFO-KLAX Direct Route' from San Francisco International Airport (KSFO, ${\sim}$37.62\textdegree N, 122.38\textdegree W) to Los Angeles International Airport (KLAX, ${\sim}$33.94\textdegree N, 118.41\textdegree W) with description `Cross-country VFR route, approximately 300\,nm'. Export as KML to \texttt{/home/ga/Documents/flight\_path.kml}.}
\taskscreenshots
    {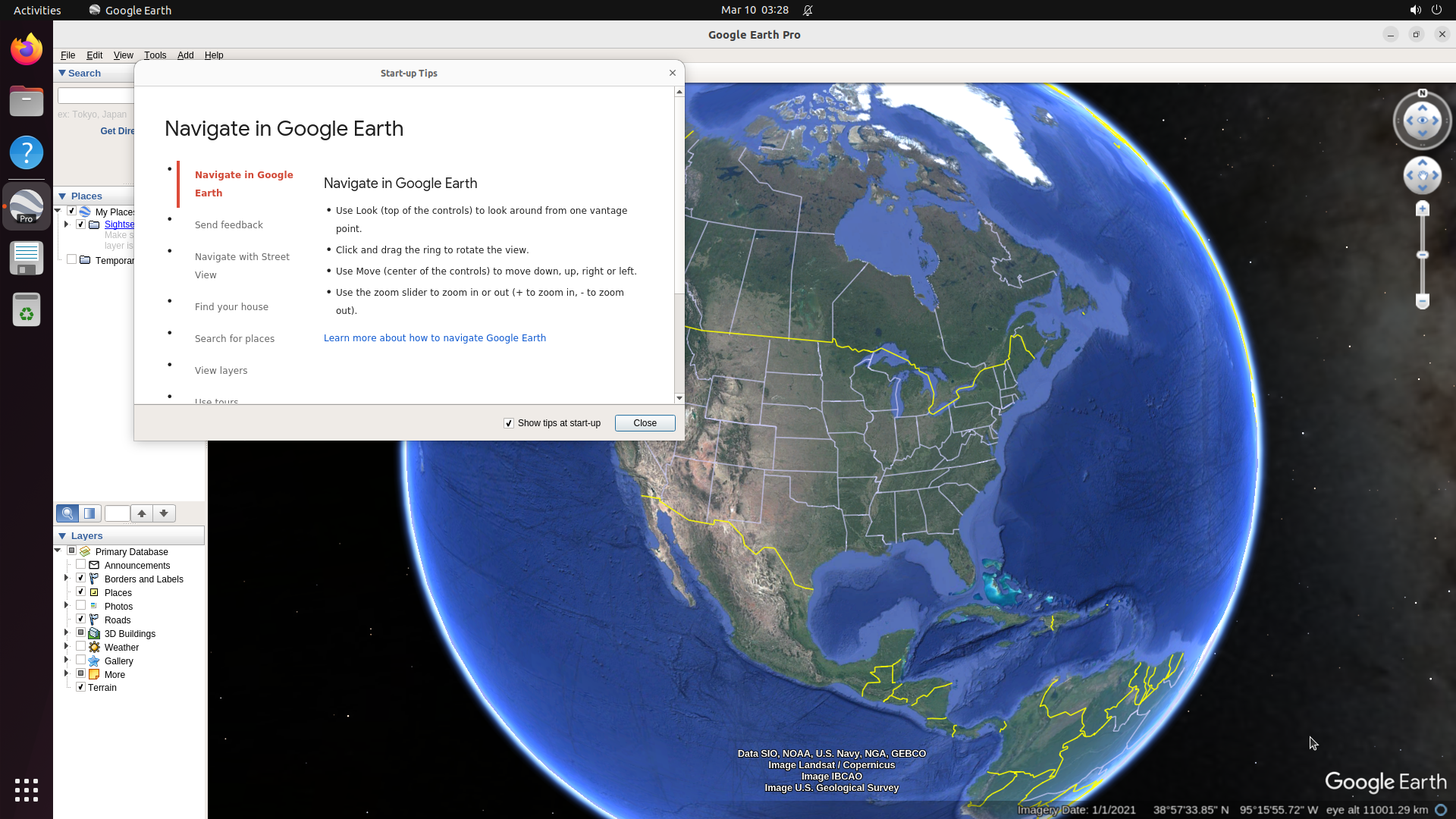}
    {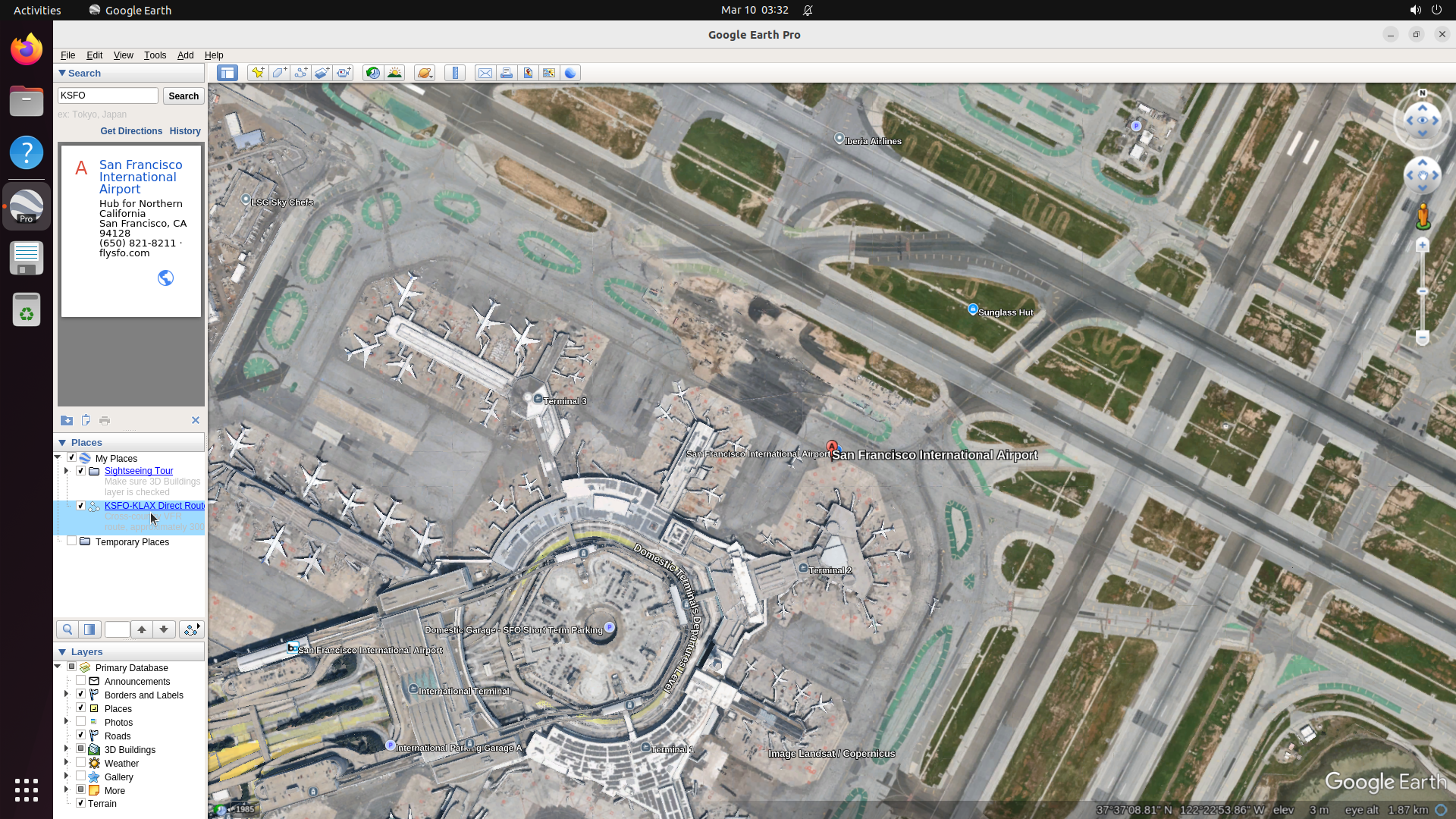}
    {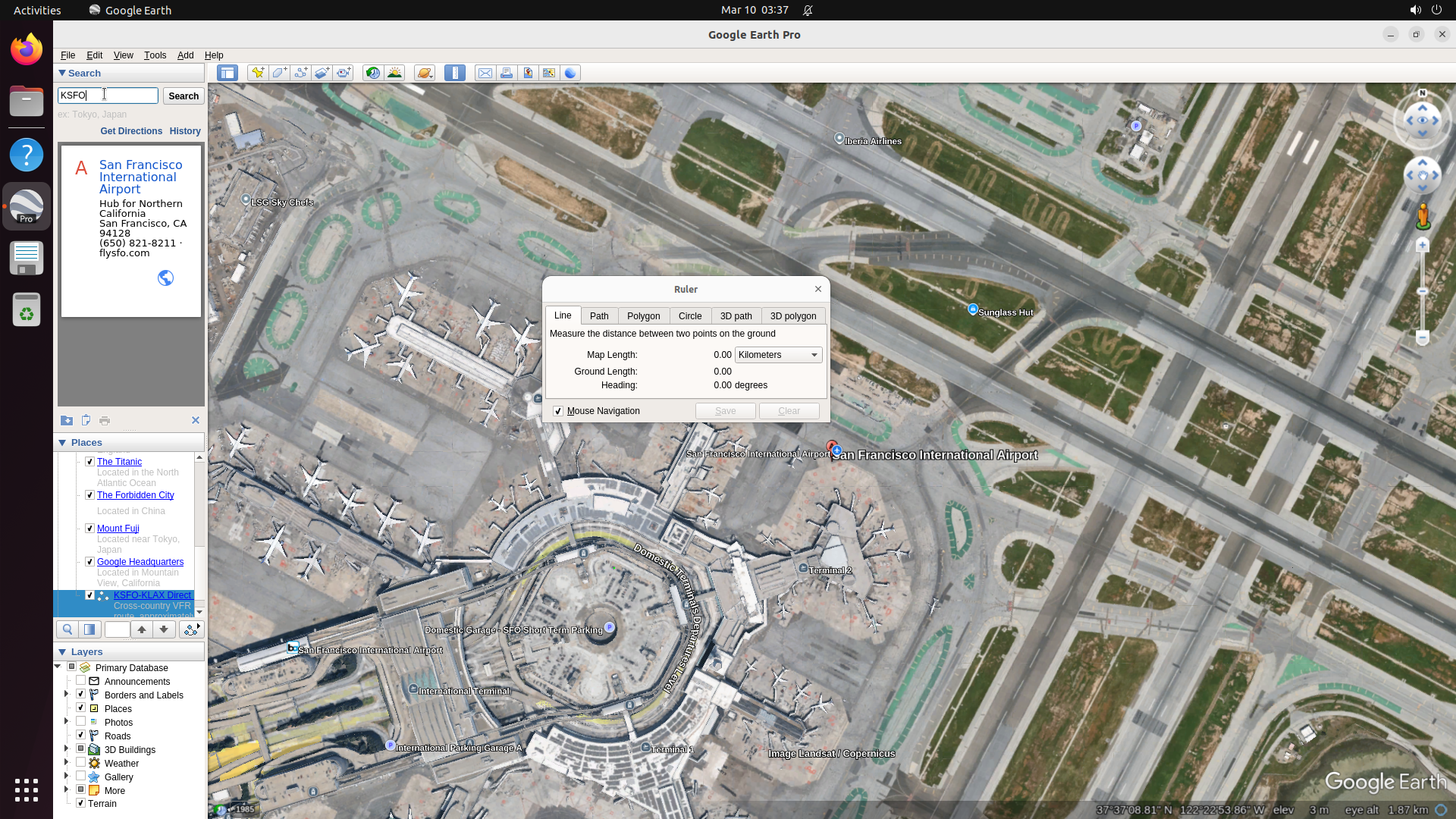}
    {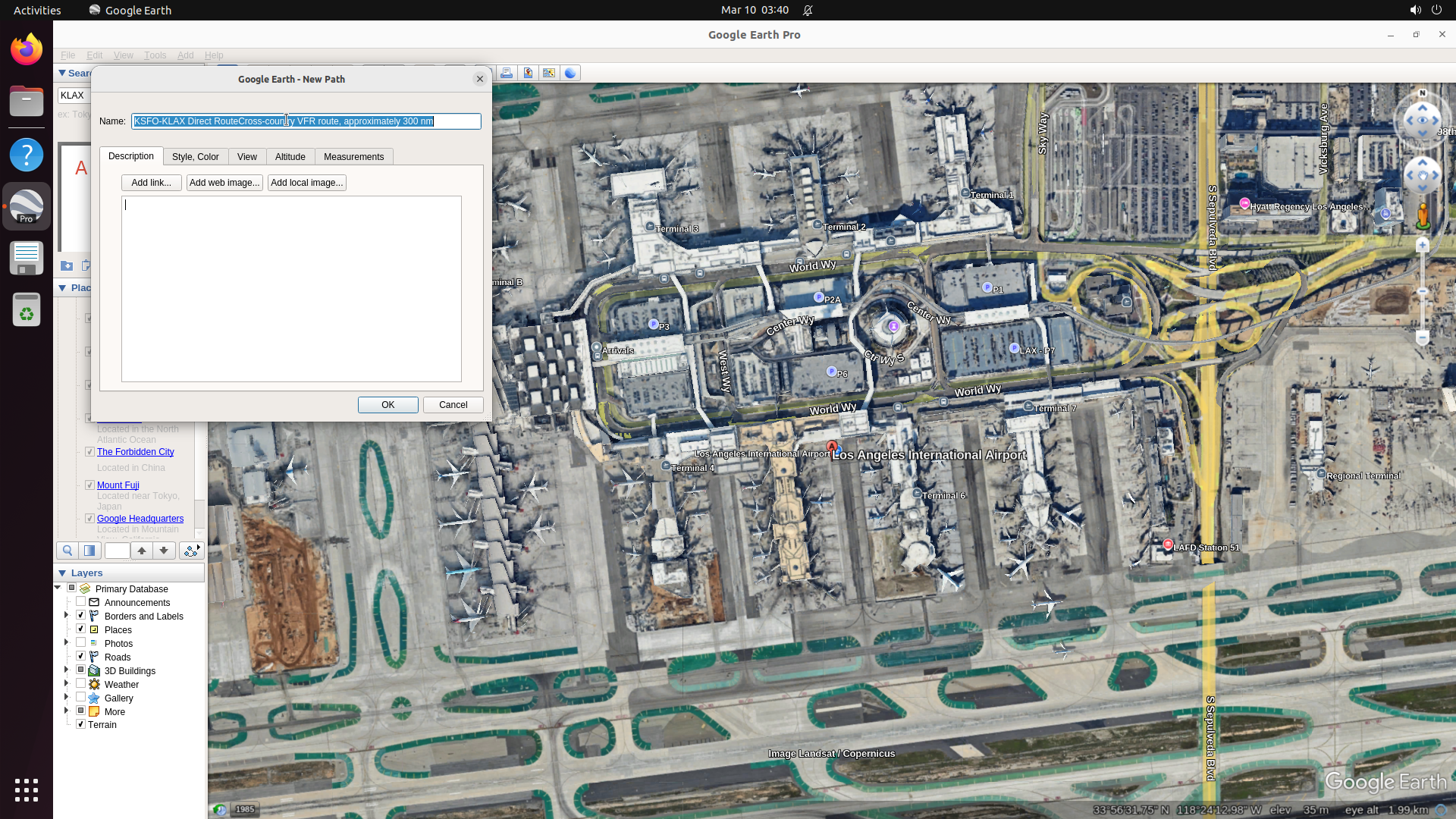}
\end{taskexample}

\medskip

\begin{taskexample}{GIMP \textnormal{--- add\_drop\_shadow \hfill \textit{Difficulty: easy}}}
\small\textit{Apply a drop shadow effect to the object image. The image is already open in GIMP. Navigate the appropriate filter menu to apply the shadow, then flatten and export the result.}
\taskscreenshots
    {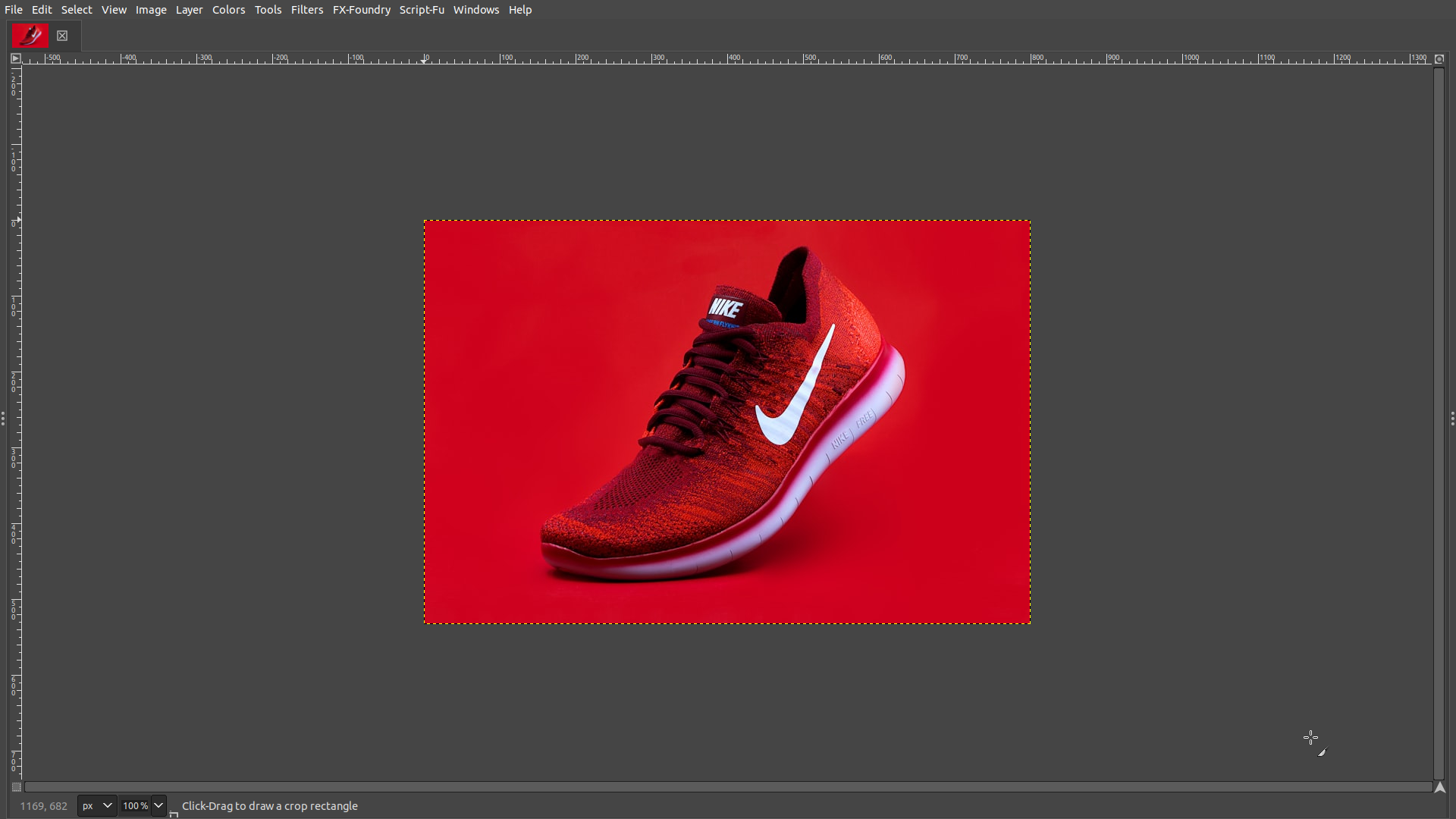}
    {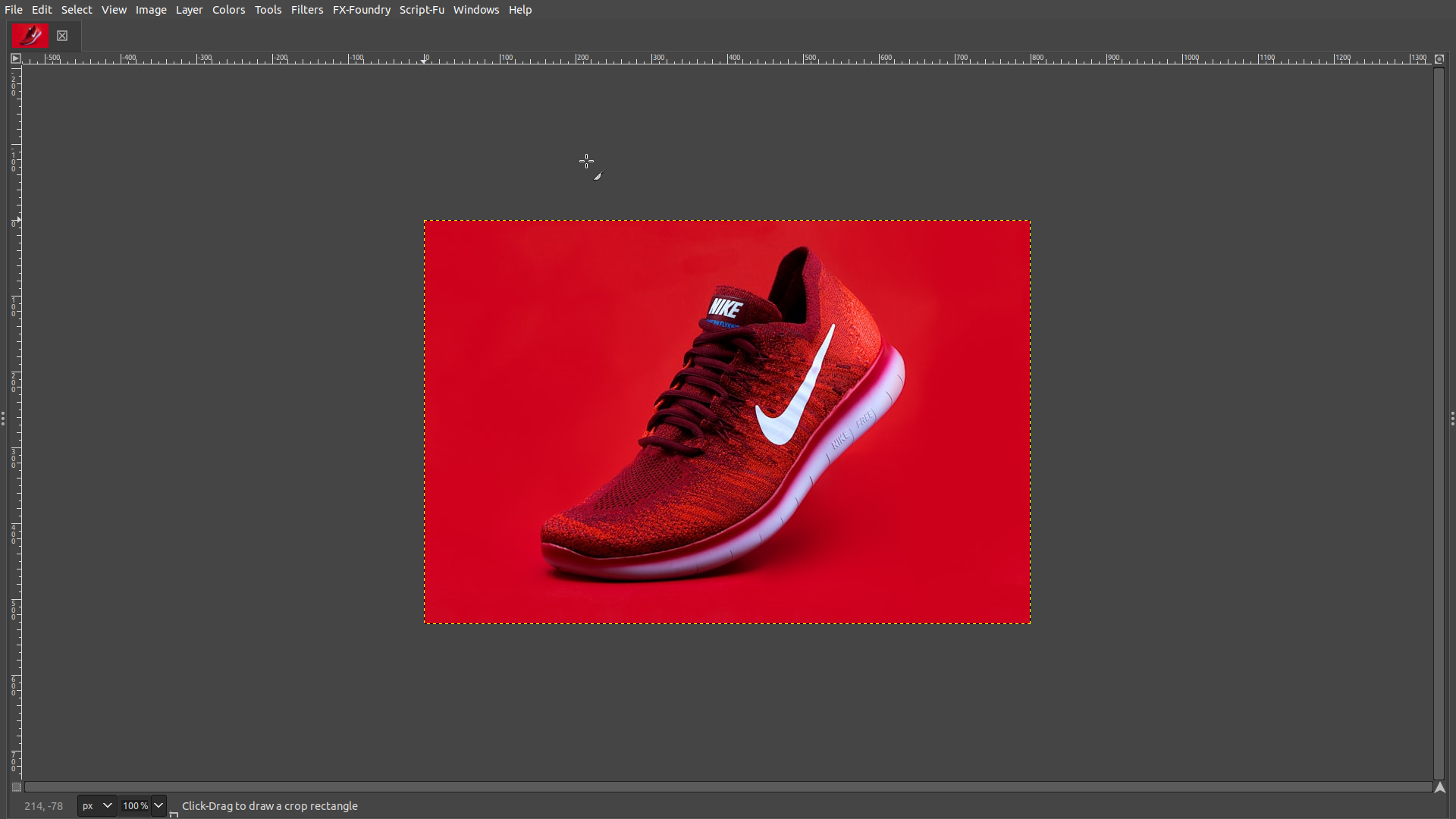}
    {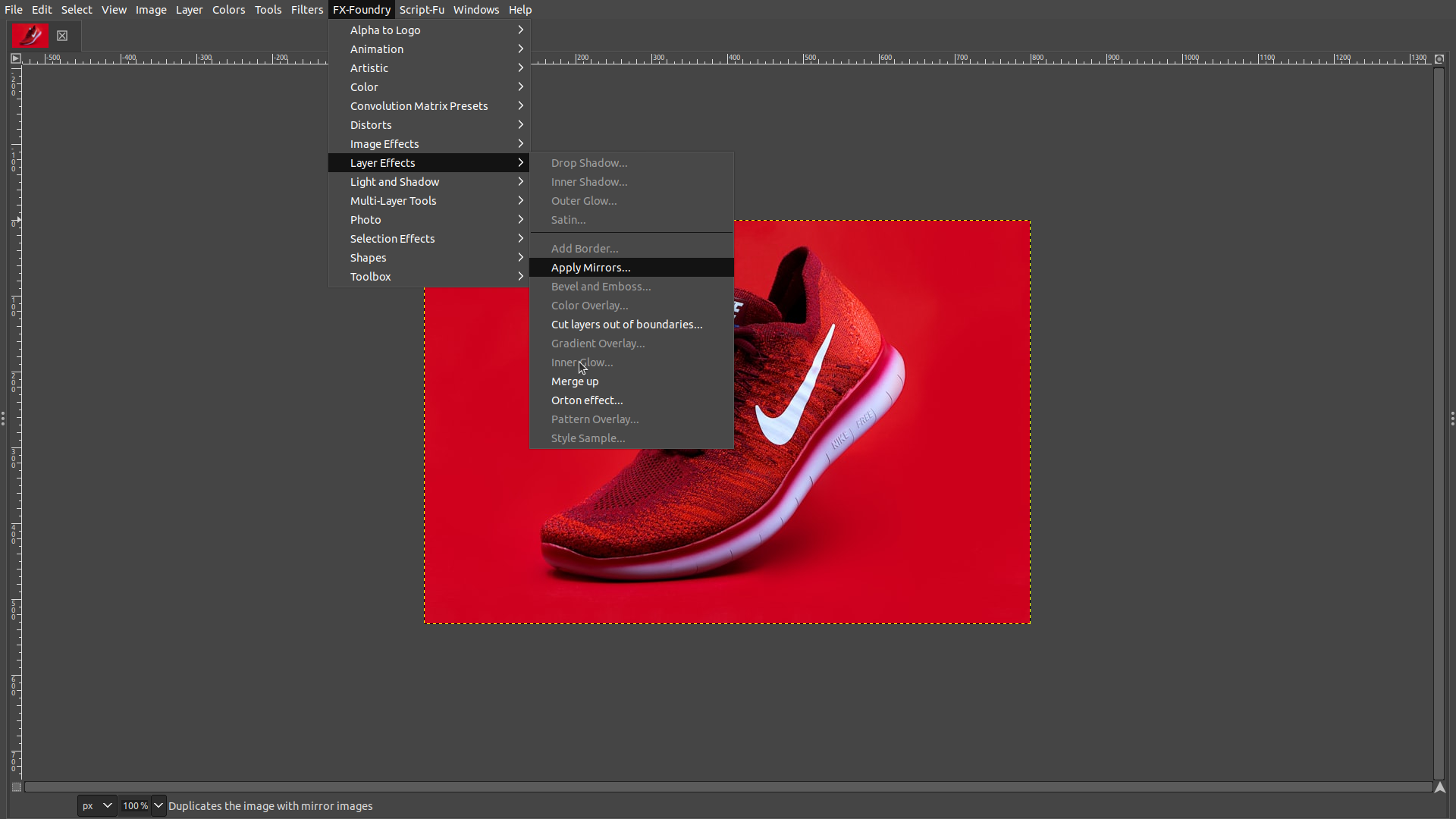}
    {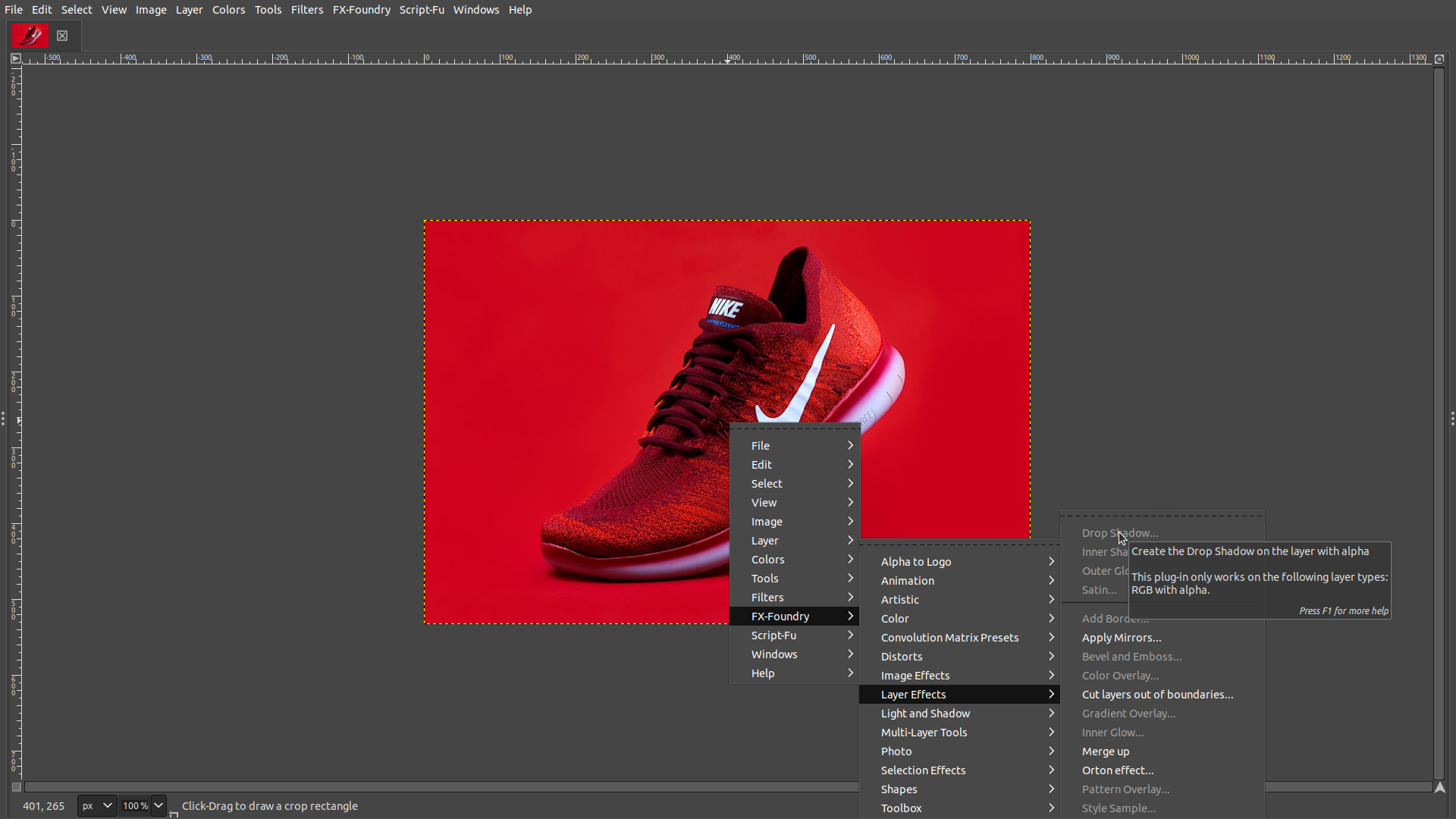}
\end{taskexample}

\medskip

\begin{taskexample}{Eclipse IDE \textnormal{--- add\_junit\_tests \hfill \textit{Difficulty: medium}}}
\small\textit{Add JUnit 5 tests to the Calculator project in Eclipse IDE. Create a test class named \texttt{CalculatorTest.java} at \texttt{src/test/java/.../CalculatorTest.java} to verify the \texttt{add()}, \texttt{subtract()}, \texttt{multiply()}, and \texttt{divide()} methods. Include a test for the division-by-zero edge case using \texttt{assertThrows(ArithmeticException.class, ...)} and ensure all tests pass.}
\taskscreenshots
    {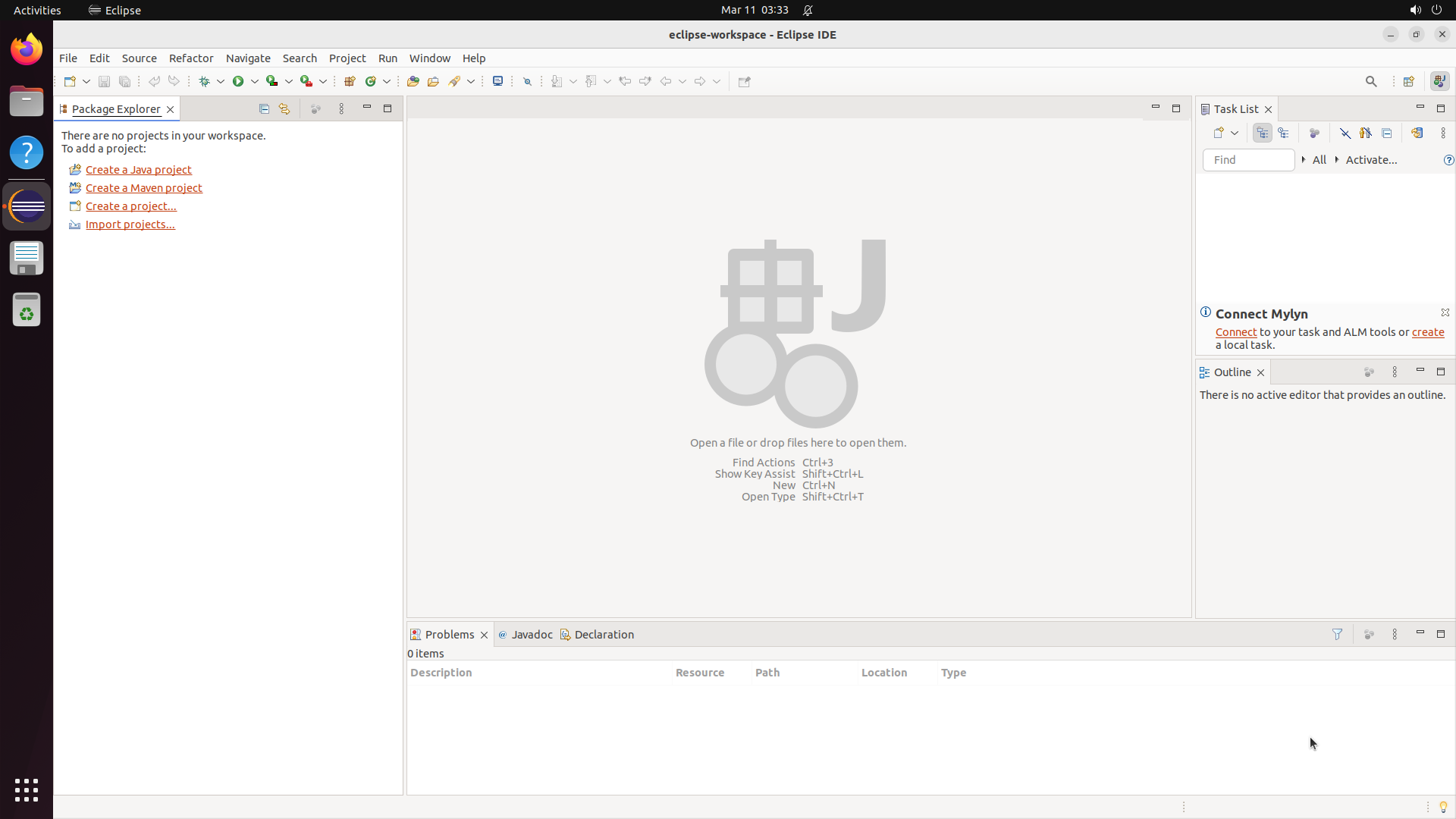}
    {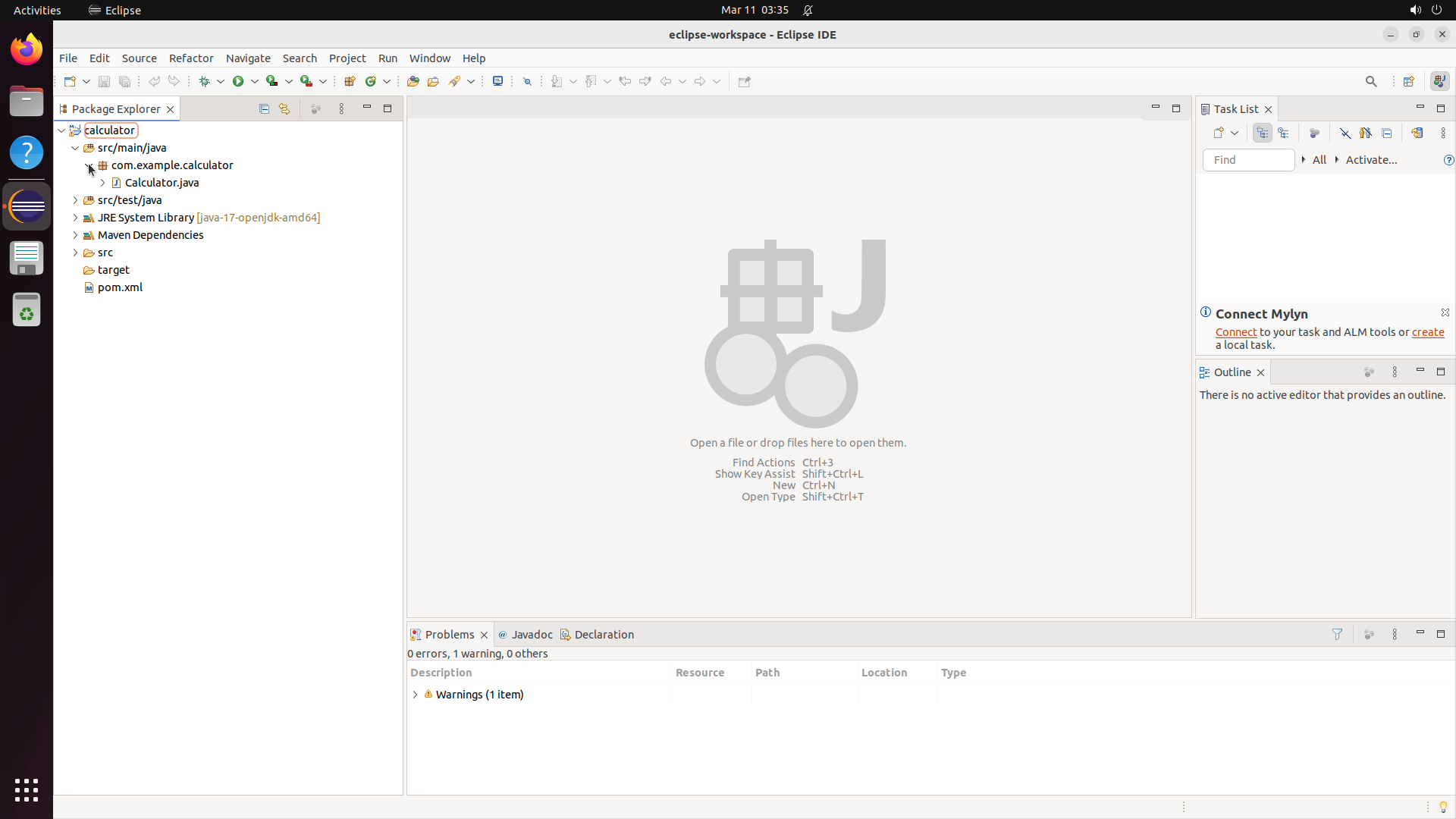}
    {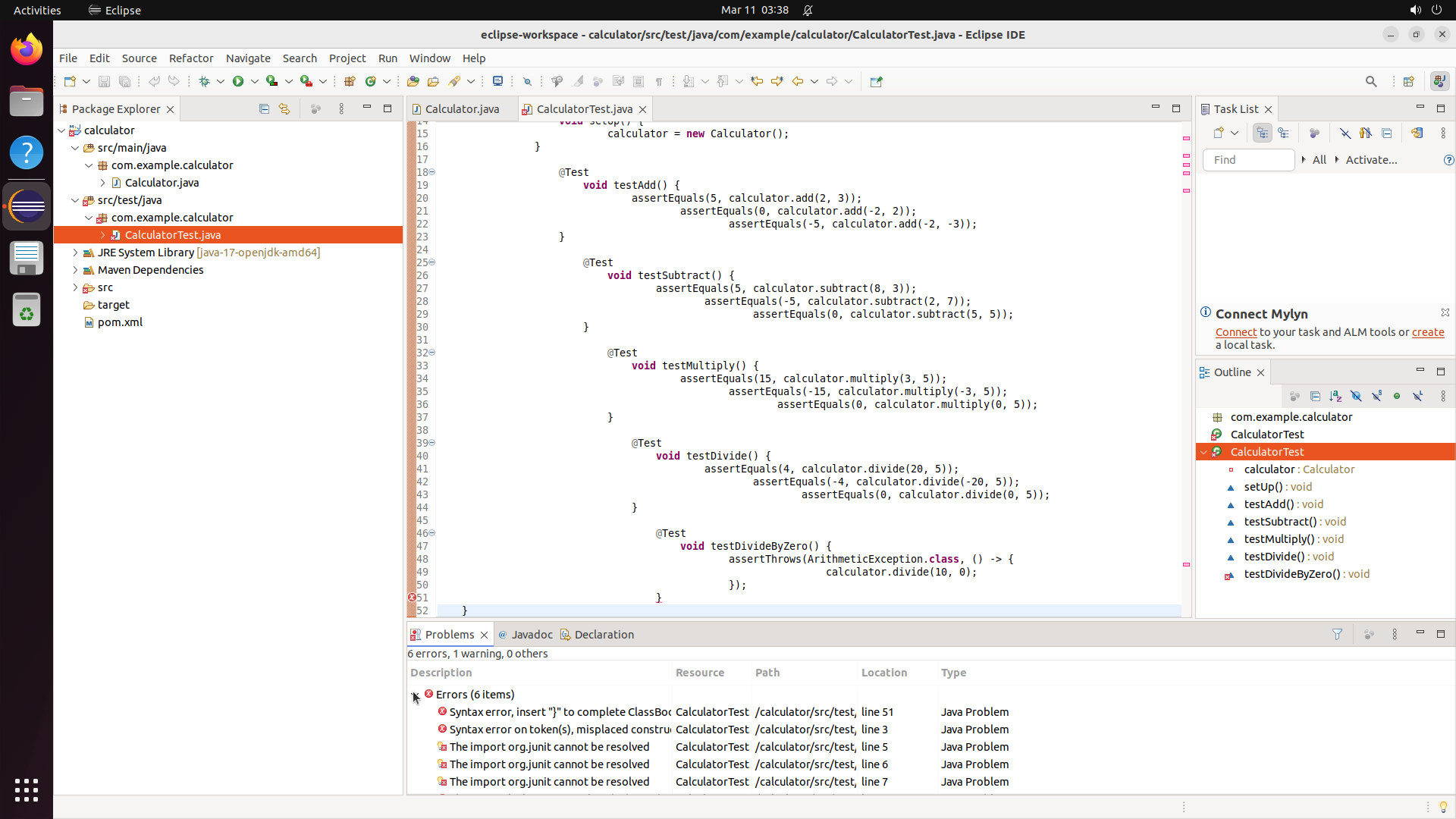}
    {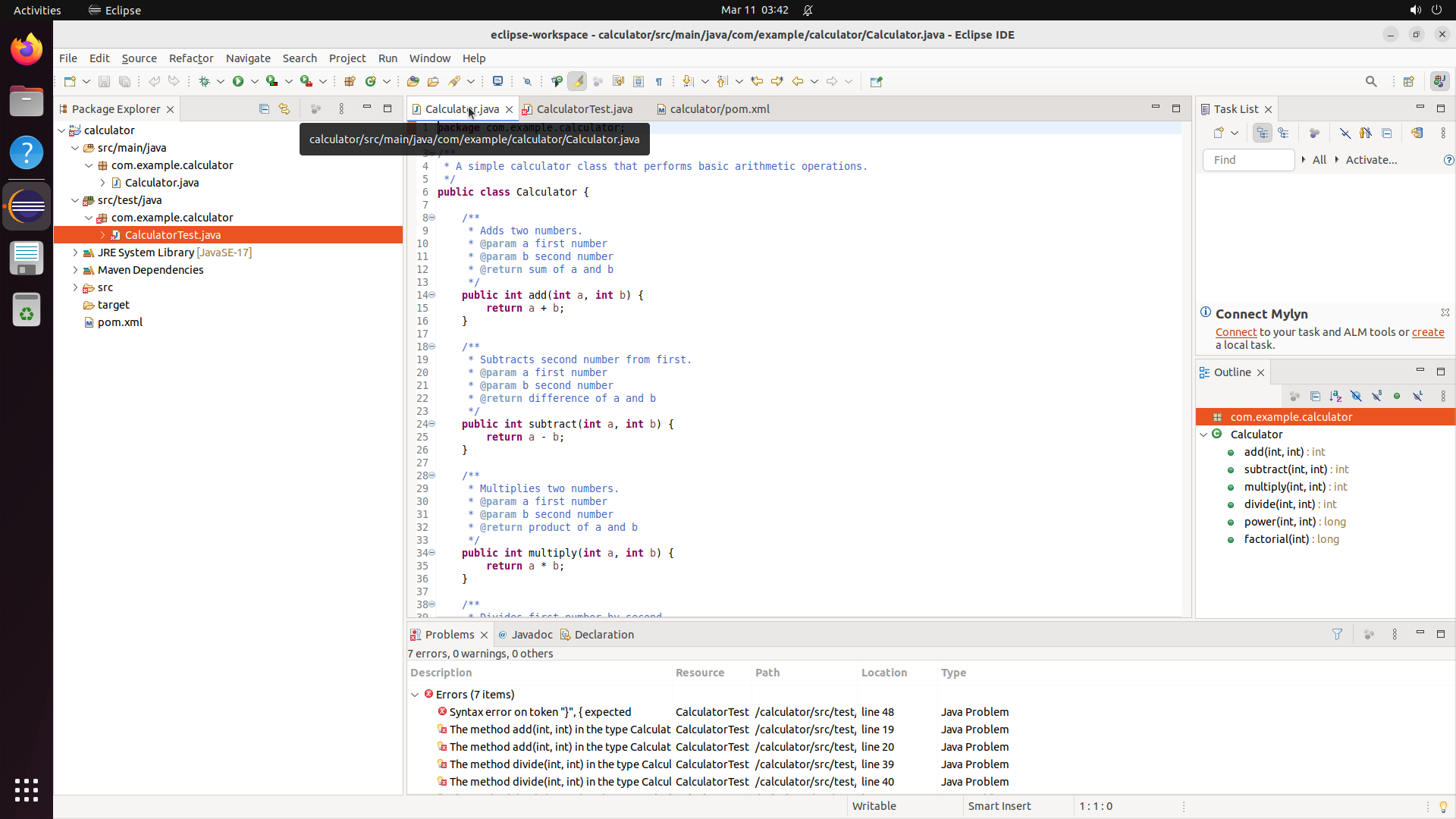}
\end{taskexample}

\medskip

\begin{taskexample}{Ardour 6 \textnormal{--- adr\_session\_prep \hfill \textit{Difficulty: hard}}}
\small\textit{Prepare an Automated Dialogue Replacement (ADR) recording session. Create three audio tracks named `Guide Track', `ADR Record', and `Beep Track'. Import narration audio onto the Guide Track at 0:00 and mute it. Record-arm the ADR Record track. Create a countdown cue on the Beep Track with exactly three distinct audio regions. Set the loop range from 5.0\,s to 15.0\,s. Save the session.}
\taskscreenshots
    {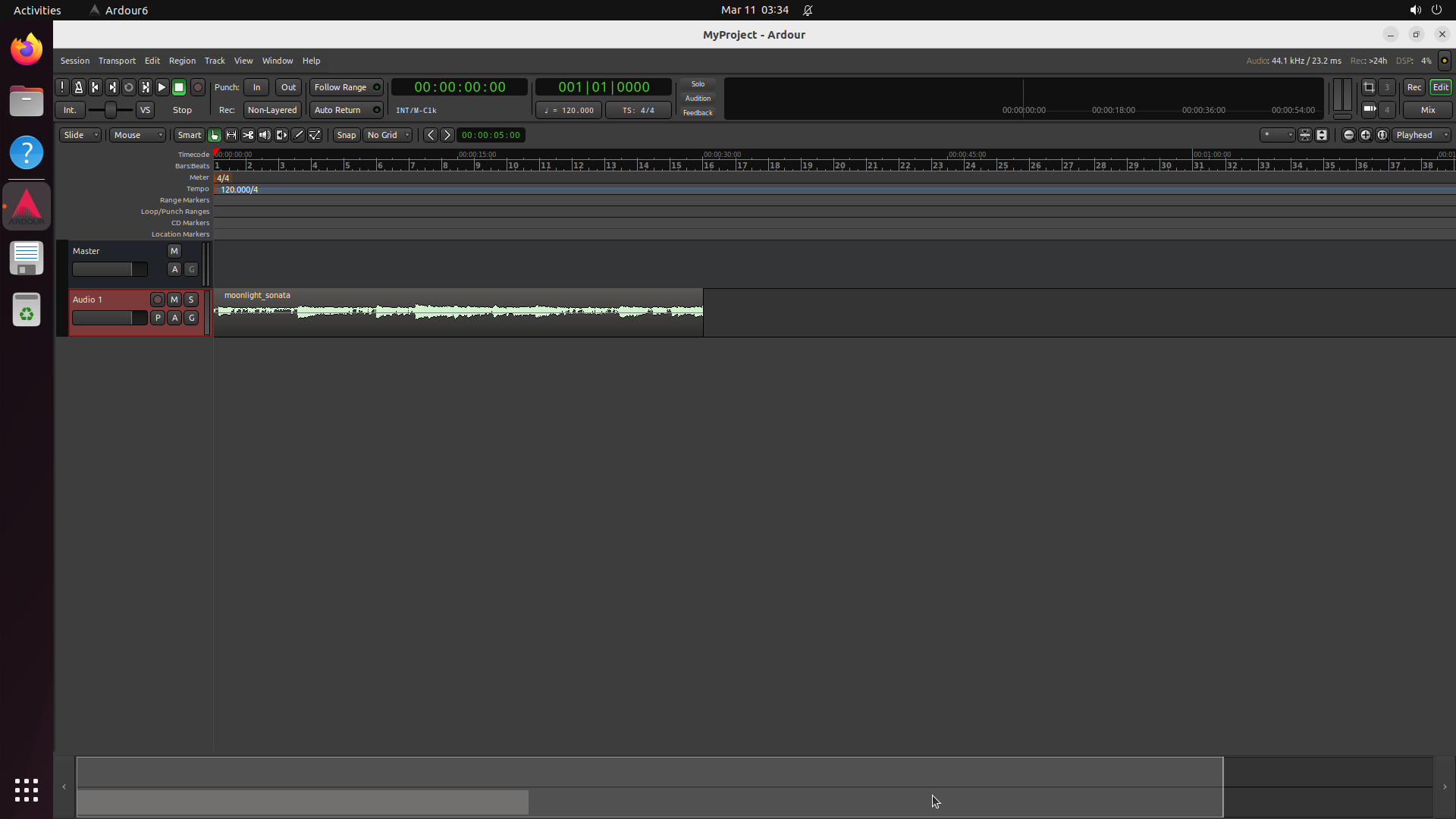}
    {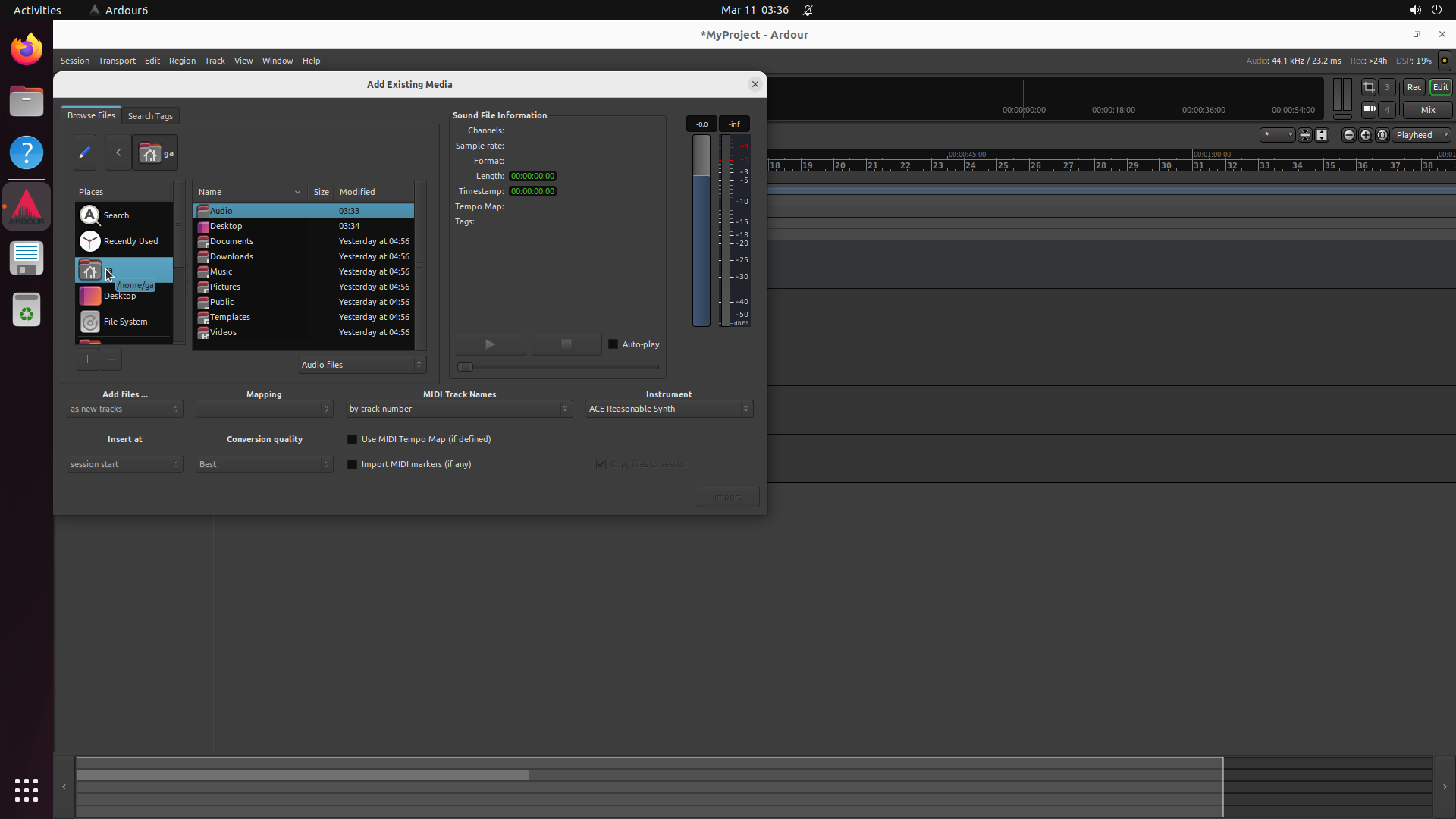}
    {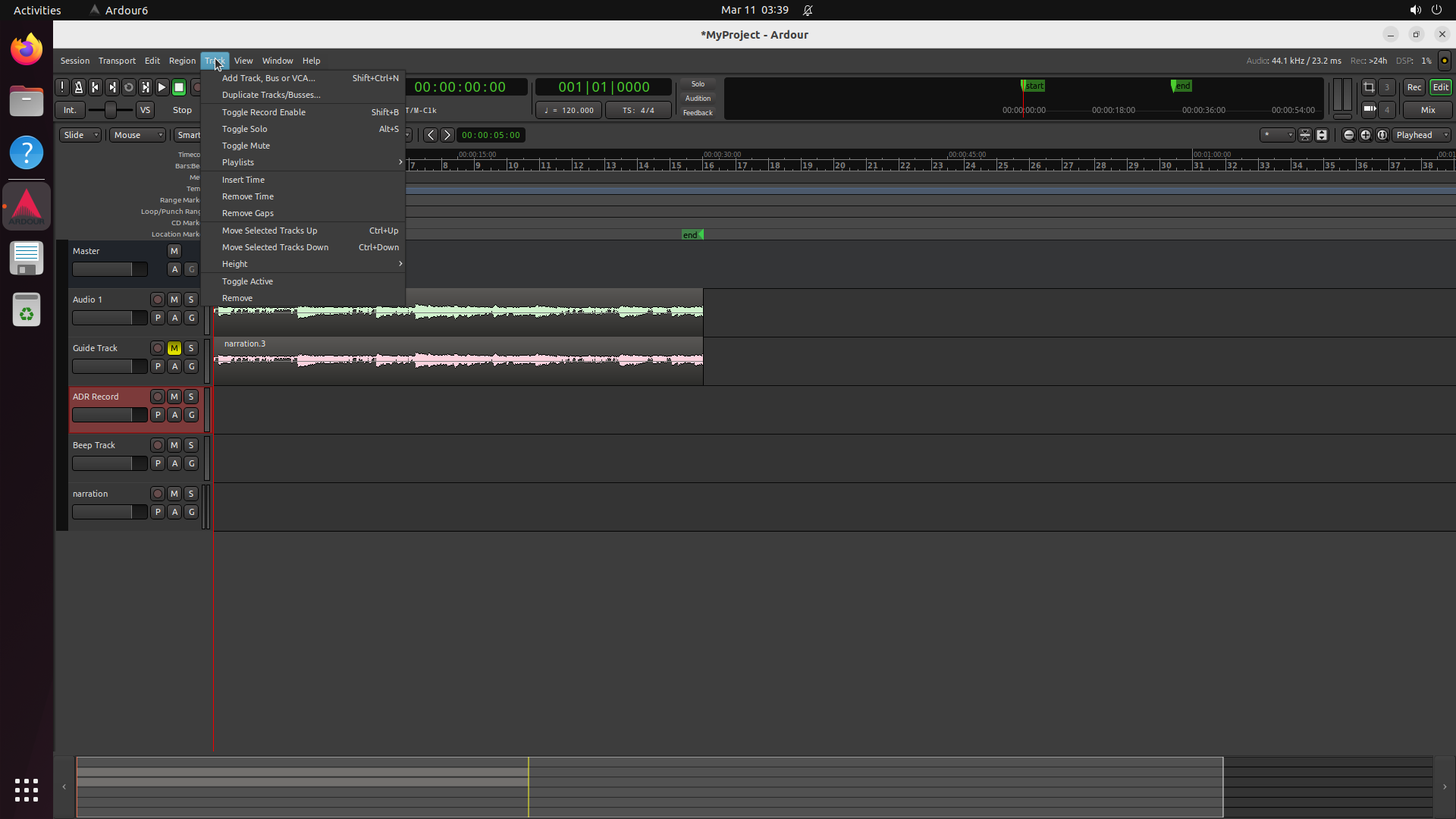}
    {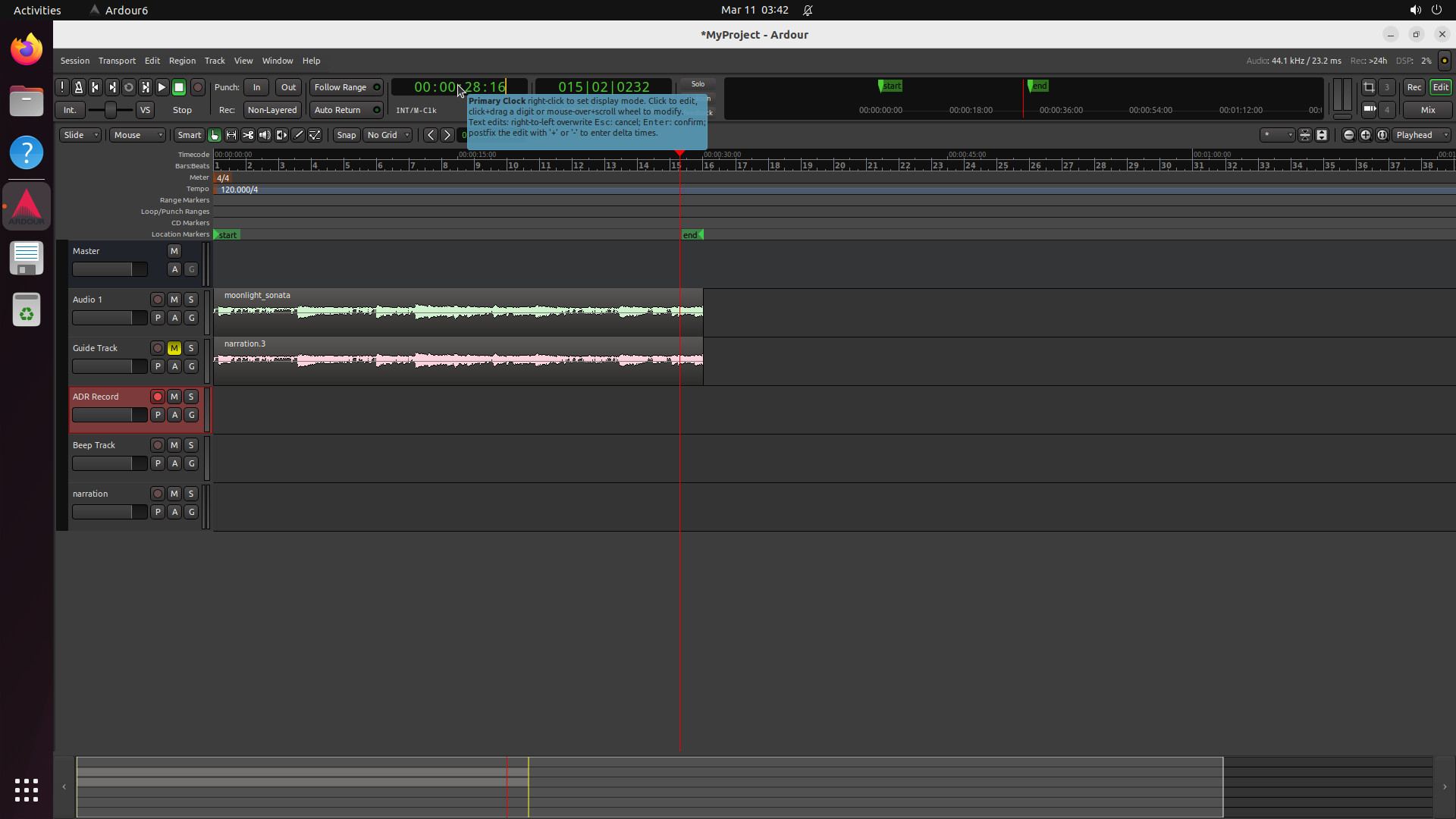}
\end{taskexample}

\medskip

\begin{taskexample}{GeoGebra \textnormal{--- archimedes\_pi\_polygon\_exhaustion \hfill \textit{Difficulty: hard}}}
\small\textit{Create an interactive GeoGebra applet demonstrating Archimedes' method of exhaustion for approximating $\pi$. Construct a unit circle centered at the origin with an integer slider $n$ (3--48) controlling the number of sides for both a regular inscribed $n$-gon and a circumscribed $n$-gon. Display the numerical lower bound (inscribed polygon's half-perimeter) and upper bound (circumscribed polygon's half-perimeter) for $\pi$. Save as \texttt{archimedes\_pi.ggb}.}
\taskscreenshots
    {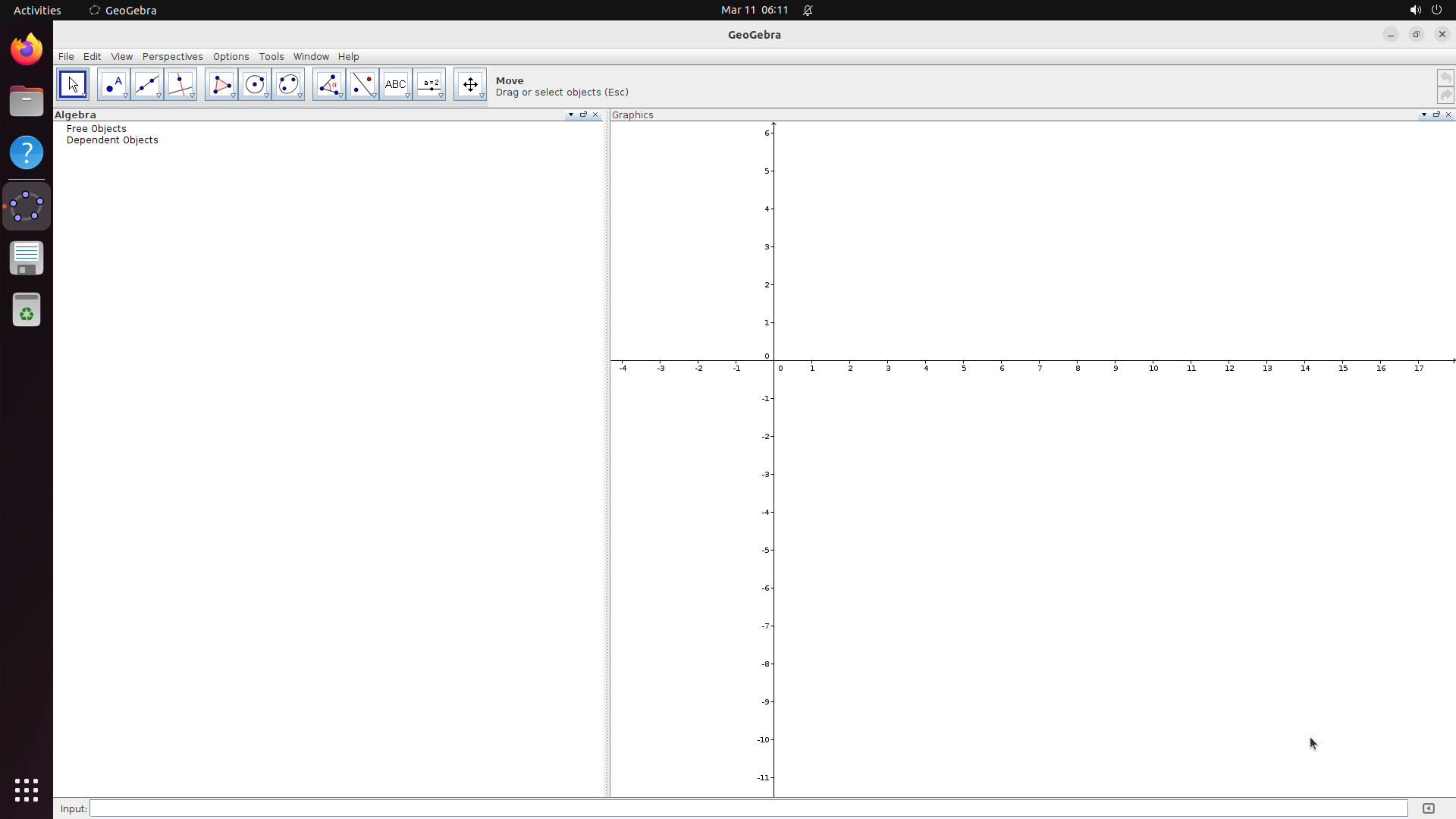}
    {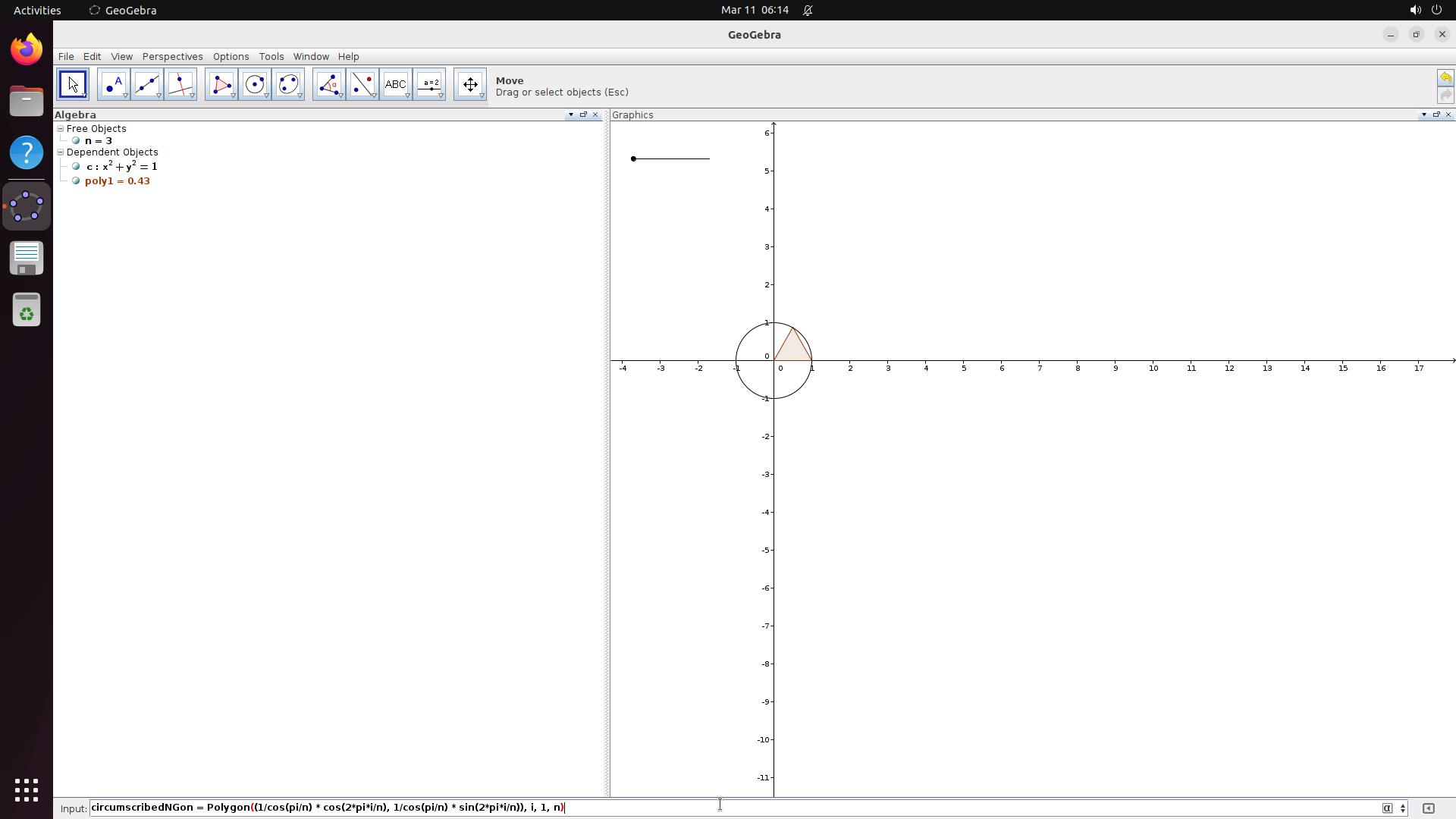}
    {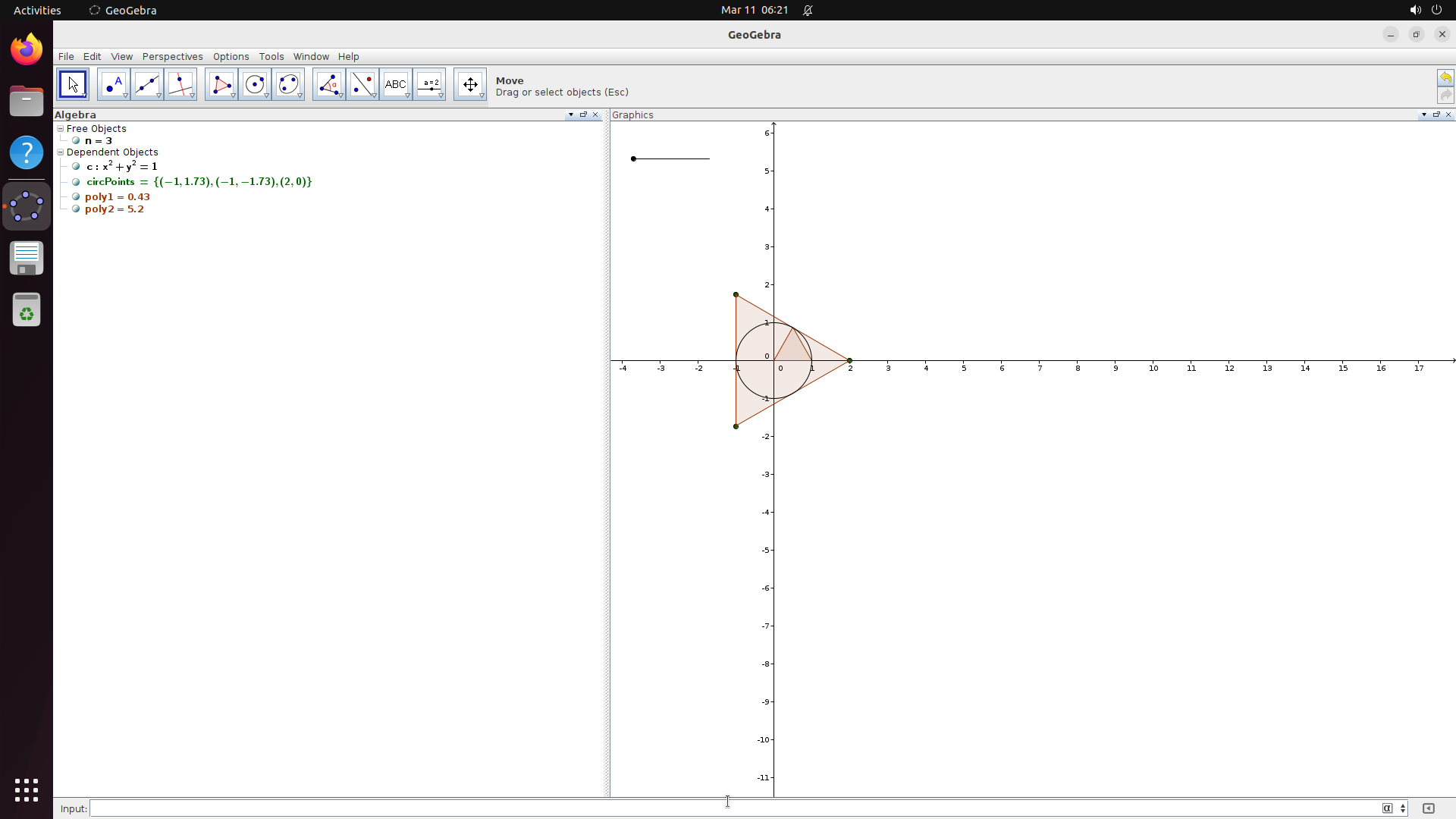}
    {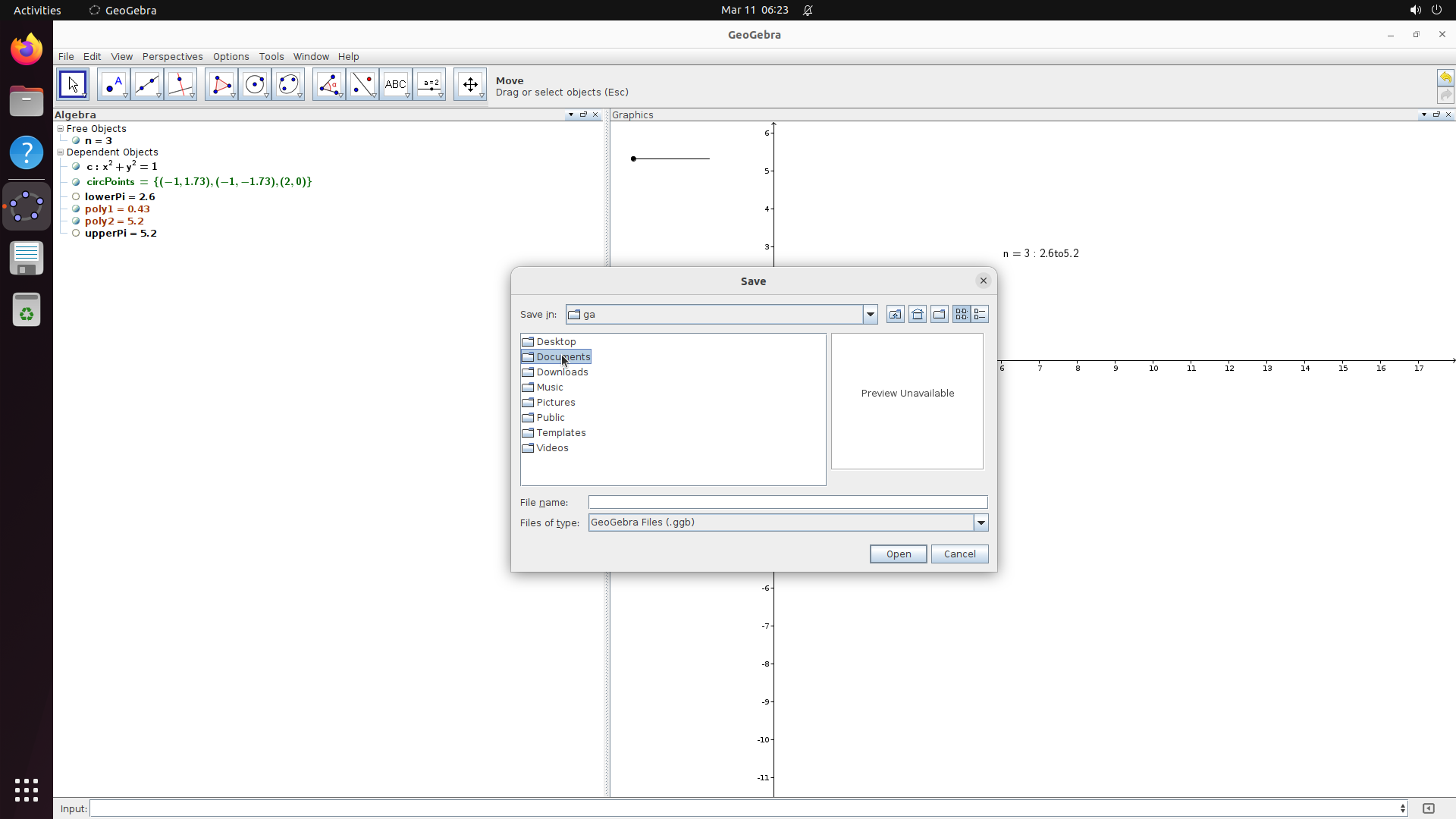}
\end{taskexample}

\medskip

\begin{taskexample}{Autopsy (Digital Forensics) \textnormal{--- chain\_of\_custody\_tamper\_audit \hfill \textit{Difficulty: hard}}}
\small\textit{Internal Affairs suspects a rogue detective accessed a sealed evidence locker and modified a digital device after seizure. Validate the cryptographic chain of custody for three seized USB drives (\texttt{seized\_usb\_1.dd}--\texttt{3.dd}) by comparing their current SHA-256 hashes against the official acquisition log. Identify the tampered drive and any files modified or created after the seizure date (2023-05-01). Create an Autopsy case ``Evidence\_Audit\_2024'' and write a forensic audit report containing the tampered image, original vs.\ current hashes, planted files, and a conclusion.}
\taskscreenshots
    {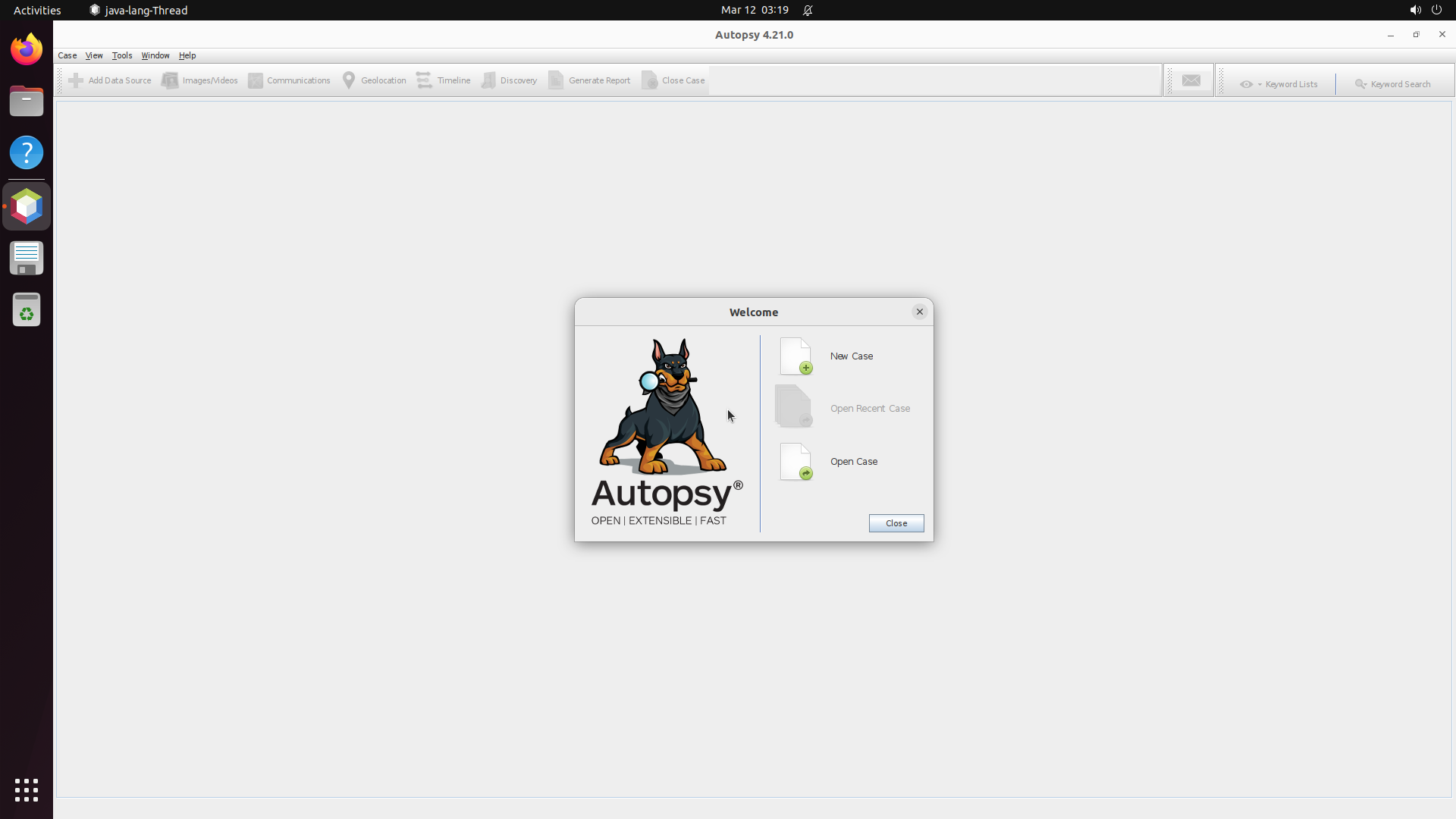}
    {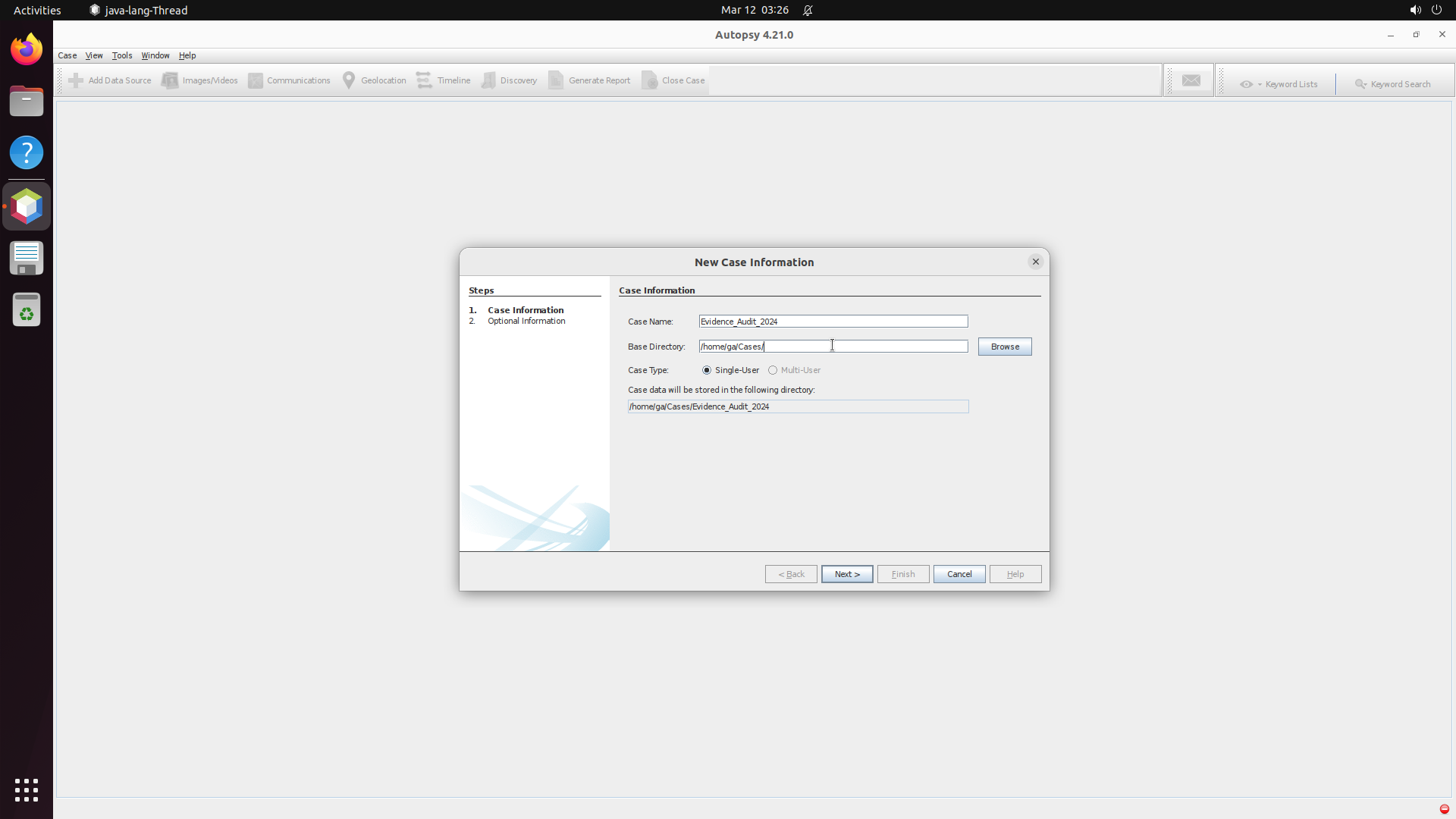}
    {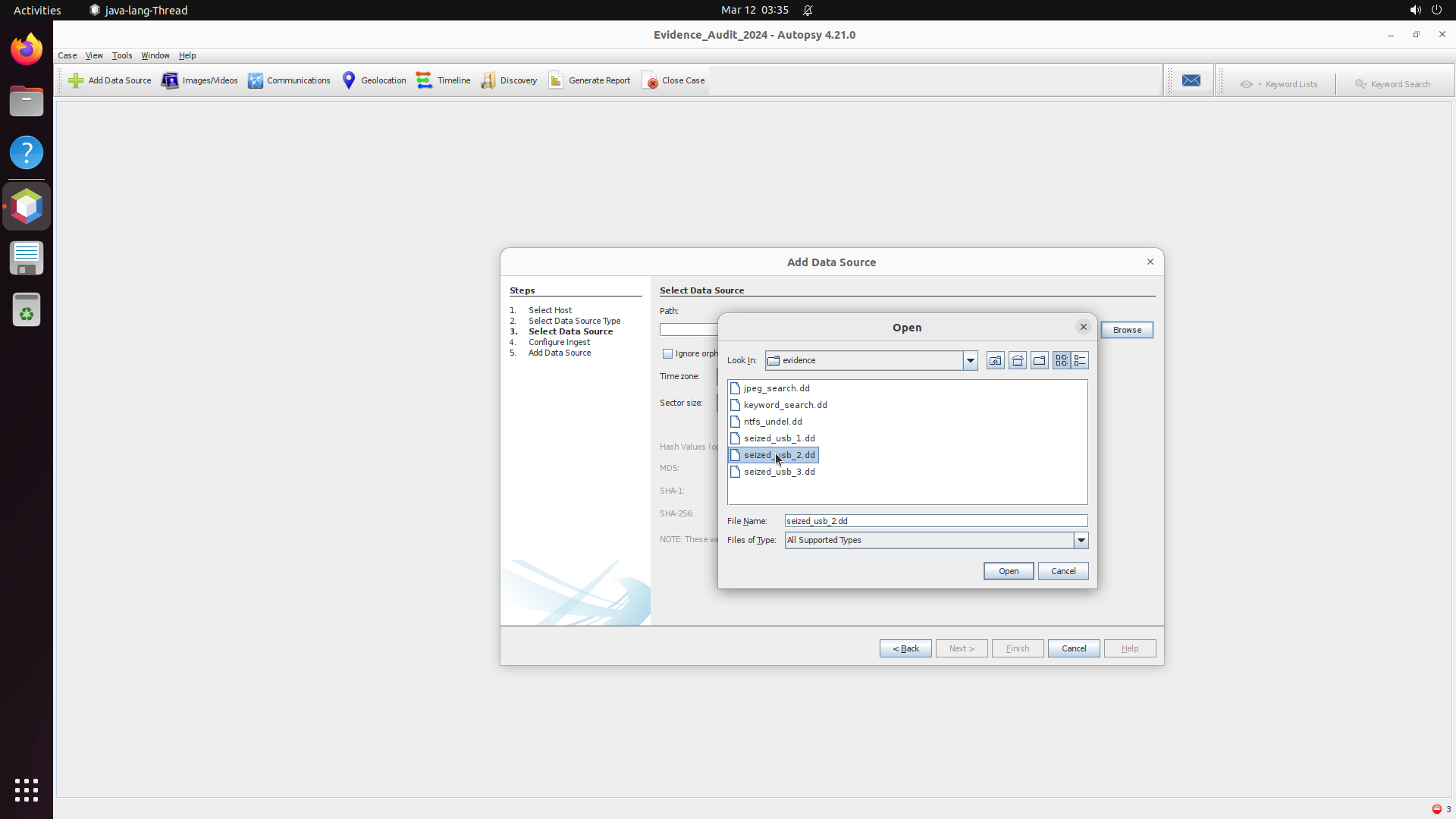}
    {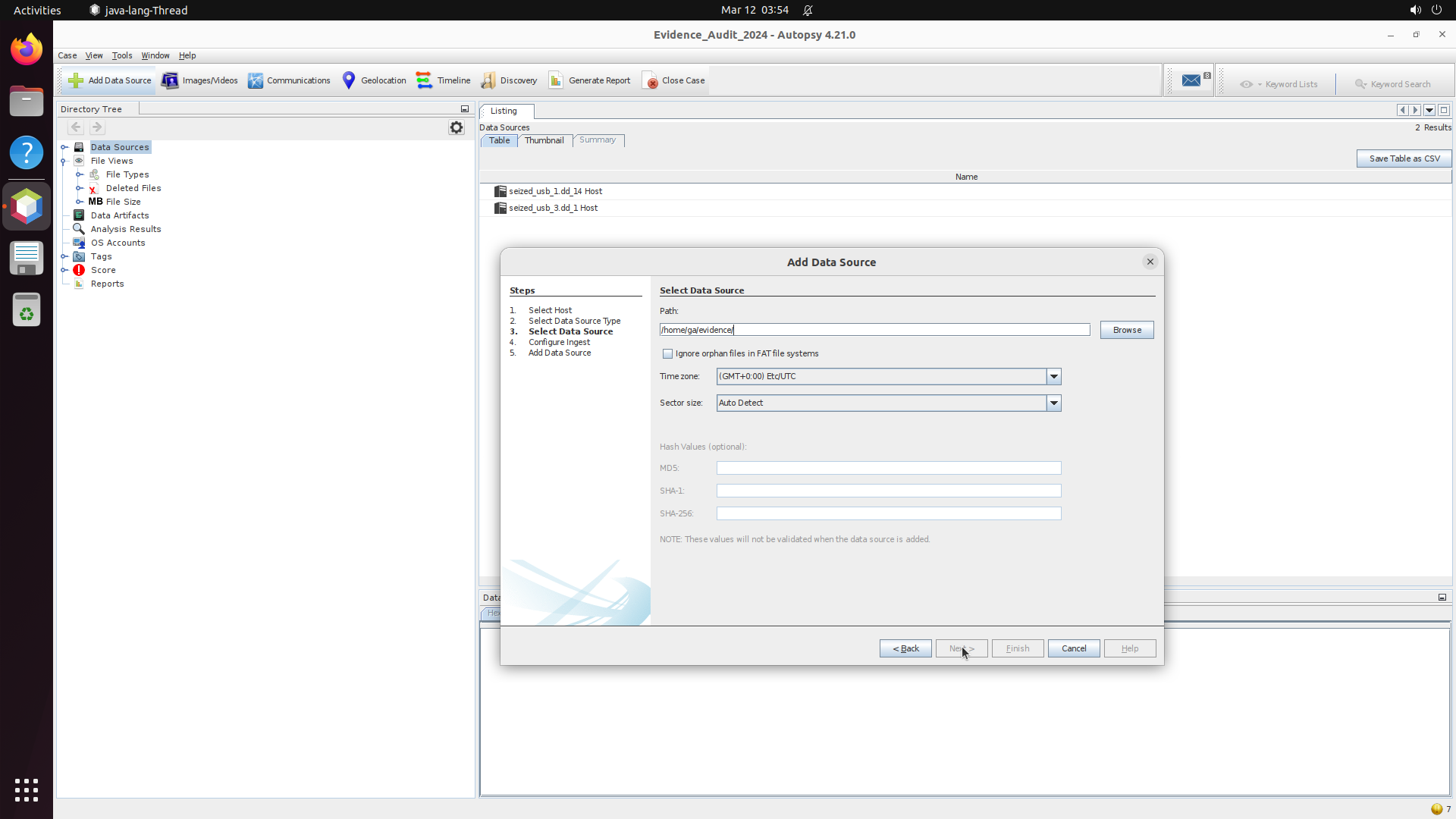}
\end{taskexample}

\medskip

\begin{taskexample}{DBeaver (Database Administration) \textnormal{--- chinook\_acquisition\_merger \hfill \textit{Difficulty: hard}}}
\small\textit{The company has acquired `QuickTunes' and their lead data needs to be merged into the main database. Merge leads from the acquisitions database into the Chinook \texttt{customers} table: split \texttt{full\_name} into first/last, map ISO country codes (\texttt{US}$\to$\texttt{USA}, \texttt{CA}$\to$\texttt{Canada}, \texttt{MX}$\to$\texttt{Mexico}), set \texttt{SupportRepId}=3, and prevent importing any lead whose email already exists. Export the newly added customers to CSV and save the migration SQL script.}
\taskscreenshots
    {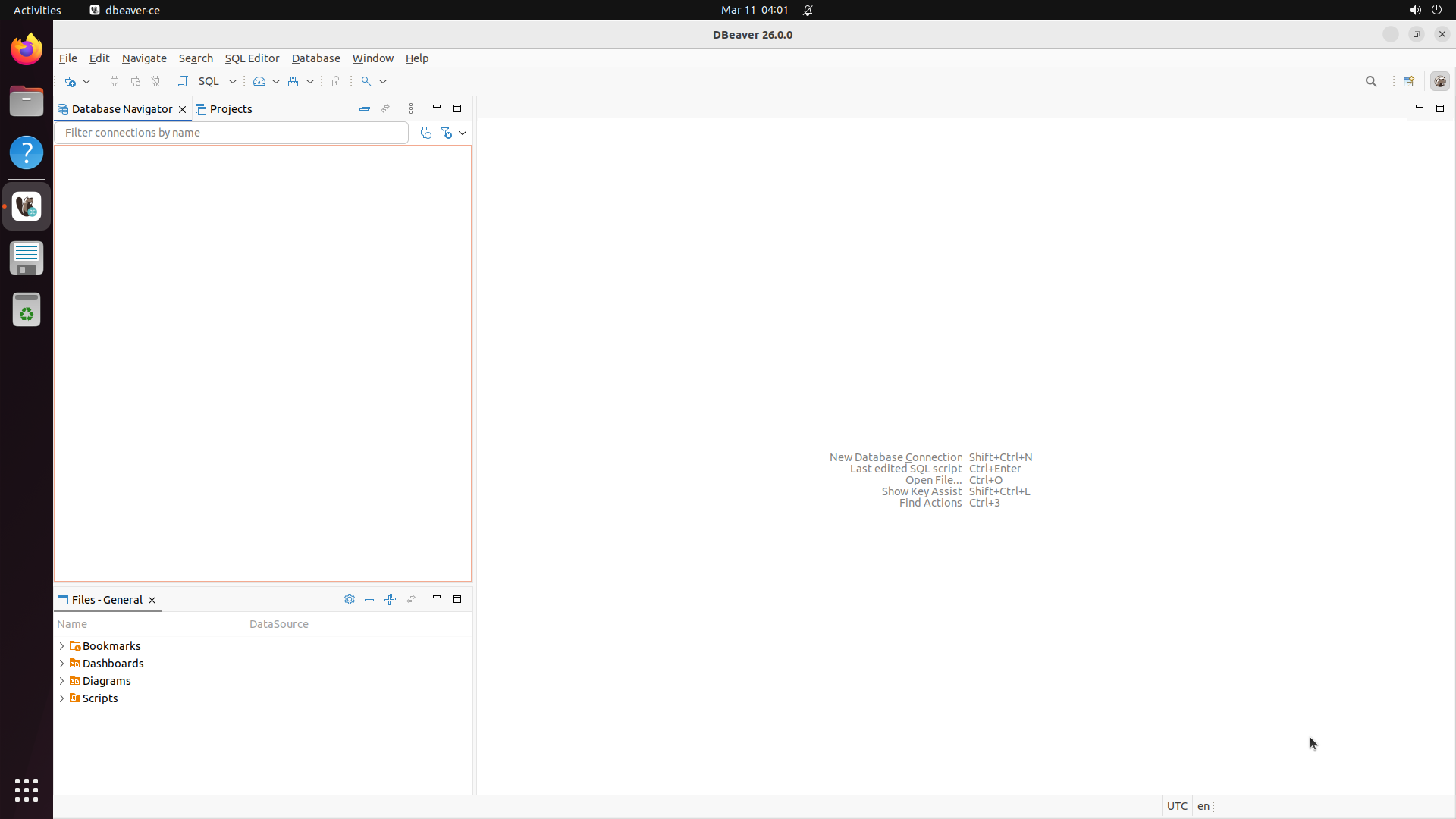}
    {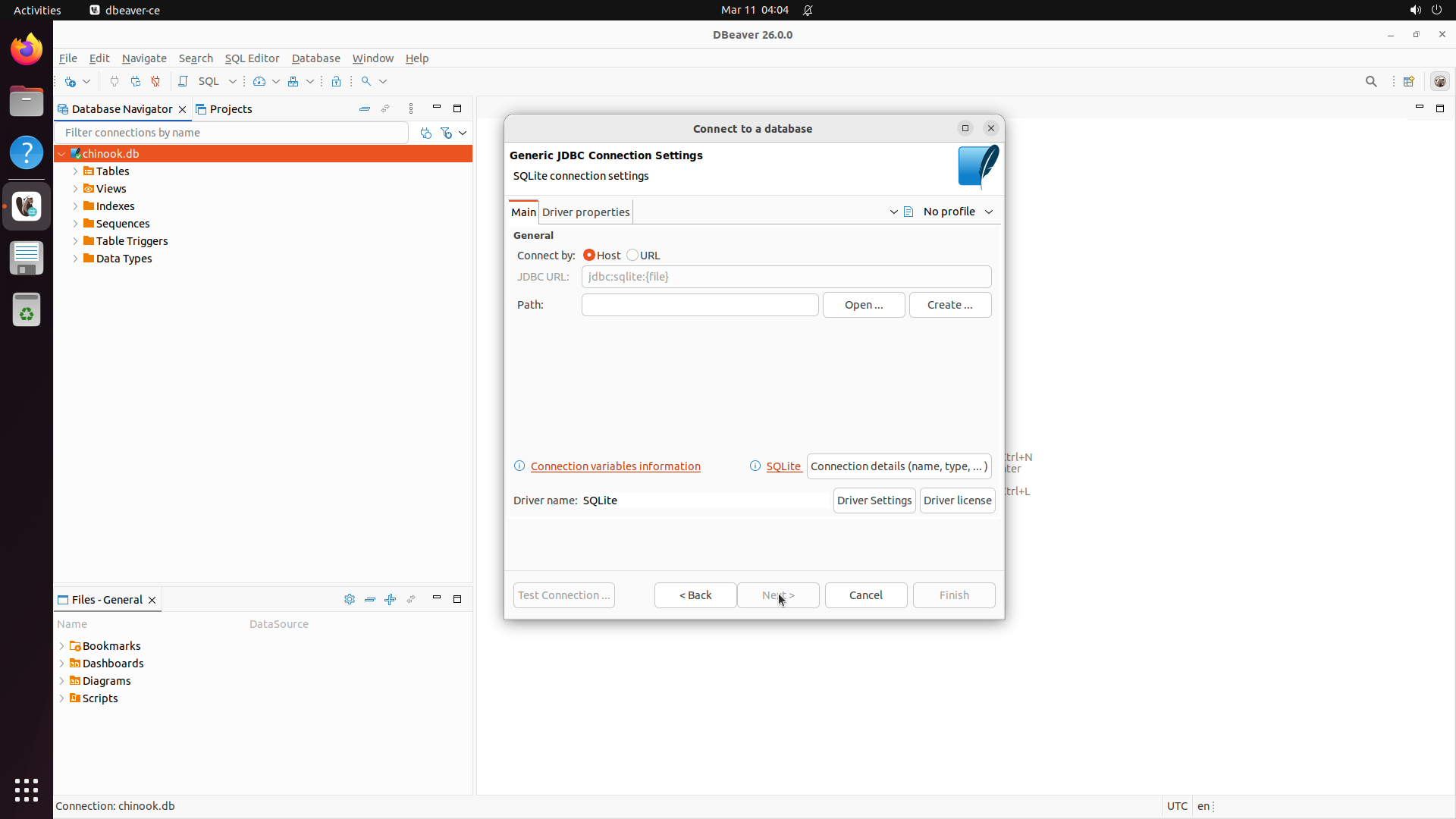}
    {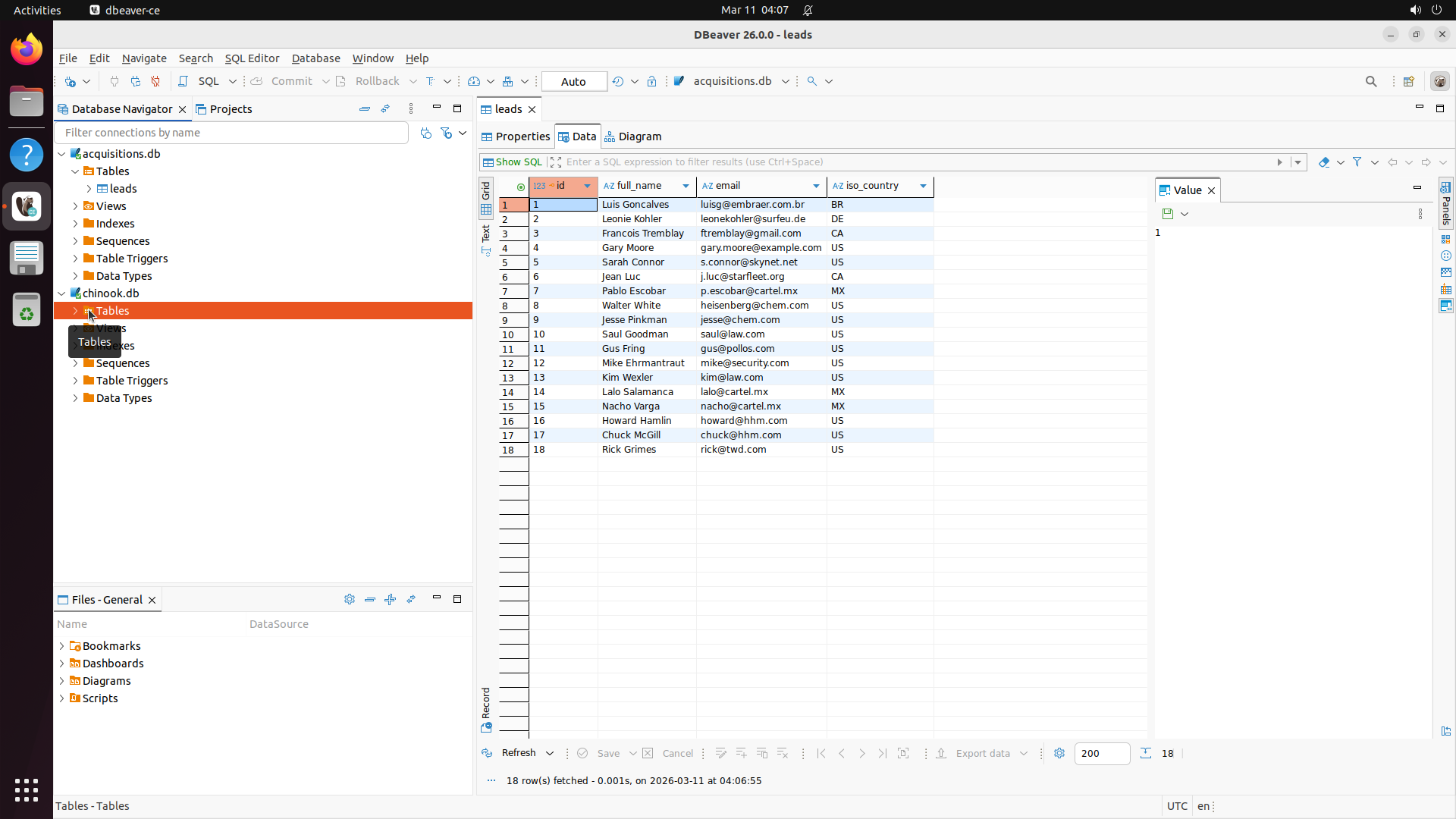}
    {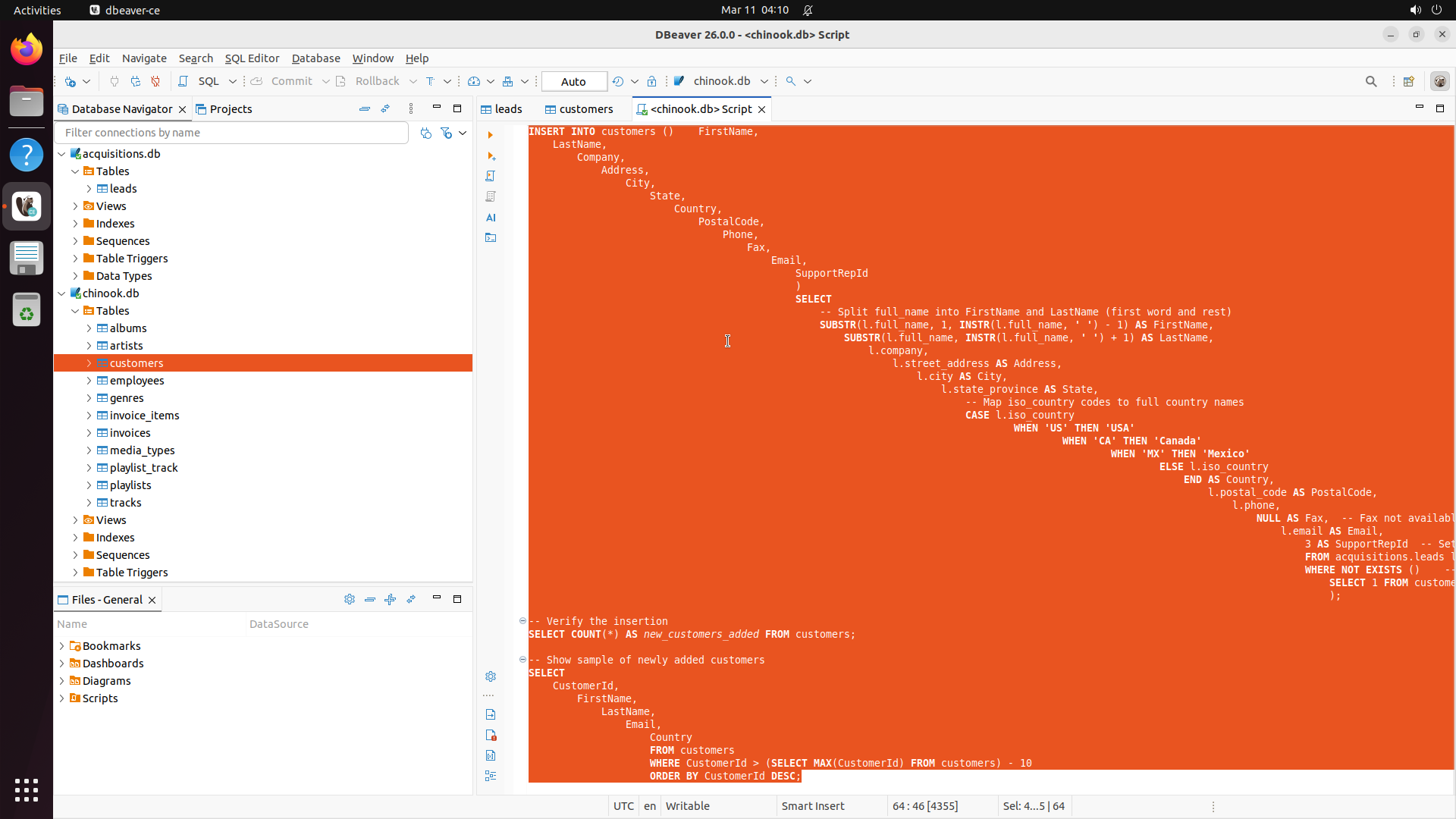}
\end{taskexample}

\medskip

\begin{taskexample}{QGIS (Geospatial Analysis) \textnormal{--- analyze\_crop\_health\_ndvi\_zonal\_stats \hfill \textit{Difficulty: medium}}}
\small\textit{Calculate the Normalized Difference Vegetation Index (NDVI) for a farm and determine the average health of each crop field. Use multispectral imagery (Band~1: Red, Band~2: NIR) and field boundary vectors. Apply the NDVI formula $(\mathit{NIR} - \mathit{Red})/(\mathit{NIR} + \mathit{Red})$, save the NDVI raster, then compute zonal statistics (mean NDVI per field) and export a GeoJSON with per-field yield potential estimates.}
\taskscreenshots
    {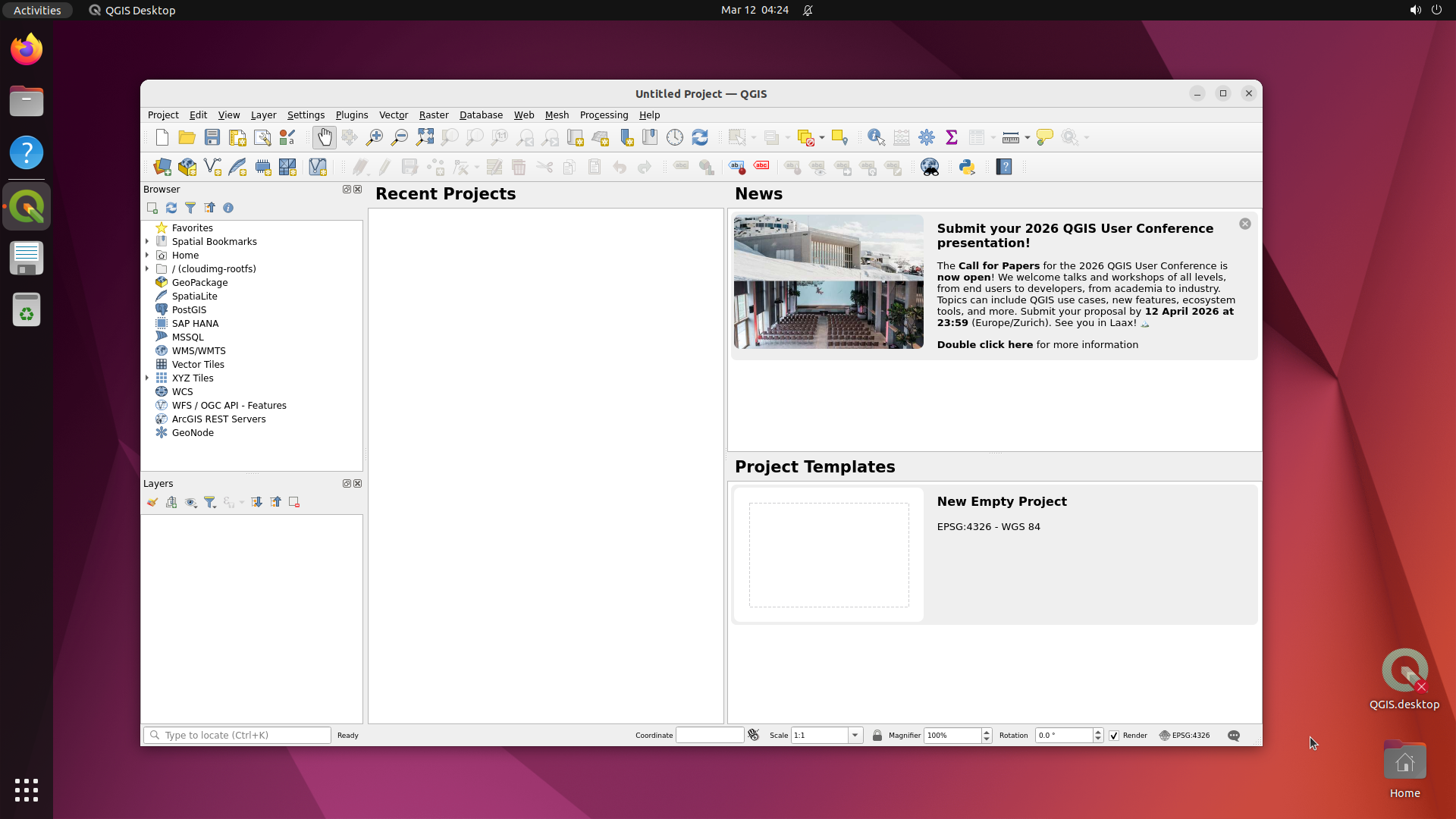}
    {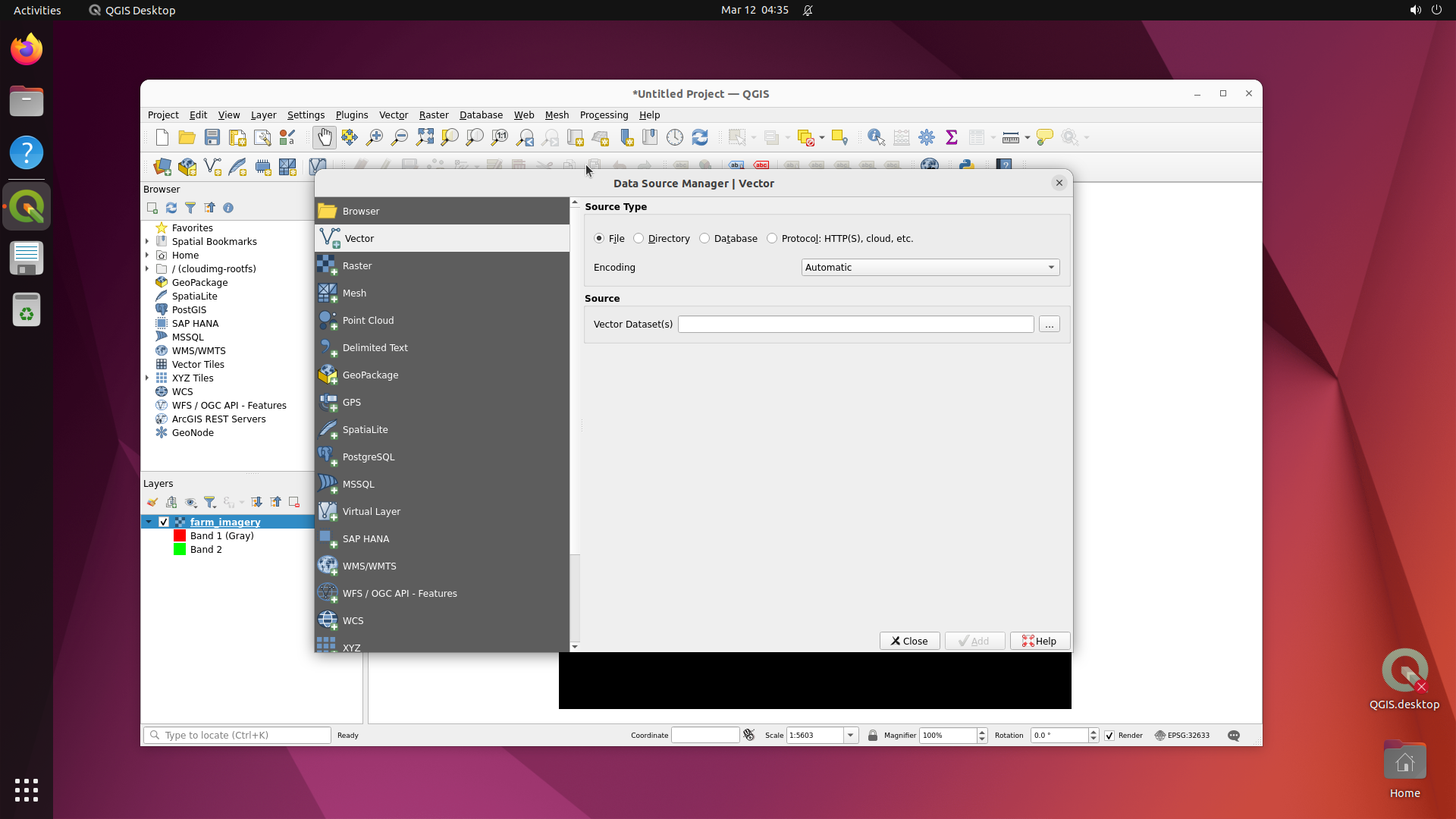}
    {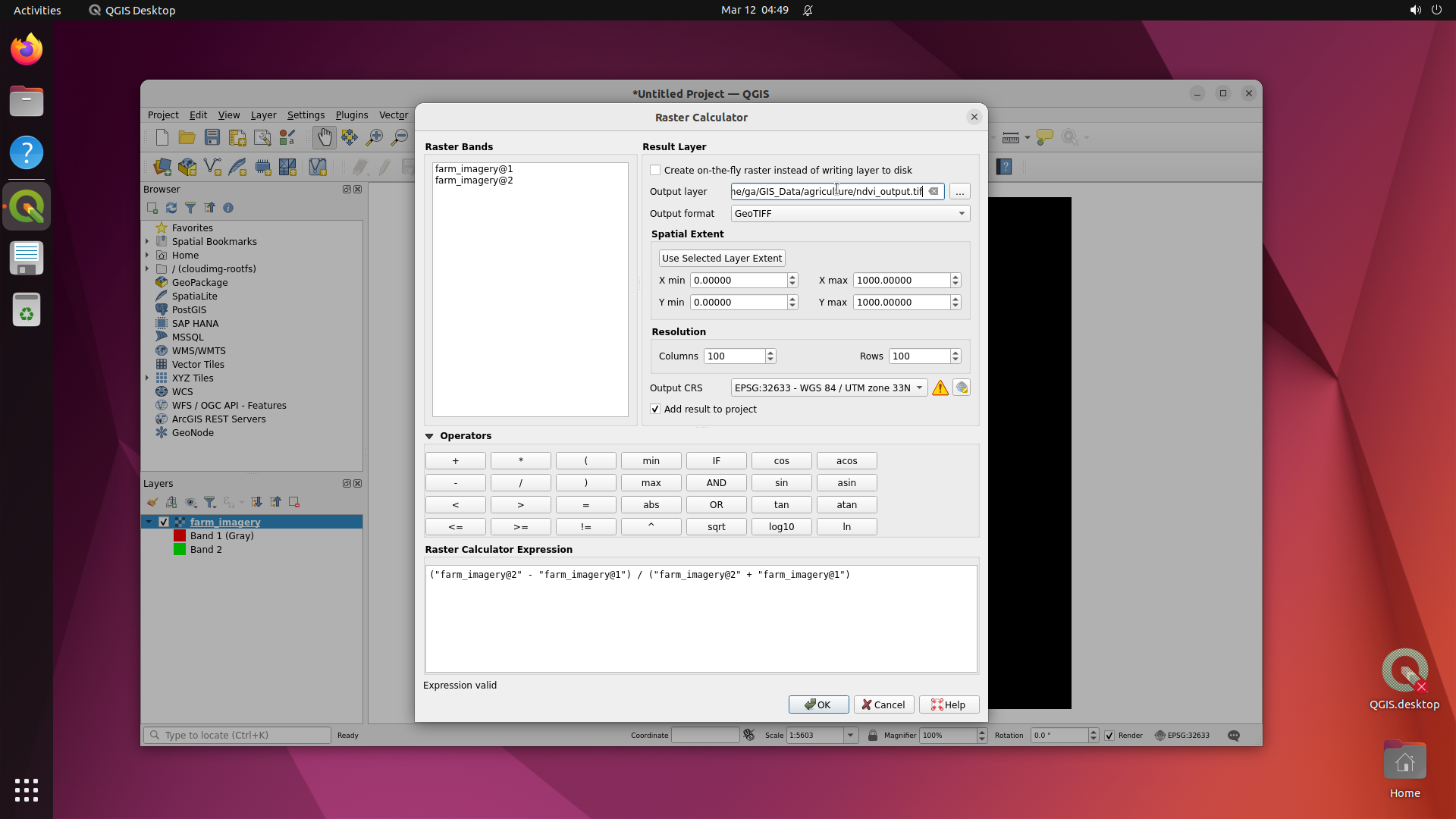}
    {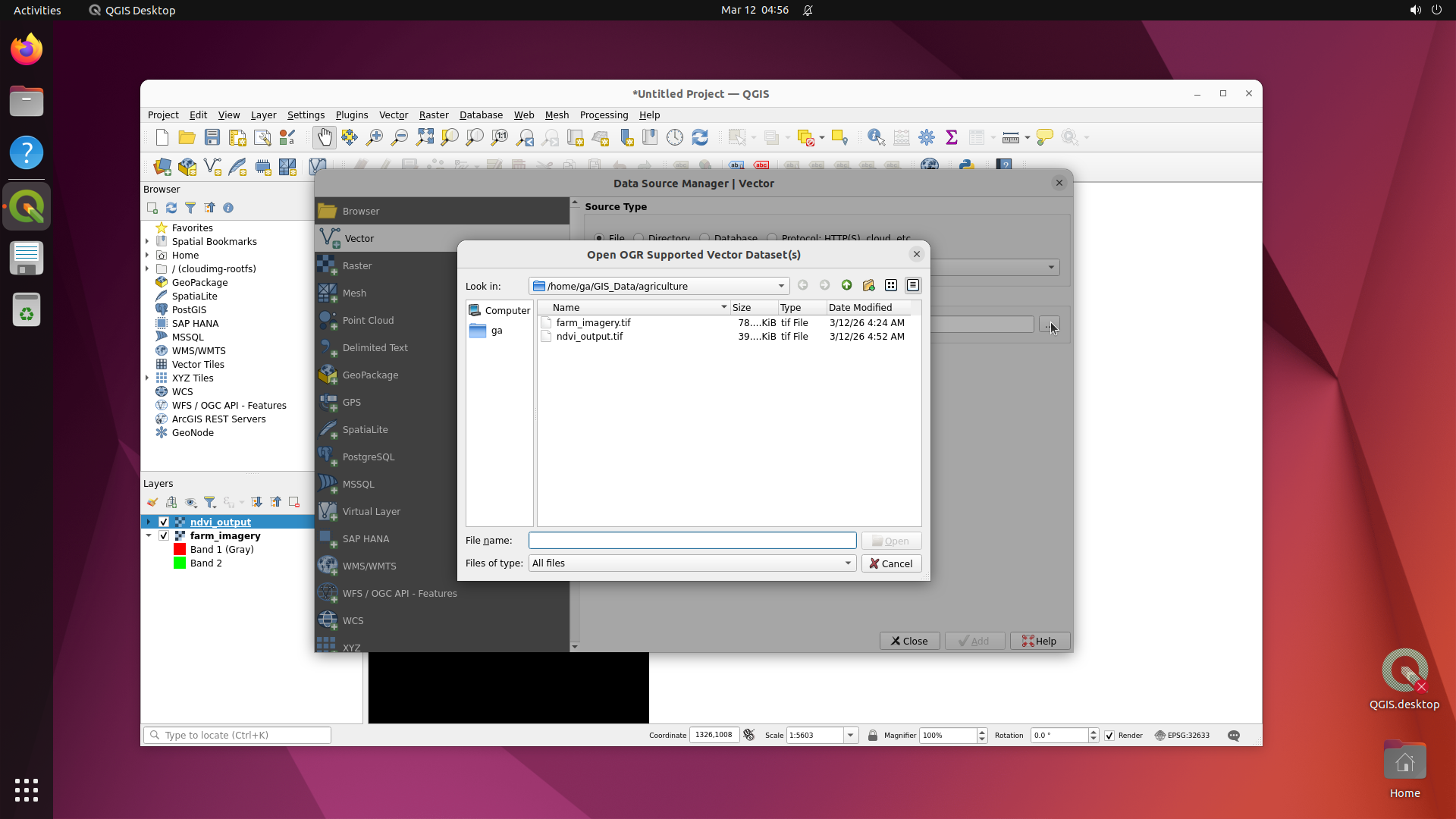}
\end{taskexample}

\medskip

\begin{taskexample}{OpenLCA (Life Cycle Assessment) \textnormal{--- ccs\_process\_retrofit \hfill \textit{Difficulty: medium}}}
\small\textit{Create a Carbon Capture and Storage (CCS) variant of a natural gas electricity generation process using the USLCI database. Copy an existing process, rename it to include `CCS', reduce the CO\textsubscript{2} fossil emission output to 10\% of its original value, add a process description note regarding the retrofit, and save a report listing the original process name, new process name, and CO\textsubscript{2} emission values before and after modification.}
\taskscreenshots
    {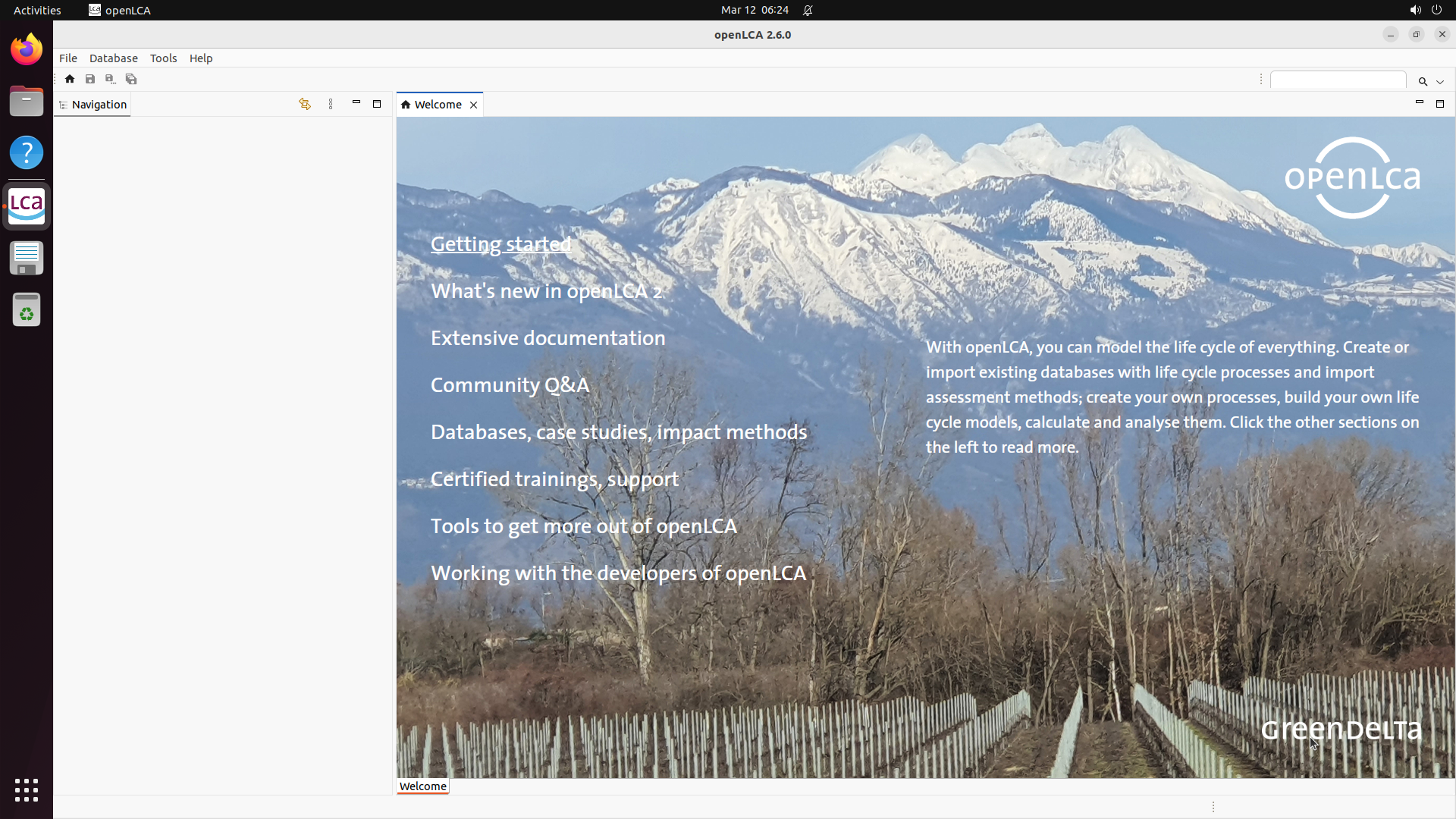}
    {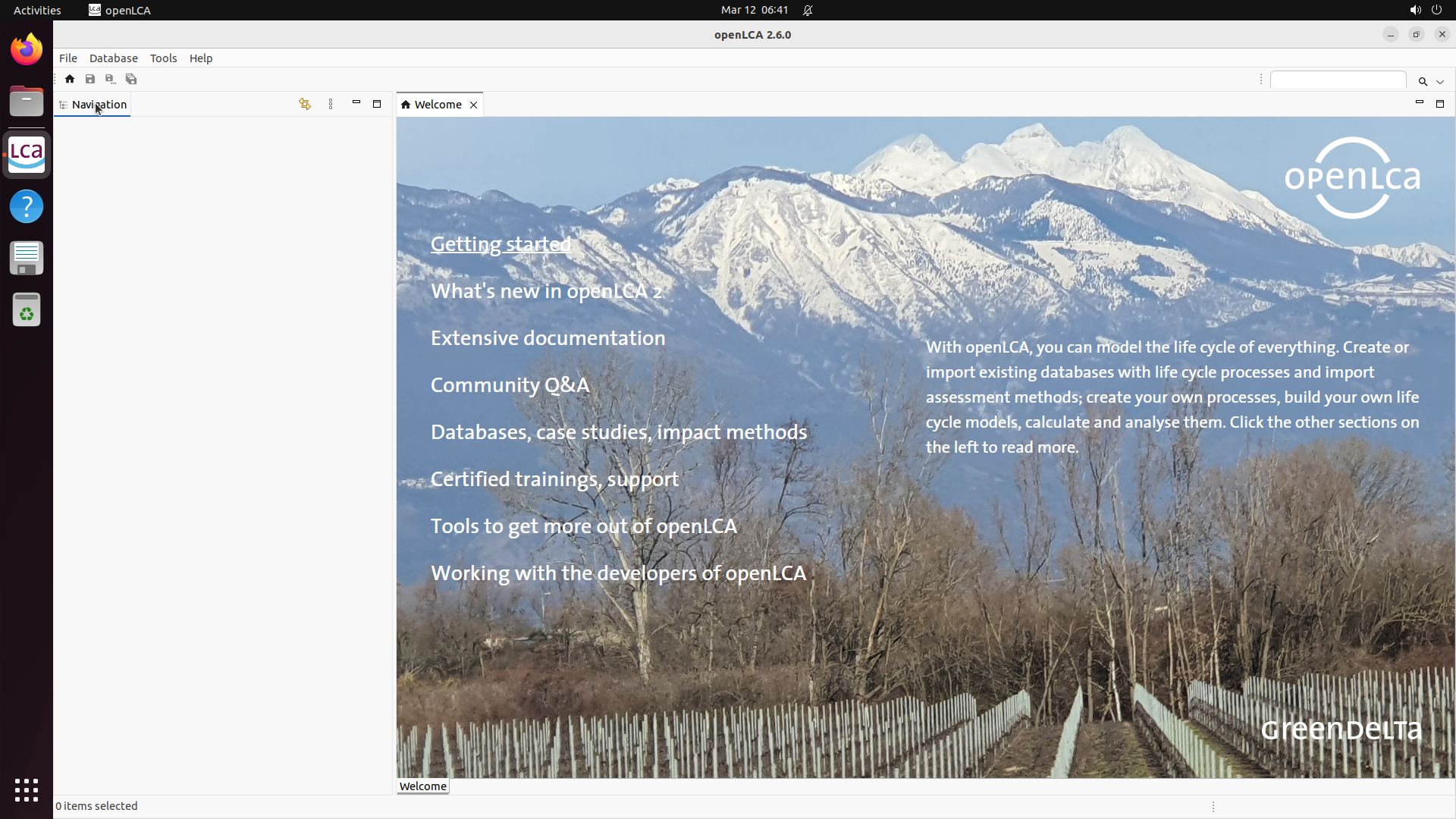}
    {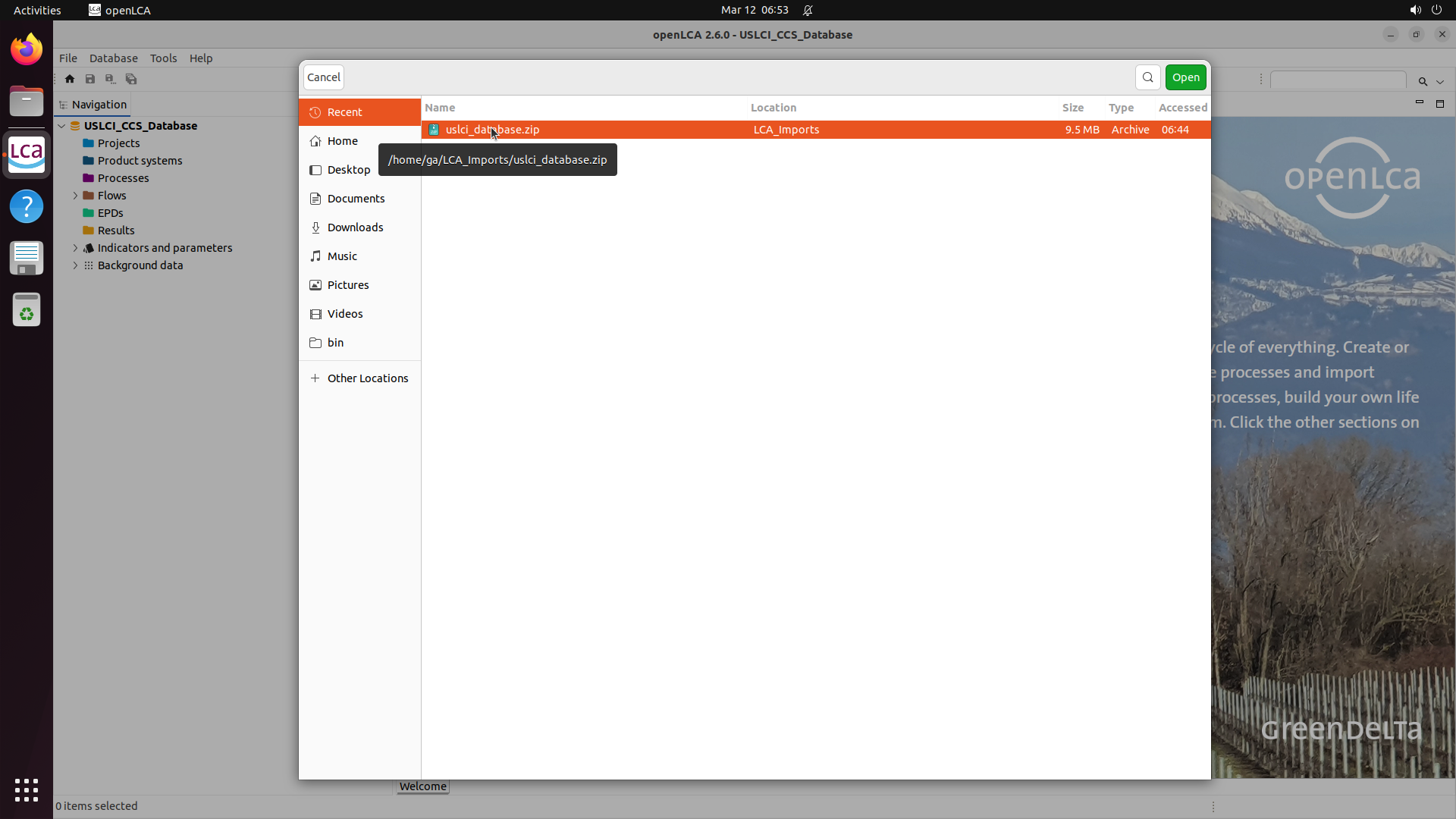}
    {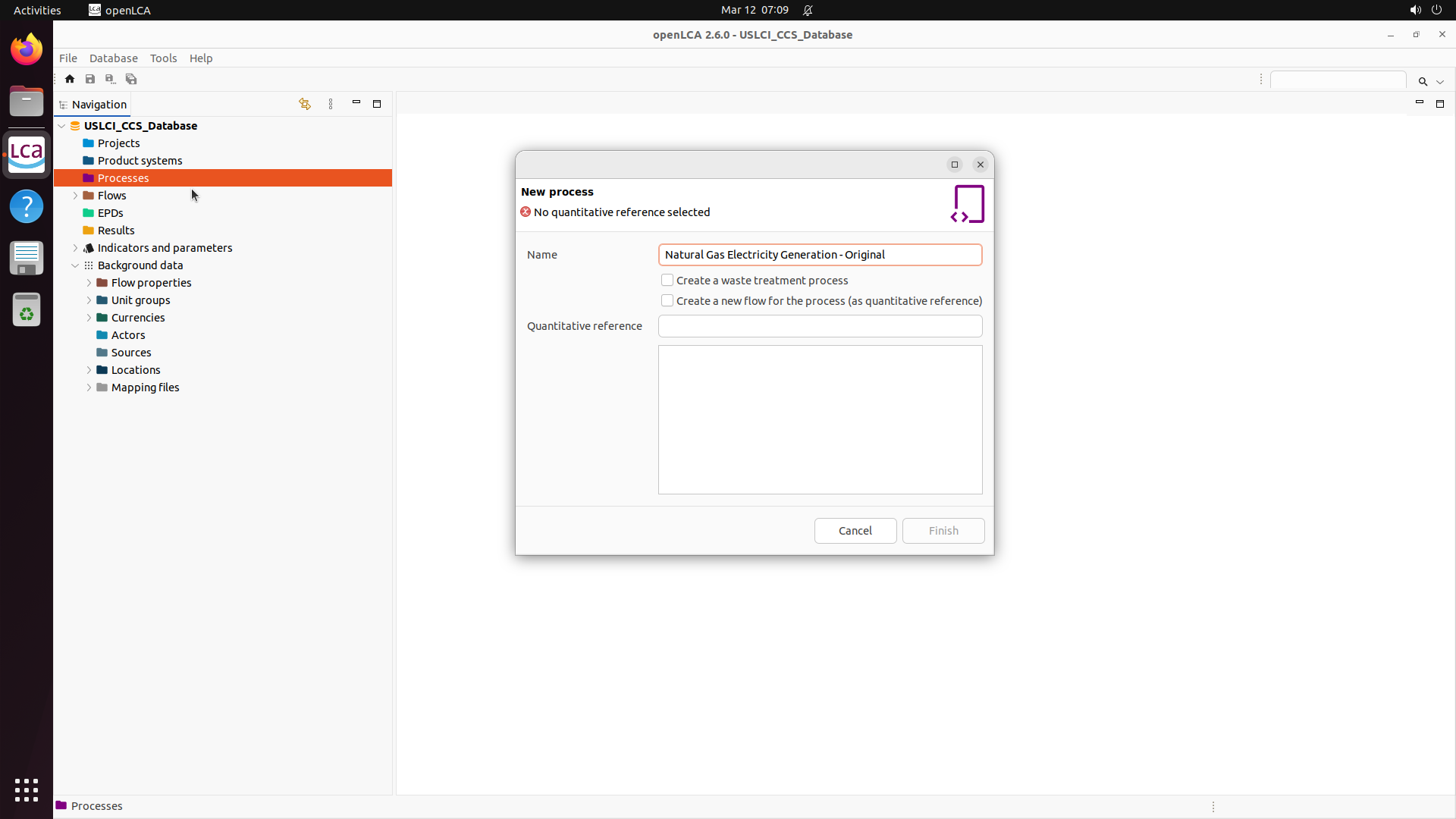}
\end{taskexample}

\medskip

\begin{taskexample}{BlenderBIM (Building Information Modeling) \textnormal{--- bcf\_issue\_authoring \hfill \textit{Difficulty: hard}}}
\small\textit{As a BIM Coordinator performing a model review: load an IFC building model into BlenderBIM, navigate to locate any \texttt{IfcDoor}, initialize a BCF (BIM Collaboration Format) project, create a topic titled ``Door Clearance Issue'' assigned to \texttt{architect@example.com}, add a comment (``Please widen to 900\,mm''), capture the 3D viewport as a viewpoint with camera coordinates and snapshot image, and export the BCF project as a \texttt{.bcfzip} coordination file.}
\taskscreenshots
    {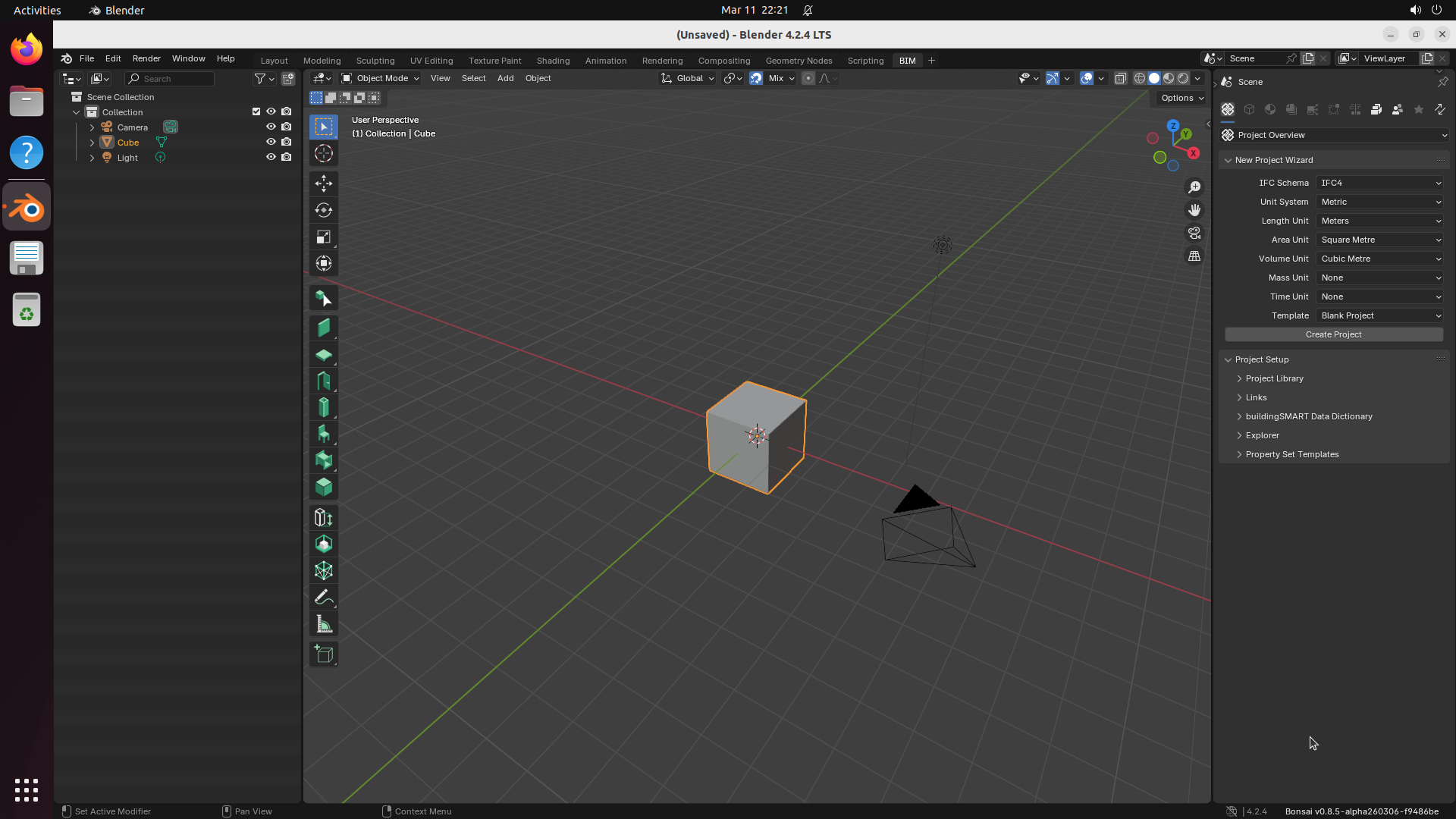}
    {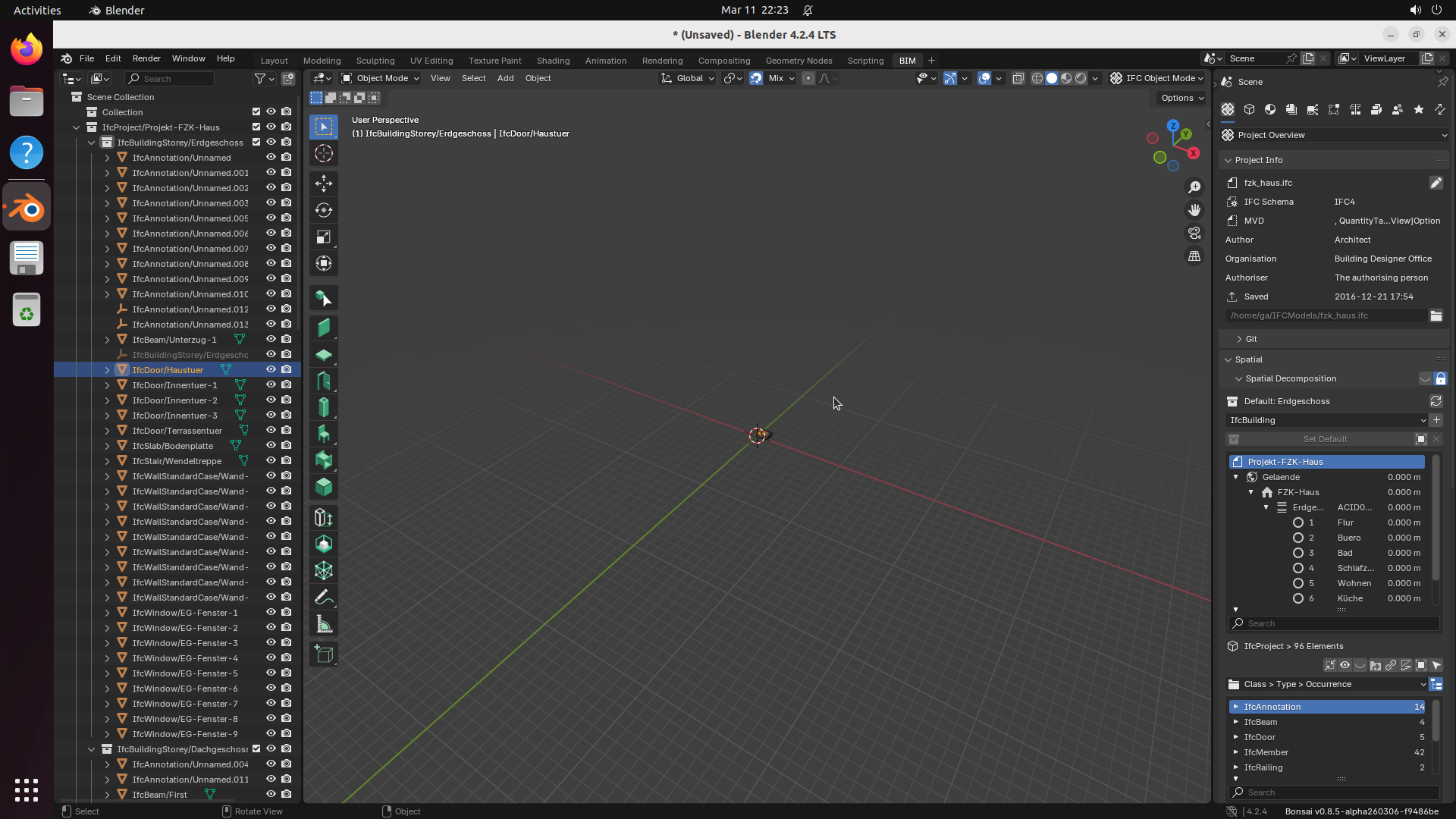}
    {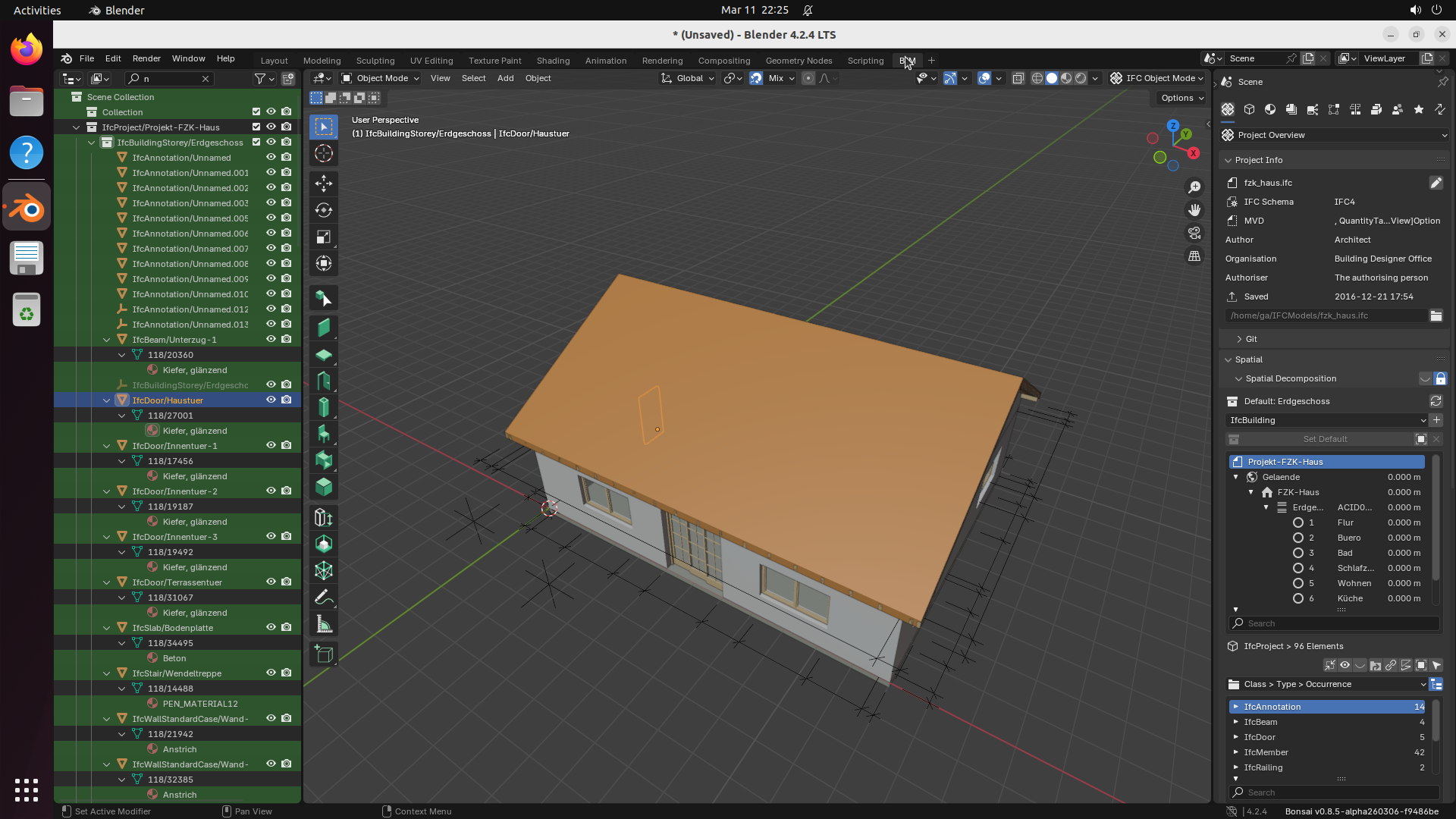}
    {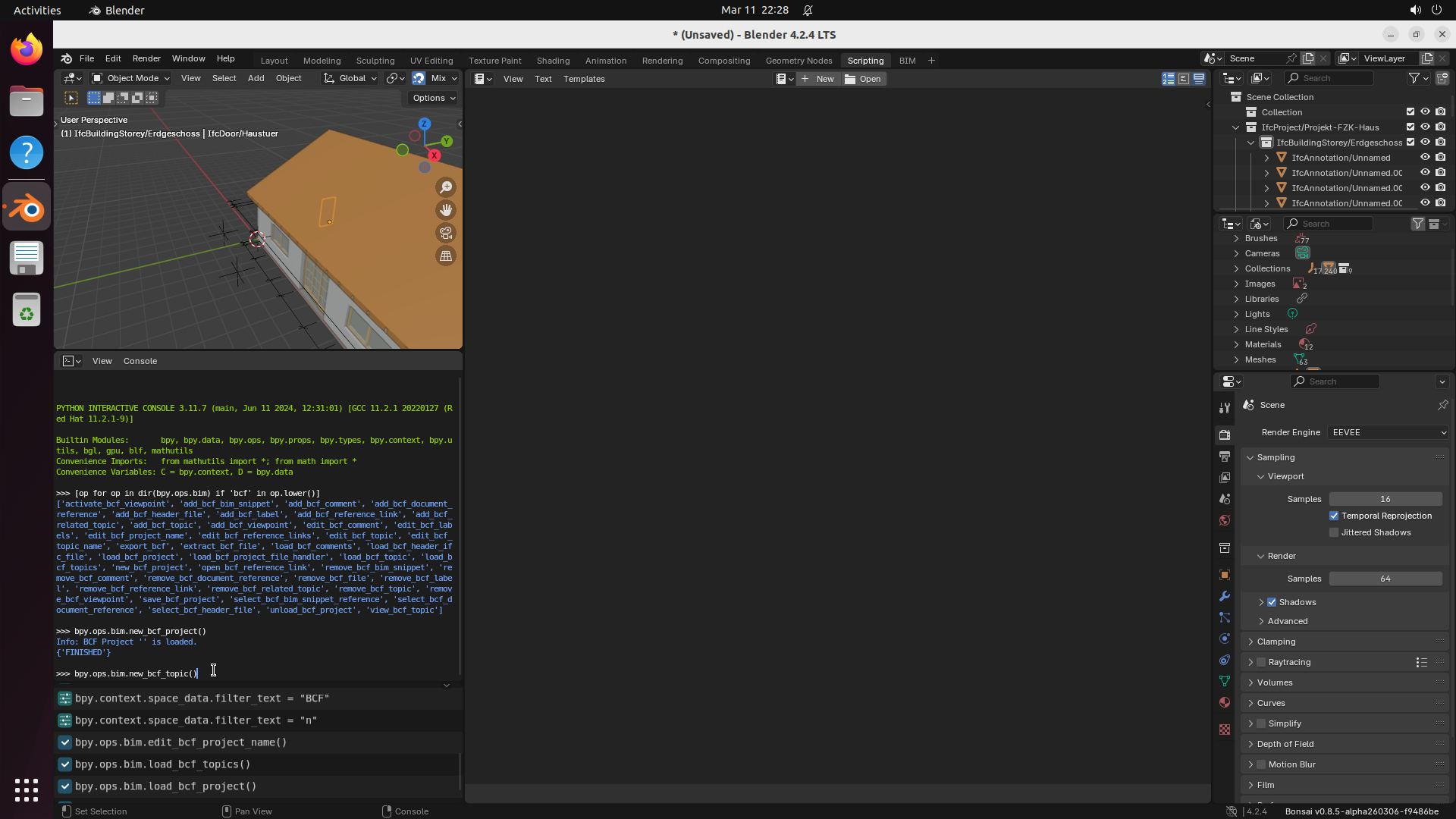}
\end{taskexample}

\medskip

\begin{taskexample}{RStudio (Statistical Computing) \textnormal{--- ames\_elasticnet\_housing \hfill \textit{Difficulty: hard}}}
\small\textit{Build a predictive model for Ames housing prices using regularized regression (Ridge, LASSO, and Elastic Net). Perform 10-fold cross-validation for all three models on the AmesHousing dataset with log-transformed sale price. Generate outputs including a preprocessing summary ($\geq$20 variables), a model comparison table (lambda values and CV RMSE converted back to dollars), the top 15 predictors by absolute coefficient magnitude, and diagnostic plots showing cross-validation curves, coefficient paths, and predicted vs.\ actual values.}
\taskscreenshots
    {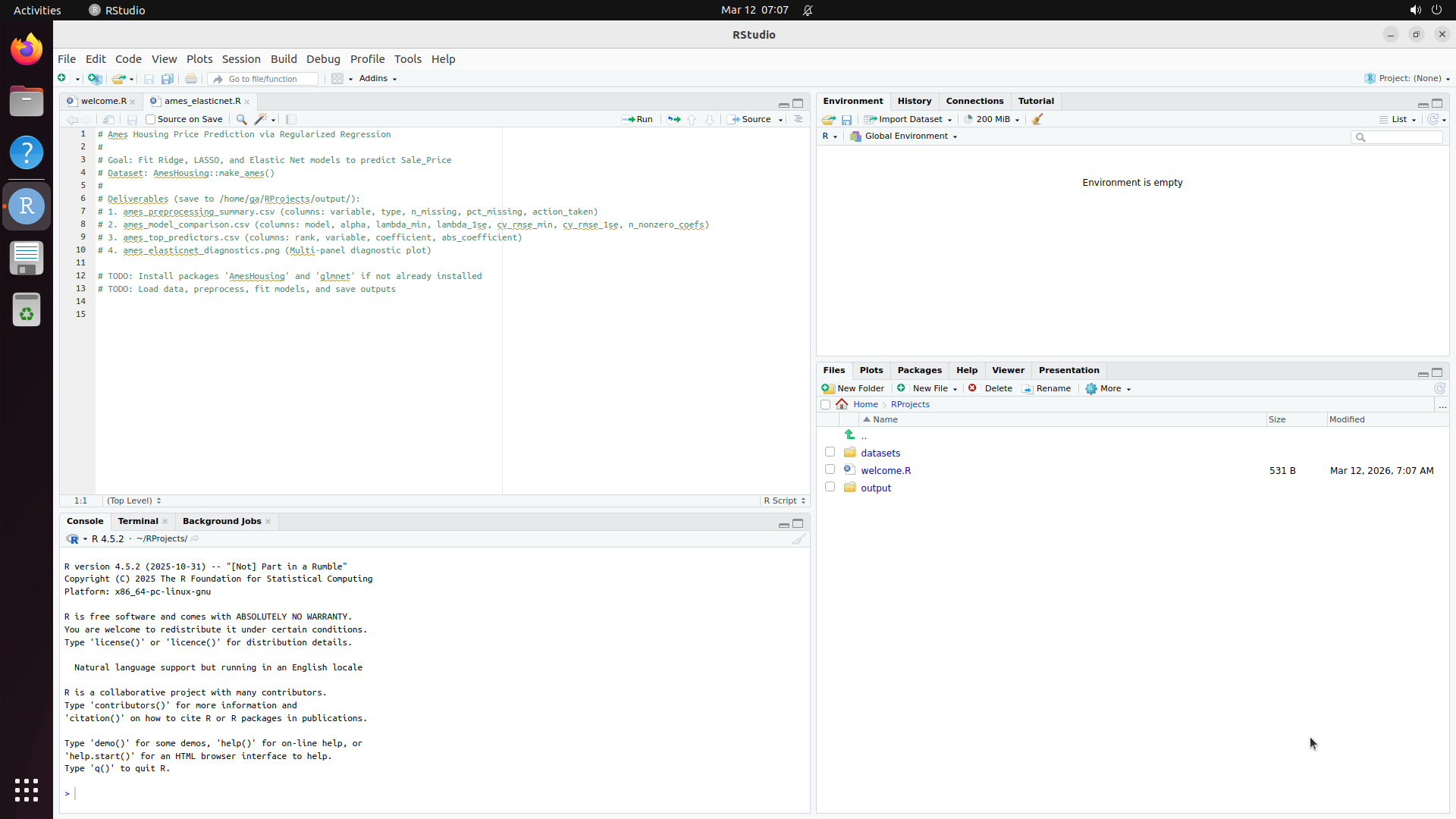}
    {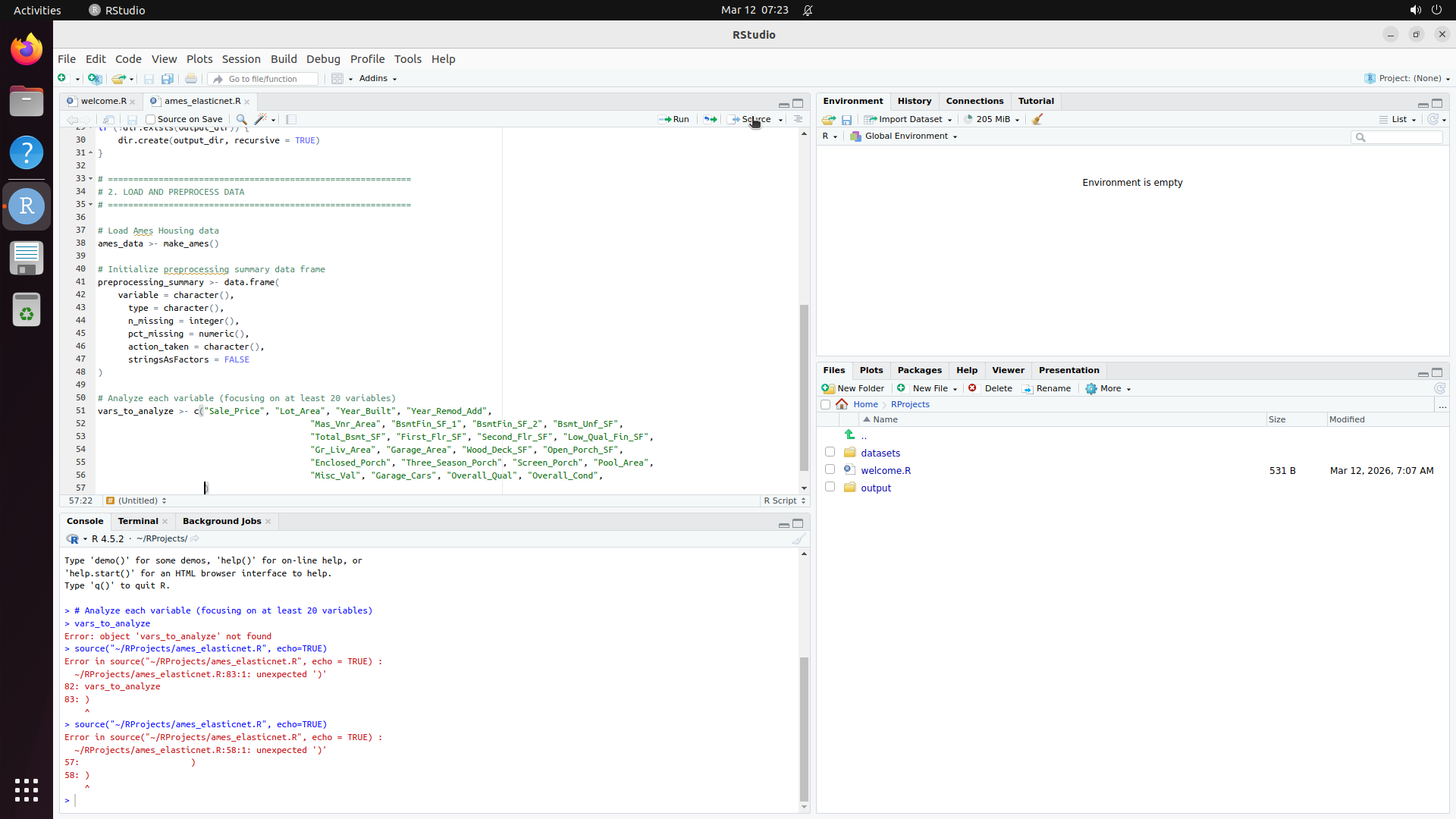}
    {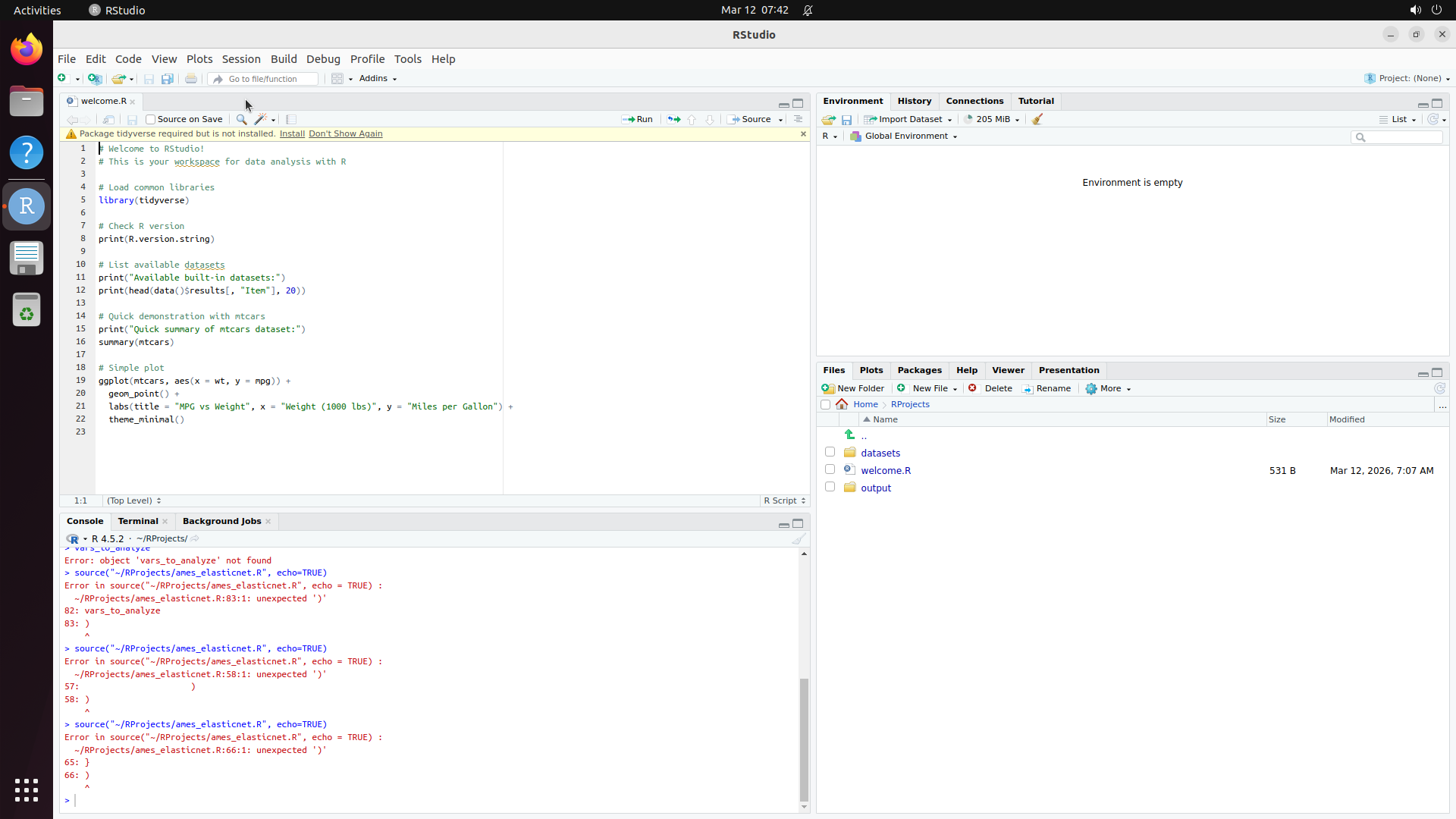}
    {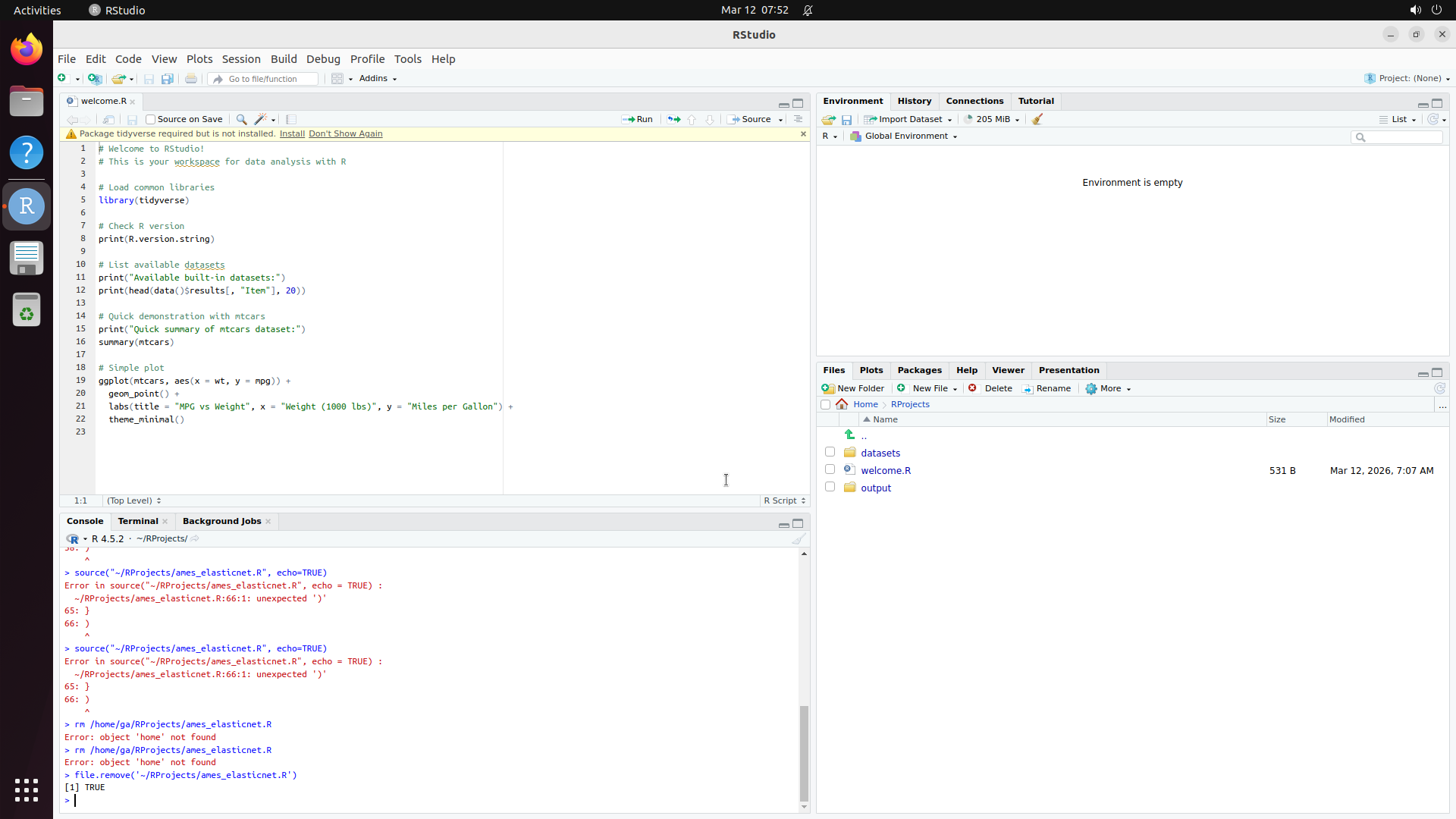}
\end{taskexample}

\section{Contamination Filtering Details}
\label{app:contamination}

With over 10,000 tasks spanning 200+ software environments, \bench{} must ensure that train and test sets do not contain near-duplicate or highly overlapping tasks whose similarity would inflate evaluation scores.
We implement an automated contamination filtering pipeline that (1)~scores all pairwise task similarities via an LLM judge, (2)~constructs a contamination graph from high-similarity pairs, (3)~identifies connected components, and (4)~assigns entire components to the same split.

\subsection{Pairwise Similarity Scoring}

For each pair of tasks within the same software environment, we prompt an LLM to classify their similarity on an 8-point ordinal scale.
Table~\ref{tab:similarity_scale} defines each level.

\begin{table}[h]
\centering
\caption{Similarity scale used by the LLM judge.  Scores $\geq 4$ are flagged as potentially contaminating.}
\label{tab:similarity_scale}
\small
\begin{tabular}{@{}clp{8.2cm}@{}}
\toprule
\textbf{Score} & \textbf{Label} & \textbf{Description} \\
\midrule
1 & \textsc{Not Similar} & Completely unrelated tasks (different goals, different domains). \\
2 & \textsc{Somewhat Similar} & Minor thematic overlap but fundamentally different tasks. \\
3 & \textsc{Some Steps Similar} & Tasks share some common substeps (e.g., both navigate to a location) but have distinctly different end goals. This is common and acceptable. \\
\midrule
4 & \textsc{Very Similar} & Tasks are extremely similar---knowing how to do one would directly help with the other. Nearly interchangeable. \\
5 & \textsc{Same Rephrased} & Essentially the same task with different wording. \\
6 & \textsc{Duplicate} & Identical or near-identical tasks. \\
7 & \textsc{Subset} & Task~1 is a strict subset of Task~2 (Task~2 = Task~1 + additional steps toward the same goal). \\
8 & \textsc{Superset} & Task~1 is a strict superset of Task~2 (Task~1 = Task~2 + additional steps toward the same goal). \\
\bottomrule
\end{tabular}
\end{table}

The prompt instructs the judge to be \emph{liberal}: categories 4--8 should only be assigned when tasks are truly interchangeable or one strictly contains the other.
Sharing common substeps (e.g., ``search for a location'' or ``take a screenshot'') is explicitly directed to category~3.
The full comparison prompt is reproduced in Appendix~\ref{app:prompt_contamination}.

\subsection{Graph Construction and Split Assignment}

\niparagraph{Contamination graph.}
We set a contamination threshold of $\tau = 4$.
Any task pair $(t_i, t_j)$ receiving a similarity score $\geq \tau$ generates an undirected edge in a \emph{contamination graph} $G$, where each node is a task.

\niparagraph{Connected components.}
We compute the connected components of $G$ via breadth-first search.
By construction, every pair of tasks within a component is transitively linked through at least one chain of contaminating similarities.
Assigning all tasks in a component to the same split guarantees zero contaminating pairs across the train/test boundary.

\niparagraph{Greedy bin-packing.}
Components are sorted by size (largest first) and greedily assigned to whichever split---train or test---is furthest from its target ratio.
After assignment, an automated verification pass confirms that no contaminating edge spans the split.

\subsection{Aggregate Statistics}

Table~\ref{tab:contamination_stats} summarizes the filtering outcome across \bench{}.

\begin{table}[h]
\centering
\caption{Contamination filtering statistics for \bench{}.}
\label{tab:contamination_stats}
\small
\begin{tabular}{@{}lr@{}}
\toprule
\textbf{Metric} & \textbf{Value} \\
\midrule
Total pairwise comparisons & 434,699 \\
Flagged pairs (score $\geq 4$) & 2,847 \quad (0.65\%) \\
\midrule
\multicolumn{2}{@{}l}{\textit{Score distribution}} \\
\quad Score 1--2 (not similar) & 1.1\% \\
\quad Score 3 (some steps similar) & 98.2\% \\
\quad Score $\geq 4$ (contaminating) & 0.65\% \\
\midrule
\multicolumn{2}{@{}l}{\textit{Breakdown of flagged pairs}} \\
\quad Score 4 (\textsc{Very Similar}) & 1,354 \\
\quad Score 5 (\textsc{Same Rephrased}) & 517 \\
\quad Score 6 (\textsc{Duplicate}) & 197 \\
\quad Score 7 (\textsc{Subset}) & 467 \\
\quad Score 8 (\textsc{Superset}) & 312 \\
\midrule
Connected components & 10,618 \\
\quad Singleton components & 94.3\% \\
\midrule
\multicolumn{2}{@{}l}{\textit{Final split}} \\
\quad Train tasks & 9,720 \\
\quad Test tasks & 2,500 \\
\bottomrule
\end{tabular}
\end{table}

The overwhelming majority of task pairs (98.2\%) receive a score of~3 (\textsc{Some Steps Similar}), reflecting the expected pattern that tasks within the same software share navigational substeps but pursue distinct goals.
Only 0.65\% of comparisons are flagged as contaminating.
Among flagged pairs, score~4 (\textsc{Very Similar}) is the most common, while true duplicates (score~6) are comparatively rare, indicating that the task generation pipeline produces diverse tasks with only occasional near-overlaps.

The 94.3\% singleton rate among connected components confirms that contamination is sparse: most tasks have no contaminating neighbor and can be freely assigned to either split.
The remaining components group small clusters of related tasks that are kept together in a single split.

\subsection{Manual Verification}

In addition to the automated pipeline, we provide a web-based verification dashboard that displays flagged task pairs side by side, along with their similarity scores and the LLM judge's reasoning.
Annotators can review borderline cases (particularly those at the threshold boundary of score~4) and override the automated judgment if necessary.
This human-in-the-loop step serves as a final safeguard to ensure the integrity of the train/test split.

The full comparison prompt is reproduced in Appendix~\ref{app:prompt_contamination}.
\section{Software Categorization}
\label{app:software_categories}

We classify each of the \numsoftwares{}+ software products along two axes to analyze how software properties affect agent performance (\S\ref{sec:benchmark}).

\niparagraph{Visual complexity.}
We rate each software's interface as \emph{low}, \emph{medium}, or \emph{high} based on the visual density and spatial complexity of its primary workspace:
\begin{itemize}[leftmargin=*, itemsep=2pt]
    \item \textbf{Low} (83 software): Flat UI with standard web forms, text-heavy layouts, clear labels and buttons. Examples: WordPress, LibreOffice Writer, Chrome, ERPNext, OpenEMR.
    \item \textbf{Medium} (91 software): Multi-panel layouts with tabbed interfaces, data tables, charts, and moderate visual density. Examples: VS Code, DBeaver, Wireshark, LibreOffice Calc, Jenkins.
    \item \textbf{High} (34 software): Spatial canvases, 3D viewports, dense scientific visualizations, map layers, waveform displays, or multi-panel imaging. Examples: Blender, GIMP, QGIS, 3D Slicer, AstroImageJ, Ardour.
\end{itemize}

\niparagraph{Domain knowledge.}
We classify whether the software requires domain-specific expertise as \emph{general} or \emph{specialized}:
\begin{itemize}[leftmargin=*, itemsep=2pt]
    \item \textbf{General} (104 software): A computer-literate person can accomplish tasks by reading the UI. No field-specific training is needed. Examples: web browsers, email clients, CRM/ERP systems, office suites, project management tools.
    \item \textbf{Specialized} (104 software): Requires understanding a specific professional or scientific domain---its concepts, terminology, or methods---to know what to do. Examples: 3D Slicer (medical imaging), AstroImageJ (astrophotometry), HEC-RAS (hydraulic modeling), PyMOL (molecular visualization), Wireshark (network protocols).
\end{itemize}

The categorization is based on the software's interface and domain, not the specific tasks generated for it. The full assignment of all software to both axes is available in the supplementary code (\texttt{analysis/categorize\_envs.py}).
\section{Occupational Coverage of \bench{}}
\label{app:soc_coverage}

To validate that \bench{} covers a broad range of occupations, we classify each of the 13K+ tasks into one of the 22 SOC (Standard Occupational Classification) major occupation groups. We prompt Gemini 3 Flash to first identify the specific occupation that would perform the task as part of their regular professional work, then map that occupation to a SOC group. All 22 groups are represented. Below we show three representative tasks per group.

\begin{tcolorbox}[
  title={SOC 1/22: Management},
  enhanced,
  breakable,
  colback=white,
  colframe=promptframe,
  colbacktitle=prompttitlebg,
  coltitle=white,
  fonttitle=\bfseries\small,
  arc=2pt,
  boxrule=0.5pt,
  left=6pt, right=6pt, top=6pt, bottom=6pt,
  toptitle=3pt, bottomtitle=3pt,
]
\textbf{Suite Crm} $\vert$ \textit{Sales Operations Manager}
\smallskip

\small You are a sales operations manager preparing for a Q4 technology sector campaign. You need to flag all 'Technology' industry accounts with a 'Hot' rating so the sales team can prioritize their outreach. Using the mass...

\vspace{4pt}\hrule\vspace{4pt}

\textbf{Microsoft Excel} $\vert$ \textit{School Business Administrator}
\smallskip

\small You are a school business administrator for Botetourt County Public Schools preparing the annual Title I compliance report. Microsoft Excel is open with school\_district.xlsx. The 'School\_Data' sheet has 11 schools (7 ...

\vspace{4pt}\hrule\vspace{4pt}

\textbf{Manageservice} $\vert$ \textit{IT Support Shift Lead}
\smallskip

\small As the Shift Lead, perform a bulk update on the 5 unassigned requests related to 'WOW Carts' (Workstations on Wheels). Assign all 5 requests to the 'Clinical IT Support' group and classify them under the 'Medical Hard...
\end{tcolorbox}
\vspace{4pt}

\begin{tcolorbox}[
  title={SOC 2/22: Business and Financial Operations},
  enhanced,
  breakable,
  colback=white,
  colframe=promptframe,
  colbacktitle=prompttitlebg,
  coltitle=white,
  fonttitle=\bfseries\small,
  arc=2pt,
  boxrule=0.5pt,
  left=6pt, right=6pt, top=6pt, bottom=6pt,
  toptitle=3pt, bottomtitle=3pt,
]
\textbf{Tor Browser} $\vert$ \textit{Intelligence Analyst}
\smallskip

\small Create a bookmark folder named 'OSINT Search Tools' containing three keyword bookmarks: 'Ahmia Onion Search' (URL: https://ahmia.fi/search/?q=\%s, Keyword: @ahmia), 'DuckDuckGo Private' (URL: https://duckduckgogg42xjoc...

\vspace{4pt}\hrule\vspace{4pt}

\textbf{Sygic Gps} $\vert$ \textit{Logistics Coordinator}
\smallskip

\small Find and display a list of hospitals in Kabul, Afghanistan (34.5553° N, 69.2075° E). The final view must show hospitals located specifically within Kabul rather than near your current location.

\vspace{4pt}\hrule\vspace{4pt}

\textbf{Oracle Sql Developer} $\vert$ \textit{Compensation Analyst}
\smallskip

\small You are a Compensation Analyst for a retail chain. The HR\_COMPLIANCE schema in Oracle (connected as hr\_compliance/Compliance2024 to XEPDB1) contains 3 months of scheduling and time-punch data. 'Fair Workweek' laws str...
\end{tcolorbox}
\vspace{4pt}

\begin{tcolorbox}[
  title={SOC 3/22: Computer and Mathematical},
  enhanced,
  breakable,
  colback=white,
  colframe=promptframe,
  colbacktitle=prompttitlebg,
  coltitle=white,
  fonttitle=\bfseries\small,
  arc=2pt,
  boxrule=0.5pt,
  left=6pt, right=6pt, top=6pt, bottom=6pt,
  toptitle=3pt, bottomtitle=3pt,
]
\textbf{Virtualmin} $\vert$ \textit{System Administrator}
\smallskip

\small The virtual server 'broken-app.test' is currently malfunctioning. Diagnose and repair the configuration errors using Virtualmin's 'Validate Virtual Servers' tool until the server passes validation with no errors. Ensu...

\vspace{4pt}\hrule\vspace{4pt}

\textbf{Wireshark} $\vert$ \textit{Incident Response Analyst}
\smallskip

\small You are an incident response analyst investigating potential port scanning activity. From the capture file currently loaded in Wireshark (200722\_tcp\_anon.pcapng), isolate only the TCP SYN packets (initial connection a...

\vspace{4pt}\hrule\vspace{4pt}

\textbf{Veracrypt} $\vert$ \textit{Systems Administrator}
\smallskip

\small Create a robust shell script named 'safe\_log\_entry.sh' in /home/ga/Documents/. The script must accept a volume password as the first argument, mount the VeraCrypt volume at /home/ga/Volumes/audit\_volume.hc, append the...
\end{tcolorbox}
\vspace{4pt}

\begin{tcolorbox}[
  title={SOC 4/22: Architecture and Engineering},
  enhanced,
  breakable,
  colback=white,
  colframe=promptframe,
  colbacktitle=prompttitlebg,
  coltitle=white,
  fonttitle=\bfseries\small,
  arc=2pt,
  boxrule=0.5pt,
  left=6pt, right=6pt, top=6pt, bottom=6pt,
  toptitle=3pt, bottomtitle=3pt,
]
\textbf{Stellarium} $\vert$ \textit{Spacecraft Communications Engineer}
\smallskip

\small You are a Spacecraft Communications Engineer for the Deep Space Network (DSN). Simulate the line-of-sight visibility of Mars during the Perseverance rover landing on February 18, 2021, at 20:30:00 UTC to verify the te...

\vspace{4pt}\hrule\vspace{4pt}

\textbf{Topocal} $\vert$ \textit{Land Surveyor}
\smallskip

\small Compute the coordinates of three new property corners (Points 101, 102, and 103) using TopoCal's COGO/Radiation tools. Import the control dataset 'jefferson\_county\_control.csv' from the Desktop or Documents\textbackslash\{\}TopoCal. U...

\vspace{4pt}\hrule\vspace{4pt}

\textbf{Openvsp} $\vert$ \textit{Aerospace Engineer}
\smallskip

\small A conceptual aerodynamicist is designing a Standard Class 15-meter sailplane. To achieve an optimal elliptical lift distribution, the wing planform has been divided into 4 distinct sections from root to tip. Read the ...
\end{tcolorbox}
\vspace{4pt}

\begin{tcolorbox}[
  title={SOC 5/22: Life, Physical, and Social Science},
  enhanced,
  breakable,
  colback=white,
  colframe=promptframe,
  colbacktitle=prompttitlebg,
  coltitle=white,
  fonttitle=\bfseries\small,
  arc=2pt,
  boxrule=0.5pt,
  left=6pt, right=6pt, top=6pt, bottom=6pt,
  toptitle=3pt, bottomtitle=3pt,
]
\textbf{Qground Control} $\vert$ \textit{Remote Sensing Scientist}
\smallskip

\small You are a Remote Sensing Scientist upgrading a mapping drone with a high-resolution Sony A7R IV camera payload. To achieve survey-grade accuracy, the autopilot must trigger the camera and record the exact microsecond ...

\vspace{4pt}\hrule\vspace{4pt}

\textbf{Openbci Gui Temp Codex} $\vert$ \textit{Neuroscientist}
\smallskip

\small Configure a dual-region analysis (Frontal vs Occipital) in the Spectrogram widget using a Synthetic session in OpenBCI GUI. Set the Top Plot to include only Channel 1 and Channel 2, and the Bottom Plot to include only...

\vspace{4pt}\hrule\vspace{4pt}

\textbf{Slicer3D} $\vert$ \textit{Medical Physicist}
\smallskip

\small Calculate the Signal-to-Noise Ratio (SNR) for the loaded MRHead brain MRI. The SNR is defined as the mean of a 'Signal' segment in the brain white matter divided by the standard deviation of a 'Noise' segment in the b...
\end{tcolorbox}
\vspace{4pt}

\begin{tcolorbox}[
  title={SOC 6/22: Community and Social Service},
  enhanced,
  breakable,
  colback=white,
  colframe=promptframe,
  colbacktitle=prompttitlebg,
  coltitle=white,
  fonttitle=\bfseries\small,
  arc=2pt,
  boxrule=0.5pt,
  left=6pt, right=6pt, top=6pt, bottom=6pt,
  toptitle=3pt, bottomtitle=3pt,
]
\textbf{Arkcase} $\vert$ \textit{Probation Officer}
\smallskip

\small You are a probation officer conducting your monthly caseload review in ArkCase. Identify all non-compliant supervisees in the Complaints module whose 'Last Contact' dates are before December 1, 2025. For each non-comp...

\vspace{4pt}\hrule\vspace{4pt}

\textbf{Odoo Scheduling} $\vert$ \textit{Career Counselor}
\smallskip

\small Schedule a calendar meeting titled 'Career Coaching Session - Emma Thompson' for the contact 'Emma Thompson'. The meeting should be set for next Wednesday at 3:00 PM for 1 hour, with Emma Thompson as an attendee. Log ...

\vspace{4pt}\hrule\vspace{4pt}

\textbf{Firefox} $\vert$ \textit{Community Health Worker}
\smallskip

\small You are a community health worker investigating toxic chemical releases using the EPA's Toxic Release Inventory (TRI) tools (https://www.epa.gov/toxics-release-inventory-tri-program). Research Harris County, Texas and...
\end{tcolorbox}
\vspace{4pt}

\begin{tcolorbox}[
  title={SOC 7/22: Legal},
  enhanced,
  breakable,
  colback=white,
  colframe=promptframe,
  colbacktitle=prompttitlebg,
  coltitle=white,
  fonttitle=\bfseries\small,
  arc=2pt,
  boxrule=0.5pt,
  left=6pt, right=6pt, top=6pt, bottom=6pt,
  toptitle=3pt, bottomtitle=3pt,
]
\textbf{Arkcase} $\vert$ \textit{Legal Assistant}
\smallskip

\small An 'Initial Legal Review' task for a specific complaint case was due yesterday and is now overdue. You need to intervene. Retrieve the Case Number from '\textasciitilde\{\}/Documents/escalation\_info.txt' and locate the 'Initial Legal R...

\vspace{4pt}\hrule\vspace{4pt}

\textbf{Wps Office Writer} $\vert$ \textit{Paralegal}
\smallskip

\small You are a paralegal at a corporate law firm. Open /home/ga/Documents/vendor\_agreement\_draft.docx, a draft Vendor Services Agreement between CloudFirst Industries, LLC and Meridian Technology Solutions, Inc. Clean up t...

\vspace{4pt}\hrule\vspace{4pt}

\textbf{Nuxeo Platform} $\vert$ \textit{Paralegal}
\smallskip

\small Perform a legal discovery operation in Nuxeo Platform. Identify all documents in the 'Projects' workspace that meet both criteria: the Title contains 'Agreement' (case-insensitive) and the content/body contains 'Acme ...
\end{tcolorbox}
\vspace{4pt}

\begin{tcolorbox}[
  title={SOC 8/22: Educational Instruction and Library},
  enhanced,
  breakable,
  colback=white,
  colframe=promptframe,
  colbacktitle=prompttitlebg,
  coltitle=white,
  fonttitle=\bfseries\small,
  arc=2pt,
  boxrule=0.5pt,
  left=6pt, right=6pt, top=6pt, bottom=6pt,
  toptitle=3pt, bottomtitle=3pt,
]
\textbf{Active Inspire} $\vert$ \textit{Art Teacher}
\smallskip

\small Create a 3-page flipchart visual analysis of 'The Great Wave off Kanagawa' and save it to /home/ga/Documents/Flipcharts/artwork\_analysis.flipchart. The first page must feature the image 'great\_wave.jpg' (from /home/ga...

\vspace{4pt}\hrule\vspace{4pt}

\textbf{Moodle} $\vert$ \textit{Biology Professor}
\smallskip

\small You are the instructor for BIO302 Advanced Cell Biology at State University. Set up a complete student progression tracking system for the course. Configure completion conditions for the following activities: 'Lab Saf...

\vspace{4pt}\hrule\vspace{4pt}

\textbf{Safe Exam Browser} $\vert$ \textit{Instructional Coordinator}
\smallskip

\small Log into SEB Server as super-admin (password: admin). For the 'Chemistry 201 - Midterm' exam configuration, enable and set the custom Quit Confirmation Message to: 'WARNING: Quitting will SUBMIT your exam permanently....
\end{tcolorbox}
\vspace{4pt}

\begin{tcolorbox}[
  title={SOC 9/22: Arts, Design, Entertainment, Sports, and Media},
  enhanced,
  breakable,
  colback=white,
  colframe=promptframe,
  colbacktitle=prompttitlebg,
  coltitle=white,
  fonttitle=\bfseries\small,
  arc=2pt,
  boxrule=0.5pt,
  left=6pt, right=6pt, top=6pt, bottom=6pt,
  toptitle=3pt, bottomtitle=3pt,
]
\textbf{Gimp} $\vert$ \textit{Graphic Designer}
\smallskip

\small Apply a pixelate filter to the image to create a blocky, mosaic-style effect.

\vspace{4pt}\hrule\vspace{4pt}

\textbf{Sweet Home 3D} $\vert$ \textit{Interior Designer}
\smallskip

\small You are an entrepreneur opening 'Spoke \& Bean', a hybrid bicycle shop, mechanical repair bay, and community espresso bar. Transform the open-plan commercial unit in bike\_shop\_starter.sh3d into a functional mixed-use s...

\vspace{4pt}\hrule\vspace{4pt}

\textbf{Gimp Osw} $\vert$ \textit{Graphic Designer}
\smallskip

\small Remove the background from the dog image in GIMP.
\end{tcolorbox}
\vspace{4pt}

\begin{tcolorbox}[
  title={SOC 10/22: Healthcare Practitioners and Technical},
  enhanced,
  breakable,
  colback=white,
  colframe=promptframe,
  colbacktitle=prompttitlebg,
  coltitle=white,
  fonttitle=\bfseries\small,
  arc=2pt,
  boxrule=0.5pt,
  left=6pt, right=6pt, top=6pt, bottom=6pt,
  toptitle=3pt, bottomtitle=3pt,
]
\textbf{Oscar Emr} $\vert$ \textit{Physician}
\smallskip

\small Dr. Chen wants to speed up prescribing for Strep Throat. Using the credentials (username 'oscardoc', password 'oscar', PIN '1117'), create a prescription favorite named 'Strep Throat - Amox 500' for patient 'Mario Ros...

\vspace{4pt}\hrule\vspace{4pt}

\textbf{Invesalius3} $\vert$ \textit{Radiologic Technologist}
\smallskip

\small Using the loaded CT Cranium DICOM dataset in InVesalius 3, generate a 3D surface of the skull (bone mask) and create a rotating animation of the model. Export the animation as either a series of at least 12 PNG screen...

\vspace{4pt}\hrule\vspace{4pt}

\textbf{Slicer3D} $\vert$ \textit{Radiologist}
\smallskip

\small Generate a subtraction enhancement map named 'EnhancementMap' (T1\_Contrast minus T1) and a binary 'EnhancementMask' from the loaded brain MRI volumes. Export the enhancement map to \textasciitilde\{\}/Documents/SlicerData/BraTS/enhance...
\end{tcolorbox}
\vspace{4pt}

\begin{tcolorbox}[
  title={SOC 11/22: Healthcare Support},
  enhanced,
  breakable,
  colback=white,
  colframe=promptframe,
  colbacktitle=prompttitlebg,
  coltitle=white,
  fonttitle=\bfseries\small,
  arc=2pt,
  boxrule=0.5pt,
  left=6pt, right=6pt, top=6pt, bottom=6pt,
  toptitle=3pt, bottomtitle=3pt,
]
\textbf{Oscar Emr} $\vert$ \textit{Medical Assistant}
\smallskip

\small A safety recall has been issued for the medication 'Atenolol' due to potential impurities. You need to identify any active patients currently prescribed this medication and flag their charts. Log in to OSCAR EMR (User...

\vspace{4pt}\hrule\vspace{4pt}

\textbf{Openemr} $\vert$ \textit{Medical Assistant}
\smallskip

\small Record a historical DTaP immunization for Jayson Fadel (DOB: 1992-06-30). Date administered: 2019-03-15, administered by: Outside Provider, site: Left Deltoid, manufacturer: Sanofi Pasteur, lot number: D2894AA, expira...

\vspace{4pt}\hrule\vspace{4pt}

\textbf{Freemed} $\vert$ \textit{Medical Assistant}
\smallskip

\small Log into FreeMED (Username: admin, Password: admin) and record a new laboratory order for patient Marcus Vance. The order must include both a 'Lipid Panel' and a 'Hemoglobin A1c' saved to his clinical record.
\end{tcolorbox}
\vspace{4pt}

\begin{tcolorbox}[
  title={SOC 12/22: Protective Service},
  enhanced,
  breakable,
  colback=white,
  colframe=promptframe,
  colbacktitle=prompttitlebg,
  coltitle=white,
  fonttitle=\bfseries\small,
  arc=2pt,
  boxrule=0.5pt,
  left=6pt, right=6pt, top=6pt, bottom=6pt,
  toptitle=3pt, bottomtitle=3pt,
]
\textbf{Opencad} $\vert$ \textit{Police Officer}
\smallskip

\small Log in to OpenCAD (http://localhost) as the Admin User (admin@opencad.local / Admin123!) and create a Person BOLO for Marcus Holloway. Include the following details: Male, African American, approx 6'1", 185 lbs, short...

\vspace{4pt}\hrule\vspace{4pt}

\textbf{Tor Browser} $\vert$ \textit{Private Investigator}
\smallskip

\small You are a Private Investigator building a persistent OSINT dashboard. Configure Tor Browser to retain browsing history and create a local HTML dashboard at /home/ga/Documents/osint\_dashboard.html featuring a heading a...

\vspace{4pt}\hrule\vspace{4pt}

\textbf{Google Earth} $\vert$ \textit{Search and Rescue Coordinator}
\smallskip

\small Identify at least 3 potential helicopter landing zones within 5 km of Rifugio Lagazuoi in the Dolomites, Italy (46.5289°N, 12.0078°E). For each LZ, create a placemark with a systematic name (e.g., LZ-Alpha, LZ-Bravo),...
\end{tcolorbox}
\vspace{4pt}

\begin{tcolorbox}[
  title={SOC 13/22: Food Preparation and Serving Related},
  enhanced,
  breakable,
  colback=white,
  colframe=promptframe,
  colbacktitle=prompttitlebg,
  coltitle=white,
  fonttitle=\bfseries\small,
  arc=2pt,
  boxrule=0.5pt,
  left=6pt, right=6pt, top=6pt, bottom=6pt,
  toptitle=3pt, bottomtitle=3pt,
]
\textbf{Libreoffice Calc} $\vert$ \textit{Baker}
\smallskip

\small Scale the cookie recipe from 24 to 75 cookies. Use practical rounding for proportional ingredient amounts and round all egg quantities up to the nearest whole number.

\vspace{4pt}\hrule\vspace{4pt}

\textbf{Floreant Pos} $\vert$ \textit{Food Service Manager}
\smallskip

\small Perform an end-to-end workflow starting in the BACK OFFICE (PIN: 1111) by creating a 'Burger Toppings' modifier group containing 'Bacon' (\$1.50), 'Avocado' (\$2.00), and 'Extra Cheese' (\$0.75). Configure a new 'Build-Y...

\vspace{4pt}\hrule\vspace{4pt}

\textbf{Chrome} $\vert$ \textit{Sommelier}
\smallskip

\small Configure the cellar Chrome workstation according to the Beverage Team Browser Standard. First, import the bookmarks from \textasciitilde\{\}/Desktop/cellar\_bookmarks.html. Then, organize them into 4 specific folders on the bookmark ba...
\end{tcolorbox}
\vspace{4pt}

\begin{tcolorbox}[
  title={SOC 14/22: Building and Grounds Cleaning and Maintenance},
  enhanced,
  breakable,
  colback=white,
  colframe=promptframe,
  colbacktitle=prompttitlebg,
  coltitle=white,
  fonttitle=\bfseries\small,
  arc=2pt,
  boxrule=0.5pt,
  left=6pt, right=6pt, top=6pt, bottom=6pt,
  toptitle=3pt, bottomtitle=3pt,
]
\textbf{Vtiger Crm} $\vert$ \textit{Landscaping Supervisor}
\smallskip

\small You are a landscaping company supervisor who needs to register a new materials supplier in Vtiger CRM so that purchase orders can be created for upcoming spring projects. Create a new Vendor record with the following ...

\vspace{4pt}\hrule\vspace{4pt}

\textbf{Libreoffice Calc} $\vert$ \textit{Groundskeeper}
\smallskip

\small Create a plant watering tracker using date formulas, the TODAY() function, conditional formatting for overdue plants, and priority sorting. Save the spreadsheet to \textasciitilde\{\}/Documents/plant\_watering\_tracker.xlsx.
\end{tcolorbox}
\vspace{4pt}

\begin{tcolorbox}[
  title={SOC 15/22: Personal Care and Service},
  enhanced,
  breakable,
  colback=white,
  colframe=promptframe,
  colbacktitle=prompttitlebg,
  coltitle=white,
  fonttitle=\bfseries\small,
  arc=2pt,
  boxrule=0.5pt,
  left=6pt, right=6pt, top=6pt, bottom=6pt,
  toptitle=3pt, bottomtitle=3pt,
]
\textbf{Sweet Home 3D} $\vert$ \textit{Licensed Cosmetologist}
\smallskip

\small You are a licensed cosmetologist opening your first boutique hair salon in a converted residential villa. Design a functional and professional salon layout in Sweet Home 3D. The design must feature a styling floor wit...

\vspace{4pt}\hrule\vspace{4pt}

\textbf{Garmin Basecamp} $\vert$ \textit{Outdoor Guide}
\smallskip

\small Garmin BaseCamp is running with `fells\_loop.gpx` data pre-imported. Find the halfway point by trail distance of the `fells\_loop` track and create a waypoint at that location named `Lunch\_Stop` with the notes `Halfway ...

\vspace{4pt}\hrule\vspace{4pt}

\textbf{Wger} $\vert$ \textit{Fitness Trainer}
\smallskip

\small As the admin user (username: admin, password: adminadmin) on http://localhost, log a workout session for today's date using the 'Full Body Workout' routine. The session should include a 'General' impression and the no...
\end{tcolorbox}
\vspace{4pt}

\begin{tcolorbox}[
  title={SOC 16/22: Sales and Related},
  enhanced,
  breakable,
  colback=white,
  colframe=promptframe,
  colbacktitle=prompttitlebg,
  coltitle=white,
  fonttitle=\bfseries\small,
  arc=2pt,
  boxrule=0.5pt,
  left=6pt, right=6pt, top=6pt, bottom=6pt,
  toptitle=3pt, bottomtitle=3pt,
]
\textbf{Copper Point Of Sale} $\vert$ \textit{Sales Representative}
\smallskip

\small Create a sales quote for the corporate customer 'Greenfield Office Solutions'. Include 20 units of 'Copy Paper A4 (500 sheets)', 10 units of 'Ballpoint Pen Blue (Box of 12)', and 8 units of 'Manila Folder Letter Size ...

\vspace{4pt}\hrule\vspace{4pt}

\textbf{Erpnext} $\vert$ \textit{Sales Operations Coordinator}
\smallskip

\small You are a sales operations coordinator at Wind Power LLC. At a recent renewable energy trade show, you met a potential customer named Marcus Chen from Greenfield Renewable Solutions. You need to enter this lead into t...

\vspace{4pt}\hrule\vspace{4pt}

\textbf{Bcwebcam} $\vert$ \textit{Cashier}
\smallskip

\small Use bcWebCam to scan a product barcode directly into a local web POS system. 1. Ensure bcWebCam is running and its 'Keyboard Emulation' (keyboard wedge) output mode is enabled in its settings. 2. A local POS system is...
\end{tcolorbox}
\vspace{4pt}

\begin{tcolorbox}[
  title={SOC 17/22: Office and Administrative Support},
  enhanced,
  breakable,
  colback=white,
  colframe=promptframe,
  colbacktitle=prompttitlebg,
  coltitle=white,
  fonttitle=\bfseries\small,
  arc=2pt,
  boxrule=0.5pt,
  left=6pt, right=6pt, top=6pt, bottom=6pt,
  toptitle=3pt, bottomtitle=3pt,
]
\textbf{Aerobridge} $\vert$ \textit{Administrative Assistant}
\smallskip

\small Register a new drone operating company in the Aerobridge system (http://localhost:8000/admin/) using username 'admin' and password 'adminpass123'. Create the company record with the following details:  - Full Name: Va...

\vspace{4pt}\hrule\vspace{4pt}

\textbf{Libreoffice Calc} $\vert$ \textit{Administrative Assistant}
\smallskip

\small Create a warranty tracking system that calculates expiration dates and days remaining. Include automated status logic and visual alerts for warranty expirations. Save the tracking system to /home/ga/Documents/warranty...

\vspace{4pt}\hrule\vspace{4pt}

\textbf{Bcwebcam} $\vert$ \textit{Office Clerk}
\smallskip

\small Configure bcWebCam so that minimizing the window sends it to the system tray (notification area) instead of the taskbar, and then execute the minimization so the application is hidden but still running.   Instructions...
\end{tcolorbox}
\vspace{4pt}

\begin{tcolorbox}[
  title={SOC 18/22: Farming, Fishing, and Forestry},
  enhanced,
  breakable,
  colback=white,
  colframe=promptframe,
  colbacktitle=prompttitlebg,
  coltitle=white,
  fonttitle=\bfseries\small,
  arc=2pt,
  boxrule=0.5pt,
  left=6pt, right=6pt, top=6pt, bottom=6pt,
  toptitle=3pt, bottomtitle=3pt,
]
\textbf{Farmos Field Kit} $\vert$ \textit{Farmer}
\smallskip

\small Create a Harvest log in farmOS Field Kit to document today's egg collection. Name the log 'Daily Egg Collection - Red Barn' with a quantity of 22 dozen. In the notes, record: 'Collected from nest boxes. 4 cracked eggs...

\vspace{4pt}\hrule\vspace{4pt}

\textbf{Qground Control} $\vert$ \textit{Agricultural UAV Technician}
\smallskip

\small You are an Agricultural UAV Technician preparing a spray drone for autonomous operations. The target field is bordered by a 15-meter tall eucalyptus windbreak on the north side, making the standard direct-descent retu...

\vspace{4pt}\hrule\vspace{4pt}

\textbf{Ekylibre} $\vert$ \textit{Farm Manager}
\smallskip

\small Register a new fertilizer product variant in the Ekylibre catalog named 'Ammonitrate 33.5'. The variant must be assigned a suitable nature (e.g., 'Engrais minéral' or 'Matière') and use 'Kilogram' (kg) as the unit.
\end{tcolorbox}
\vspace{4pt}

\begin{tcolorbox}[
  title={SOC 19/22: Construction and Extraction},
  enhanced,
  breakable,
  colback=white,
  colframe=promptframe,
  colbacktitle=prompttitlebg,
  coltitle=white,
  fonttitle=\bfseries\small,
  arc=2pt,
  boxrule=0.5pt,
  left=6pt, right=6pt, top=6pt, bottom=6pt,
  toptitle=3pt, bottomtitle=3pt,
]
\textbf{Subsurface} $\vert$ \textit{Commercial Diver}
\smallskip

\small Update the buddy field to 'Michael Chen' for the first dive in the logbook (Dive \#2, December 4, 2010 at Sund Rock, Hoodsport, WA) and save the changes.

\vspace{4pt}\hrule\vspace{4pt}

\textbf{System Advisor Model} $\vert$ \textit{Solar Photovoltaic Installer}
\smallskip

\small A solar installer in Tucson, Arizona wants to determine the optimal panel tilt angle for a residential PV system. The customer has a 5 kW system planned for a fixed south-facing roof, but the roof pitch is adjustable ...

\vspace{4pt}\hrule\vspace{4pt}

\textbf{Emoncms} $\vert$ \textit{Solar Photovoltaic Installer}
\smallskip

\small A solar PV installation is reporting negative power values because the Current Transformer (CT) sensor was installed backwards. Correct the polarity for node 'garage\_solar', input 'power' by inverting the values. Log ...
\end{tcolorbox}
\vspace{4pt}

\begin{tcolorbox}[
  title={SOC 20/22: Installation, Maintenance, and Repair},
  enhanced,
  breakable,
  colback=white,
  colframe=promptframe,
  colbacktitle=prompttitlebg,
  coltitle=white,
  fonttitle=\bfseries\small,
  arc=2pt,
  boxrule=0.5pt,
  left=6pt, right=6pt, top=6pt, bottom=6pt,
  toptitle=3pt, bottomtitle=3pt,
]
\textbf{Vtiger Crm} $\vert$ \textit{Medical Equipment Repairer}
\smallskip

\small Update the asset with Serial Number 'SN-US-2024-9981' (Sonosite Edge II - Radiology Dept) by changing its Status to 'Out of Service'. Create a related HelpDesk Ticket for this asset with the Title 'Dead Transducer Pro...

\vspace{4pt}\hrule\vspace{4pt}

\textbf{Crimson} $\vert$ \textit{SCADA Technician}
\smallskip

\small You are a SCADA technician configuring HMI-level fallback pump sequencing logic for a municipal wastewater plant's Lift Station A (LS-A) in Red Lion Crimson 3.0. Using the IO tag register and Ten States engineering el...

\vspace{4pt}\hrule\vspace{4pt}

\textbf{Graphite} $\vert$ \textit{Network Operations Center (NOC) Analyst}
\smallskip

\small You are a Network Operations Center (NOC) analyst preparing for a Monday morning ops review. Create a Graphite dashboard named 'Weekly Ops Review' containing three graphs comparing current metrics against data from 7 ...
\end{tcolorbox}
\vspace{4pt}

\begin{tcolorbox}[
  title={SOC 21/22: Production},
  enhanced,
  breakable,
  colback=white,
  colframe=promptframe,
  colbacktitle=prompttitlebg,
  coltitle=white,
  fonttitle=\bfseries\small,
  arc=2pt,
  boxrule=0.5pt,
  left=6pt, right=6pt, top=6pt, bottom=6pt,
  toptitle=3pt, bottomtitle=3pt,
]
\textbf{Chrome} $\vert$ \textit{Prepress Technician}
\smallskip

\small Configure the prepress terminal's Chrome browser for handling heavy graphics files securely. Read the specification document at \textasciitilde\{\}/Desktop/prepress\_terminal\_spec.txt. You must: 1) Force PDFs to download instead of open...

\vspace{4pt}\hrule\vspace{4pt}

\textbf{Erpnext} $\vert$ \textit{Quality Control Inspector}
\smallskip

\small Wind Power LLC has experienced inconsistent quality with the 'Shaft' components received from Eagle Hardware. Your job is to enforce mandatory incoming quality checks. Set up a Quality Inspection Template named 'Shaft...

\vspace{4pt}\hrule\vspace{4pt}

\textbf{Libreoffice Calc} $\vert$ \textit{Brewer}
\smallskip

\small Organize the homebrewing data and calculate the ABV for each batch using the formula ABV = (OG - FG) × 131.25. Apply conditional formatting to highlight batches within the target ABV range of 4.5-6.5\% and save the res...
\end{tcolorbox}
\vspace{4pt}

\begin{tcolorbox}[
  title={SOC 22/22: Transportation and Material Moving},
  enhanced,
  breakable,
  colback=white,
  colframe=promptframe,
  colbacktitle=prompttitlebg,
  coltitle=white,
  fonttitle=\bfseries\small,
  arc=2pt,
  boxrule=0.5pt,
  left=6pt, right=6pt, top=6pt, bottom=6pt,
  toptitle=3pt, bottomtitle=3pt,
]
\textbf{Sygic Gps} $\vert$ \textit{Delivery Driver}
\smallskip

\small Search for Kabul Airport (Hamid Karzai International Airport) and add it to your Favorites. Ensure the Favorites list is open to show the saved location.

\vspace{4pt}\hrule\vspace{4pt}

\textbf{Subsurface} $\vert$ \textit{Commercial Diver}
\smallskip

\small Update dive \#2 (December 4, 2010, at Sund Rock) in the dive log at /home/ga/Documents/dives.ssrf by adding a weight system entry with type 'Integrated' and weight '4.5 kg'. Save the updated dive log to /home/ga/Docume...

\vspace{4pt}\hrule\vspace{4pt}

\textbf{Chrome} $\vert$ \textit{Ship Officer}
\smallskip

\small Configure the bridge computer's Chrome browser for an ultra-low-bandwidth, high-latency satellite internet connection (e.g., Iridium Certus). First, delete all 8 high-bandwidth entertainment bookmarks (YouTube, Netfli...
\end{tcolorbox}
\vspace{4pt}
\section{Experimental Setup Details}
\label{app:experimental_setup}

\subsection{Models Used Across the Pipeline}
\label{app:pipeline_models}

Table~\ref{tab:pipeline_models} lists the models used at each stage of the \method{} pipeline, along with the harness (agentic framework) used to run them.

\begin{table}[ht]
    \centering
    \small
    \begin{tabular}{@{} p{3.8cm} p{2.8cm} p{2.5cm} p{2.2cm} @{}}
    \toprule
    \textbf{Pipeline Stage} & \textbf{Model} & \textbf{Harness} & \textbf{Section} \\
    \midrule
    \multicolumn{4}{@{}l}{\textit{GDP-Grounded Software Selection}} \\
    Category extraction & GPT-5-High & -- & \S\ref{sec:selection} \\
    Category deduplication & Gemini-3-Flash-Preview & -- & \S\ref{sec:selection} \\
    Product enumeration & GPT-5-High & -- & \S\ref{sec:selection} \\
    Access-barrier evaluation & Gemini-3-Pro-Preview & -- & \S\ref{sec:selection} \\
    GDP attribution & GPT-5-High  & -- & \S\ref{sec:selection} \\
    \midrule
    \multicolumn{4}{@{}l}{\textit{Environment Creation (\S\ref{sec:scaling_software})}} \\
    Creation agent ($\text{Agent}_C$) & Claude Opus 4.5/4.6 & Claude Code & \S\ref{sec:scaling_software} \\
    Audit agent ($\text{Agent}_{\text{audit}}$) & Claude Opus 4.5/4.6 & Claude Code & \S\ref{sec:scaling_software} \\
    Memory summarization ($\text{Agent}_{\text{summ}}$) & Claude Opus 4.5/4.6 & Claude Code & \S\ref{sec:scaling_software} \\
    \midrule
    \multicolumn{4}{@{}l}{\textit{Task Generation (\S\ref{sec:tasks})}} \\
    Task proposer (seed tasks) & Claude Opus 4.5/4.6 & Claude Code & \S\ref{sec:tasks} \\
    Task amplifier & Gemini-3-Pro-Preview & -- & \S\ref{sec:tasks} \\
    VLM task filter & Gemini-3-Flash-Preview & --  & \S\ref{sec:tasks} \\
    Privileged info extraction & Gemini-3-Pro-Preview & Gemini CLI & \S\ref{sec:verifiers} \\
    Checklist generation & Gemini-3-Pro-Preview & -- & \S\ref{sec:verifiers} \\
    \midrule
    \multicolumn{4}{@{}l}{\textit{\bench{}-Long Generation}} \\
    Task design and implementation & Claude Opus 4.5/4.6 & Claude Code & App.~\ref{app:bench_long} \\
    \midrule
    \multicolumn{4}{@{}l}{\textit{Training}} \\
    Teacher (trajectory generation) & Kimi-K 2.5 & Based on Qwen3VL & \S\ref{sec:training} \\
    Student (distillation target) & Qwen3-VL-2B & From OSWorld Repository  & \S\ref{sec:training} \\
    \midrule
    \multicolumn{4}{@{}l}{\textit{Evaluation}} \\
    VLM verifier & Gemini-3-Flash-Preview & -- & \S\ref{sec:evaluation} \\
    Test-Time Auditing agent & Gemini-3-Flash-Preview & Custom Harness & \S\ref{sec:test_time_agent} \\
    \bottomrule
    \end{tabular}
    \caption{Models and harnesses used at each stage of the \method{} pipeline. Empty cells indicate no harness was used.}
    \label{tab:pipeline_models}
    \end{table}

\subsection{Evaluated Models}
\label{app:evaluated_models}

We evaluate four frontier models on \bench{}-Long: Gemini~3~Flash, Kimi-K~2.5, Claude~Sonnet~4.6, and GPT-5.4. We do not evaluate Claude~Opus~4.6 on \bench{}-Long due to cost constraints. Furhter, \$5-per-task budget across 200 tasks, would imply opus would use much fewer steps ($\sim100$) than other models. Furthermore, Opus~4.6 and Sonnet~4.6 achieve nearly identical performance on OSWorld (72.7 vs.\ 72.5), suggesting that the additional cost would likely yield limited additional signal on our benchmark.

\niparagraph{Agent harnesses.}
For GPT-5.4 and Claude Sonnet 4.6, we use their official agent harnesses from their respective documentation.\footnote{GPT-5.4: \url{https://developers.openai.com/api/docs/guides/tools-computer-use}}\footnote{Claude Sonnet 4.6: \url{https://github.com/anthropics/claude-quickstarts/tree/main/computer-use-demo/computer_use_demo}}
For Gemini 3 Flash and Kimi-K 2.5, official harnesses were not publicly available at the time of our experiments, so we adapted the Qwen3-VL harness from the OSWorld repository.\footnote{\url{https://github.com/xlang-ai/OSWorld/blob/main/mm_agents/qwen3vl_agent.py}}
For the Qwen3-VL student model, we use the OSWorld harness directly.
\section{Propose-and-Amplify Ablation: Qualitative Analysis}
\label{app:seed_ablation}

To understand why seed tasks improve amplified task quality (\S\ref{sec:tasks}), we compare tasks generated with and without seed examples on three representative software applications: Firefox (web browser), AstroImageJ (astronomical image analysis), and Moodle (learning management system). We analyze the differences along three dimensions.

\niparagraph{Realism.}
With seeds, tasks are grounded in professional workflows: Firefox tasks involve researching real websites (Grants.gov, NSF Award Search) from the perspective of specific roles such as an investigative journalist or a development director; AstroImageJ tasks reference real astronomical objects (Eagle Nebula, WASP-12b) and real techniques (photon transfer CCD gain measurement, light curve detrending); Moodle tasks reflect actual institutional operations (grade auditing, custom role creation with specific capabilities). Without seeds, tasks shift toward feature demonstrations: Firefox tasks focus on browser features (importing a CA certificate, exporting a TLS certificate), AstroImageJ tasks become generic processing operations (align images, create a master dark frame), and the domain-specific professional context is largely absent. Moodle is a partial exception, likely because its structure is well-represented in LLM training data.

\niparagraph{Difficulty and horizon.}
With seeds, tasks require 50--80 steps with multi-stage workflows that chain several operations: research, download, organize, and synthesize (Firefox); load data, process, measure, and report (AstroImageJ); configure database, set permissions, and verify via logs (Moodle). Without seeds, tasks typically require 30--50 steps and involve single-feature operations. For instance, a Firefox TLS certificate export task reduces to visiting a webpage, clicking the padlock, and downloading the certificate.

\niparagraph{Setup script quality.}
With seeds, setup scripts range from 50 to 164 lines and perform substantial data preparation: downloading real astronomical data from Hubble archives, generating synthetic FITS files with physically plausible noise models (Poisson noise, read noise, dark current), and populating databases with multi-table SQL inserts. Without seeds, setup scripts are 40--120 lines and often just launch the application and open a URL, with minimal data preparation. This makes tasks more fragile (dependent on external network resources) and less reproducible. One exception is a Firefox custom CA import task (without seeds), which creates a full PKI infrastructure in 140 lines, showing that the model can occasionally produce high-quality setups without seeds, but does so inconsistently.

\niparagraph{Summary.}
The seed tasks teach the amplifier two things the prompt alone cannot: (1)~what realistic professional work looks like for a specific software, and (2)~how to prepare a rich initial state with real or realistic data. Without these examples, the model falls back to its generic knowledge of the software's features, producing tasks that are simpler, less realistic, and less reproducible.
\section{Trajectory Behavioral Analysis}
\label{app:trajectory_analysis}

To understand how agents behave on \bench{}, we run an automated behavioral analysis pipeline over 2,981 trajectories (701 passed, 2,280 failed) from our evaluation runs. The pipeline operates in three stages.

\niparagraph{Stage 1: Per-trajectory behavioral summary.}
Each trajectory (the full sequence of screenshots, actions, and model responses) is fed to an LLM, which decomposes the agent's behavior into natural phases and produces a high-level behavioral summary. Crucially, the LLM is not told whether the trajectory passed or failed, and is given no controlled vocabulary or predefined categories---it describes what happened in its own words. This avoids biasing the analysis toward expected failure modes.

\niparagraph{Stage 2: Pattern discovery.}
The per-trajectory summaries are shuffled into random mixed-environment batches and fed to an LLM, which identifies recurring behavioral patterns across each batch. The instruction requires environment-agnostic language: no software names, no UI element names, no application-specific terminology. After all batches are processed, a consolidation pass merges overlapping patterns into a canonical set of 15 deduplicated patterns. Each pattern is a short description of a recurring behavior (e.g., ``the agent enters retry loops when actions do not take effect'').

\niparagraph{Stage 3: Pattern matching.}
For each trajectory, we send its behavioral summary alongside the 15 canonical patterns to an LLM and ask which patterns are present. Each step in a trajectory can be tagged with multiple patterns (e.g., a step may involve both UI exploration and a retry), so fractions across patterns do not sum to one. For each pattern, we compute two metrics: \emph{step fraction} (what fraction of a trajectory's steps exhibit the pattern, averaged across trajectories) and \emph{presence rate} (what fraction of trajectories exhibit the pattern at least once). Both metrics are computed separately for passed and failed trajectories.

\begin{figure}[h]
    \centering
    \includegraphics[width=\textwidth]{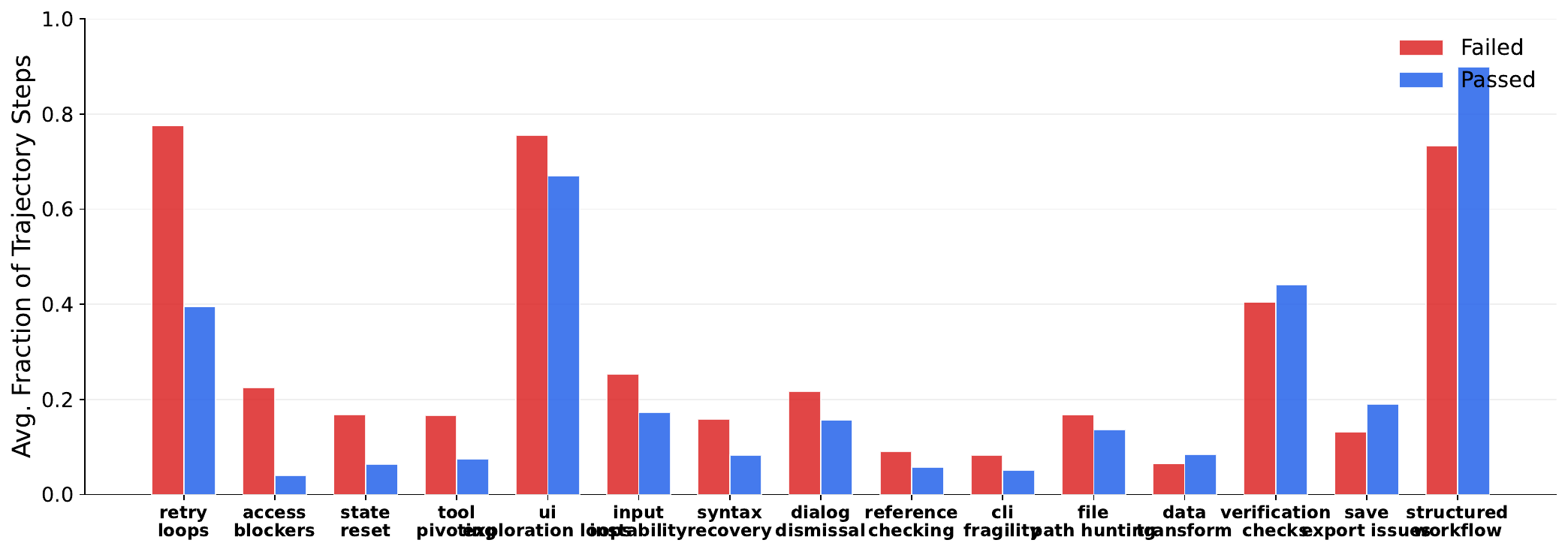}
    \caption{Step-weighted pattern intensity across all 15 discovered behavioral patterns. For each pattern, bars show the average fraction of trajectory steps exhibiting that pattern, split by passed vs.\ failed trajectories. Patterns are sorted by the gap between failed and passed (largest gap on the left).}
    \label{fig:trajectory_step_full}
\end{figure}

\begin{figure}[h]
    \centering
    \includegraphics[width=\textwidth]{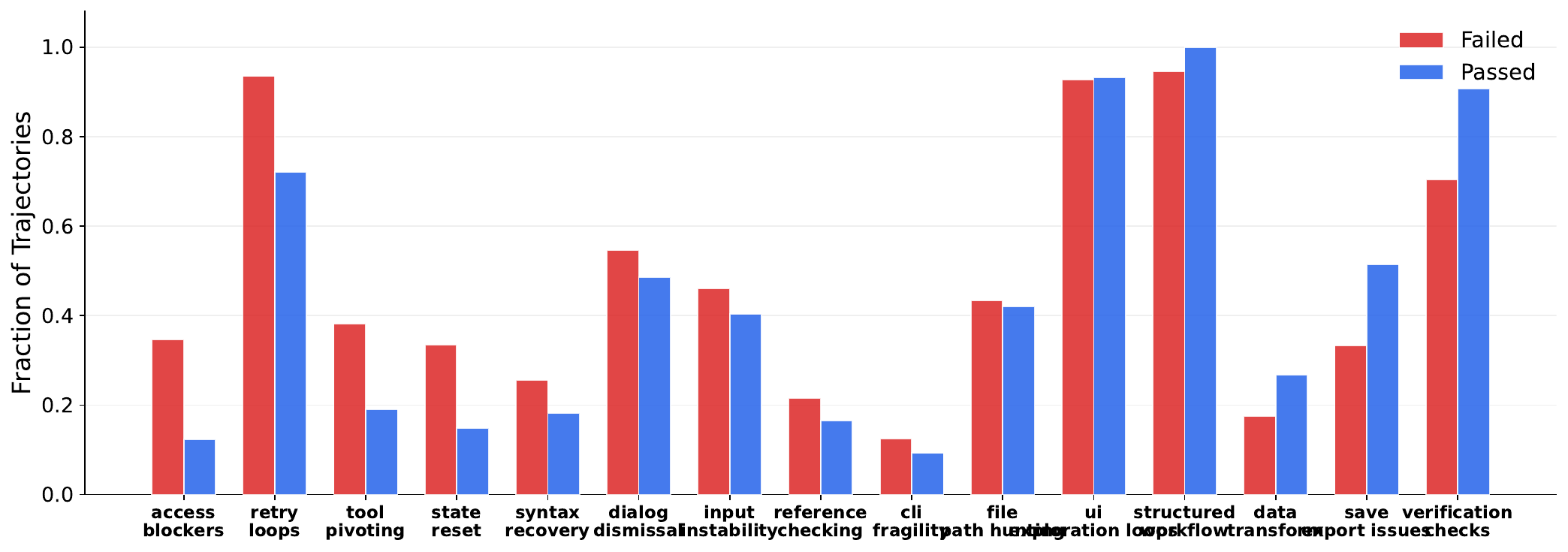}
    \caption{Pattern presence rate across all 15 discovered behavioral patterns. For each pattern, bars show the fraction of trajectories that exhibit the pattern at least once. Patterns are sorted by the gap between failed and passed.}
    \label{fig:trajectory_presence_full}
\end{figure}

\niparagraph{Results.}
Figure~\ref{fig:trajectory_step_full} shows the step-fraction view and Figure~\ref{fig:trajectory_presence_full} shows the presence-rate view across all 15 patterns. The three patterns highlighted in the main paper (\S\ref{sec:benchmark}) are \emph{retry loops}, \emph{UI exploration}, and \emph{verification checks}. Beyond these, several additional patterns show notable gaps between passed and failed trajectories. For instance, \emph{access blockers} (authentication failures, unreachable services) occupy 23\% of steps in failed trajectories but only 4\% in passed ones. \emph{Tool pivoting}---where the agent abandons the primary GUI and switches to CLI or alternative tools---is present in 38\% of failed trajectories but only 19\% of passed ones. On the positive side, \emph{save/export} steps are present in 52\% of passed trajectories but only 33\% of failed ones, reflecting that successful agents more often reach the final stage of the task.

\section{Extended Related Work}
\label{app:related_work}

This appendix expands on the related work discussion in the main paper (\S\ref{sec:related_work}).

\niparagraph{Benchmarks and datasets for computer-use agents.}
Existing work on evaluating computer-use agents can be divided into static datasets that provide scale and interactive benchmarks that test actual task completion.
Static datasets such as Mind2Web~\cite{deng2023mind2web}, Android in the Wild~\cite{rawles2023androidinthewild}, and AndroidControl~\cite{androidcontrol2024} offer thousands of annotated episodes across hundreds of applications, but evaluation is limited to action-matching against recorded traces rather than execution-based verification, so valid alternative strategies are penalized.
Interactive web benchmarks range from synthetic micro-tasks~\cite{liu2018miniwob} to realistic self-hosted environments~\cite{webarena2024,visualwebarena2024,workarena2024,workarenaplus2024}, but cover at most six websites and are restricted to the browser.
On the desktop, OSWorld~\cite{osworld2024} provides 369 tasks across 9 applications on Linux; Windows Agent Arena~\cite{windowsagentarena2025}, Spider2-V~\cite{spider2v2024}, AssistGUI~\cite{assistgui2024}, TheAgentCompany~\cite{theagentcompany2024}, and ScienceBoard~\cite{scienceboard2025} extend coverage to Windows, data science, and scientific domains but remain limited to 100--494 tasks and 5--20 applications each.
On mobile, AndroidWorld~\cite{androidworld2025} provides 116 interactive tasks across 20 apps.
ProgrammingWithPixels~\cite{programmingwithpixels2025} scales to 5,400 task instances but within a single application (VS~Code).
Across prior interactive benchmarks, environment creation is typically manual, which limits their scale, and none simultaneously provides a training split, long-horizon tasks exceeding 100 steps, or broad occupational coverage.
\bench{} bridges the gap between the scale of static datasets and the execution-based evaluation of interactive benchmarks: by automating environment creation through the creation-audit loop (\S\ref{sec:scaling_software}), it provides 10K+ interactive tasks across 200+ software applications on four platforms, with train/test splits, a long-horizon benchmark requiring 200+ steps, and GDP-grounded coverage of all 22 SOC occupation groups.

\niparagraph{Automated environment and task generation.}
A growing body of work generates tasks or trajectories within pre-existing environments.
AgentTrek~\cite{xu2024agenttrek} synthesizes web trajectories from online tutorials, OS-Genesis~\cite{osgenesis2025} derives tasks retrospectively from agent exploration, and several other methods propose tasks from GUI observations or evolve curricula within fixed environments~\cite{pae2024,bagel2024,webrl2025}.
However, these approaches are bounded by the set of environments that already exist; they cannot create new ones.
A parallel line of work uses LLMs to generate environments, but within narrow domains: text-based planning tasks~\cite{agentgen2024}, 3D indoor scenes for embodied AI~\cite{holodeck2024}, tool-use API compositions~\cite{agentworldmodel2026}, or code-editing setups from GitHub repositories~\cite{swegym2024}.
None of these targets real GUI software that requires installation, configuration with domain-appropriate data, and interactive verification.
For task generation at scale, the seed-then-amplify paradigm is well established: Self-Instruct~\cite{selfinstruct2023} bootstraps instructions from a small seed set, and subsequent works evolve complexity~\cite{wizardlm2024} or use multi-agent pipelines~\cite{agentinstruct2024}, but these generate text instruction-response pairs rather than executable environment tasks.
\method{} addresses all three gaps: its creation-audit loop (\S\ref{sec:scaling_software}) converts real software into interactive environments via coding agents verified by an independent auditor, its propose-and-amplify strategy (\S\ref{sec:tasks}) generates tasks by having an agent actually run the software to produce high-quality seeds that are then amplified and filtered via execution, and a shared memory across agents ensures learnings accumulate so newer environments are created faster.

\niparagraph{Training computer-use agents.}
Trajectory distillation from strong models has emerged as an effective recipe for training GUI agents: AgentTrek~\cite{xu2024agenttrek} distills from web tutorial replays, Explorer~\cite{explorer2025} scales to over 94K web trajectories, and PC~Agent-E~\cite{pcagente2025} augments 312 human demonstrations to train a 72B model that surpasses Claude~3.7~Sonnet on WindowsAgentArena-V2.
Beyond distillation, alternative training strategies have also shown promise: DigiRL~\cite{digiRL2024} achieves a 49.5-point improvement over SFT with a 1.3B model, and UI-TARS~\cite{uitars2025} combines enhanced perception, unified action modeling, and iterative reflective training to achieve state-of-the-art results across multiple benchmarks.
Open vision-language backbones such as Qwen2-VL~\cite{wang2024qwen2vl} and Qwen2.5-VL~\cite{qwenteam2025qwen25vl} are common foundations for recent open GUI-agent systems~\cite{osatlas2025,opencua2025}.
However, existing training pipelines are typically limited to relatively small sets of applications, and scaling laws for GUI agent trajectory distillation remain underexplored.
Our distillation experiments across 200 software applications show log-linear scaling (${\sim}$3.5 points per data doubling), demonstrate that a 2B student can outperform models 2$\times$ its size, and reveal that cross-software generalization is limited (22--27\% recovery), motivating scalable environment creation.

\niparagraph{Evaluation of computer-use agents.}
Existing interactive benchmarks often use hand-written programmatic verifiers that check the final system state~\cite{osworld2024,webarena2024}, which can be reliable but are labor-intensive to author and maintain and typically provide binary pass/fail.
TheAgentCompany~\cite{theagentcompany2024} introduces checkpoint-based partial credit but still requires custom evaluator code per task.
The LLM-as-a-judge paradigm~\cite{mtbench2023} and its extensions to agent evaluation~\cite{agentasjudge2024} offer a more general alternative to script-based evaluation.
VLM-based evaluation has been explored in the CUA setting for filtering training trajectories~\cite{xu2024agenttrek}, step-level trajectory assessment~\cite{seagent2025}, and autonomous trajectory evaluation~\cite{panetal2024guieval}, while checklist-based evaluation has shown strong correlation with human preference for text generation~\cite{wildbench2024} and per-subgoal VLM evaluation has been explored in robotics~\cite{stepeval2025}.
Our checklist-based VLM verifier extends this line of work by incorporating privileged information extracted from environment setup scripts, enabling the verifier to check agent outputs against known ground-truth answers without per-task evaluation code.
We additionally introduce integrity checks that detect workflow bypasses such as fabricating report data or using the terminal instead of the intended GUI, an issue related to the broader problem of reward hacking in agent evaluation~\cite{evilgenie2025}.

\niparagraph{Economic impact of AI and occupation-grounded benchmarks.}
A substantial body of work studies which occupations are susceptible to AI automation, starting from Frey and Osborne's occupation-level risk estimates~\cite{freyosborne2017} and Eloundou et al.'s O*NET-based LLM exposure framework~\cite{eloundou2023gpts}.
Felten et al.~\cite{felten2021aioe} introduced the widely adopted AI Occupational Exposure index, and Acemoglu~\cite{acemoglu2024macroai} formalized GDP-level impact estimation via task-level cost savings.
These studies use occupational data, often from O*NET, to connect AI capabilities to labor-market outcomes, but focus on measuring exposure rather than directing benchmark design.
Wang et al.~\cite{wang2026doesagentdevelopmentreflect} recently quantified this gap, finding that across 43 agent benchmarks and 72K tasks, coverage is heavily programming-centric, with much less representation in many economically significant domains outside computing.
GDPval~\cite{gdpval2025} takes a step toward economic grounding by evaluating models on tasks from 44 occupations across 9 GDP-contributing industries, but is limited to one-shot evaluation rather than interactive agentic tasks.
\method{} inverts the standard direction: rather than measuring which occupations are exposed to AI, it uses per-software GDP attribution to determine which software to include in an agent benchmark, covering all 22 SOC major occupation groups with interactive, execution-verified tasks.

\stopcontents[appendix]

\end{document}